\documentclass[a4paper,11pt,twoside]{book}

\usepackage{hyperref,graphics,epsfig,pdfsync}
\usepackage[utf8]{inputenc}
\usepackage[english]{babel}
\usepackage{amsfonts}
\usepackage{graphicx}
\usepackage{amsmath,amsthm, amssymb} 
\usepackage[square]{natbib}
\usepackage[utf8]{inputenc}
\usepackage[toc,page]{appendix}

\usepackage{palatino} 

\usepackage{graphicx}

\usepackage{newfloat}
\usepackage{caption}

\usepackage{tabularx}

\usepackage{booktabs}

\usepackage{multirow}

\usepackage{array}
\newcolumntype{M}[1]{>{\centering\arraybackslash}m{#1}}
\newcolumntype{N}{@{}m{0pt}@{}}

\newcommand{\specialcell}[2][c]{%
  \begin{tabular}[#1]{@{}c@{}}#2\end{tabular}}

\usepackage{algorithmicx}
\usepackage{algpseudocode} 
\DeclareFloatingEnvironment[fileext=frm,placement={!ht},name=Algorithm]{algorithm}

\usepackage{enumerate, subfigure, color}

\makeatletter
\def\blfootnote{\xdef\@thefnmark{}\@footnotetext}
\makeatother

\newcommand{\ind}{1{\hskip -2.5 pt}\hbox{I}}

\def\argmax{\operatornamewithlimits{arg\,max}}
\def\argmin{\operatornamewithlimits{arg\,min}}

\DeclareMathOperator{\E}{\mathbb{E}}
\renewcommand{\P}{\mathbb{P}}

\newcommand{\rp}{\right)}

\newcommand{\iii}{\begin{enumerate}}
\newcommand{\fff}{\end{enumerate}}
\newcommand{\iiii}{\begin{itemize}}
\newcommand{\ffff}{\end{itemize}}
\newcommand{\mfi}{\begin{eqnarray*}}
\newcommand{\mff}{\end{eqnarray*}}
\newcommand{\mfni}{\begin{eqnarray}}
\newcommand{\mfnf}{\end{eqnarray}}

\newtheorem{thm}{Theorem}

\newtheorem{rem}{Remark}
\newtheorem{ex}{Example}

\providecommand{\nor}[1]{\left\lVert {#1} \right\rVert}

\providecommand{\scal}[2]{\left\langle{#1},{#2}\right\rangle}

\newcommand{\R}{\mathbb R}
\newcommand{\CC}{\R}

\newcommand{\N}{\mathbb N}

\newcommand{\hh}{\mathcal H}
\newcommand{\EE}{\mathcal E}
\newcommand{\eps}{\varepsilon}
\newcommand{\EM}{\widehat{\mathcal E}}

\newcommand{\la}{\lambda}

\newcommand{\F}{\mathcal F}
\newcommand{\V}{\mathcal V}
\newcommand{\fn}{\widehat{f}}
\newcommand{\En}{\widehat{\mathcal E}}

\newcommand{\X}{\mathcal X}
\newcommand{\Y}{\mathcal Y}
\newcommand{\Z}{\mathcal Z}

\newcommand{\eqals}[1]{\begin{align*}#1\end{align*}}
\newcommand{\eqal}[1]{\begin{align}#1\end{align}}
\newcommand{\bpr}{\begin{proof}}
\newcommand{\epr}{\end{proof}}
\newcommand{\be}{\begin{equation}}
\newcommand{\ee}{\end{equation}}

\newtheorem{definition}{Definition}
\newcommand{\bd}{\begin{definition}}
\newcommand{\ed}{\end{definition}}

\newcommand{\bi}{\begin{itemize}}
\newcommand{\ei}{\end{itemize}}

\newtheorem{ass}{Assumption}
\newcommand{\ba}{\begin{ass}}
\newcommand{\ea}{\end{ass}}

\newtheorem{remark}{Remark}
\newcommand{\br}{\begin{remark}}
\newcommand{\er}{\end{remark}}

\newtheorem{example}{Example}
\newcommand{\bex}{\begin{example}}
\newcommand{\eex}{\end{example}}

\newtheorem{proposition}{Proposition}
\newcommand{\bp}{\begin{proposition}}
\newcommand{\ep}{\end{proposition}}

\newtheorem{lemma}{Lemma}
\newcommand{\blm}{\begin{lemma}}
\newcommand{\elm}{\end{lemma}}

\newtheorem{theorem}{Theorem}
\newcommand{\bt}{\begin{theorem}}
\newcommand{\et}{\end{theorem}}

\newtheorem{corollary}{Corollary}
\newcommand{\bcor}{\begin{corollary}}
\newcommand{\ecor}{\end{corollary}}

\newcommand{\Nystrom}[1]{{Nystr\"om}}

\newcommand{\rhox}{{\rho_{\X}}}
\newcommand{\Ltwo}{{L^2(\X,\rhox)}}

\newcommand{\K}{K}
\newcommand{\mK}{K_n}

\newcommand{\yn}{\widehat{y}}

\newcommand{\tS}{\tilde{S}}
\newcommand{\tSn}{\tS_n}

\newcommand{\tkappa}{\tilde{\kappa}}

\newcommand{\frho}{f_\rho}

\newcommand{\SNR}{\textrm{SNR}}

\newcommand{\mR}{\mathbb{R}}
\newcommand{\mN}{\mathbb{N}}
\newcommand{\mE}{\mathbb{E}}
\newcommand{\mcH}{\mathcal{H}}
\newcommand{\mcF}{\mathcal{F}}

\newcommand{\mcE}{\mathcal{E}}

\DeclareMathOperator*{\arginf}{arg\,inf}
\newcommand{\eref}[1] {(\ref{#1})}

\providecommand{\scal}[2]{\langle{#1},{#2}\rangle}

\usepackage{textcomp}

\newtheorem{alg}{Algorithm}
\newtheorem{as}{Assumption}

\usepackage{adjustbox/adjustbox}

\usepackage[table]{xcolor}
\definecolor{Gray}{gray}{0.85}

\usepackage{graphicx}
\usepackage[percent]{overpic} 
\usepackage{bm}

\newcommand{\regressor}{\ensuremath{\Phi(x)}}
\newcommand{\regressorNum}[1]{\ensuremath{\Phi(x_{#1})}}
\newcommand{\outputSet}{\ensuremath{\mathbf{y}}}
\newcommand{\outputQuant}{\ensuremath{y}}
\newcommand{\nrOfSamples}{n}
\newcommand{\inertialParameters}{\bm{\pi}}
\newcommand{\estimatedInertialParameters}{\hat{\bm{\pi}}}
\newcommand{\RLSinputMat}{Z}
\newcommand{\RLSoutputMat}{U}
\newcommand{\RLSinputSample}{z}
\newcommand{\RLSoutputSample}{u}

\author{Raffaello Camoriano}
\title{\textsc{Large-scale Kernel Methods\\ and Applications to Lifelong Robot Learning}}
\date{}

\begin{document}


\maketitle

\thispagestyle{empty}
\begin{center}
\begin{minipage}{0.75\linewidth}
    \centering
    \includegraphics[height=2cm]{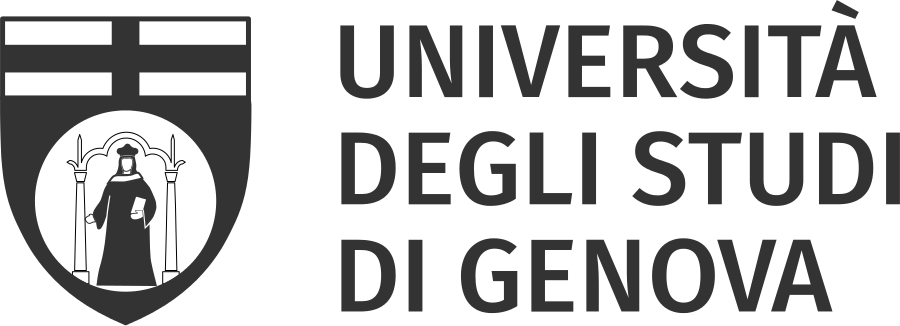}\hfill
    \includegraphics[height=2cm]{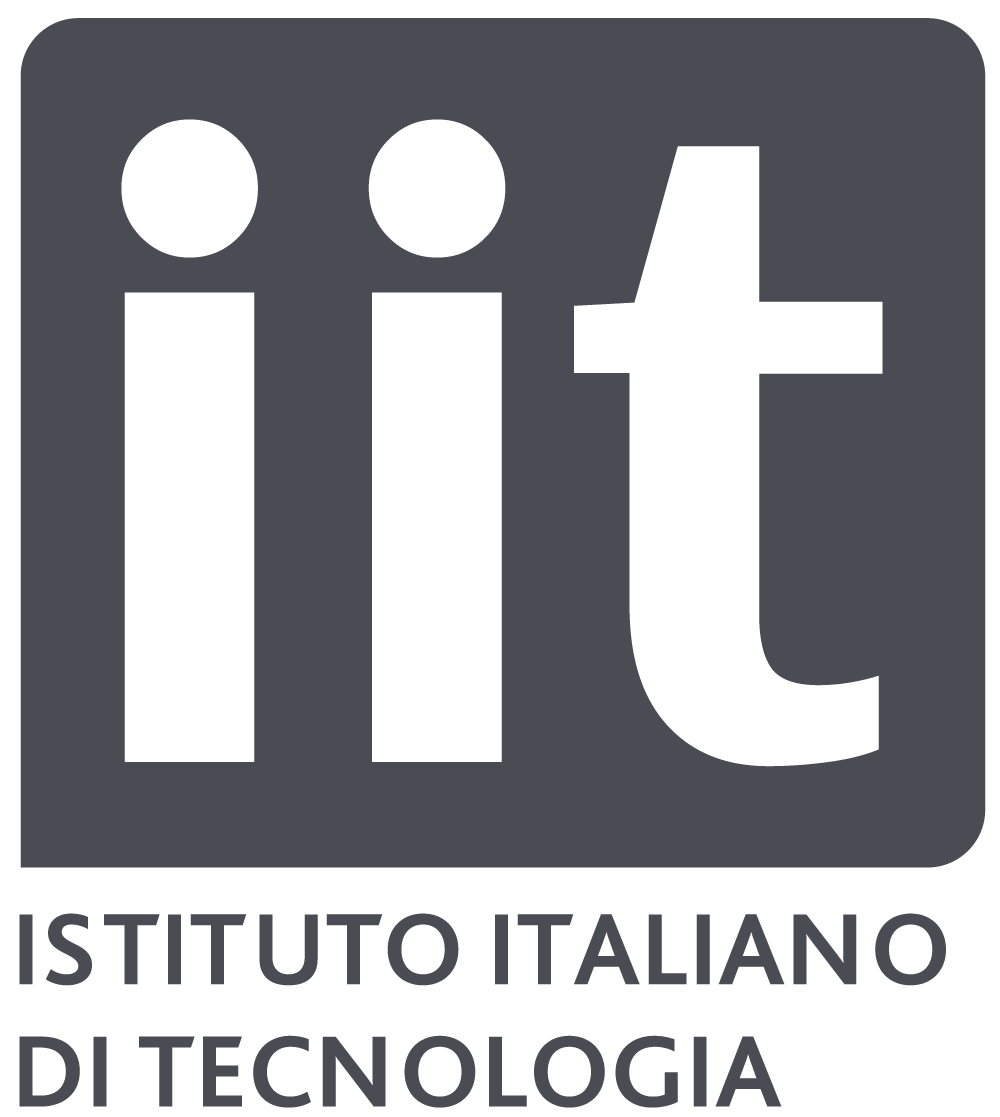}\par
    \vspace{1.5cm}
    {\sc{\LARGE Large-scale Kernel Methods\\ and Applications to Lifelong Robot Learning \par}}
    \vspace{1cm}
    {by\\ \vspace{0.5cm} \Large Raffaello Camoriano\par}
    \vspace{1cm}
    {A thesis submitted in partial fulfillment of the requirements for the degree of\par}
    \vspace{0.3cm}
    {\sc{\large Doctor of Philosophy\par}}
    \vspace{0.3cm}
    {Doctoral Course in Bioengineering and Robotics}\\
    {XXIX Cycle}\\
    {Curriculum in Humanoid Robotics}\\
    \vspace{0.5cm}
    {Advisors:\\Prof. Giorgio Metta\\ Prof. Lorenzo Rosasco}\\
    \vspace{0.5cm}
    {iCub Facility\\Istituto Italiano di Tecnologia (IIT)}\\
    \vspace{0.3cm}
    {Laboratory for Computational and Statistical Learning\\Istituto Italiano di Tecnologia (IIT)}\\
    \vspace{0.3cm}
    {Dipartimento di Informatica, Bioingegneria, Robotica e Ingegneria dei Sistemi (DIBRIS)\\Universit\`a degli Studi di Genova (UNIGE)}\\
    \vspace{1cm}
    {\Large March 2017}\\
    \vspace{1cm}
    {$\copyright$ 2014--2017 Raffaello Camoriano}
\end{minipage}
\end{center}

%

\chapter*{Abstract}
In the last few decades, the digital revolution has produced an unparalleled abundance of data in many fields.
The capability of automatically identifying recurring patterns and extracting information from data has acquired a prominent role in the value creation chain of today's economy and in the advancement of science.
As the size and richness of available datasets grow larger, the opportunities for solving increasingly challenging problems with algorithms learning directly from data grow at the same pace.
Notable examples of learning problems are visual object recognition, speech recognition, and lifelong robot learning, empowering robotic agents to learn continuously from experience in dynamic environments.
Consequently, the capability of learning algorithms to work with large amounts of data has become a crucial scientific and technological challenge for their practical applicability.
Hence, it is no surprise that large-scale learning is currently drawing plenty of research effort in the machine learning research community.
In this thesis, we focus on kernel methods, a theoretically sound and effective class of learning algorithms yielding nonparametric estimators.
Kernel methods, in their classical formulations, are accurate and efficient on datasets of limited size, but do not scale up in a cost-effective manner.
Recent research has shown that approximate learning algorithms, for instance random subsampling methods like Nystr\"om and random features, with time-memory-accuracy trade-off mechanisms are more scalable alternatives.
In this thesis, we provide analyses of the generalization properties  and computational requirements of several types of 
such approximation schemes.
%
In particular, we expose the tight relationship between statistics and computations, with the goal of tailoring the accuracy of the learning process to the available computational resources.
Our results are supported by experimental evidence on large-scale datasets and numerical simulations.
We also study how large-scale learning can be applied to enable accurate, efficient, and reactive lifelong learning for robotics.
In particular, we propose algorithms allowing robots to learn continuously from experience and adapt to changes in their operational environment.
The proposed methods are validated on the iCub humanoid robot in addition to other benchmarks.

\chapter*{Acknowledgments}
The fruitful path of my doctoral studies would not have been possible without the help, the support, and the invaluable insights of a number of extraordinary and open-minded people.
I start by thanking my advisors, Giorgio Metta and Lorenzo Rosasco, whose guidance and consideration inspired me to grow as a scientist, and also, I hope, as a person.
My deep (or shallow, as they may prefer) gratitude also goes to Alessandro Rudi, with whom I have had the privilege to closely and extensively work in these years on exciting projects, and Giulia Pasquale, who has always been available to discuss about any matter regarding our shared experience as Ph. D. students and friends.
I would also like to warmly thank all my other co-authors: Tomás Angles, Alessandro Chiuso, Carlo Ciliberto, Junhong Lin, Diego Romeres, Lorenzo Natale, Francesco Nori, Silvio Traversaro, and Mattia Zorzi.
I have been lucky enough to be part of three unique research groups: iCub Facility (IIT), the Laboratory for Computational and Statistical Learning (LCSL, IIT-MIT), and the Slipguru group (DIBRIS, University of Genoa).
It would take too long to name all the exceptional people with whom I have had the privilege and pleasure to discuss about scientific and more worldly topics throughout these years.
I will only mention some of them in random order, aware of the fact that I am probably forgetting too many (in this case, I beg your pardon!): Silvia Villa, Bang C\^{o}ng Vu, Guillaume Garrigos, Saverio Salzo, Alessandro Verri, Matteo Barbieri, Samuele Fiorini, Gian Maria Marconi, Elisa Maiettini, Fabio Anselmi, Gemma Roig, Luigi Carratino, Enrico Cecini, Andrea Tacchetti, Matteo Santoro, Beno\^{i}t Dancoisne, Georgios Evangelopoulos, Maximilian Nickel, Ugo Pattacini, Vadim Tikhanoff, Alessandro Roncone, Francesco Romano, Tanis Mar, Matej Hoffmann, Massimo Regoli, Anand Suresh, Andrea Schiappacasse, Francesca Odone, Ernesto De Vito, Annalisa Barla, Fabio Solari, Manuela Chessa, Nicoletta Noceti, Alessia Vignolo, Damiano Malafronte, Federico Tomasi, Sriram Kumar, Prabhu Kumar, Stefano Saliceti, Ali Paikan, Daniele Pucci, Elena Ceseracciu, Sean Ryan Fanello, and Arjan Gijsberts.
I am also very thankful to the reviewers of this thesis, Alessandro Chiuso and Alessandro Lazaric, for their patience and useful comments, and to all the reviewers of the works I submitted to the attention of the scientific community.
On a more personal level, I am and will always be wholeheartedly grateful to my parents, Carla and Gian Pietro, my grandmother Gaby, my entire family, my girlfriend Sara and her family, and my best friends Antonio, Claire, Edoardo, Enrico, Francesco, Ilaria, and all of those who decisively helped me to complete this extremely challenging and arduous path.

This journey has definitely been the best time of my life, and I feel honored to have shared it with every single one of you.

\chapter*{Notation}
\noindent
$\displaystyle
\begin{array}{ll}
\N & \mbox{natural numbers}\\
\R & \mbox{real numbers}\\
\R_+ & \mbox{positive real numbers}\\
z & \mbox{column vector (if not differently specified)}\\
Z & \mbox{matrix}\\
z^\top, Z^\top & \mbox{transpose of vector $z$ or matrix $Z$}\\
x \in \X & \mbox{input sample and input space}\\
X & \mbox{input matrix (each row is a sample $x^\top$)}\\
y \in \Y & \mbox{output label (or vector) and output space}\\
Y & \mbox{output matrix (each row is an output label --- or vector --- $y^\top$)}\\
\Z = \X \times \Y& \mbox{data space}\\
n \in \N & \mbox{number of training samples}\\
d \in \N & \mbox{dimensionality of the input space $\X$}\\
T \in \N & \mbox{dimensionality of the output space $\Y$}\\
S_n & \mbox{set of $n$ input-output pairs $\{x_i,y_i\}_{i=1}^n$}\\
\scal{\cdot}{\cdot}_\hh & \mbox{inner product in space $\hh$}\\
\|\cdot\|_p & \mbox{$p$-norm, with $p=2$ if not specified}\\
f(\cdot)\in \F & \mbox{function belonging to a space of functions}\\
K(\cdot,\cdot) & \mbox{kernel function}\\
K_n& \mbox{empirical kernel matrix of size $n\times n$ associated to $K$}\\
\hh_K & \mbox{reproducing kernel Hilbert space (RKHS) associated to $K$}\\
\Phi(\cdot) & \mbox{feature map}\\
I_n & \mbox{$n\times n$ identity matrix}\\
\mathbf{0} & \mbox{zero vector}\\
\hat y & \mbox{predicted output}\\
\ell(\cdot,\cdot) & \mbox{loss function}\\
\rho(\cdot) & \mbox{probability distribution}\\
{\mathcal N}(\mu,\sigma^2) & \mbox{Gaussian distribution with mean $\mu$ and variance $\sigma^2$}\\
{\mathcal U}(a,b) & \mbox{uniform distribution in the interval $[a,b]$}\\
\text{sign}(a) & \mbox{sign of $a$}\\
\text{rank}(Z) & \mbox{rank of $Z$}\\
\text{Tr}(Z) & \mbox{trace of $Z$}\\
\E & \mbox{expectation of a random variable}
\end{array}
$

%
\chapter*{Acronyms}
\noindent
$\displaystyle
\begin{array}{ll}
\mbox{ALS} & \mbox{Approximate Leverage Scores}\\
\mbox{ERM} & \mbox{Empirical Risk Minimization}\\
\mbox{ES} & \mbox{Early Stopping}\\
\mbox{F/T} & \mbox{Force/Torque}\\
\mbox{FFT} & \mbox{Fast Fourier Transform}\\
\mbox{GPR} & \mbox{Gaussian Process Regression}\\
\mbox{i. i. d.} & \mbox{independent and identically distributed}\\
\mbox{KOLS} & \mbox{Kernel Ordinary Least Squares}\\
\mbox{KRLS} & \mbox{Kernel Regularized Least Squares}\\
\mbox{KRR} & \mbox{Kernel Ridge Regression}\\
\mbox{LGP} & \mbox{Local Gaussian Processes}\\
\mbox{LWPR} & \mbox{Locally Weighted Projection Regression}\\
\mbox{NKRLS} & \mbox{Nystr\"om Kernel Regularized Least Squares}\\
\mbox{NP} & \mbox{Nonparametric}\\
\mbox{NYTRO} & \mbox{NYstr\"om iTerative RegularizatiOn}\\
\mbox{P} & \mbox{Parametric}\\
\mbox{PCA} & \mbox{Principal Component Analysis}\\
\mbox{PCR} & \mbox{Principal Component Regression}\\
\mbox{RBD} & \mbox{Rigid Body Dynamics}\\
\mbox{RBF} & \mbox{Radial Basis Function}\\
\mbox{RF} & \mbox{Random Features}\\
\mbox{RF-KRLS} & \mbox{Random Features Kernel Regularized Least Squares}\\
\mbox{RFRRLS} & \mbox{Random Features Recursive Regularized Least Squares}\\
\mbox{RKHS} & \mbox{Reproducing Kernel Hilbert Space}\\
\mbox{RLS(C)} & \mbox{Regularized Least Squares (for Classification)}\\
\mbox{(R)MSE} & \mbox{(Root) Mean Square Error}\\
\mbox{RRLS} & \mbox{Recursive Regularized Least Squares}\\
\mbox{SGD} & \mbox{Stochastic Gradient Descent}\\
\mbox{S(I)GM} & \mbox{Stochastic (Incremental) Gradient Method}\\
\mbox{SLT} & \mbox{Statistical Learning Theory}\\
\mbox{SP} & \mbox{Semiparametric}\\
\mbox{SVM} & \mbox{Support Vector Machine}\\
\mbox{(T)SVD} & \mbox{(Truncated) Singular Value Decomposition}
\end{array}
$


\chapter*{List of Publications}
\noindent Camoriano, R.$^*$, Angles, T.$^*$, Rudi, A., and Rosasco, L. (2016a). NYTRO: When Subsampling Meets Early Stopping. In \textit{Proceedings of the 19th Inter-
national Conference on Artificial Intelligence and Statistics (AISTATS)}, pages 1403–1411.\\

\noindent Camoriano, R.$^*$, Pasquale, G.$^*$, Ciliberto, C., Natale, L., Rosasco, L., and Metta, G. (2016b). Incremental Robot Learning of New Objects with Fixed Update Time. \textit{arXiv preprint} arXiv:1605.05045.\\

\noindent Camoriano, R., Traversaro, S., Rosasco, L., Metta, G., and Nori, F. (2016c).
Incremental semiparametric inverse dynamics learning. In \textit{IEEE
International Conference on Robotics and Automation (ICRA)}, pages 544–550.
IEEE.\\

\noindent Camoriano, R.$^*$, Pasquale, G.$^*$, Ciliberto, C., Natale, L., Rosasco, L., and Metta, G. (2017). Incremental Robot Learning of New Objects with Fixed Update Time. To appear in \textit{IEEE International Conference on Robotics and Automation (ICRA).}\\

\noindent Lin, J., Camoriano, R., and Rosasco, L. (2016). Generalization Properties
and Implicit Regularization for Multiple Passes SGM. In \textit{International Conference on Machine Learning (ICML)}.\\

\noindent Romeres, D., Zorzi, M., Camoriano, R., and Chiuso, A. (2016). Online semiparametric learning for inverse dynamics modeling. In \textit{Decision and Control (CDC), 2016 IEEE 55th Conference on}, pages 2945–2950. IEEE.\\

\noindent Rudi, A., Camoriano, R., and Rosasco, L. (2015). Less is More: Nystr\"om
Computational Regularization. In \textit{Advances in Neural Information Processing Systems (NIPS) 28}, pages 1657–1665. Curran Associates, Inc.\\

\noindent Rudi, A., Camoriano, R., and Rosasco, L. (2016). Generalization Properties
of Learning with Random Features. \textit{ArXiv e-prints}.
\blfootnote{$^*$ Equal contribution.}
\clearpage

\tableofcontents

%

\chapter*{Introduction}
\label{Chap:intro}
\addcontentsline{toc}{chapter}{\nameref{Chap:intro}}
Research in machine learning began in the 1950s within the field of artificial intelligence, at the intersection between statistics and computer science, with the objective of devising computer programs able to solve problems by learning from experience \citep{turing1950computing,rosenblatt1958perceptron,minsky1969perceptrons,crevier1993ai,russell2003artificial}.
Many tasks in visual object recognition, speech recognition, and system identification and estimation, among others, can be framed in terms of learning the associated input-output mappings directly from examples.
In other words, if a system solving the problem cannot be deterministically defined a-priori, it might be possible to find it by training a predictive model on data relevant to the task, made available by sensors and, possibly, a supervising agent.
In recent years, the availability of large-scale datasets in many fields, such as robotics and computer vision, has posed an unprecedented opportunity towards the conception of novel learning systems enabling artificial agents to solve increasingly challenging problems without being explicitly pre-programmed.
At the same time, a crucial scientific and technological challenge for the feasibility of a learning system is the scalability of the learning algorithm with respect to the number of training samples and their dimensionality.
%
Among the numerous machine learning approaches available nowadays, we focus on kernel methods, a set of nonparametric methods with a firm theoretical grounding and the potential for high predictive accuracy.
Exact kernel methods \citep{schlkopf2002learning,shawe-taylor2004,hofmann2008}, are effective and efficient on datasets of limited size, and cannot take full advantage of currently available large-scale datasets\footnote{For large-scale datasets, we mean datasets with a nuber of examples $n\gtrsim 10^5$}.
Recent approximate algorithms with time-memory-accuracy trade-off mechanisms \citep{conf/nips/BottouB07} have proven to be flexible alternatives to exact kernel methods for learning nonlinear models over large datasets, representing a substantial step forward in the field (e. g. Random Features \citep{conf/nips/RahimiR07,rahimi2008uniform} and Nystr\"om methods \citep{gittens2013revisiting}).\\
%
%
The main objective of this thesis is to present several types of scalable randomized large-scale learning algorithms, accompanied by rigorous analyses of their generalization properties in the statistical learning theory (SLT) framework \citep{cucker02onthe,cucker2007learning}
 and computational requirements, supported by experimental evidence on benchmark datasets and numerical simulations.
We highlight the central role of the relationship between statistics and computations in our analyses, theoretically and empirically attesting that parameters affecting time and memory complexities can also control predictive accuracy.
This results in very flexible and scalable learning algorithms which can be tailored to the statistical properties of a specific learning task and to the available computational resources.
Moreover, we investigate the application of large-scale learning algorithms to robotics, in particular to lifelong robot learning \citep{thrun1995lifelong}.
This allows for the continuous updating of the predictive models within a robotic system in an open-ended time span, based on an ever-growing amount of collected training samples, possibly with bounded time and memory complexities.
We propose large-scale learning algorithms for lifelong learning tasks in visual object recognition for robotics and robot dynamics learning.

\section*{Large-scale Machine Learning}
In their classical formulations, kernel methods rapidly become computationally intractable as the training set size increases.
In the prominent case of Kernel Regularized Least Squares (KRLS) \citep{schlkopf2002learning}, the computational complexity for training and testing strongly depends on the number of samples $n$ composing the training set and the number of features $d$.
In particular, training costs $O(n^2d+n^3)$ in time and $O(n^2)$ in memory complexity, while testing costs $O(n)$ in time.
These computational complexities are overwhelming for current computer platforms if $n$ is too large.
The memory cost is the first bottleneck encountered in practical applications, since the kernel matrix cannot fit in the RAM memory and computationally intensive memory mapping techniques become necessary.
On the other hand, linear learning algorithms such as Regularized Least Squares (RLS) \citep{rifkin2002everything} require the computation and storage of a covariance matrix of size $d \times d$ for training.
This means that training a linear model does not suffer from a very large number of training samples.
In fact, its training computational time, $O(d^3+nd^2)$, is linear in $n$.
Notably, testing only requires $O(d)$, which does not depend on $n$.
However, in many learning problems the nonlinearity of the estimator is a fundamental property, and this is one of the main reasons supporting the choice of nonlinear learning algorithms like KRLS with respect to linear ones.
For instance, the aforementioned inverse dynamics learning problem for robotics is a regression problem implying strongly nonlinear relations between the input features (joint positions, velocities and accelerations) and the predicted outputs (forces and torques), see \citep{rasmussen2006,reference/robo/FeatherstoneO08,conf/humanoids/TraversaroPMNN13}.
It is therefore clear that the ideal objective would be to devise kernelized learning algorithms, with their nonlinear capabilities, while retaining, or at least not degrading too much, the advantageous computational properties of linear learning algorithms.
To this aim, many computational strategies to scale up kernel methods have been recently proposed \citep{conf/icml/SmolaS00,conf/nips/WilliamsS00,conf/nips/RahimiR07,conf/icml/YangSAM14,conf/icml/LeSS13,conf/icml/SiHD14,conf/colt/ZhangDW13}.
In this thesis, we will focus on three types of randomized large-scale learning algorithms, namely:
\begin{itemize}
\item Data-independent subsampling schemes, including random features approaches.
\item Data-dependent subsampling schemes, including Nystr\"om methods.
\item Stochastic gradient methods (SGM), in various formulations.
\end{itemize}
For these randomized approaches, we provide novel generalization analyses and experimental benchmarkings, as reported in \citep{rudi2015less,camoriano2016nytro,rudi2016genrf,lin2016}.
Now, we will briefly introduce these families of methods, which will be treated in greater detail in the corresponding chapters.
Let us begin with random features (RF) \citep{conf/nips/RahimiR07}, a subsampling scheme allowing to approximate kernel methods by means of features which are randomly generated in a data-independent fashion.
These features can then be fed to a linear algorithm such as RLS, globally resulting in an approximated nonlinear learning algorithm (RF-RLS).
The accuracy of RF-RLS can be made arbitrarily close to the one of KRLS by increasing the number of random features \citep{rudi2016genrf}.
The introduction of random features \citep{conf/nips/RahimiR07} has been an important step towards the application of classical kernel methods to large scale learning problems (see for example \citep{huang2014kernel, 2014arXiv1411.4000L}).
However, there are very few theoretical results about the statistical properties of RF-approximated kernel methods.
We provide a statistical analysis of the accuracy of KRLS approximated by Random Features.
In particular, we study the conditions under which it attains the same statistical properties of exact KRLS in terms of accuracy and analyze the implications on the computational complexity of the algorithm.
Notably, we show that optimal learning rates can be achieved with a number of features smaller than the number of examples.
As a byproduct, we also show that learning with random features can be seen as a way for controlling the statistical properties of the estimator, rather than only to speed up computations.\\
Another important class of large-scale algorithms are the Nystr\"{o}m methods \citep{conf/nips/WilliamsS00}, based on a low-rank approximation of the kernel matrix.
We present (as also reported in \citep{rudi2015less}) new optimal learning bounds for Nystr\"{o}m methods with generic subsampling schemes. 
Notably, we show that the amount of subsampling, besides computations, also affects accuracy. 
Leveraging on this, we propose a new efficient incremental version of Nystr\"{o}m KRLS with integrated model selection. 
We provide extensive experimental results supporting our findings.\\
Subsequently, we investigate the interplay between iterative regularization/early stopping learning algorithms \citep{engl1996regularization,zhang2005boosting, bauer2007regularization,earlyStopping,CapYao06} and Nystr\"om subsampling schemes.
Iterative regularized learning algorithms, such as Landweber iteration or the $\nu$-method \citep{de2006spectral},
are a family of methods which compute a sequence of solutions to the learning problem with increasingly high precision.
They provide a valid alternative to Tikhonov regularization methods in terms of time complexity, especially in regimes in which the amount of noise in the data is large.
Iterative methods are particularly efficient when used in combination with early stopping \citep{earlyStopping}.
 Still, in the dual formulation they require overwhelming amounts of memory for kernel matrix storage as the sample complexity increases. 
Following the idea of Nystr\"{o}m kernel approximation, widely applied to KRLS algorithm with Tikhonov regularization \citep{conf/nips/WilliamsS00,rudi2015less}, we propose (as reported in \citep{camoriano2016nytro}) to combine Nystr\"{o}m subsampling techniques with early stopping in the kernelized iterative regularized learning context, in order to reduce memory and time requirements.
The resulting learning algorithm, named NYTRO (NYstr\"om iTerative RegularizatiOn), is presented, together with the analysis of its generalization properties and extensive experiments.
\\
Finally, we consider the stochastic gradient method (SGM), often called stochastic gradient descent.
SGM is a widely used algorithm in machine learning, especially in large-scale settings, due to its simple formulation and reduced computational complexity
 \citep{bousquet2008tradeoffs}.
Despite being commonly used for solving practical learning problems, SGM's generalization properties 
have not been exhaustively investigated before.
We study (as we also reported in \citep{lin2016}) the generalization properties of SGM for learning with convex loss functions and linearly parameterized functions.
We show that, in the absence of penalizations or constraints, the stability and approximation properties of the algorithm can be controlled by tuning either the step-size or the number of passes over the training set. 
In this view, these
parameters can be seen to control a form of implicit regularization. 
Numerical results complement our theoretical findings.

\section*{Lifelong Learning for Robotics}

The iCub humanoid robot \citep{Metta2010} is a suitable match for lifelong robot learning applications, due to its large variety and number of sensors and degrees of freedom (DOFs), the support for extended  data generation processes with rich dynamics, an advanced visual system, and a number of open problems in perception and interaction in which learning can play a fundamental role.
One of the goals of this thesis is to present the application of large-scale learning methods to lifelong visual object recognition in robotics.
In particular, 
we propose (see also \citep{camoriano2016incrementalclass,camoriano2017incrementalclass}) a novel object recognition algorithm based on Regularized Least Squares for Classification (RLSC) \citep{rifkin2003regularized}, in which examples are presented incrementally in a real-world robot supervised learning scenario. 
The model learns incrementally to recognize classes of objects, taking into account the imbalancedness of the the dataset at each step and adapting its statistical properties to changing sample complexity.
New classes can be added on-the-fly to the model, with no need for retraining.
Incoming labeled pictures are used for increasing the accuracy of the predictions and adapting them to changes in environmental conditions (e. g., lighting conditions).\\
Last, but not least, we present another robotics application of interest, that is inverse dynamics learning \citep{conf/humanoids/TraversaroPMNN13,Nguyen-TuongP10,GijsbertsM11,romeres2016online}.
We propose novel approaches for inverse dynamics learning applied to robotics, in particular to the iCub humanoid robot (for the scope of this work see \citep{camoriano2016incremental}, and, for an alternative method, \citep{romeres2016online}).
Classical inverse dynamics estimators rely on rigid body dynamics (RBD) models of the kinematic chain of interest (e.g., a limb of the iCub).
This modeling strategy enables good generalization performance in the robot workspace.
Yet, it poses two issues: 1) The estimated inertial parameters of the RBD model might be inaccurate, especially after days of operation of the robot. The reasons include changes in the physical properties of the components, sensor drift, wear and tear, and thermal phenomena.
2) RBD models exclude the effects of limb and joint flexibility, and possibly others, on the inverse dynamics mapping.
These dynamic effects can be substantial in real-world scenarios.
On the other hand, machine learning approaches are capable of learning the input-output mapping directly from data, without the restrictions imposed by prior assumptions about the rigid physical structure of the system.
We combine standard physics-based modeling techniques \citep{reference/robo/FeatherstoneO08,conf/humanoids/TraversaroPMNN13} with nonparametric modeling based on large-scale kernel methods based on random features \citep{GijsbertsM11}, in order to improve the model's predictive accuracy and interpretability.
Unlike \citep{Nguyen-TuongP10}, the proposed approach learns incrementally during robot operation with fixed update complexity, making it suitable for real-time applications and lifelong robot learning tasks. 
The model can be adjusted in time via incremental updates, increasing the accuracy of its predictions and adapting to changes in physical conditions.
The system has been implemented using the GURLS \citep{tacchetti2013gurls} machine learning software library.
\section*{Organization}
The thesis is organized as follows.
Part \ref{part:materials} introduces essential machine learning concepts employed throughout the work.
In particular, Chapter~\ref{Chap:SLT} provides an introduction to Statistical Learning Theory (SLT), Chapter~\ref{Chap:kernelMethods} presents kernel methods, while Chapter~\ref{Chap:spectralReg} describes the fundamentals of spectral regularization.
Subsequently, Part~\ref{part:largeScale} focuses on scaling up kernel methods for large-scale learning applications, providing novel generalization analyses.
Data-dependent subsampling schemes are treated in Chapter~\ref{chap:lessismore}, with focuses on optimal learning bounds for Nystr\"om KRLS in Section~\ref{sec:lessismore}, and on the combination of early stopping and Nystr\"om subsampling (the proposed NYTRO algorithm) in Section~\ref{sec:nytro}.
Furthermore, data-independent subsampling schemes, in particular the optimal learning bounds for random features KRLS (RF-KRLS), are discussed in Chapter~\ref{chap:randfeats}.
Part~\ref{part:lifelong} is concerned with iterative learning algorithms and lifelong learning, including specific applications to robotics.
In particular, Chapter~\ref{chap:sgd} presents the new statistical analysis of SGM, Chapter~\ref{chap:incclass} describes our novel incremental multiclass classification algorithm with extension to new classes in constant time, and, finally, Chapter~\ref{chap:invdyn} deals with our recently proposed incremental semiparametric inverse dynamics learning method with constant update complexity.
The thesis is concluded by final remarks in Chapter~\ref{chap:concl}.
\clearpage
\pagebreak

\part{Mathematical Setting}
\label{part:materials}

	\chapter{Statistical Learning Theory}
	\label{Chap:SLT}
		The field of machine learning studies algorithms and systems capable of extrapolating information and learning predictive models from data, rather than being specifically programmed to solve a given task.
These techniques are particularly useful for solving a wide range of tasks for which it is hard or unfeasible to define a comprehensive set of decision rules.
Examples of use are widespread, and include visual object recognition, speech recognition, control policy learning, fraud detection, dynamic pricing, and product ranking.
Machine learning
tackles this kind of problems by learning from examples.
It can be subdivided in the following sub-fields:
\begin{itemize}
\item Supervised learning: The algorithm learns the mapping between inputs and desired outputs, based on a set of input-output pairs associated with the task, provided by a supervising entity. A performance measure (e.g. a loss function) is required.
\item Unsupervised learning: Data are made available to the learning algorithm, whose goal is to discover hidden structure in it. The provided examples do not include output labels. Main techniques falling in this definition include, among others: Clustering, feature learning, and dimensionality reduction techniques such as principal component analysis (PCA).
\item Reinforcement learning: The learning system has access to a dynamic environment, in which it has to perform a goal for which no supervision is provided. The actions applied to the environment result in a reward signal to be maximized in order to achieve the goal. Application examples include autonomous driving and general game playing.
\end{itemize}
In this work, the focus will be on supervised learning, in particular in the large-scale context.
In fact, when a significantly large number of data samples\footnote{We refer to input-output pairs as data points, samples or examples.} is available, learning algorithms can become too computationally expensive for execution.
This represents a relevant technological issue, hampering the way towards the applicability of machine learning methods.
For this reason, it is necessary to design novel large-scale learning algorithms capable of taking full advantage of the remarkable volume of data available today.
This unfolds many opportunities of scientific research, especially regarding the study of  the interplay between the statistical and computational properties~\citep{bottou2007large} of these methods, and their generalization properties.
Throughout this work, we will analyze the statistical properties of supervised learning algorithms in the Statistical Learning Theory (SLT) \citep{cucker02onthe} framework, introduced next.

\section{Supervised Learning}

\begin{figure}[t]
\begin{center}
\includegraphics[width=3in]{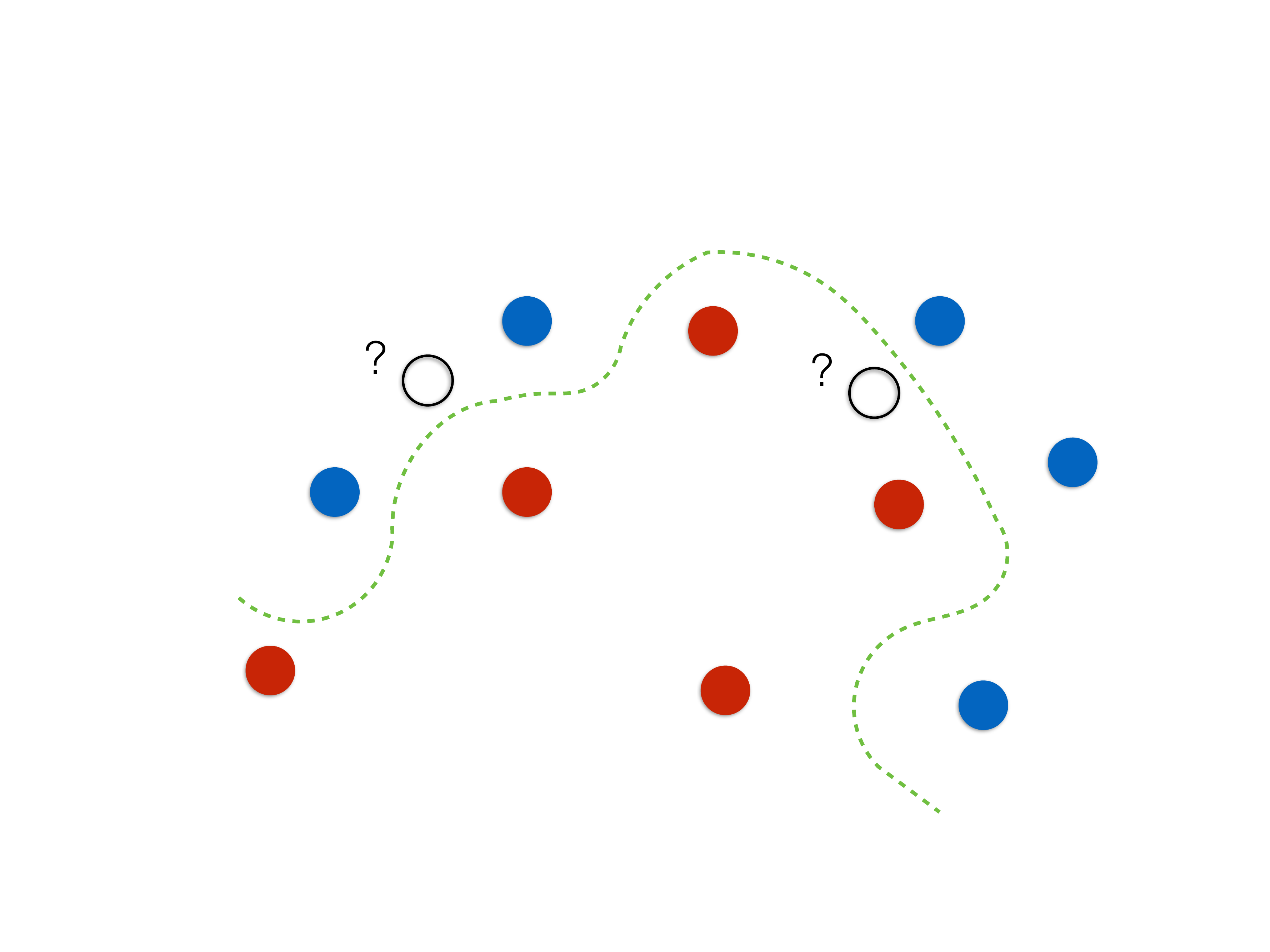}
\end{center}
\caption{In binary classification, given examples (blue and red points) the problem is to learn a classifier (represented by the dashed line) correctly predicting the label of new input samples (white points).}
\label{Fig:class_boundary}
\end{figure}

Supervised learning assumes to have access to a set of $n$ input-output pairs which are instances of the considered learning task.
This finite set of examples is called {\em training set}:
$$ 
S_n= \{(x_1,y_1),\dots,(x_n,y_n)\}.
$$
The objective of a supervised learning algorithm is {\em learning from examples}, that is by computing an estimator $f$, based on the training set, mapping \textit{previously unseen} inputs $x_{\text{new}}$ into the corresponding outputs $y_{\text{new}}$ (see Figure \ref{Fig:class_boundary}), as follows:
$$
f: x_{\text{new}} \mapsto  y_{\text{new}}.
$$

\begin{figure}[t]
\begin{center}
\includegraphics[width=1.8in]{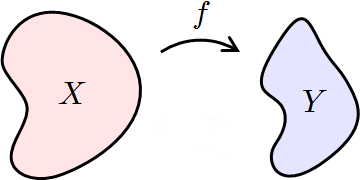}
\end{center}
\caption{Input and outputs space related by a functional relationship.}
\label{Fig:functionalRelationship}
\end{figure}

The data are usually subject to {\em uncertainty} (e.g., due to noise).
Statistical learning models this by assuming the existence of an underlying probabilistic model for the data, discussed in the following.\\
We will now more formally introduce the two fundamental concepts defining a learning task.
First, examples are drawn from the \textit{data space}, a probability space $\mathcal X \times \mathcal \Y$ with measure $\rho$.
$\mathcal \X$ is called \textit{input space}, while $\mathcal \Y$ is called \textit{output space} (see Figure \ref{Fig:functionalRelationship}).
Secondly, a measurable {\em loss function}, 
$$\ell: \mathcal Y \times \mathcal Y \to [0, \infty),$$
is defined to evaluate the quality of the predictions of the estimator.

On the basis of these key concepts, we can define the \textit{expected error} (more details are reported in Section~\ref{sec:expectedErr})
$${\cal E}(f) = \int_{\mathcal  X \times \mathcal Y} \ell(y, f(x)) d\rho(x,y).$$
The problem of learning is to find an estimator $f$ minimizing the expected error, that is solving
$$
\inf_{ f \in \mathcal F} {\cal E}(f),
$$ 
with $\rho$ fixed and unknown.
Here, $\mathcal F$ is the set of functions $f: \mathcal X \to \mathcal Y$ such that the expected risk ${\cal E}(f)$ is well defined.
Access to $\rho$ is limited to a finite {\em training set} 
composed of samples identically and independently distributed (i. i. d.) according to $\rho$
$$S_n =\{ (x_1, y_1), \dots, (x_n, y_n)\} \sim \rho^n.$$
Since $\rho$ is known only via the training set $S_n$, finding an exact solution to the expected risk minimization is generally not possible.
One way to overcome this problem is to define an \textit{empirical error} measured on the available data to treat the learning problem in a computationally feasible way.
We first discuss in greater detail the concepts introduced above.

\section{Data Space}

The input/output examples belong to the {\em data space} $\mathcal{Z} = \mathcal{X} \times \mathcal{Y}$.
The \textit{input space} $\mathcal{X}$ can take several forms, depending on the specific learning problem. We report some of the most common cases:
\begin{itemize}
\item Linear spaces
	\begin{itemize}
	\item Euclidean spaces: $\mathcal{X}\subseteq \R^d, \quad d \in \N$.
	\item Space of matrices: $\mathcal{X}\subseteq \R^{a \times b}, \quad a,b \in \N$.
	\end{itemize}
\item Structured spaces
	\begin{itemize}
	\item  Probability distributions: Given a finite set $\Omega$ of dimension $d$, we can consider as input space the space of all possible probability distributions over $\Omega$, which is $\mathcal X = \{x \in  \R^d_+ ~:~ \sum_{j=1}^d x^j=1\}$, $d\in \N$. This also holds more in general for any set  $\Omega$, even if not finite.
	\item Strings/Words: Consider an alphabet $\Sigma$ of symbols. The input space can be the space of all possible words composed of $p\in \N$, as follows: $\mathcal X=\Sigma^p$.
	\item $\mathcal X$ is a space of graphs.
	\end{itemize}
\end{itemize}

Similarly, the definition of output space $\mathcal{Y}$ results in several different categories of learning problems, some of the most common of which are:
\begin{itemize}
\item Linear spaces
	\begin{itemize}
	\item Regression: $\mathcal{Y}\subseteq \mathbb R$.
	\item Multi-output (multivariate) regression: $\mathcal{Y}\subseteq \mathbb R^T, T>1$.
	\item Functional regression: Outputs are functions, $\mathcal{Y}$ is a Hilbert space.
	\end{itemize}
\item Structured spaces
	\begin{itemize}
	\item $\mathcal{Y}$ is a space of probability distributions.
	\item $\mathcal{Y}$ is a space of strings.
	\item $\mathcal{Y}$ is a space graphs.
	\end{itemize}
\item Other spaces
	\begin{itemize}
	\item Binary classification: $\mathcal{Y}=\{-1,1\}$, or any pair of different numbers.
	\item Multi-category (multiclass) classification: Each example belongs to one among $T$ categories, $\mathcal{Y}=\{1,2, \dots, T\}, T \in \N$.
	\item Multilabel classification: Each example is associated to any subset of $T$ output categories, $\mathcal{Y}=2^{\{1,2, \dots, T\}}, T \in \N$.
	\end{itemize}
\end{itemize}
\begin{rem}[Multi-task Learning]
One of the most general supervised learning settings is called Multi-task Learning, for which $\mathcal Z = (\mathcal X_1,\mathcal Y_1) \times(\mathcal X_2,\mathcal Y_2) \times ... \times (\mathcal X_T,\mathcal Y_T) $ and the training set $S$ is composed of one training set $S_i$ for each task $i$ of the $T$ tasks: $S = {S_1, S_2, ..., S_T}, T\in \N$.
\end{rem}

\section{Probabilistic Data Model}
\label{sec:probDataModel}
\begin{figure}[t]
\begin{center}
\includegraphics[width=2.8in]{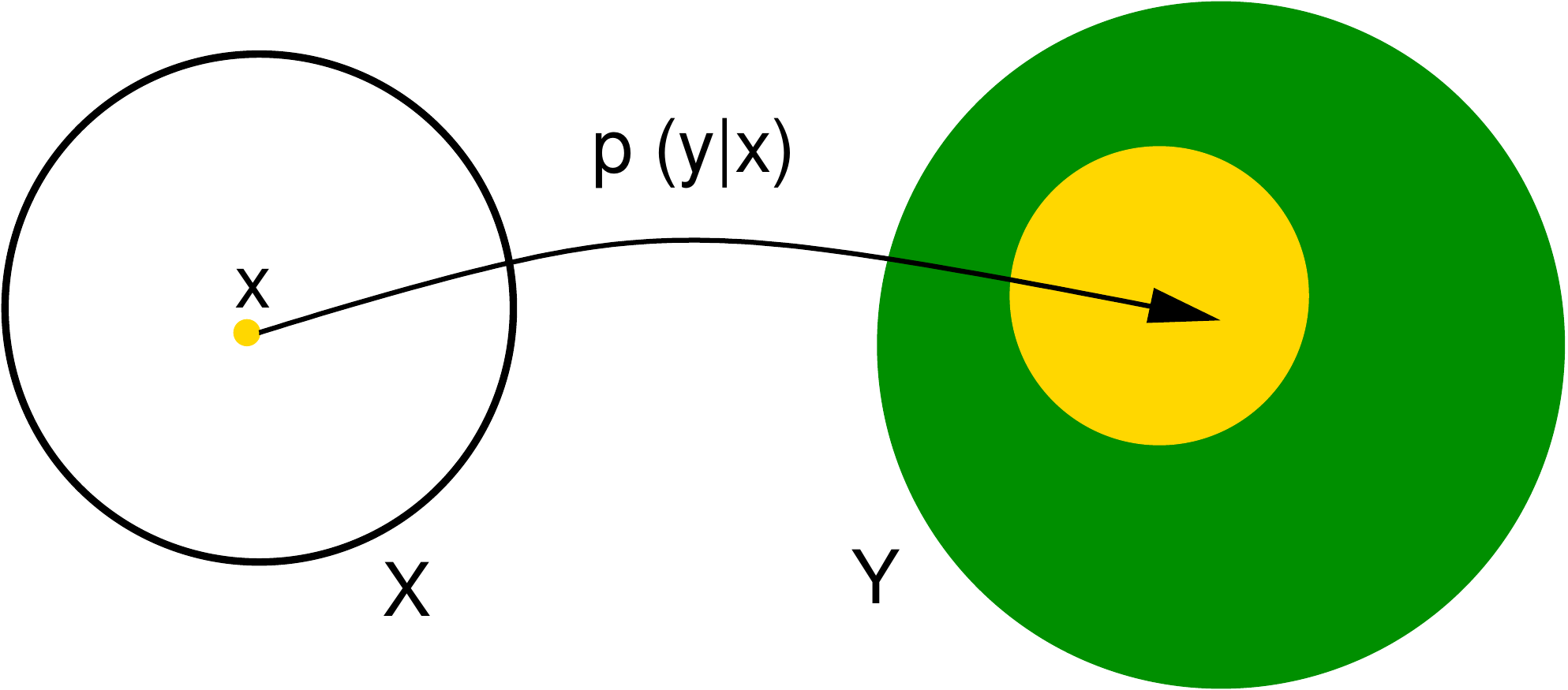}
\end{center}
\caption{Given an input $x$,  the yellow area is the support of its associated distribution of possible outputs $\rho(y|x)$. The green area represents the support of the distribution of all possible outputs $\rho(y)$.}
\label{Fig:Conditional}
\end{figure}

\begin{figure}[t]
\begin{center}
\includegraphics[width=2.8in]{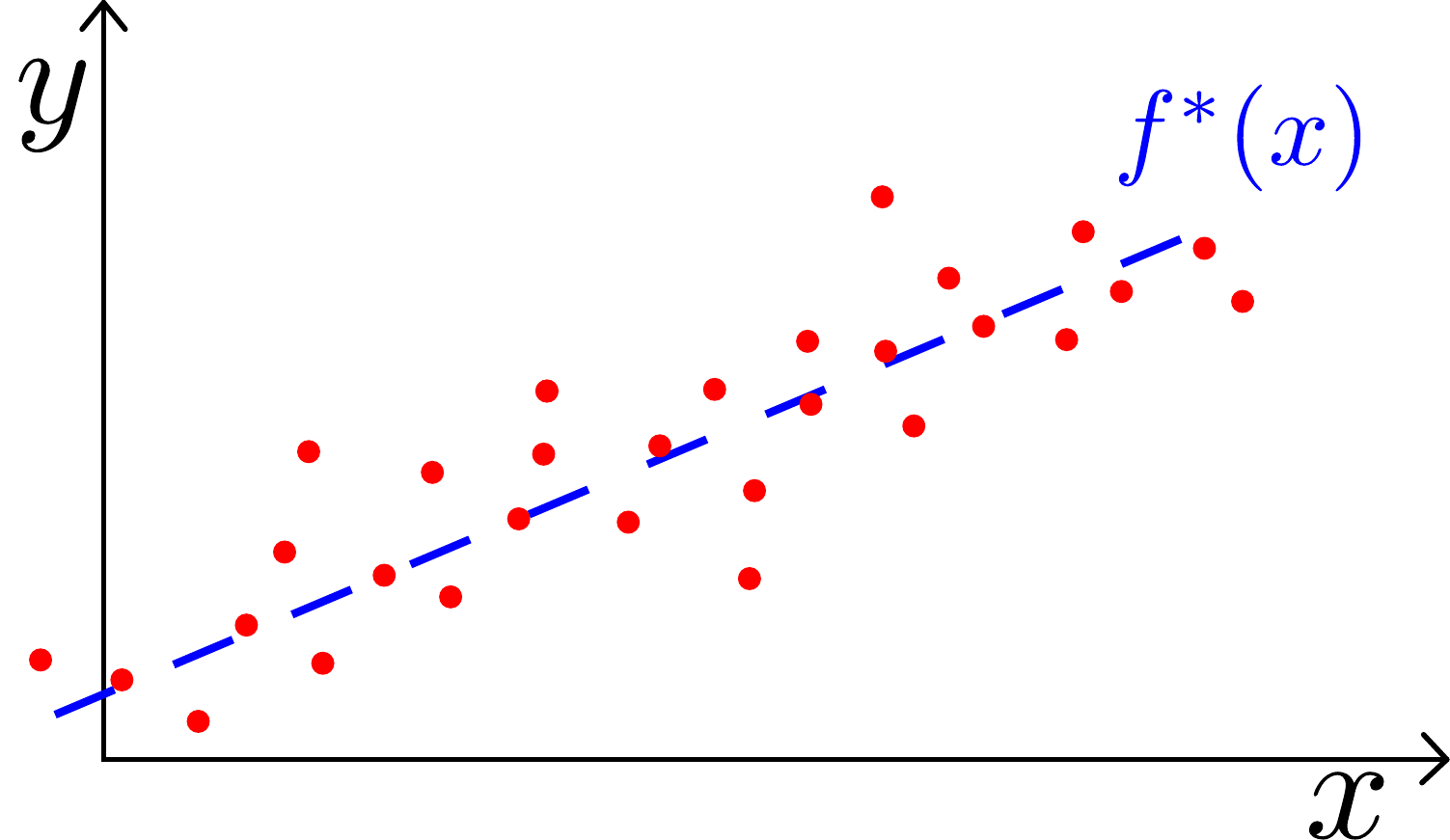}
\end{center}
\caption{Fixed unknown linear function $f^*$ and noisy examples sampled from the $y=f^*(x)+\epsilon$ model.}
\label{Fig:Regression}
\end{figure}

In supervised learning, in order to estimate an input/output relation, we assume that a model expressing this relationship exists. In SLT, the assumed model is probabilistic, in the sense that the data samples constituting the training and the test sets are assumed to be sampled independently from the same fixed and unknown data distribution $\rho(x,y)$\footnote{In this case, we say that data samples are independent and identically distributed (i. i. d. assumption).}
 on the data space $\mathcal{Z}$.
Thus, $\rho$ encodes the uncertainty in the data, for instance caused by noise, partial information or quantization.
By also assuming that $\rho(x,y)$ can be factorized as
$$\rho(x,y)=\rho_{\mathcal{X}}(x)\rho(y|x) \quad \forall (x,y) \in \mathcal{Z},$$
the various sources of uncertainty can be separated.
In particular, the marginal distribution $\rho_{\mathcal{X}}(x)$ models the uncertainty in the sampling of the input points.
Instead, $\rho(y|x)$ is the conditional distribution modeling the {\em non deterministic} input-output mapping, as shown in Figure \ref{Fig:Conditional}.
We now consider two instances of data model, associated with regression and classification.
\begin{ex}[Fixed and Random Design Regression]
In statistics, the commonly assumed data model for regression is the following,
$$y_i=f^*(x_i)+\epsilon_i,$$
where $\mathcal{X} = \{x_1,  x_2, ..., x_n\}$ is a deterministic discrete set of inputs, $f^*$ is a fixed unknown function and $\epsilon$ is random noise.
For example, $f^*$ could be a linear function $f^*(x)=x^\top w^*$ with $w^*\in \R^d$, and the $\epsilon$ component could be Gaussian noise, distributed according to ${\mathcal N}(0,\sigma )$, $\sigma\in [0,\infty)$.
The aforementioned model is named \textit{fixed design regression}, since $\mathcal{X}$ is fixed and deterministic.
By contrast, in SLT the so-called \textit{random design} setting is often considered, according to which the training samples are not given a-priori, but according to a probability distribution $\rho_{\mathcal X}$ (for example, uniformly at random).
See Figure \ref{Fig:Regression} for an example.
\end{ex}

\begin{figure}[h!]
\begin{center}
\includegraphics[width=2.8in]{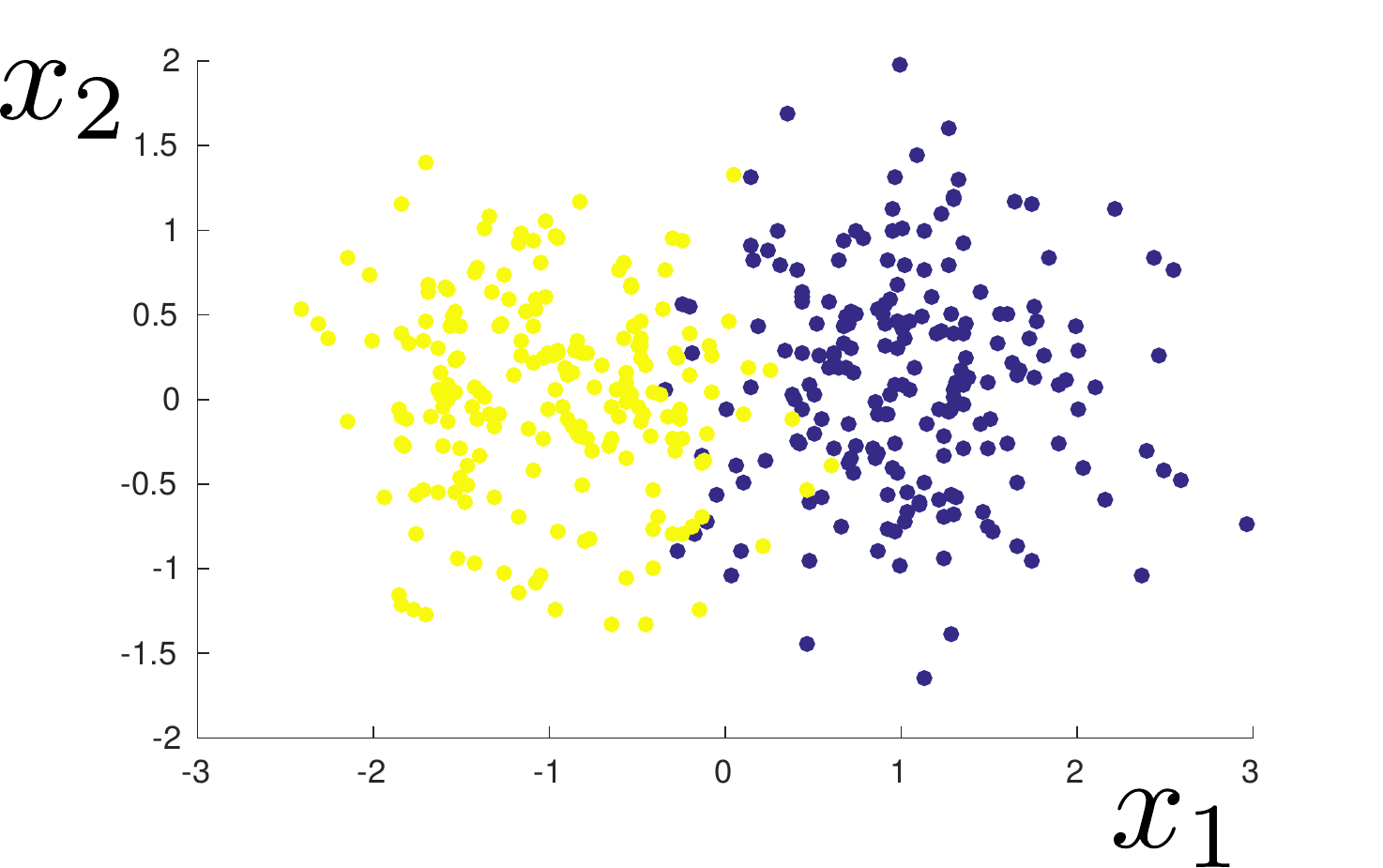}
\end{center}
\caption{2-D example of a dataset sampled from a mixed Gaussian distribution. Samples of the yellow class are realizations of a Gaussian centered at $(-1,0)$, while samples of the blue class are realizations of a Gaussian centered at $(+1,0)$. Both Gaussians have standard deviation $\sigma = 0.6$.}
\label{Fig:Classification}
\end{figure}

\begin{ex}[Binary classification]
In binary classification, input-output samples are sampled randomly according to a distribution $\rho$ over $\X \times \{-1, 1\}$.
In a simple example of binary classification problem, $\rho$ is a mixture of two Gaussians, each corresponding to a class, $\rho(x,y) =  \rho(x|y=-1) + \rho(x|y=-1)$, with
$$\rho(x|y=-1)=\frac 1 c {\mathcal N}(-1,\sigma_- ), \, \sigma_-\in [0,\infty)$$ 
$$\rho(x|y=1)=\frac 1 c {\mathcal N}(+1,\sigma_+), \, \sigma_+\in [0,\infty)$$
where $c \in \R$ is a suitable normalization factor such that $\rho(x,y)$ is a probability distribution, that is s. t. $\int \rho(x,y) dx dy = 1$.
An example dataset drawn according to the aforementioned $\rho(x,y)$ is shown in Figure \ref{Fig:Classification}
\end{ex}

\section{Loss Function}

\begin{figure}
\centering
\includegraphics[width=0.8\linewidth]{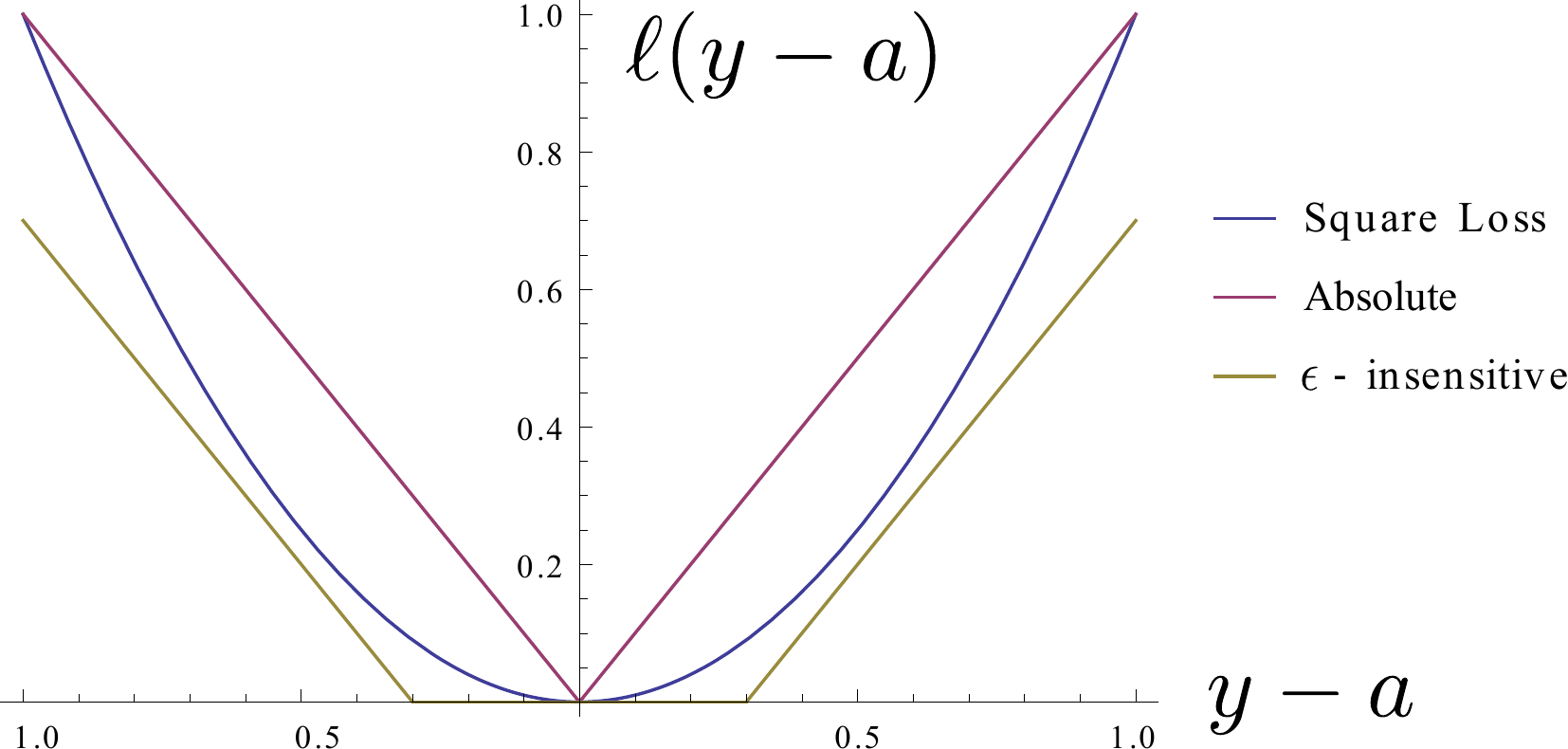}
\caption{Loss functions for regression.}
\label{fig:regressionloss}
\end{figure}

We have seen that the problem of learning an estimator amounts to finding a function $f$ ``best" approximating the underlying input-output relationship.
To do so, we need a measure of the predictive accuracy of the learned estimator with reference to the specific learning task in exam.
In particular, the most natural way to do so is to define a point-wise {\em loss function} 
$$
\ell:\mathcal{Y}\times \mathcal{Y}\to [0,\infty),
$$
measuring the loss $\ell(y, f(x))$ the learning system undergoes by predicting $f(x)$ instead of the actual output $y$.
For instance, typical losses for regression problems differ from the ones used for classification.
We will now recall some of the most usual ones.\\
Loss functions for regression
usually depend on the deviation $y-a$ between the real output $y$ and the predicted value $a = f(x)$.
\begin{itemize}
	\item Square loss: It is the most commonly employed loss function for regression, defined as $\ell(y,a) = (y-a)^2$, with $a,b \in \mathcal Y$.
	\item Absolute loss: $\ell(y,a) = | y-a |$
	\item $\epsilon$-insensitive loss: $\ell(y,a) = max\{| y-a |- \epsilon,0\}$, with $\epsilon \in (0,+\infty)$. It is used in Support Vector Machines (SVMs) for regression \citep{smola1997support}.
\end{itemize}
Loss functions for regression are shown in Figure \ref{fig:regressionloss}.

\begin{figure}
\centering
\includegraphics[width=0.8\linewidth]{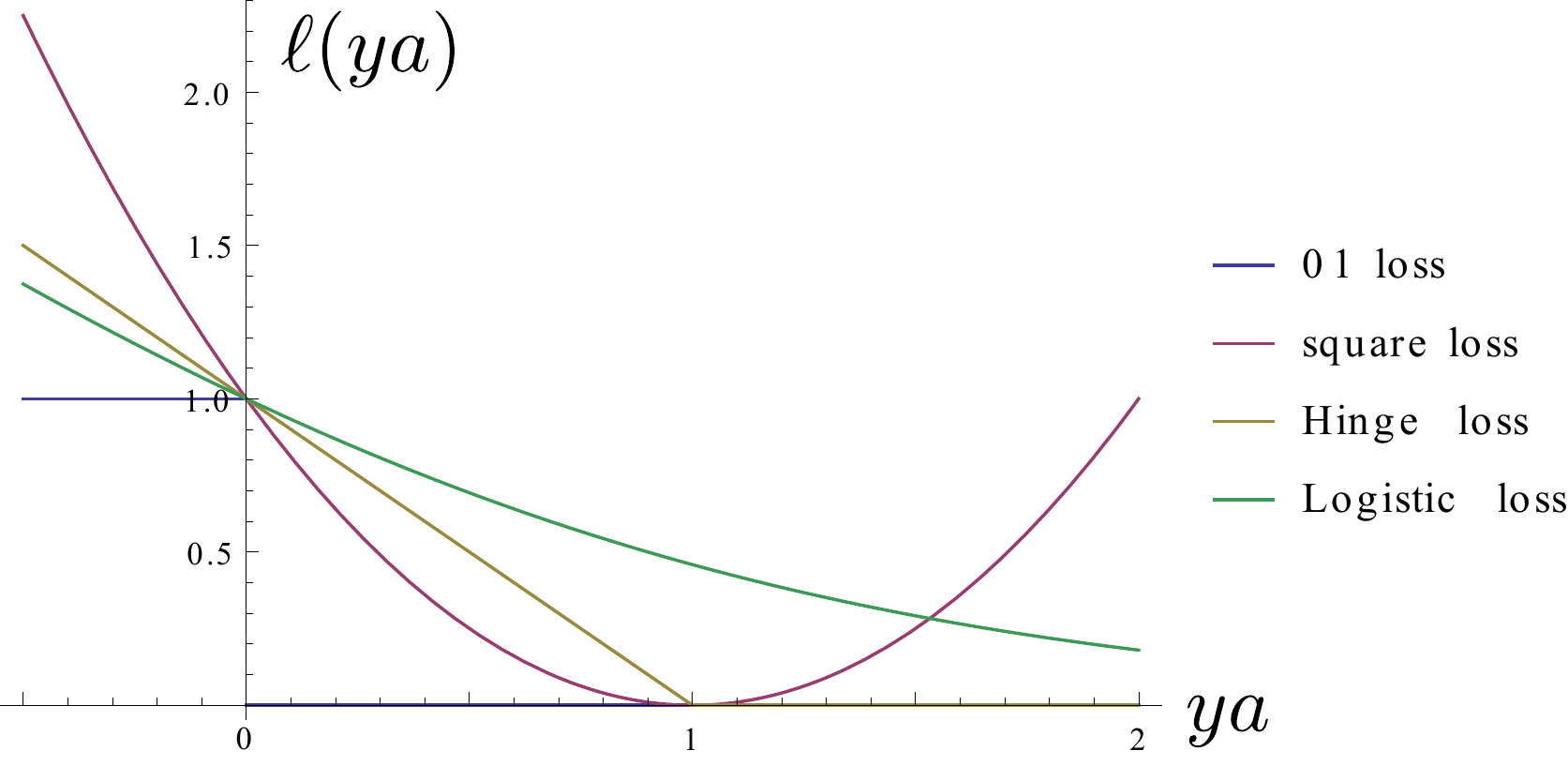}
\caption{Margin-based loss functions for classification.}
\label{fig:classloss}
\end{figure}

\noindent Loss functions for classification: For the sake of simplicity, we consider the setting in which $\mathcal Y = \{+1,-1\}$.
\begin{itemize}
\item Misclassification loss (0-1 loss, counting loss): It is probably the most natural choice for measuring the accuracy of a classifier.
It assigns a cost $1$ if the predicted label is incorrect, and $0$ otherwise. 
It can be defined as $\ell(y,a) = \ind_{y \neq a}$, where $\ind_{(\cdot)}$ is the indicator function, $y$ is the real output associated to the input $x$, and $a = \text{sign}(f(x))$ is the output label predicted by the learned estimator $f$.
\item Surrogate losses: The 0-1 loss is non-convex and makes optimization very hard.
To overcome this issue, convex \textit{surrogate loss functions} acting as convex relaxations of the 0-1 loss are used.
To introduce them, we consider the real-valued prediction $a = f(x)$, without taking the sign.
Given a pair $(x,y)$ and an estimator $f$, we define the quantity $ya$, called \textit{margin}.
Surrogate losses are often defined in terms of margin: $\ell(y,a) = \ell(ya)$.
We report the most commonly used ones below:
	\begin{itemize}
	\item Hinge loss: Used in Support Vector Machines (SVMs) for classification, it is defined as $\ell(y,a) = |1-ya|_+ = max\{ 1-ya, 0 \}$.
	\item Exponential loss: $\ell(y,a) = e^{-ya}$, used in Boosting algorithms.
	\item Logistic loss: $\ell(y,a) = log(1+e^{-ya})$, used in logistic regression.
	\item Square loss: It can be used also for classification, and it can be written in terms of margin as $\ell(y,a) = (y-a)^2 = (1-ya)^2$.
	\end{itemize}
\end{itemize}
See Figure \ref{fig:classloss} for a pictorial representation of the aforementioned loss functions for classification.

\section{Expected Error}
\label{sec:expectedErr}
The loss function $\ell$, alone, is not enough for quantifying how well an estimator performs on any possible future data sample drawn from $\rho$.
To this end, we define the \textit{expected error} (also {\em expected risk} or \textit{expected loss}) of an estimator $f$ given a loss $\ell$ as
\begin{equation}\label{eq:expectedError}
{\mathcal E}(f)=\int_{\mathcal  X \times \mathcal Y} \ell(y, f(x)) d\rho(x,y),
\end{equation}
which expresses the average loss the estimator is expected to incur in on samples drawn from $\rho$.
Hence, in the SLT framework the ``best"  possible estimator is the {\em target function} $f^*:\mathcal X\to \mathcal Y$ minimizing ${\mathcal E}(f)$,
$$
\inf_{f \in \mathcal F}  {\mathcal E}(f),
$$
where $\mathcal F$ is named \textit{target space} and is the space of functions for which ${\mathcal E}(f)$ is well-defined.
Notice that different loss functions yield different target functions, as reported in Appendix~\ref{app:lossTarget}.
It is essential to note that this minimization cannot be computed in general, because the data distribution $\rho$ is not available.
The objective of a learning algorithm is to find an estimator $f$ as close as possible to $f^*$ and behaving well on unseen data, despite having access to just a finite realization $S_n \sim \rho^n$, the training set.
An algorithm fulfilling this criterion is said to \textit{generalize well}, which is the cornerstone concept of SLT and machine learning in general.

\section{Generalization Error Bound}
A learning algorithm $\mathcal{A}$ can be thought of as a mapping $\mathcal{A}: S_n \mapsto \fn$ from the the training set $S_n$ to the associated estimator $\fn$.
To introduce the concepts of generalization and consistency, we assume that $\mathcal{A}$ is deterministic, even if this assumption can be relaxed, as we will see in the following chapters.
Here, the only source of uncertainty is therefore due to stochasticity in the data, modeled by $\rho$.

We have seen that the target function associated to a learning problem is the one minimizing the expected risk.
A good learning algorithm is able to find an estimator $\fn$ yielding an expected error as close as possible to the one of the target function.
However, to perform a more precise analysis of a learning algorithm we need to quantify this behavior, considering that the learned estimator $\fn$ depends on the training set $S_n$.
A widely studied property of learning algorithms is the \textit{generalization error bound}, which describes a bound on the error with probability $1-\delta$.
More formally, given a distribution $\rho$,
$\forall \delta \in[0,1], n >0$, there exists a function $\epsilon(\delta,n)$, called \textit{learning rate} (or \textit{learning error}), such that,
\begin{equation}\label{eq:learningRate}
\P\left( \EE(\fn)-\inf_{f\in\mathcal F} \EE(f)\le \epsilon(\delta,n) \rp \ge 1- \delta.
\end{equation}

\section{Overfitting and Regularization} 
 
Ideally, an estimator $\fn$ should  {\em mimic}  the target function $f^*$, in the sense that its expected error should get close to the one of $f^*$ as $n \rightarrow \infty$.
The latter requirement needs some care, since  $\fn$ depends on the training set $S_n$ and hence is random.
As we have seen in \eqref{eq:learningRate}, one possibility is to require an algorithm to have a good learning rate.
Given a rich enough hypotheses space $\hh$, 
a good learning algorithm should be able to describe well (fit) the data
and at the same time 
disregard noise. 
Indeed, a key to ensure good generalization is to avoid \textit{overfitting}, that characterizes estimators which are highly dependent on the training data.
This can happen if $\hh$ is too rich, thus making the model ``learn the noise'' in the training data.
In this case, the estimator is said to have high \textit{variance}.
On the other hand, if $\hh$ is too simple the estimator is less dependent on the training data and thus more robust to noise, but $\hh$ may not be rich enough to approximate well the target function $f^*$.
\textit{Regularization} is  a general class of techniques that allow to restore stability and ensure generalization.
It considers a sequence of hypotheses spaces $\hh_\la$, parameterized by a \textit{regularization parameter} $\la$, with
$$
\lim_{\la \rightarrow 0} \hh_{\la} = \F.
$$

At this point, a natural question is whether an optimal regularization parameter in terms of learning algorithm performance exists, and, if so, how it can be found in practice.
We next characterize the corresponding minimization problem to uncover one of the most 
fundamental aspects of machine learning.
As we saw in the previous sections, the generalization  performance of a learning algorithm 
is as much high as
\begin{equation}\label{Perf}
\EE(\fn _\lambda) -  \inf_{ \F} \EE(f)
\end{equation}
is low.
To get an insight on how to choose $\lambda$, we theoretically analyze  how this choice influences performance. 
For a fixed $\la$, we can decompose \eqref{Perf} as
\begin{equation}\label{BiasVarianceDecomp}
 \EE(\fn _\lambda)- \inf_{ \F} \EE(f)
 =  
\underbrace{ \EE(\fn _\lambda)- \inf_{ \hh_\la} \EE(f) }_{Variance}
 + 
\underbrace{  \inf_{ \hh_\la} \EE(f)- \inf_{ \F} \EE(f) }_{Bias}.
\end{equation}
Indeed, on the one hand we introduce a {\em variance} term to control the complexity of the hypotheses space.
On the other hand, we allow the class complexity to grow to reduce the \textit{bias} term.
The optimal $\lambda^*$ is the one minimizing \eqref{BiasVarianceDecomp}, the so-called bias variance trade off.\\
Specifically, by bias we mean the deviation
$$ \inf_{ \hh_\la} \EE(f)- \inf_{ \F} \EE(f).$$
This is often called the {\em approximation error} and does not depend on the data, but only on how well the class $\hh_\la$ ``approximates'' $\F$.
There are many instances of the above setting in which it is possible to design
regularization schemes so that the approximation error \textit{decreases for decreasing} $\la$.
In fact, in such schemes a small $\la$ corresponds to a reduced penalization of rich function classes.\\
In contrast, the term 
$$\EE(\fn_\lambda)- \inf_{ \hh_\la} \EE(f)$$ 
is called {\em sample}, or {\em estimation error}, or simply the \textit{variance} term.
It is data dependent and stochastic, and measures the variability of the output of the algorithm, for a given complexity, with respect to an ideal algorithm having access to all the data.
In most regularized learning algorithms a small $\la$ corresponds to a complex model which is more sensitive to data stochasticity, thus leading to a larger variance term.\\

\section{Bias Variance Trade-off and Cross-validation}

\begin{figure}[t!]
\begin{center}
\includegraphics[width=2.7in]{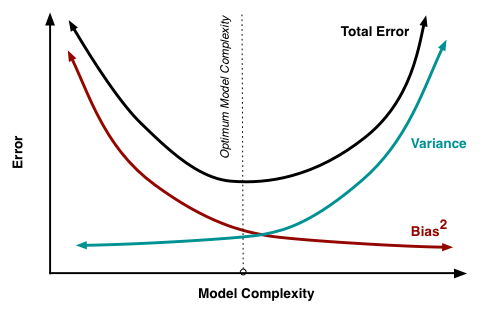}
\end{center}
\caption{The Bias-variance trade-off. In Tikhonov regularization the parameter $\la$ controls the model's complexity. A small $\la$ yields high complexity, and vice-versa.}\label{Fig:BiasVariance}
\end{figure}

We briefly discuss some practical considerations regarding the regularization parameter choice  for regularized learning algorithms.
As previously introduced, variance decreases with $\la$, while bias increases with it.
A larger bias is preferred when training samples are few and/or noisy to achieve a better control of the variance, while it can be decreased for larger datasets.
For any given training set, the best choice of $\lambda$ would be the one guaranteeing the optimal trade-off between bias and variance (that is the value minimizing their sum), as shown in Figure \ref{Fig:BiasVariance}.
However, the theoretical analysis is not directly useful in practice since the data distribution,
 hence the expected loss, is not accessible.
 In practice, data driven procedures are used to find a proxy for the expected loss, the simplest of which is called \textit{hold-out cross-validation}. 
 Part of the training set is held-out and used to compute a (hold-out) error acting as a proxy of the expected error.
  An empirical bias variance trade-off is achieved choosing the value of $\la$ with the minimum hold-out error.
  When data are scarce, the hold-out procedure, based on a simple "two ways split" of the training set, might be unstable.
  In this case, so called $V$-fold cross validation 
 is preferred, which is based on multiple data splitting. More precisely, the data are divided in $V \in [2,...,n] $ 
 (non overlapping) sets.
 Each set is held-out and used to compute a hold-out error, which is eventually averaged to obtain the final $V$-fold  cross-validation error.
 The extreme case where $V=n$ is called \textit{leave-one-out} cross-validation.\\

\section{Empirical Risk Minimization \& Hypotheses Space} 
\label{sec:ERM}

A good learning algorithm is able to find an estimator $f$ approximating the target function $f^*$.
From the computational viewpoint, though, the minimization of the expected risk yielding $f^*$,
$$
\inf_{f \in \mathcal{F}} \EE(f),
$$
is unfeasible, since $\rho$ is unknown and the algorithm can only access a finite training set $S_n$.
An effective approach to learning algorithm design in this setting is to minimize the error on the finite training set instead of the whole distribution.
To this end, given a loss $\ell$ we define the {\em empirical risk} (or {\em empirical error})
$$
\EM(f)=\frac 1 n \sum_{i=1}^n \ell(y_i, f(x_i)),
$$
acting as a proxy for the expected error defined in~\eqref{eq:expectedError}.
In practice, to turn the above idea into an actual algorithm we need to select a suitable hypotheses space  $\mathcal H \subset \mathcal F$ of candidate estimators, such that $\EM(f)$ is well defined $\forall f \in \mathcal{H}$.
The hypotheses space should be such that computations are feasible and, at the same time, it should be {\em rich} enough to approximate $f^*$.
To sum up, the ERM problem can be written as
\begin{equation}\label{HExpErrMin}
\inf_{f\in \hh} \EM(f).
\end{equation}
One possible method for controlling the size of the hypotheses space $\hh$ is \textit{Tikhonov regularization} \citep{TikhonovOriginal}.
This method adds a so-called \textit{regularizer} to the empirical risk minimization problem in \eqref{HExpErrMin}, which allows to control the size of $\hh = \hh_{\la}$ via the regularization parameter $\la$.
A regularizer is a functional $R:\hh\to [0,\infty)$ that penalizes estimators which are too ``{\em complex}".
In this case, we could replace~\eqref{HExpErrMin} by
\begin{equation}\label{RExpErrMin}
\inf_{f\in \hh} \EM(f) +\lambda R(f),
\end{equation}
for some \textit{regularization parameter} $\lambda>0$.
, with 
$$
\fn _\lambda = \arginf_{f\in \hh} \EM(f) +\lambda R(f).
$$
In particular, we will use the squared norm in $\hh$ as a regularizer, obtaining
\begin{equation}\label{eq:TikhoII}
\fn _\lambda = \arginf_{f\in \hh} \EM(f) +\lambda \|f\|_\hh^2,
\end{equation}
with $\|f\|_\hh^2 = \langle f,f \rangle_{\hh}$.

\section{Regularized Least Squares}
\label{sec:RLS}
In this section, we introduce Regularized Least Squares (RLS), a learning algorithm based on Tikhonov regularization employing the square loss.\\
The learning algorithm is defined as
\begin{equation}\label{eq:Tikhorls}
\min_{w\in \mathbb R^d}
\frac 1 n \sum_{i=1}^n (y_i-  w^\top x_i)^2
+\lambda w^\top w,\quad  \lambda \ge 0,
\end{equation}
considering as hypotheses space the class of linear functions, that is 
\begin{equation}\label{linearFunctionsSpace}
\mathcal H=\{f:\mathbb R^d\to \mathbb R~:~ \exists w\in \ \mathbb R^d ~\text{such that}~ f(x)=x^\top w, ~\forall x \in \mathbb R^d\}.
\end{equation}
Each function $f$ is defined by a vector $w$, and we let $f_w(x)=x^\top w$.
A motivation for considering the above scheme is to view  the empirical risk 
$$
\En(f_w) = 
\frac 1 n \sum_{i=1}^n (y_i-  w^\top x_i)^2,
$$
with $f_w(x) = w^\top x$, as a proxy for the expected risk
$$
\EE(f_w) =  
\int \rho(x,y)(y-  w^\top x)^2 dxdy,
$$
which is not computable.
Note that finding a function $f_w$ reduces to finding the corresponding vector $w$.
The term $w^\top w$ is the regularizer and helps preventing overfitting by controlling the stability of the solution. 
The parameter 
$\lambda$ balances the empirical error term and the regularizer. 
As we will see in the following, this seemingly simple example will be the basis for much more complicated solutions.

\subsection{Computations}

It is convenient to introduce the  input matrix $X \in \R^{n \times d}$, whose rows are the input samples, and the  output vector  $Y \in \R^{n}$ whose entries are the corresponding outputs\footnote{Note that in the 1-vs-all multiclass classification setting the output vector $Y$ becomes an $n \times T$ matrix.}. With this notation, the empirical risk can be expressed as
$$
\frac 1 n \sum_{i=1}^n (y_i-  w^\top x_i)^2=\frac 1 n \|Y- X w\|^2.
$$
A direct computation shows that the gradients with respect to $w$ of the empirical risk and the regularizer are, respectively,
$$
-\frac 2 n X^\top (Y- X w), \quad \text{and} \quad 2 w.
$$
Then, setting the gradient to zero,  we have that the solution of regularized least squares solves the linear system
$$
(X^\top X+ n \lambda  I_d)w= X^\top Y,
$$
where $I_d$ is the $d \times d$ identity matrix.
Several comments are in order.
First, several methods can be used to solve the above linear system, Cholesky decomposition being the method of choice, since the matrix
 $X^\top X+n\lambda I_d$ is symmetric and positive definite. 
The complexity of the method is essentially $O(nd^2)$ for training
and $O(d)$ for testing. The parameter $\lambda$ controls the {\em invertibility} of the matrix  
$(X^\top X+ n \lambda  I_d)$.

	\chapter{Kernel Methods}
	\label{Chap:kernelMethods}	
				
In this section, we introduce the key concepts of \textit{feature map} and \textit{kernel}, that allow to generalize RLS to nonlinear models. 

\section{Feature Maps}

A \textit{feature map} is a map 
$$\Phi: \X \mapsto \V$$
 from the input space $\X$ into a new space $\V$ called \textit{feature space}, endowed with a scalar product
denoted by $\langle \cdot,\cdot \rangle_{\V}$.
The feature space can be infinite dimensional.
 
 \subsection{Beyond Linear Models}
\begin{figure}[t!]
\begin{center}
\includegraphics[width=3.in]{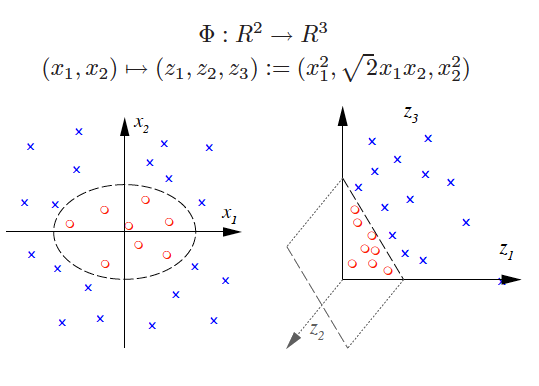}
\end{center}
\caption{A pictorial representation of the potential effect of considering a feature map $\Phi : \R^2 \mapsto \R^3$ in a simple binary classification problem.}\label{fig:feat map}
\end{figure}

The simplest case is when $\V=\R^p$, and we can view the entries $\Phi(x)^j$, $j=1,\dots, p$ as novel measurements on the input points.
For instance, consider $\X=\R ^2$ and the feature map $x=(x_1, x_2)\mapsto \Phi(x)=(x_1^2,\sqrt{2} x_1x_2,x_2^2)$. With this choice, if we now consider 
$$
f_w(x)=w^\top \Phi(x)=\sum_{j=1}^p w^j\Phi(x)^j ,
$$
we effectively have that the function is no longer linear in the original input space $\X$, but it is a polynomial of degree $2$. 
Clearly, the same reasoning holds for much more general choices of measurements (features), in fact {\em any} finite set of measurements. 
Although seemingly simple,  the above  observation allows to consider very general models.
Figure \ref{fig:feat map} gives a geometric interpretation of the potential effect of considering a feature map.
Points which are not easily classified by a linear model in the input space $\X$
can be easily classified  by a {\em linear model in the feature space} $\V$.

\subsection{Computations}
While feature maps allow to design nonlinear  models, the computations are essentially the same as in the linear case.
Indeed,  it is easy to see that the computations considered for linear models, under different loss functions, remain unchanged, 
as long as we change $x \in \mathbb R ^d$ into $\Phi(x)\in \mathbb R ^p$.
For example, for least squares we simply need to replace the $n  \times d$ 
input matrix $X$ with a  new $n \times p$ feature matrix $\Phi_n$, where each row is the image of an input point in the feature space, as defined by the feature map.
Thus, the only changes in the time and memory complexities of the learning algorithms of choice lay in the replacement of $d$ with $p$, and are noticeable only if $p \gg d$.

\section{Representer Theorem}
In this section we discuss how the above reasoning can be further generalized. The key result  is that 
the solution of regularization problems of the form \eqref{eq:TikhoII} can always be written as
\begin{equation}
\label{eq:repre}
\widehat w^\top =\sum_{i=1}^n x_i^\top \alpha_i, 
\end{equation} 
where $x_1, \dots, x_n$ are the training inputs and $\alpha=(\alpha_1, \dots, \alpha_n)$ are a set of coefficients.
The above result is an instance of the so-called representer theorem. 
We first discuss this result in the context of RLS.

\subsection{Representer Theorem for RLS}
The result follows by noting that the following equality holds,
\begin{equation}
\label{reprls_trick}
(X^\top X+\lambda n I_d)^{-1}X^\top =X^\top (XX^\top +\lambda n I_n)^{-1},
\end{equation}
so that we have, 
$$
w=X^\top  \underbrace{(XX^\top +\lambda n I_n)^{-1}Y}_{c}=\sum_{i=1}^n x_i^\top c_i.
$$
Equation~\eqref{reprls_trick} follows from considering the singular value decomposition (SVD) of $X$, that is $X=U\Sigma V^\top $. 
Therefore, we have $X^\top =V\Sigma^\top U^\top $, so that 
$$
(X^\top X+\lambda n I_d)^{-1}X^\top = V (\Sigma^\top \Sigma + n \lambda I_d)^{-1} \Sigma^\top   U^\top    
$$
and 
$$
X^\top (XX^\top +\lambda n I_n)^{-1}=V\Sigma^\top (\Sigma \Sigma^\top + n \lambda I_n)^{-1} U^\top .
$$
Note that the equality
$$
(\Sigma^\top \Sigma + n \lambda I_d)^{-1} \Sigma^\top =  \Sigma^\top (\Sigma \Sigma^\top + n \lambda I_n)^{-1}
$$
can be trivially verified.

\subsection{Representer Theorem Implications}
Using \eqref{eq:repre} and \eqref{reprls_trick}, it is possible to show how the vector $\alpha$ of coefficients can be computed considering different loss functions.
In particular, for  the square loss the vector of coefficients   satisfies the following linear system 
\begin{equation}\label{KRLSLinSys}
(K_n+\lambda n I_n)\alpha=Y
\end{equation}
where $K_n \in \R^{n \times n}$ is a matrix with entries $(K_n)_{i,j}=x_i^\top x_j$. 
The matrix $K_n$  is called the {\em kernel matrix} 
and is symmetric and positive semi-definite.

%
%

\section{Kernels}
\label{sec:kernels}
Given an input space $\X$, a symmetric positive definite definite function $K: \X \times \X \to \R$ is a \textit{kernel function} if there exists a feature map $\Phi: \X \mapsto
\hh$ for which
$$
K(x,x') = \scal{\Phi(x)}{\Phi(x')}_\hh,\quad \forall x,x' \in \X,
$$
where $\hh$ is the unique Hilbert space of functions  from $\X$ to $\R$ defined by $K$ as the completion of the linear span 
$\{K(\cdot,x)~:~ x \in \X \}$ with respect to the  inner product
$$
\scal{K(\cdot,x)}{K(\cdot,x')}_\hh = K(x,x'), \quad x, x' \in \X,
$$ 
see \citep{aronszajn1950theory}.
The space $\hh$ is called the reproducing kernel Hilbert space (RKHS) associated to the reproducing kernel function $K$.
The kernel matrix $(K_N)_{i,j}=K(x_i,x_j) $ associated to the positive definite kernel function $K$ is positive semi-definite for all $x_1, \dots, x_N\in \X$, $N \in \N$.
Moreover, the following two properties hold for $K$ and $\hh$:
\begin{enumerate}
\item For all $x\in \X$, $K_x(\cdot)=K(x, \cdot)$ belongs to $\hh$.
\item The so called reproducing property holds: $f(x)=\scal{f}{K_x}_\hh$, for all $f\in \hh,\, x\in \X$ \citep{steinwart2008support}.
\end{enumerate}
As we saw in the previous section, one of the main advantages for using the representer theorem is that the solution of the problem depends on the input points 
only through inner products $x^\top x'$. Kernel methods can be seen as replacing the inner product with a more general function
$K(x,x')$. In this case, the representer theorem~\eqref{eq:repre}, that is $f_w(x)=w^\top x=\sum_{i=1}^n x_i^\top x\alpha_i$,
 becomes
\begin{equation}
\label{eq:repre2}
\fn(x)=\sum_{i=1}^n K(x_i,x)\alpha_i,
\end{equation} 
and in this way we can directly derive \textit{kernelized} versions of many linear learning algorithms, including RLS, as we will see in the next section. 


%
%
Popular examples of positive definite kernels include:
\begin{itemize}
\item Linear kernel $K(x,x')=x^\top x'$
\item Polynomial kernel $K(x,x')=(x^\top x'+1)^d$
\item Gaussian kernel $K(x,x')=e^{-\frac{\|x-x'\|^2}{2\sigma^2}}$
\end{itemize}
The last two kernels have a tuning parameter, the degree $d$ and Gaussian bandwidth $\sigma$, respectively.
%


\section{Kernel Regularized Least Squares}
\label{sec:krr}
We now analyze the kernelized version of RLS, namely Kernel Regularized Least Squares (KRLS)  or Kernel Ridge Regression (KRR).
Given data points 
$$
(x_1, y_1), \dots (x_n,y_n) \in  (\X\times \R)^n,
$$  
and a kernel $K$,  KRLS is defined by the minimization problem,
\be\label{eq:def-krr}
\widehat{f}_\la = \argmin_{f \in \hh} \frac{1}{n} \sum_{i=1}^n (f(x_i)-y_i)^2 \; + \; \la \nor{f}^2_\hh, \quad \la > 0,
\ee
where $\nor{f}_\hh^2 = \scal{f}{f}_\hh$, for all $f \in \hh$.
The representer theorem \citep{kimeldorf1970correspondence,scholkopf2001generalized} shows that,  while the minimization is taken over a possibly infinite dimensional space, the minimizer of the above problem is of the form,
\be\label{eq:krr-clform}
\widehat{f}_\la(x) = \sum_{i=1}^n \alpha_i K(x,x_i), \quad {\alpha} = (\mK + \la n I_n)^{-1} Y,
\ee
where ${\alpha} = (\alpha_1,\dots,\alpha_n)$, $\mK \in \R^{n\times n}$ with $(\mK)_{ij} = K(x_i,x_j)$ and $Y = (y_1,\dots,y_n)$.\\
In terms of computational complexity we can say that:
\begin{itemize}
\item Time complexity is $O(n^3 +n^2 d)$, where $n^2 d$ refers to the computation of $K_n$ and $n^3$ to the inversion of the $n \times n$ matrix $(K_n+\lambda n I_n)$.
\item Memory complexity is $O(n^2)$, due to the storage of the kernel matrix $K_n$.
\end{itemize}
In practice, KRLS's memory complexity is its main practical bottleneck, since state-of-the-art computers cannot trivially store in RAM full kernel matrices for $n > 10^5$.
Consequently, in large-scale applications, in which $n$ can be considerably larger, so-called ``exact'' kernel methods using the full kernel matrix, such as KRLS, are not a viable option.
In Part \ref{part:largeScale}, we will see how kernel methods can be extended to large-scale settings and study their generalization properties.

	\chapter{Spectral Regularization}
	\label{Chap:spectralReg}	
		In this section, we recall the concept of \textit{spectral regularization}, a general formalism describing a large class of regularization methods giving rise to consistent kernel methods.

\section{Introduction}
\label{sec:spectralIntro}
Spectral regularization originates from the inverse problems literature, in particular from methods to invert matrices in a numerically stable way.
The idea of applying spectral regularization to
statistics \citep{Wahba/90} and machine learning \citep{vapnik1982estimation,poggio1989theory,schlkopf2002learning,devito2005learning,gerfo2008spectral} is based on the observation that the same principles allowing for numerically stable inversion can be shown to prevent overfitting in the SLT framework.
Different spectral methods have a common derivation, but result in different regularized learning algorithms with specific computational complexities and statistical properties.
As we saw in Chapter \ref{Chap:SLT}, a learning problem can be framed as the minimization of the expected risk on a suitable hypotheses space $\hh$.
The idea is that the solution satisfies
$$
\inf_{f\in \hh} \EE(f).
$$
In the following, we consider the square loss for simplicity.
In practical algorithms, the empirical risk $\EM(f) =  \frac{1}{n} \sum_{i=1}^n (y_i-f(x_i))^2$ is introduced as a proxy of the expected risk, and empirical risk minimization, 
$$
\inf_{f\in \hh} \EM(f) = 
\inf_{f\in \hh} \frac{1}{n}\sum_{i=1}^n (y_i-f(x_i))^2,
$$
is performed.
The unregularized solution to ERM (corresponding to the so called Kernel Ordinary Least Squares --- KOLS --- solution) can be written as
$$
\fn(x) = \sum_{i=1}^n \alpha_i K(x,x_i),
$$
where $K$ is a suitable kernel function and the coefficients vector $\alpha \in \R^n$ is the solution of the inverse problem
\begin{equation}\label{eq:kolsSpectral}
K_n \alpha = Y.
\end{equation}
The solution can be subject to numerical instability caused by noise and sampling.
For example, in the learning setting the kernel matrix can be
decomposed as
$$
K_n = Q \Sigma Q^\top,
$$
where $\Sigma = \text{diag}(\sigma_1 , . . . , \sigma_n )$
is the eigenvalues matrix, $\sigma_1 \geq \sigma_2 \geq ...\sigma_n \geq 0$
are the eigenvalues in decreasing order, and $
q_1 , . . . , q_n$ are the corresponding eigenvectors.
Then,
\begin{align}
\begin{split}
\alpha &= K_n^{-1} Y \\
&= Q \Sigma^{-1} Q^\top Y \\
&=\sum_{i=1}^n \frac{1}{\sigma_i}\langle  q_i,Y\rangle q_i,
\end{split}
\end{align}
since $[\Sigma^{-1}]_{ii} = 1/\sigma_i$.
It is therefore clear that in correspondence of small eigenvalues, small perturbations of
the data due to sampling and noise can cause large changes in the solution.
The spectral regularization literature includes a rich variety of methods allowing to invert linear operators with high condition number\footnote{The condition number of a normal matrix $A$ is defined as $\kappa(A) = \vert \sigma_{\text{max}}\vert  / \vert \sigma_{\text{min}}\vert $, where $\sigma_{\text{max}}, \sigma_{\text{min}}$ are the maximal and minimal eigenvalues of $A$ respectively. Kernel matrices are normal.} in a stable way.
In general, spectral regularization methods act on the eigenvalues of the matrix to stabilize its inversion.
This is done by replacing the original unbounded operator with a \textit{regularization operator} \citep{engl1996regularization}, which allows to control the condition number via a regularization parameter.
We will see some examples of this in the following, starting from the case of Tikhonov regularization.

\section{Tikhonov Regularization}
In Tikhonov regularization \citep{TikhonovOriginal}, an explicit penalization term $R$ is added to the ERM objective function to enforce smoothness of the solution and prevent overfitting, as follows
\begin{equation}
\inf_{f\in \hh} \EM(f) + \la R(f)= 
\inf_{f\in \hh} \frac{1}{n} \sum_{i=1}^n (y_i-f(x_i))^2 + \la \|f\|_\hh^2.
\end{equation}
We will now observe that Tikhonov regularization has an effect from a
numerical point of view.
In fact, it yields the linear system
\begin{equation}\label{eq:tikSpectral}
(K_n + n\la I_n)\alpha = Y,
\end{equation}
which stabilizes the possibly ill-conditioned matrix inversion problem of \eqref{eq:kolsSpectral}.
In particular, by considering the eigendecomposition associated to the regularized problem in \eqref{eq:tikSpectral} we obtain that
\begin{align}
\label{eq:spectralTikhonov}
\begin{split}
\alpha &= (K_n + n\la I_n)^{-1} Y \\
&= Q(\Sigma + n\la I_n)^{-1} Q^\top Y \\
&=\sum_{i=1}^n \frac{1}{\sigma_i+n \la}\langle  q_i,Y\rangle q_i.
\end{split}
\end{align}
Regularization filters out the undesired components associated to small eigenvalues, increasing the condition number of the regularized linear operator.
Eigenvalues are affected as follows:
\begin{itemize}
\item If $\sigma \gg \la n$, then $\frac{1}{\sigma_i +n\la} \approx \frac{1}{\sigma_i}$
\item If $\sigma \ll \la n$, then $\frac{1}{\sigma_i +n\la} \approx \frac{1}{n\la}$
\end{itemize}
Consequently, the condition number is controlled as follows:
$$
\kappa(K_n) = \frac{\vert \sigma_{\text{max}}\vert }{\vert \sigma_{\text{min}}\vert }
\Longrightarrow
\kappa(K_n + n\la I_n) = \frac{\vert \sigma_{\text{max}}\vert }{\vert  n \la \vert }
$$
A good range for the regularization parameter $\la$ falls between the smallest and the largest eigenvalues of $K_n$.

\subsection{Regularization Filters}
We can generalize the notion of spectral regularization beyond the Tikhonov case by introducing the concept of  \textit{regularization filter} \citep{bertero1998introduction} $G_\la :\R^{n \times n} \mapsto \R^{n \times n}$ acting on the eigenvalues of the kernel matrix, defined as
$$
G_\la (K_n) = Q \bar G_\la(\Sigma) Q^\top,
$$
with $\bar G_\la :\R^{n \times n} \mapsto \R^{n \times n}$.
$\bar G_\la$, in turn, is defined in terms of the \textit{scalar filter function} $g_{\la}: \R \mapsto \R$ as
$$
[\bar G_\la(\Sigma)]_{ii} = g_{\la}(\sigma_i).
$$
What $g_{\la}$ does is simply to invert the  eigenvalues in a controlled way to enforce smoothness of the solution.
For instance, in the case of Tikhonov regularization filtering, we have that
$$
g_{\la}(\sigma) = \frac{1}{\sigma + n \la}
$$
is the corresponding scalar function.
Thus, as seen in Equation \eqref{eq:spectralTikhonov}, the coefficients vector can be computed as
\begin{align}
\begin{split}
\alpha &= G_\la(K_n)  \\
&= \sum_{i=1}^n g_{\la}(\sigma_i) \langle  q_i,Y\rangle q_i \\
&= \sum_{i=1}^n \frac{1}{\sigma_i + n \la} \langle  q_i,Y\rangle q_i.
\end{split}
\end{align}
As we will see in the following sections, this formalism is very flexible and can be used to characterize different regularized learning algorithms in a unified way.
This class of algorithms is known collectively as spectral regularization.
Each algorithm is defined by a suitable filter function $G_\la$, and is not necessarily based on penalized ERM.
The notion of filter function was studied in machine learning
and gave a connection between function approximation in
signal processing and approximation theory. 
\begin{rem}[Filter function properties]
Not every scalar function defines a regularization scheme.
Roughly speaking, a good filter function must have the following
properties:
\begin{itemize}
\item As $\la$ goes to $0$, $g_{\la} (\sigma) \rightarrow 1/\sigma$ so that
$G_{\la} (K_n) \rightarrow K_n^{-1}$.
\item $\la$ controls the magnitude of the (smaller) eigenvalues of
$G_{\la} (K_n)$.
\end{itemize}
\end{rem}



\section{Spectral Cut-off}
		
This method is one of the oldest regularization techniques and
is also known as Truncated Singular Value Decomposition (TSVD) and Principal Component Regression (PCR).
Its nature is simple to explain:
Given the eigendecomposition $K_n = Q\Sigma Q^\top$, a regularized
inverse of the kernel matrix is built by discarding all the
eigenvalues smaller than the prescribed threshold $n \la$.
The associated regularization filter $\bar G_\la$ is defined as
$$
\bar G_\la(\Sigma) = \textrm{diag}(\sigma_1^{-1},\dots,\sigma_{m}^{-1}, 0,\dots, 0),
$$
where $m \in \N$ is the largest index for which $\sigma_i \geq n\la$,
and corresponds to the scalar filter function
$$
g_\la (\sigma) = 
\begin{cases} \frac{1}{\sigma} & \mbox{if } \sigma \geq n\la \\ 
0 & \mbox{otherwise}
\end{cases}.
$$
Interestingly enough, one can show that spectral cut-off is equivalent to
the following procedure:
\begin{itemize}
\item Unsupervised projection of the data using (kernel) PCA \citep{scholkopf1998nonlinear}.
\item ERM on the projected data without explicit
regularization.
\end{itemize}
Note that the only free parameter is the number of components $m$ we retain
for the projection, which depends on the threshold $n\la$.
Therefore, we can say that in this algorithm the regularization operation coincides with the projection on the $m$ largest eigencomponents.

\section{Iterative Regularization via Early Stopping}
\label{sec:itReg}

In the previous sections, we have seen how explicit penalization (Tikhonov) and projection (spectral cut-off) can implement regularization mechanisms.
Here, we outline another regularization strategy based on iterative regularization, whose driving principle is to recursively compute a sequence of solutions to the learning problem.
The first few iterations yield simple solutions, while executing too many iterations may result in increasingly complex solutions, potentially leading to overfitting phenomena.
Therefore, early termination of the iterations (early stopping) has a regularizing effect.
We now describe this idea in greater detail by considering one of its most clear-cut instances, the Landweber iteration.

\subsection{Landweber Iteration}

The Landweber iteration (or iterative Landweber algorithm) can be seen as the minimization of the empirical risk
$$
\En (\widehat f) = \frac{1}{n} \|Y-K_n c\|_2^2
$$
via gradient descent.
The Landweber iteration defines a sequence of solutions as follows:
\begin{equation}
\alpha_i = \alpha_{i-1} + \eta (Y - K_n \alpha_{i-1} ),
\end{equation}
with $\alpha_0 = 0$.
If the largest eigenvalue of $K_n$ is smaller than $n$, the above
iteration converges if we choose the step size $\eta = 2/n$.
It can be proven by induction that the solution at iteration $t$ is
$$
\alpha = \eta \sum_{i=0}^{t-1}(In - \eta K_n)^i Y.
$$
Note that the well-known relation
$$
\sum_{i=0}^\infty (1-a)^i=a^{-1} \quad \forall a\in (0,1)
$$
also holds replacing $a$ with a
matrix.
The resulting formula is called Neumann series,
$$
\sum_{i=0}^\infty (I-A)^i=A^{-1},
$$
and holds for any invertible matrix $A$ such that $\|A\| <1$.
If we consider the kernel matrix (or rather $I_n - \eta K_n$), we obtain that an approximate inverse for it can be defined considering a truncated Neumann series, that is
\begin{equation}
K_n^{-1} = \eta \sum_{i=0}^\infty (I_n - \eta K_n )^i \approx \eta \sum_{i=0}^{t-1} (I_n - \eta K_n )^i = G_t(K_n).
\end{equation}
The filter function $G_t(K_n)$ of the Landweber iteration corresponds to a
truncated power expansion of $K_n^{-1}$, and the associated scalar filter function is
$$
g_t(\sigma) = \eta \sum_{i=0}^{t-1} (I_n - n \sigma)^i.
$$
The regularization parameter is the number of iterations $t$.
Roughly speaking, $t \sim 1/\la$.
In fact, 
\begin{itemize}
\item Large values of $t$ correspond to minimization of the
empirical risk and tend to overfit.
\item Small values of $t$ tend to oversmooth (recall that $c_0 = 0$).
\end{itemize}
Early stopping of the iteration allows to find an optimal trade-off between oversmoothing and overfitting solutions, which corresponds to a regularization effect.

\part{Kernel Methods for Large-scale Learning}
\label{part:largeScale}

\chapter{Speeding up by Data-dependent Subsampling}
\label{chap:lessismore}
	
\section{Introduction}
In Chapter \ref{Chap:kernelMethods}, we have seen how kernel methods provide an elegant framework to develop nonparametric 
statistical approaches to learning \citep{schlkopf2002learning}.
Prohibitive memory requirements of exact kernel methods, making these methods unfeasible when dealing with large datasets, have also been discussed.
Indeed, this observation has motivated a variety of  computational strategies to develop large-scale kernel methods \citep{conf/icml/SmolaS00,conf/nips/WilliamsS00,conf/nips/RahimiR07,conf/icml/YangSAM14,conf/icml/LeSS13,conf/icml/SiHD14,conf/colt/ZhangDW13}. Approximation schemes based on generative probabilistic models have also been proposed in the Gaussian Processes literature (see for example \citep{quinonero2005unifying}), and are beyond the scope of this work.
\\
In this chapter, we devote our attention to subsampling methods, that we broadly refer to as Nystr\"om 
approaches.
These methods replace the empirical kernel matrix, 
needed by standard kernel methods, with a smaller matrix obtained by column subsampling \citep{conf/icml/SmolaS00,conf/nips/WilliamsS00}. Such  procedures are shown to often dramatically reduce memory/time requirements while preserving good practical performances \citep{conf/nips/KumarMT09,conf/icml/LiKL10,Zhang:2008:INL:1390156.1390311,conf/nips/DaiXHLRBS14}.\\
In Section \ref{sec:lessismore} we study our recently proposed \citep{rudi2015less} optimal learning bounds of subsampling schemes such as the Nystr\"om method, while in Section \ref{sec:nytro} we investigate the generalization properties of NYTRO, a novel regularized learning algorithm combining subsampling and early stopping \citep{camoriano2016nytro}.

	\section{Less is More: Regularization by Subsampling}
	\label{sec:lessismore}
	\subsection{Setting}

%
The goal of this section is two-fold. First, and foremost,  we aim at providing a theoretical characterization of  the generalization properties of Nystr\"om methods
in a statistical learning setting. Second,  we wish to understand the
role played by the subsampling level both from a statistical and a computational point of view. As discussed in the following, this latter question leads to a natural variant of
Kernel Regularized Least Squares (KRLS)
, where the subsampling level controls both regularization and computations. 

From a theoretical perspective, the effect of  Nystr\"om approaches has been primarily characterized considering the discrepancy between a given empirical kernel matrix and its subsampled version \citep{Drineas:2005:NMA:1046920.1194916,gittens2013revisiting,Wang:2013:ICM:2567709.2567748,journals/jmlr/DrineasMMW12,conf/innovations/CohenLMMPS15,conf/aistats/WangZ14,Kumar:2012:SMN:2503308.2343678}. While interesting in their own right, these latter results do not directly  yield information on the generalization properties of the obtained algorithm. Results in this direction, albeit suboptimal,  were first derived in \citep{journals/jmlr/CortesMT10} (see also \citep{6547995,conf/nips/YangLMJZ12}), and more recently in \citep{conf/colt/Bach13,alaoui2014fast}. 
In these latter papers, sharp error analyses in expectation are derived  in a fixed design  regression setting for a form of Kernel Regularized Least Squares. 
In particular, in  \citep{conf/colt/Bach13} a basic uniform sampling approach is studied, while in \citep{alaoui2014fast} a subsampling scheme based on the notion of leverage score is considered.   
The main technical contribution of our study is an extension of these latter results to the statistical learning setting, where  the design is random and  high probability estimates are considered. 
The more general setting makes the analysis considerably more complex. 
Our main result gives optimal finite sample bounds for both uniform and leverage score based subsampling strategies. 
These methods are shown to achieve the same (optimal) learning error as KRLS, recovered as a special case, while allowing substantial computational gains. 
Our analysis highlights the interplay  between the Tikhonov regularization and subsampling parameters, suggesting 
that the latter can be used to control simultaneously  regularization and computations. 
This strategy implements a form of {\em computational regularization} in the sense that the computational resources are tailored to the generalization properties in the data. This idea is developed  considering an incremental strategy to efficiently compute   learning solutions for different subsampling levels. 
The  procedure thus obtained,  which is a simple variant of classical Nystr\"om Kernel Regularized Least Squares (NKRLS) with uniform sampling,  allows for efficient  model selection and achieves state of the art results on a variety of benchmark large-scale datasets. \\
The rest of the Section is organized as follows. In Subsection \ref{sect:krls-nyst}, we introduce the setting and algorithms we consider.
In Subsection \ref{sect:theo-analysis}, we present our main theoretical contributions. In Subsection \ref{sect:incAlgoExperiments}, we discuss computational aspects and experimental results.
%
%
%
%
%
\subsection{Supervised Learning with KRLS and  \Nystrom{}} \label{sect:krls-nyst}
We consider a learning setting based on the one outlined in Chapter \ref{Chap:SLT}.
Let $\X\times \R$ be a probability space with distribution $\rho$, where we view $\X$ 
and $\R$ as the input and output spaces, respectively. 
The learning goal is to minimize the {\em expected risk},
\be\label{eq:expmin}
\min_{f\in \hh} \EE(f), \quad\quad  \EE(f) = \int_{\X\times \R} (f(x)-y)^2 d\rho(x,y),
\ee
provided $\rho$ is known only through a training set.
In the following, we consider kernel methods, as introduced in Section \ref{sec:kernels} in the case of random design regression, outlined in Section \ref{sec:probDataModel}, in which
\be\label{eq:rdreg}
y_i=f_*(x_i)+\eps_i, \quad i=1, \dots, n,
\ee
with $f_*$ a fixed {\em regression} function, $\eps_1, \dots, \eps_n$ a sequence of  random variables seen as noise, and $x_1, \dots, x_n$ random inputs.
%
In Section \ref{sec:krr}, we have seen that a classical way to derive an empirical solution to problem~\eqref{eq:expmin} is to consider the KRLS learning algorithm, based on Tikhonov regularization, 
\be\label{eq:krls-problem}
\min_{f \in \hh} \frac{1}{n}\sum_{i=1}^{n}(f(x_i)-y_i)^2 + \la \nor{f}^2_\hh, \la > 0.
\ee
We recall that a solution $\fn_\la$ to problem~\eqref{eq:krls-problem} exists, it is unique and the representer theorem \citep{schlkopf2002learning} shows that it can be written as 
 \be\label{eq:rep}
\fn_\la(x) = \sum_{i=1}^n \hat\alpha_i K(x_i, x) \quad \textrm{ with } \quad \hat\alpha = ({\K_n} + \la n I_n)^{-1} y, 
\ee
where $x_1,\dots, x_n$ are the training set points, $y = (y_1,\dots,y_n)$ and ${\K_n}$ is the empirical kernel matrix. Note that this result implies that we can restrict the minimization in~\eqref{eq:krls-problem} to the space,
 \be\label{eq:hhn}
\hh_n = \{f \in \hh~|~ f = \sum_{i=1}^n \alpha_i K({x}_i, \cdot),\; \alpha_1, \dots, \alpha_n \in \R\}.
\ee
As already discussed,  storing the kernel matrix ${\K_n}$, and solving the linear system in~\eqref{eq:rep},  can become computationally unfeasible as  $n$ increases.
In the following, we consider strategies to find more efficient solutions, based on the idea of replacing $\hh_n$ with 
$$\hh_m = \{f ~|~ f = \sum_{i=1}^m \alpha_i K(\tilde{x}_i, \cdot),\; \alpha \in \R^m\},$$
where $m\le n$ and $\{\tilde{x}_1, \dots, \tilde{x}_m\}$ is a subset of the input  points in the training set. The solution $\hat{f}_{\la, m}$ of the corresponding minimization problem can now be written as, 
\be \label{eq:repny}
\hat{f}_{\la, m}(x) = \sum_{i=1}^m \tilde{\alpha}_i K(\tilde{x}_i, x)\quad \textrm{with}\quad
\tilde{\alpha} = (K_{nm}^\top K_{nm} + \la n K_{mm})^\dag K_{nm}^\top y,
\ee 
where $A^\dag$ denotes the Moore-Penrose pseudoinverse of a matrix $A$, and $(K_{nm})_{ij} = K(x_i, \tilde{x}_j)$, $(K_{mm})_{kj} = K(\tilde{x}_k, \tilde{x}_j)$ with $i \in \{1,\dots,n\}$ and $j,k \in \{1,\dots,m\}$  \citep{conf/icml/SmolaS00}\footnote{Note that the estimator $\hat{f}_{\la, m}$ only depends on the $m$ selected points, while if we used subsampling to first construct an approximation of the matrix $K_n$ and then compute a vector $\alpha \in \R^n$ the estimator would be a combination of the kernel centered on all training points, thus less efficient.}.
The above approach is related to \Nystrom{} methods and   different approximation strategies  correspond to different ways to select the inputs subset. 
While our framework applies to a broader class of strategies, see Section C.1 of \citep{rudi2015less}, in the following we primarily consider two techniques.
\begin{itemize}
\item {\bf Plain \Nystrom{}}. The points $\{\tilde{x}_1, \dots, \tilde{x}_m\}$ are sampled uniformly at random without replacement from the training set.
\item {\bf Approximate leverage scores (ALS) \Nystrom{}}.
Recall that the {\em leverage scores} associated to the training set points $x_1, \dots, x_n$  are 
 \be \label{eq:levscoredef}
 (l_i(t))_{i=1}^n, \quad l_i(t) = (\K_n (\K_n + t n I)^{-1})_{ii}, \quad i \in \{1, \dots, n\}
 \ee 
for any $t > 0$, where $(\K_n)_{ij} = K(x_i, x_j)$. In practice, leverage scores are onerous to compute and approximations
$(\hat l_i(t))_{i=1}^n$ can be considered \citep{journals/jmlr/DrineasMMW12,alaoui2014fast,conf/innovations/CohenLMMPS15}.
 In particular, in the following we are interested in suitable approximations defined as follows:
\bd[$T$-approximate leverage scores]\label{def:approx-lev-scores}
Let $(l_i(t))_{i=1}^n$ be the leverage scores associated to the training set for a given $t$. Let $\delta > 0$, $t_0 > 0$ and $T \geq 1$. We say that $(\hat l_i(t))_{i=1}^n$ are $T$-approximate leverage scores with confidence $\delta$, when with probability at least $1-\delta$, 
$$
\frac{1}{T} l_i(t) \leq \hat l_i(t) \leq T l_i(t) \quad \forall i\in\{1,\dots,n\}, t \geq t_0.
$$
\ed
Given $T$-approximate leverage scores\footnote{Algorithms for approximate leverage scores computation were proposed in \citep{journals/jmlr/DrineasMMW12,alaoui2014fast,conf/innovations/CohenLMMPS15}.} for $t > \la_0$,  $\{\tilde{x}_1, \dots, \tilde{x}_m\}$  are sampled from the training set  independently with replacement, and  with  probability to be selected given by  $P_t(i) =  \hat l_i(t) / \sum_j \hat l_j(t)$.
\end{itemize}
In the next subsection, we state and discuss our main result showing that the KRLS formulation based on  plain or approximate leverage scores \Nystrom{} provides optimal empirical solutions to problem~\eqref{eq:expmin}.
%
\subsection{Theoretical Analysis}\label{sect:theo-analysis}
We now state and discuss our main results, for which several assumptions are needed.
The first  basic assumption is that problem~\eqref{eq:expmin} admits at least a  solution.
\ba\label{as:exists-fh} 
There exists an $f_\hh \in \hh$ such that 
$$\EE(f_\hh)=\min_{f\in \hh}\EE (f).$$
\ea
Note that,  while the minimizer might not be unique, our results apply to the case in which $f_\hh$ is the unique minimizer with minimal norm.
Also,  note that the above condition is weaker than assuming the regression function in~\eqref{eq:rdreg} to belong to $\hh$.
Finally, we note that our study can be adapted to  the case in which minimizers do not exist, but the analysis is considerably more involved and is therefore left to future work.
\\
The second assumption is a basic condition on the probability distribution.
\ba\label{as:noise}
Let $z_x$ be the random variable $z_x = y - f_\hh(x)$, with $x \in \X$, and $y$ distributed according to $\rho(y|x)$. Then, there exists $M, \sigma > 0$ such that $\mathbb{E} |z_x|^p \leq \frac{1}{2}p!M^{p-2}\sigma^2$ for any $p \geq 2$, almost everywhere on $\X$.
\ea
The above assumption is needed to control random quantities and is related to a {\em noise} assumption  in the regression model~\eqref{eq:rdreg}. It is clearly weaker than the often considered bounded output assumption \citep{steinwart2008support}, and trivially verified in classification.
\\
The last two assumptions describe the capacity (roughly speaking the {\em ``size''}) of the  hypothesis space induced by  $K$ with respect to  $\rho$ and the regularity of $f_\hh$ with respect to $K$ and $\rho$. 
To discuss them,  we first need the following definition.
\bd[Covariance operator and effective dimensions]
We define the covariance operator as 
$$
C: \hh \to \hh, \quad \scal{f}{C g}_\hh = \int_\X f(x)g(x)d\rhox(x) \;\;, \quad \forall \, f, g \in \hh.
$$
Moreover, for $\la>0$, we define the random variable
$${\cal N}_x(\la) = \scal{K_x}{(C+\la I)^{-1} K_x}_\hh,$$ 
with $x\in \X$ distributed according to $\rhox$, and let 
$$
{\cal N}(\la) = \mathbb{E}\,{\cal N}_x(\la), \quad\quad {\cal N}_\infty(\la)=\sup_{x\in \X} {\cal N}_x(\la).
$$
\ed
We add several comments. Note that $C$ corresponds to the second moment operator, but we refer to it as the covariance operator with an abuse of terminology. Moreover, note that ${\cal N}(\la)  = \text{Tr}(C(C+\la I)^{-1})$, where ``Tr'' indicates the trace of  a matrix (see \citep{caponnetto2007optimal}). This latter quantity, called effective dimension or degrees of freedom,  can be seen as a measure of the 
capacity of the hypothesis space. The quantity ${\cal N}_\infty(\la)$ can be seen to provide a uniform bound on the leverage scores in ~\eqref{eq:levscoredef}. Clearly, ${\cal N}(\la)\le {\cal N}_\infty(\la)$ for all $\la>0$.
\ba\label{as:kerrho}
The kernel $K$ is measurable, $C$ is bounded.
 Moreover, for all $\la>0$ and a $Q>0$,   
\eqal{
 & {\cal N}_\infty(\la)<\infty,\label{eq:lsbound}\\
 &{\cal N}(\la) \leq Q \la^{-\gamma}, \quad 0 < \gamma \leq 1.\label{eq:poleffdim}
}
\ea
Measurability of $K$ and boundedness of $C$  are  minimal conditions to ensure that the covariance operator is 
a well defined   linear, continuous, self-adjoint, positive operator \citep{steinwart2008support}. Condition~\eqref{eq:lsbound} is satisfied if the kernel is bounded
$\sup_{x\in \X}K(x,x)= \kappa^2<\infty$, indeed in this case $ {\cal N}_\infty(\la)\le \kappa^2/\la$ for all $\la>0$. Conversely, 
it can be seen that condition~\eqref{eq:lsbound} together with boundedness of $C$ imply  that the kernel is bounded, indeed
\footnote{If  ${\cal N}_\infty(\la)$ is finite, then ${\cal N}_\infty(\nor{C}) = \textrm{sup}_{x \in X} \nor{(C+\nor{C} I)^{-1}K_x}^2 \geq 1/2 \nor{C}^{-1}\textrm{sup}_{x \in X} \nor{K_x}^2$, therefore $K(x,x) \leq 2\nor{C}{\cal N}_\infty(\nor{C})$.}
$$
\kappa^2\le 2\nor{C}{\cal N}_\infty(\nor{C}).
$$
Boundedness of the kernel implies in particular that the operator $C$ is trace class  
and allows to  use tools from spectral theory. Condition~\eqref{eq:poleffdim} quantifies the capacity assumption and is related to 
covering/entropy number conditions (see \citep{steinwart2008support} for further details). In particular, it is known that condition~\eqref{eq:poleffdim}
is ensured if the eigenvalues $(\sigma_i)_i$ of $C$ satisfy a polynomial decaying condition $\sigma_i \sim i^{-\frac{1}{\gamma}}$.
Note that, since the operator $C$ is trace class, Condition~\eqref{eq:poleffdim} always holds for $\gamma=1$.
Here, for space constraints and in the interest of clarity we restrict  to such a polynomial  condition, but the analysis directly applies to other conditions including  exponential decay or a finite rank conditions \citep{caponnetto2007optimal}. 
Finally, we have the following regularity assumption. 
\ba\label{as:source}
 There exists $s \geq 0$, $1 \leq R < \infty$, such that $\nor{C^{-s} f_\hh}_{\hh} < R$. 
\ea
The above condition is fairly standard, and can be equivalently formulated in terms of classical concepts in approximation theory such as 
interpolation spaces \citep{steinwart2008support}. Intuitively, it quantifies the degree to which $f_\hh$ can be well approximated by functions in the RKHS $\hh$ and allows to control the bias/approximation error of a learning solution. For  $s=0$, it is always satisfied. For larger $s$,  we are assuming $f_\hh$ to belong to subspaces of $\hh$ that are the images of the  fractional compact operators $C^s$. Such spaces contain functions which, expanded on a basis of eigenfunctions of $C$, have larger coefficients in correspondence to large eigenvalues.
Such an assumption is natural in view of using techniques such as~\eqref{eq:rep}, which  can be seen as a form  of spectral filtering (see Chapter \ref{Chap:spectralReg}), that estimate stable solutions by discarding the contribution of small eigenvalues \citep{journals/neco/GerfoROVV08}.  
In the next section, we are going to quantify the quality of empirical solutions of Problem~\eqref{eq:expmin} obtained by schemes of the form~\eqref{eq:repny}, in terms of the quantities in Assumptions~\ref{as:noise},~\ref{as:kerrho},~\ref{as:source}. 
%
\subsubsection{Main results}\label{sect:main-res}
In this section, we state and discuss our main results,  starting 
with optimal finite sample error bounds for regularized least squares based on plain  and approximate leverage score based \Nystrom{} subsampling.
\bt\label{thm:opt-rates-NyKRLS}
Under Assumptions~\ref{as:exists-fh}, \ref{as:noise}, ~\ref{as:kerrho},  and ~\ref{as:source}, 
let  $\delta>0$,  $v = \min(s, 1/2)$, $p = 1 + 1/(2v + \gamma)$ and assume
$$
n \,\geq\, 1655\kappa^2 + 223\kappa^2\log\frac{6\kappa^2}{\delta} + \left(\frac{38p}{\nor{C}} \log \frac{114\kappa^2 p}{\nor{C}\delta} \right)^p.$$ 
Then,  the following inequality  holds with probability at least $1-\delta$, 
\be\label{eq:excess-risk-bounded}
 \EE(\hat{f}_{\la, m}) - \EE(f_\hh) \leq q^2\, n^{-\frac{2v+1}{2v + \gamma + 1}}, 
\ee
with
\be 
 q = 6R\left(2\nor{C}+\frac{M\kappa}{\sqrt{\nor{C}}} + \sqrt{\frac{Q\sigma^2 }{\nor{C}^\gamma}}\right)\log\frac{6}{\delta},
\ee
with  $\hat{f}_{\la, m}$  as in~\eqref{eq:repny}, $\la = \nor{C} n^{-\frac{1}{2v + \gamma + 1}}$ 
and 
\begin{enumerate}
\item for plain \Nystrom{} 
$$m \geq (67 \vee 5 {\cal N}_\infty(\la))\log\frac{12\kappa^2}{\la \delta};$$
\item for ALS \Nystrom{} and $T$-approximate leverage scores with subsampling probabilities $P_\la$, $t_0 \geq \frac{19\kappa^2}{n}\log\frac{12n}{\delta}$ and 
 $$m \geq  (334 \vee 78 T^2 {\cal N}(\la)) \log \frac{48n}{\delta}.$$
\end{enumerate}
\et
We add several comments.
First, the above results can be shown to be optimal in a minimax sense.
Indeed, minimax lower bounds proved in \citep{caponnetto2007optimal, SteinwartHS09} show that the learning rate in~\eqref{eq:excess-risk-bounded} is optimal under the considered assumptions (see Theorems~2,~3 of \citep{caponnetto2007optimal}, for a discussion on minimax lower bounds see Section~2 of \citep{caponnetto2007optimal}). 
Second, the obtained bounds can be compared to those obtained for other regularized learning techniques. 
Techniques known to achieve optimal error rates include Tikhonov regularization \citep{caponnetto2007optimal,SteinwartHS09,mendelson2010regularization}, iterative regularization by early stopping \citep{bauer2007regularization,CapYao06}, spectral cut-off regularization (a.k.a. principal component regression or truncated SVD) \citep{bauer2007regularization,CapYao06}, as well as regularized stochastic gradient methods \citep{ying2008online}.  All these techniques are essentially equivalent from a statistical point of view and differ only in the required computations. For example, iterative methods allow for a computation of solutions corresponding to different regularization levels which is more efficient than Tikhonov or SVD based approaches. 
The key observation is that  all these methods have the same $O(n^2)$ memory requirement. In this view, our results show that randomized subsampling methods can break such a memory barrier, and consequently achieve much better time complexity,  while preserving optimal learning guarantees. Finally,  we can compare our results with previous analysis of randomized kernel methods. As already mentioned, results close to those in Theorem~\ref{thm:opt-rates-NyKRLS} are given  in \citep{conf/colt/Bach13,alaoui2014fast} in a fixed design setting.  Our results extend and generalize the conclusions of these papers to a general statistical learning setting.  Relevant results are given in \citep{conf/colt/ZhangDW13} for a different  approach, based on averaging KRLS solutions obtained splitting the data in $m$ groups ({\em divide and conquer RLS}). The analysis in \citep{conf/colt/ZhangDW13} is only  in expectation, but considers random design and  
shows that the proposed method is indeed optimal provided the number of splits is chosen 
depending on the effective dimension $ {\cal N}(\la)$.
This is the only other work we are aware of establishing optimal learning rates for 
randomized kernel approaches in a statistical learning setting. In comparison with \Nystrom{} computational regularization the main disadvantage of the divide and conquer approach is computational and in the model selection phase where solutions corresponding to different regularization parameters and number of splits usually need to be computed. 
\\
The proof of Theorem~\ref{thm:opt-rates-NyKRLS} is fairly technical and lengthy. It incorporates ideas from \citep{caponnetto2007optimal} and techniques developed to study spectral filtering regularization \citep{bauer2007regularization,rudi2013sample}. In the next section, we briefly sketch some main ideas and discuss how they suggest an interesting  perspective on 
regularization techniques including subsampling.
\begin{figure}[t]
{\centering
\subfigure{
                \includegraphics[trim=0cm 1cm 0cm 0.5cm, clip=true, width=0.3\textwidth]{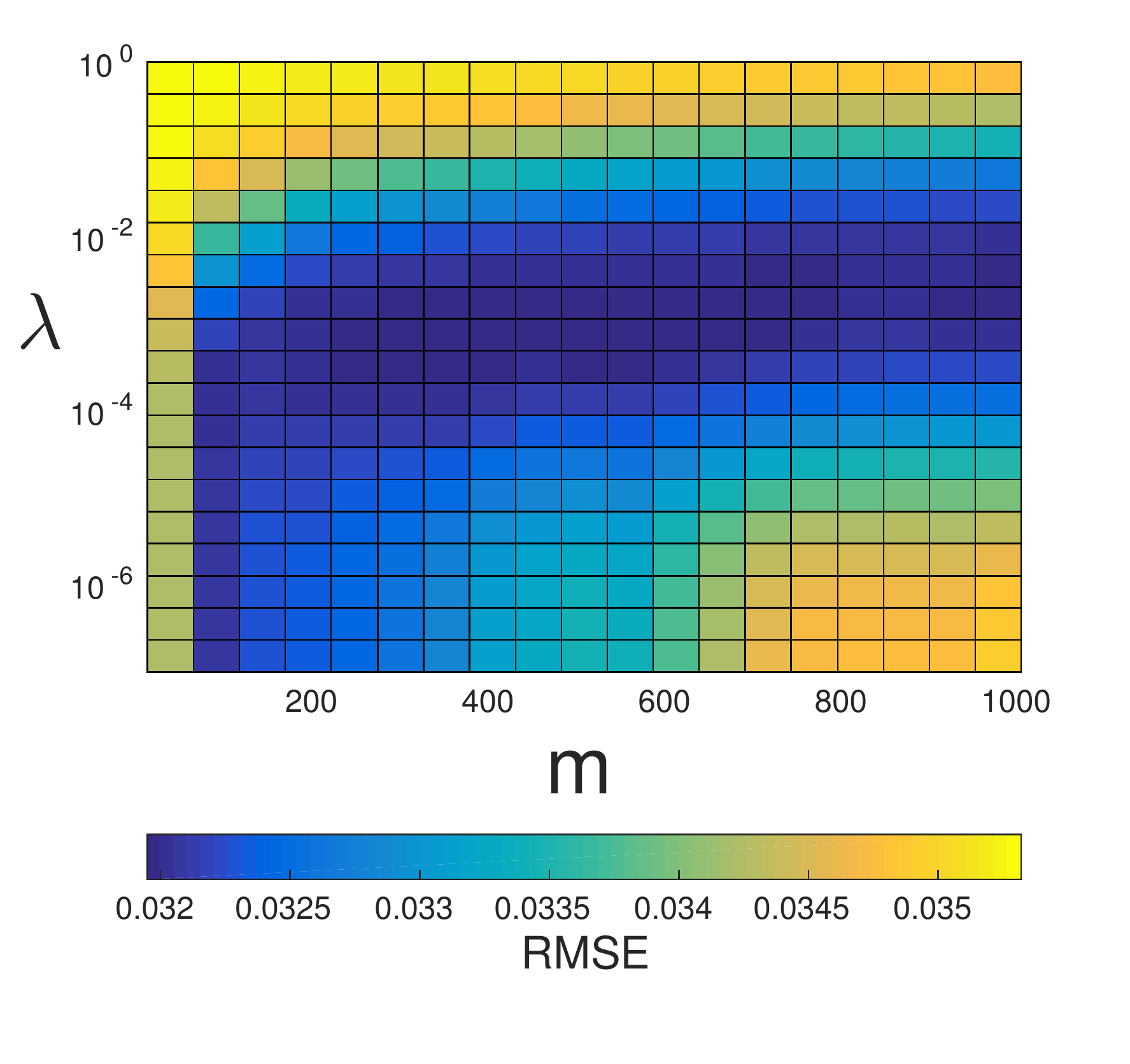}
        }
\subfigure{
                \includegraphics[trim=0cm 1cm 0cm 0.5cm, clip=true, width=0.3\textwidth]{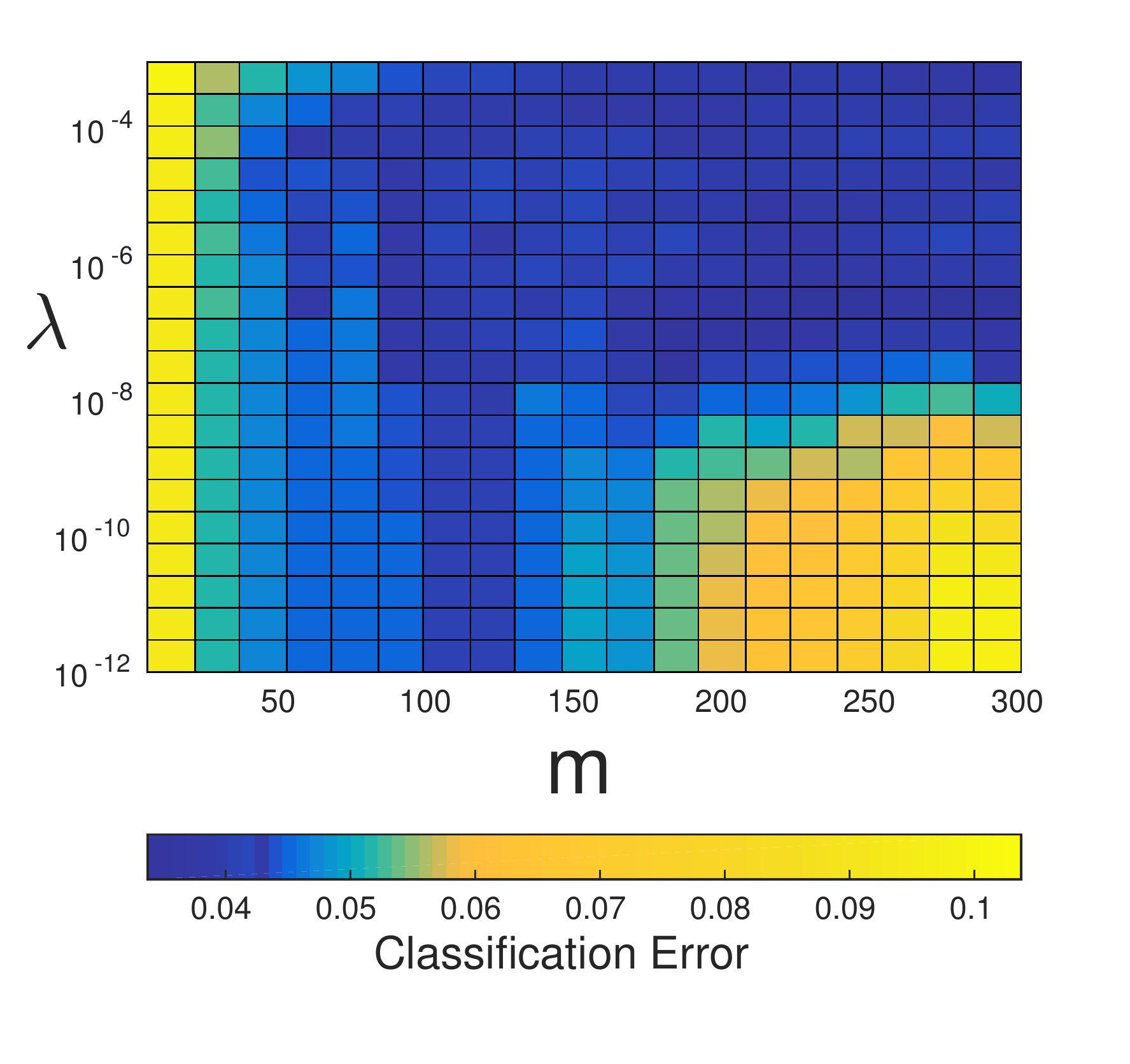}
        }
\subfigure{
                \includegraphics[trim=0cm 1cm 0cm 0.5cm, clip=true, width=0.3\textwidth]{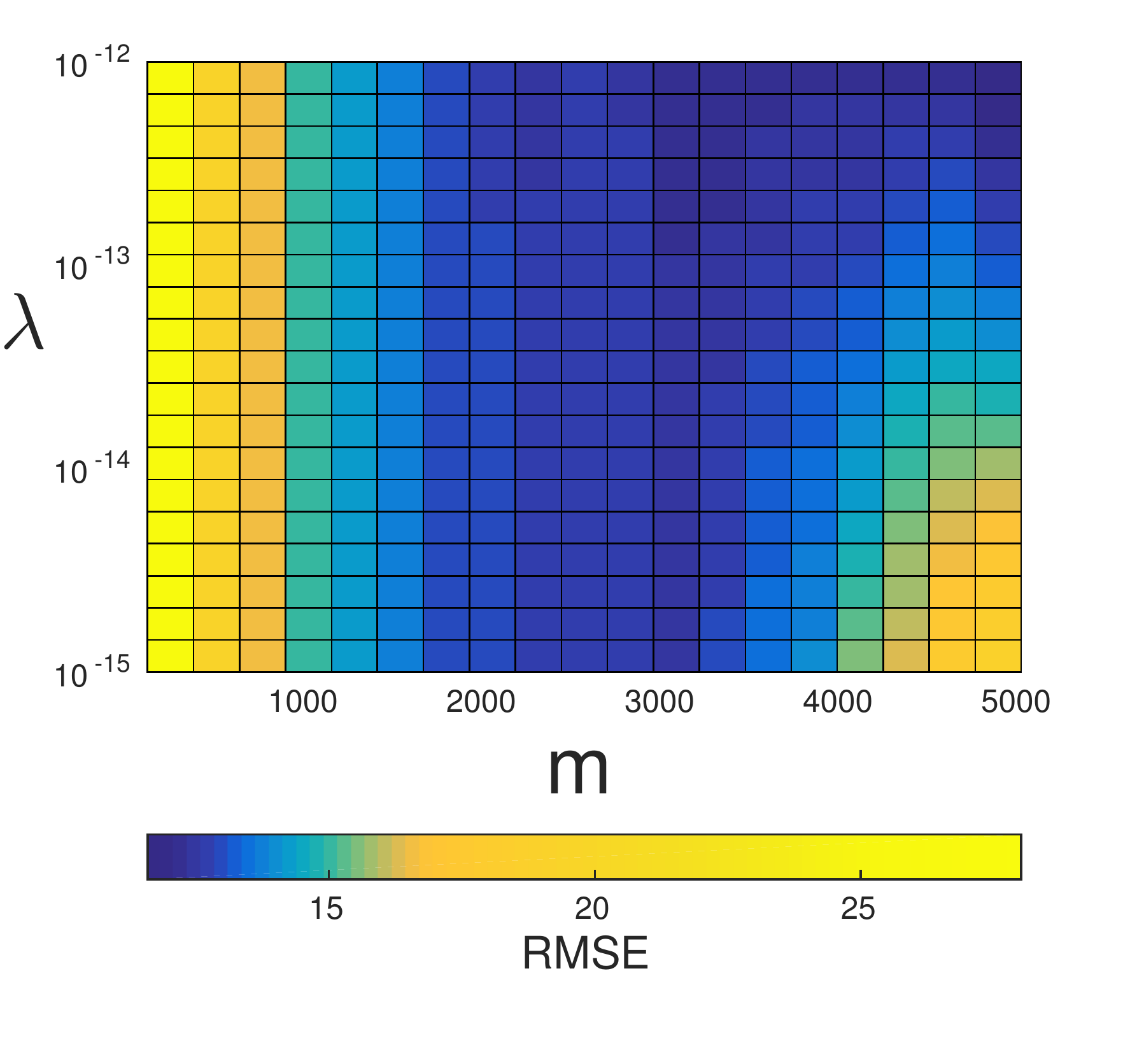}
        }
}
        \caption{Validation errors associated to $20 \times 20$ grids of values for $m$ (x axis) and $\lambda$ (y axis) on \texttt{pumadyn32nh} (left), \texttt{breast cancer} (center) and \texttt{cpuSmall} (right). \label{fig:m_lambda_plots}}
\end{figure}
\subsubsection{Proof sketch and a computational regularization perspective}
A key step in  the proof of Theorem~\ref{thm:opt-rates-NyKRLS} is an error decomposition, and corresponding bound, for any  fixed $\la$ and $m$. Indeed, it is proved in Theorem 2 of \citep{rudi2015less} and Proposition 2 of \citep{rudi2015less} that, for $\delta>0$, with probability at least $1-\delta$,
\begin{equation}\label{eq:oracle}
\begin{gathered}
\left|{\cal E}(\hat f_{\la , m}) - {\cal E}(f_{\hh})\right|^{1/2} \lesssim \\  R\left(\frac{M \sqrt{{\cal N}_\infty(\la)}}{n} + \sqrt{\frac{\sigma^2 {\cal N}(\la)}{n}}\right)\log \frac{6}{\delta} +  R {\cal C}(m)^{1/2 + v} + R\la^{1/2+v}. 
\end{gathered}
\end{equation}
The first and last term in the right hand side of the above inequality can be seen as forms of {\em sample and approximation errors} \citep{steinwart2008support} and are studied in Lemma 4 of \citep{rudi2015less} and Theorem 2 of \citep{rudi2015less}. 
The mid term can be seen as a {\em computational error} and depends on the considered subsampling scheme. 
Indeed, it is shown in Proposition 2 of \citep{rudi2015less} that ${\cal C}(m)$ can be taken as, 
$$
{\cal C}_{\rm pl}(m) = \min \left\{t > 0 ~\middle|~ (67 \vee 5 {\cal N}_\infty(t))\log\frac{12\kappa^2}{t \delta} \leq m \right\},
$$
for the plain \Nystrom{} approach, and 
$$
{\cal C}_{\rm ALS}(m) = \min \left\{\frac{19\kappa^2}{n}\log\frac{12n}{\delta} \leq t \leq \nor{C}{} ~\middle|~ 78 T^2 {\cal N}(t) \log \frac{48n}{\delta} \leq m \right\},
$$
for the approximate leverage scores approach. The bounds in Theorem~\ref{thm:opt-rates-NyKRLS} follow by: 1) minimizing in $\la$ the sum of the first and third term 2) choosing $m$ so that the computational error is of the same order of the other terms. Computational resources and regularization are then tailored to the generalization properties of the data at hand.  We add a few comments. First, note that the error bound in~\eqref{eq:oracle} holds for a large class of subsampling schemes, as discussed in Section C.1 in the appendix of \citep{rudi2015less}. 
Then specific error bounds can be derived developing computational error estimates. 
Second, the error bounds in 
Theorem 2 of \citep{rudi2015less} and Proposition 2 of \citep{rudi2015less}, and hence in Theorem~\ref{thm:opt-rates-NyKRLS}, easily generalize to a larger class of regularization schemes beyond Tikhonov approaches, namely spectral filtering \citep{bauer2007regularization}. 
For space constraints, these  extensions are deferred to a longer version of the paper. 
Third, we note that, in practice, optimal data driven parameter choices, e.g. based on hold-out estimates \citep{CapYao06}, can be used to adaptively achieve optimal learning bounds.\\
Finally, we observe that a different perspective is derived starting from inequality~\eqref{eq:oracle}, and noting that the role played by $m$ and $\la$ can also be exchanged. Letting $m$ play the role of a regularization parameter, $\la$ can be set as a function of $m$ and $m$ tuned adaptively.  For example, in the case of a plain \Nystrom{} approach, if we set 
$$\la = \frac{\log m}{m},  \quad \text{and}\quad  m = 3 n^{\frac{1}{2v + \gamma + 1}}\log n,$$
then the obtained learning solution achieves the error bound in \eqref{eq:excess-risk-bounded}\footnote{Note that in Theorem~\ref{thm:opt-rates-NyKRLS} the bound depends on $\lambda$ and $m$. By optimizing these two parameters  to minimize the upper bound we note that $m = \texttt{const}$ is sufficient for optimal bounds. Thus, we can simply write $\lambda$ as a function of $m$.}.
As above, the subsampling level can also  be chosen by cross-validation. Interestingly, in this case by tuning $m$ we naturally control computational resources and regularization.
An advantage of this latter parameterization is that, as described in the following, the solution corresponding 
to different subsampling levels is easy to update using Cholesky rank-one update formulas \citep{Golub1996}. As discussed in the next section, in practice, a joint tuning over $m$
and $\la$ can be done starting from small $m$ and appears to be advantageous both for error and computational performances. 
%
\subsection{Incremental Updates and Experimental Analysis}\label{sect:incAlgoExperiments}
In this subsection, we first describe an incremental  strategy to efficiently  explore different subsampling levels
and then perform extensive empirical tests aimed in particular at:
\begin{itemize}
\item Investigating the statistical and computational benefits of considering varying subsampling levels.
\item Compare the performance of the algorithm with respect to state of the art solutions on several large-scale benchmark datasets.
\end{itemize}
We only consider a plain \Nystrom{} approach, deferring to future work the
analysis of leverage scores based sampling techniques.
Interestingly, we will see that such a basic approach can often provide state of the art performances.
%
\begin{figure}[t]
\begin{minipage}[b]{0.5\linewidth}
\begin{algorithmic}
 \State {{\bf Input:} Dataset $(x_i, y_i)_{i=1}^n$, Subsampling $(\tilde{x}_j)_{j=1}^m$,\\ Regularization Parameter $\la$.}
 \State  {{\bf Output:} \Nystrom{} KRLS estimators $\{\tilde\alpha_1,\dots,\tilde\alpha_m\}$.}
 \State Compute $\gamma_1$; $R_1 \gets \sqrt{\gamma_1};$
 \For{$t \in \{2,\dots, m\}$}
  \State Compute $A_t, u_t, v_t$;
  \State $R_t \gets \begin{pmatrix} R_{t-1} & 0\\ 0 & 0\end{pmatrix}$;
  \State $R_t \gets {\tt cholup}(R_t, u_t, '+')$;
  \State $R_t \gets {\tt cholup}(R_t, v_t, '-')$;
  \State $\tilde{\alpha}_t \gets R_t^{-1} (R_t^{-\top} (A_t^\top Y));$
 \EndFor
\end{algorithmic}
\captionof{algorithm}{Incremental \Nystrom{} KRLS. ${\tt cholup}$ is the Cholesky rank-1 update routine.\label{alg:incr-nys-krls}}
\end{minipage}
\hfill
\begin{minipage}[b]{0.45\linewidth}
\centering
\includegraphics[width=0.9\textwidth]{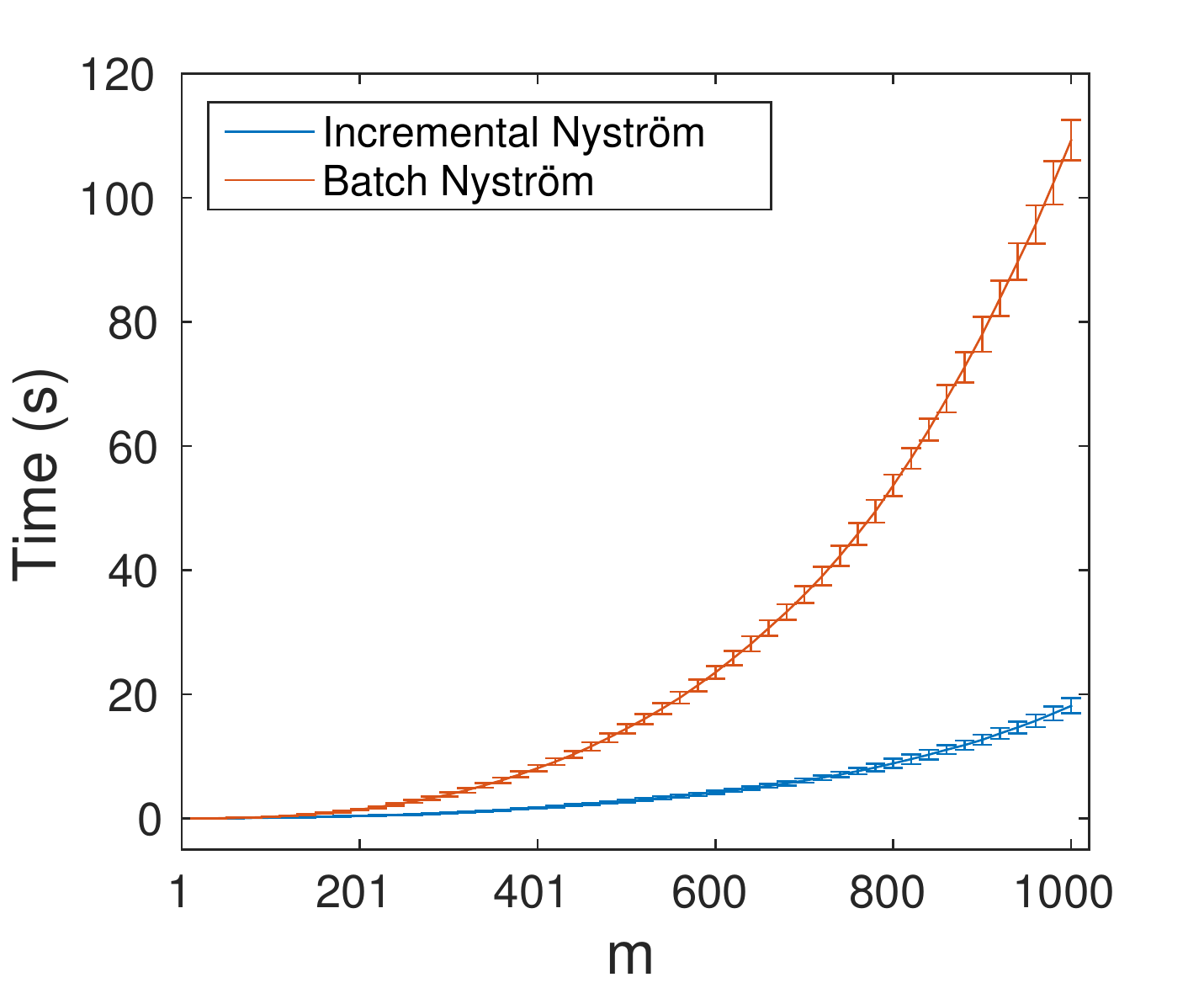}
\captionof{figure}{Model selection time on the \texttt{cpuSmall} dataset. $m \in \left[1,1000\right]$ and $T=50$, 10 repetitions.\label{fig:modSel1}}
\end{minipage}
\end{figure}
\subsubsection{Efficient incremental updates}\label{sect:eff-inc-updates}
Algorithm~\ref{alg:incr-nys-krls} efficiently computes 
solutions corresponding to different subsampling levels, by exploiting rank-one Cholesky updates \citep{Golub1996}. 
The proposed procedure allows to efficiently compute a whole regularization path of solutions, and hence perform fast model selection\footnote{The code for Algorithm~\ref{alg:incr-nys-krls} is available at \url{lcsl.github.io/NystromCoRe}.} (see Appendix~\ref{sect:algNysCompReg})\footnote{Note that Algorithm~\ref{alg:incr-nys-krls} allows to compute the regularization path incrementally in $m$ for a fixed $\lambda$. In practice, the procedure can be easily parallelized over the candidate $\lambda$ values to find the most convenient $(\lambda, m)$ combination.}.
In Algorithm~\ref{alg:incr-nys-krls}, the function {\tt cholup} is the Cholesky rank-one update formula available in many linear algebra libraries. The total cost of the algorithm is $O(nm^2 + m^3)$ time to compute $\tilde{\alpha}_2, \dots, \tilde{\alpha}_m$, while a naive non-incremental algorithm would require $O(nm^2M + m^3M)$ with $M$ is the number of analyzed subsampling levels. The following are some quantities needed by the algorithm: $A_1 = a_1$ and $A_t = (A_{t-1} \; a_t) \in \R^{n\times t}$, for any $2 \leq t \leq m$. Moreover, for any $1 \leq t \leq m$, $g_t = \sqrt{1 + \gamma_t}$ and
\eqals{
& u_t = (c_t/(1 + g_t),\,g_t),& & a_t = (K(\tilde{x}_t,x_1),\dots,K(\tilde{x}_t,x_n)),\\ 
& v_t = (c_t/(1 + g_t),\,-1),& & b_t = (K(\tilde{x}_t,\tilde{x}_1),\dots,K(\tilde{x}_t,\tilde{x}_{t-1})),\\ 
& c_t = A_{t-1}^\top a_t + \la n b_t, & & \gamma_t = a_t^\top a_t + \la n K(\tilde{x}_t,\tilde{x}_t).
}
\subsubsection{Experimental analysis}
 We empirically study the properties of Algorithm~\ref{alg:incr-nys-krls}, considering a Gaussian kernel of width $\sigma$. The selected datasets are already divided in a training and a test part\footnote{In the following we denote by $n$ the total number of points and by $d$ the number of dimensions.}. We randomly split the training part in a training set and a validation set ($80\%$ and $20\%$ of the $n$ training points, respectively) for parameter tuning via cross-validation. The $m$ subsampled points for \Nystrom{} approximation are selected uniformly at random from the training set. 
 We report the performance of the selected model on the fixed test set, repeating the process for several trials.\\
{\bf Interplay between $\la$ and $m$.} We begin with  a  set of results showing that incrementally  exploring different subsampling levels can yield 
very good performance while substantially reducing the computational requirements. We consider the
\texttt{pumadyn32nh} ($n=8192$, $d=32$), the  \texttt{breast cancer} ($n=569$, $d=30$), and   the \texttt{cpuSmall} ($n=8192$, $d=12$) datasets\footnote{\url{www.cs.toronto.edu/~delve} and \url{archive.ics.uci.edu/ml/datasets}}.  In Figure \ref{fig:m_lambda_plots}, we report the validation errors associated to a $20 \times 20$ grid of values for $\lambda$ and $m$. The $\lambda$ values are logarithmically spaced, while the $m$ values are linearly spaced. The ranges and kernel bandwidths, chosen according to preliminary tests on the data, are $\sigma = 2.66$, $\lambda \in \left[10^{-7}, 1\right]$, $m \in \left[ 10, 1000 \right]$ for \texttt{pumadyn32nh},  $\sigma = 0.9$, $\lambda \in \left[10^{-12},10^{-3}\right]$, $m \in \left[5, 300\right]$ for \texttt{breast cancer}, and $\sigma = 0.1$, $\lambda \in \left[10^{-15}, 10^{-12} \right]$, $m \in \left[100, 5000 \right]$ for \texttt{cpuSmall}. The main observation that can be derived from this first series of tests is that a small $m$ is sufficient to obtain the same results achieved with the largest $m$. 
For example, for \texttt{pumadyn32nh} it is sufficient to choose $m=62$ and $\lambda = 10^{-7}$ to obtain an average test RMSE of $0.33$ over 10 trials, which is the same as the one obtained using $m=1000$ and $\lambda = 10^{-3}$, with a 3-fold speedup of the joint training and validation phase. 
Also, it is interesting to observe that for given values of $\lambda$, large values of $m$ can decrease the performance. 
This observation is consistent with the results in Subsection~\ref{sect:main-res}, showing that $m$ can play the role of a regularization parameter.  
Similar results are obtained for  \texttt{breast cancer}, where for $\lambda=4.28 \times 10^{-6}$ and $m=300$ we obtain a $1.24\%$ average classification error on the test set over 20 trials, while for $\lambda=10^{-12}$ and $m=67$ we obtain $1.86\%$. 
For \texttt{cpuSmall}, with $m=5000$ and $\lambda=10^{-12}$ the average test RMSE over 5 trials is $12.2$, while for $m=2679$ and $\lambda=10^{-15}$ it is only slightly higher, $13.3$, but computing its associated solution requires less than half of the time and approximately half of the memory.
%
\begin{table*}[t]
\caption{Test RMSE comparison for exact and approximated kernel methods. The results for KRLS, Batch \Nystrom{}, RF and Fastfood are the ones reported in \citep{conf/icml/LeSS13}. $n_{tr}$ is the size of the training set.\label{tab:exp1.1}}
\begin{center}
\resizebox{\textwidth}{!}{%
\begin{tabular}{ccccccccccc}
\toprule 
{\em Dataset} & $n_{tr}$ & $d$ & {\em Incremental} &{\em KRLS}   & {\em Batch } & {\em RF} & {\em Fastfood} & {\em Fastfood} & {\em KRLS} & {\em Fastfood} \\ 
&  &  & {\em \Nystrom{} RBF} & {\em RBF}  & {\em \Nystrom{} RBF} & {\em RBF} & {\em RBF} & {\em FFT} & {\em Matern} & {\em Matern}  \\ 
\midrule
Insurance Company & 5822 & 85 & $0.23180\pm 4 \times 10^{-5}$ & \textbf{0.231} & 0.232 & 0.266 & 0.264 & 0.266 & 0.234 & 0.235\\ 
CPU & 6554 & 21 & \textbf{$\mathbf{2.8466 \pm 0.0497}$} & 7.271 & 6.758 & 7.103 & 7.366 & 4.544 & 4.345 & 4.211\\  
CT slices (axial) & 42800 & 384 & $\mathbf{7.1106 \pm 0.0772}$ & NA & 60.683 & 49.491 & 43.858 & 58.425 & NA & 14.868\\ 
Year Prediction MSD & 463715 & 90 & $\mathbf{0.10470 \pm 5 \times 10^{-5}}$ & NA & 0.113 & 0.123 & 0.115 & 0.106 & NA & 0.116\\ 
Forest & 522910 & 54 & $0.9638 \pm 0.0186$ & NA & \textbf{0.837} & 0.840 & 0.840 & 0.838 & NA & 0.976\\ 
\bottomrule
\hline 
\end{tabular} 
}
\end{center}
\end{table*}
\\{\bf Regularization path computation}. If the subsampling level $m$ is used as a regularization parameter, the computation of a regularization path corresponding to different subsampling levels becomes crucial during the model selection phase. A naive approach, that consists in recomputing the solutions of \eqref{eq:repny} for each subsampling level, 
 would require $O(m^2nT + m^3LT)$ computational time, where $T$ is the number of solutions with different subsampling levels to be evaluated and $L$ is the number of Tikhonov regularization parameters.
  On the other hand, by using the incremental \Nystrom{} algorithm the model selection time complexity is $O(m^2n + m^3L)$ for the whole regularization path. We experimentally verify this speedup on \texttt{cpuSmall} with 10 repetitions, setting $m \in \left[1,5000\right]$ and $T=50$. The model selection times, measured on a server with 12 $\times$ 2.10GHz Intel$^\circledR$ Xeon$^\circledR$ E5-2620 v2  CPUs and 132 GB of RAM, are reported in Figure \ref{fig:modSel1}. The result clearly confirms the beneficial effects of incremental \Nystrom{} model selection on the computational time.\\
{\bf Predictive performance comparison}.
Finally,  we consider  the performance of the algorithm on several large scale benchmark datasets considered in \citep{conf/icml/LeSS13}, see Table \ref{tab:exp1.1}. $\sigma$ has been chosen on the basis of preliminary data analysis. $m$ and $\la$ have been chosen by cross-validation, starting from small subsampling values up to $m_{max}=2048$, and considering $\la \in \left[ 10^{-12} , 1\right]$. After model selection, we retrain the best model on the entire training set and compute the RMSE on the test set. We consider 10 trials, reporting the performance mean and standard deviation. The results in Table \ref{tab:exp1.1} compare \Nystrom{} computational regularization with the following methods (as in \citep{conf/icml/LeSS13}):
\begin{itemize}
\item \textbf{Kernel Regularized Least Squares (KRLS):} Not compatible with large datasets.
\item \textbf{Random Fourier features (RF):} As in \citep{conf/nips/RahimiR07}, with a number of random features $D=2048$.
\item \textbf{Fastfood RBF, FFT and Matern kernel:} As in \citep{conf/icml/LeSS13}, with $D=2048$ random features.
\item \textbf{Batch \Nystrom{}:} \Nystrom{} method \citep{conf/nips/WilliamsS00} with uniform sampling and $m=2048$.
\end{itemize}
The above results show that the proposed incremental \Nystrom{} approach behaves really well, matching state of the art predictive performances.
	\section{NYTRO: Nystr\"om Iterative Regularization}
	\label{sec:nytro}
	\subsection{Setting}
A key feature towards scalability is being able to  tailor computational requirements to the generalization properties/statistical accuracy  allowed by the data.  In other words,  the precision with  which computations need to be performed  should be  determined  not only by the the amount, but also by the  quality  of the available data.

Early stopping (see Section \ref{sec:itReg}),  known as iterative regularization in inverse problem theory  \citep{engl1996regularization,zhang2005boosting, bauer2007regularization,earlyStopping,CapYao06}, provides a simple and sound implementation of this intuition. 
An empirical objective function is optimized in an iterative way with no explicit constraint or penalization and regularization is achieved  by suitably stopping the iteration. 
Too many iterations might lead to overfitting, while stopping too early might result in oversmoothing \citep{zhang2005boosting, bauer2007regularization,earlyStopping,CapYao06}. Then, the best stopping rule arises from a form of bias-variance trade-off \citep{hastie2001elements}. 
The key observation is that the number of iterations controls at the same time the computational complexity as well as the statistical properties of the obtained learning algorithm \citep{earlyStopping}. 
Training and model selection can hence be performed with often considerable gain in time complexity.

Despite these nice properties, early stopping procedures often share the same space complexity requirements,
hence bottle necks, of other methods, such as those based on Tikhonov regularization (see \citep[][]{TikhonovOriginal,RidgeRegression}).
As seen in previous sections, a natural way to tackle these issues is to consider randomized subsampling approaches.
In this Section, we ask whether early stopping and subsampling methods can be fruitfully combined. With the context of kernel methods in mind,  we propose and study NYTRO (NYstr\"om iTerative RegularizatiOn),  a simple algorithm combining these two ideas. After recalling  the properties and advantages of different regularization approaches in Subsection~\ref{sect:setting}, in Subsection~\ref{sect:proposedAlgorithm} we present in detail NYTRO and our main result, the characterization of its generalization properties. 
In particular, we analyze the conditions under which it attains the same statistical properties of subsampling and early stopping. Indeed, our study shows that while both techniques share similar, optimal, statistical properties, they are computationally advantageous in different regimes and NYTRO outperforms early stopping in the appropriate regime, as discussed in Subsection~\ref{sect:discussion}.
The theoretical results are validated empirically in Subsection~\ref{sect:experiments}, where NYTRO is shown to provide competitive results even at a fraction of the computational time, on a variety of benchmark datasets.

\subsection{Learning and Regularization}\label{sect:setting}
In this section we introduce the problem of learning in the fixed design setting and discuss different regularized learning 
approaches, comparing their statistical and computational properties. This section is  a survey that might be interesting in its own right, 
and reviews several results providing the context for the study in the paper.

\subsubsection{The Learning Problem}\label{sect:fix-setting}
We introduce the  learning setting we consider in this section.
Let $\X = \R^d$ be the input space and $\Y \subseteq \R$  the output space. Consider  a {\em fixed design} setting \citep{conf/colt/Bach13}, as introduced in Section \ref{sec:probDataModel}, where the input points $x_1, \dots, x_n \in \X$ are fixed, while the outputs $y_1,\dots,y_n \in \Y$ are  given  by
$$ y_i = f_*(x_i) + \epsilon_i, \quad \forall \; i \in \{1,\dots, n\}$$
where $f_* : \X \to \Y$ is a fixed function and $\epsilon_1,\dots,\epsilon_n$ are random variables. The latter can be seen  seen as noise and are  assumed to be independently and identically distributed according to a probability distribution  $\rho$ with  zero mean and variance $\sigma^2$. 
In this context, the goal is  to minimize the {\em expected risk}, that is 
\eqal{\label{eq:ideal-problem}
 &\min_{f \in \hh} {\cal E}(f), ~~
 {\cal E}(f) = {\mathbb E} \frac{1}{n}\sum_{i=1}^n \left(f(x_i) - y_i\right)^2, ~~ \forall f \in \hh,
}
on a hypotheses space $\hh$.
In a real applications, $\rho$ and $f_*$ are unknown and  accessible only by means of a single realization $(x_1, y_1), \dots, (x_n, y_n)$ called {\em training set} and an approximate solution needs to be found.
The quality of a solution $f$ is measured by the {\em excess risk}, defined as
$$
R(f) = {\cal E}(f) - \inf_{v \in \hh} {\cal E}(v), \quad \forall f \in \hh.
$$
We next discuss estimation schemes to find a solution and compare their computational and statistical properties.

\subsubsection{From (Kernel) Ordinary Least Square to Tikhonov Regularization}

 A classical approach to derive an empirical solution to Problem~\eqref{eq:ideal-problem} is ERM, as seen in Section \ref{sec:ERM},
 \eqal{\label{eq:ols-problem}
f_\textrm{ols} = \argmin_{f \in \hh} \frac{1}{n} \sum_{i=1}^n \left(f(x_i) - y_i\right)^2. 
}
Here, we are interested in the case where  $\hh$ is the RKHS 
 induced by a positive definite kernel $K:\X \times \X \to \R$.
In this case Problem~\eqref{eq:ols-problem} corresponds to Kernel Ordinary Least Squares (KOLS --- see Section \ref{sec:spectralIntro}) 
and  has the closed form solution
\eqal{\label{eq:ols}
f_{\textrm{ols}}(x) & = \sum_{i=1}^n \alpha_{\textrm{ols}, i} K(x,x_i),\quad \alpha_{\textrm{ols}} =  K_n^{\dag} y,
}
for all $x \in \X$, where $(K_n)^\dag$ denotes the pseudo-inverse of the  $ \in \R^{n\times n}$  empirical kernel matrix $[K_n]_{ij} = K(x_i, x_j)$ and $y = (y_1, \dots, y_n)$.
The cost for computing the coefficients $\alpha_{\textrm{ols}}$ is $O(n^2)$ in memory and $O(n^3 + q(\X) n^2)$ in time, where $q(\X) n^2$ is the cost for computing $K_n$ and $n^3$ the cost for obtaining its pseudo-inverse.
Here, $q(\X)$ is the cost of evaluating the kernel function. In the following,  we are concerned with the dependence on $n$ and hence view $q(\X)$
as a constant. 


The statistical properties of KOLS, and related methods,   can be characterized by suitable notions of {\em dimension}
that we recall next. 
 The simplest is  the {\em full} dimension, that is
 $$ d^* = \textrm{rank}\, K_n,$$
which measures the degrees of freedom of the kernel matrix. This latter quantity might not be stable when 
$K_n$ is ill-conditioned. A more robust notion is provided by the {\em effective dimension}
$$ d_\textrm{eff}(\la) = \text{Tr}(K_n (K_n + \la n I_n)^{-1}),\quad \la > 0.$$
Indeed, the above quantity can be shown to be related 
to  the eigenvalue decay of $K_n$ \citep{conf/colt/Bach13,alaoui2014fast} and can be considerably smaller than $d^*$, as discussed in the following.
Finally,  consider 
\eqal{\label{eq:tilde-d}
\tilde{d}(\la) = n \max_i (K_n(K_n+\la n I_n)^{-1})_{ii},\quad \la > 0.
}
It is easy to see that the following inequalities hold,
$$ d_\textrm{eff}(\la) \leq \tilde{d}(\la)\leq 1/\la,\quad d_\textrm{eff}(\la) \leq d^*\leq n,\qquad \forall \la > 0.$$ 
Aside from the above notion of dimensionality, the statistical accuracy of empirical least squares solutions
depends on a natural form of signal to noise ratio defined next.
Note that  the function that minimizes the excess risk in $\hh$ is given by 
\eqals{
f_\textrm{opt} &= \sum_{i=1}^n \alpha_{\textrm{opt},i} K(x, x_i),\quad \forall x \in \X\\
\alpha_\textrm{opt} & =  K_n^\dag \mu, \quad \textrm{with}\;\; \mu = {\mathbb E} y.
}
Then, the signal to noise ratio is defined as 
\eqal{\label{eq:snr}
\SNR = \frac{\nor{f_\textrm{opt}}^2_\hh}{\sigma^2}.
}
Provided with the above definitions, we can recall a first basic results characterizing the statistical accuracy of KOLS.
\bt\label{thm:ols}
Under the assumptions of Section~\ref{sect:fix-setting}, the following equation holds,
$$ {\mathbb E} R(f_\textrm{ols}) = \frac{\sigma^2 d^*}{n}.$$
\et
The above result shows that  the excess risk of KOLS can be bounded in terms of the full dimension, the noise level and the number of points. 
However, in general ERM {\em does not} provide the best results and regularization is needed. 
We next recall this fact,  considering first Tikhonov regularization, that is the 
%
KRLS algorithm (see Section \ref{sec:krr}) given by
\eqal{\label{eq:rls-problem}
\bar{f}_\la = \argmin_{f \in \hh} \frac{1}{n} \sum_{i=1}^n \left(f(x_i) - y_i\right)^2 + \la\nor{f}^2_\hh. 
}
We recall that the  solution of Problem~\eqref{eq:rls-problem} is 
\eqal{\label{eq:std-krls}
\bar{f}_\la(x) & = \sum_{i=1}^n \bar{\alpha}_{\la i} K(x,x_i),\quad \bar{\alpha}_\lambda =  (K_n + \la n I_n)^{-1} Y,
}
for all $x \in \X$. The intuition that regularization can be beneficial is made precise by the following result comparing
KOLS and KRLS.

\bt\label{thm:rls}
Let $\la^* = \frac{1}{SNR}$. The following inequalities hold, 
$$ {\mathbb E} R(\bar{f}_{\la^*}) \leq \frac{\sigma^2 d_\textrm{eff}(\la^*)}{n} \leq  \frac{\sigma^2 d^*}{n} = {\mathbb E} R(f_\textrm{ols}).$$
\et
We add a few comments. First, as announced, the above result quantifies the benefits of regularization.  Indeed,  it shows that there exists a $\la^*$ for which the expected excess risk of KRLS is smaller than the one of KOLS. As discussed  in Table~1 of \citep{conf/colt/Bach13}, if $d^* = n$ and the kernel is sufficiently ``rich'', namely universal \citep{micchelli2006universal},  then  
 $d_\textrm{eff}$ can be less than a fractional power of $d^*$, so that  $d_\textrm{eff} \ll d^*$ and 
 $$ {\mathbb E} R(\bar{f}_{\la^*}) \;\ll\; {\mathbb E} R(f_\textrm{ols}).$$ 
 Second, note that the choice of the regularization parameter depends on a  form of signal to noise ratio, which is usually unknown. In practice, a regularization path\footnote{The set of  solutions  corresponding to regularization parameters  in a discrete set $\Lambda \subset \R$.}  is computed and then a model selected or found by aggregation \citep{hastie2001elements}. Assuming the selection/aggregation
step to have negligible computational cost, the complexity of performing training {\em and } model selection is then
$O(n^2)$ in memory and $O\left(n^3|\Lambda|\right)$ in time. 
These latter requirements can become prohibitive when $n$ is large and the question is whether 
the same statistical accuracy of KRLS can be achieved while reducing time/memory requirements.

\subsubsection{Early Stopping and \Nystrom{} Methods}\label{sect:from-tikh-to-early}
In this section, we first recall how early stopping regularization allows to achieve the same statistical 
accuracy of KRLS  with potential saving in time complexity. 
Then, we recall how subsampling ideas can be used  in the framework of Tikhonov regularization to reduce 
 the space complexity with no loss of statistical accuracy.


\paragraph{Iterative Regularization by Early Stopping} The idea is to consider the gradient descent minimization 
of Problem~\eqref{eq:ols} for a fixed number of steps $t$, as seen in Section \ref{sec:itReg}.
We now introduce the notation used in this section for early stopping.
The  algorithm is
\eqal{ \label{eq:iterative-reg}
\breve{f}_t(x) & = \sum_{i=1}^n \breve{\alpha}_{t,i} K(x_i, x),\\
\breve{\alpha}_t &= \breve{\alpha}_{t-1} - \frac{\gamma}{n} (K_n \breve{\alpha}_{t-1} - Y),
}
where $\gamma < 1/\nor{K_n}$ and $\breve{\alpha}_0 = 0$. 
The only tuning parameter is the number of steps, which, as shown next,  controls at the same time the computational complexity and statistical accuracy of the algorithm. 
The following theorem compares the expected excess risk of early stopping with the one of KRLS.

\bt\label{thm:land-wrt-krls}
When $\gamma < 1/\nor{K_n}$ and $t \geq 2$ the following holds
$$
{\mathbb E} R\left(\breve{f}_{\gamma,t}\right) \leq c_t \; {\mathbb E} R\left(\bar{f}_{\frac{1}{\gamma t}}\right).  
$$
with $c_t = 4\left(1+\frac{1}{t-1}\right)^2 \leq 16$.
\et
The above theorem follows as a corollary of our main result given in Theorem~\ref{thm:main} and recovers
results essentially given in \citep{raskutti2014early}. 
Combining the above result  with Theorem~\ref{thm:rls}, and setting $t^*=\frac{1}{\gamma \la^*} = \frac{\SNR}{\gamma}$, we have that 
$${\mathbb E} R\left(\breve{f}_{\gamma,t^*}\right) \approx {\mathbb E} R\left(\bar{f}_{\la^*}\right) \leq {\mathbb E} R\left(f_\textrm{ols}\right).
$$
The statistical accuracy of early stopping is essentially the same as KRLS and can be vastly better than a na\"{i}ve ERM approach. Note that the cost of computing the best possible solution with early stopping 
is $O(n^2t^*) = O(n^2 \SNR)$. Thus, the computational time of early stopping is proportional to the signal to noise ratio. Hence, it could be much better than  KRLS for noisy problems, that is when $\SNR$ is small. The main bottle neck of early stopping regularization is that it has the same space requirements of KRLS. Subsampling approaches have been proposed to tackle this issue.
\paragraph{Subsampling and Regularization} 
Recall that  the solution of the standard KRLS problem belongs to $\hh_n$, as in \eqref{eq:hhn}.
We know from Subsection \ref{sect:krls-nyst} that the basic idea of  \Nystrom{} KRLS (NKRLS) is to restrict Problem~\eqref{eq:rls-problem} to  a subspace $\hh_m \subseteq \hh_n$ defined as
\eqal{\label{eq:Hm}
\hh_m = \{\sum_{i=1}^m c_i K(\cdot,\tilde{x}_i) | c_1,\dots,c_m \in \R \},
}
where $M = \{\tilde{x}_1, \dots, \tilde{x}_m\}$ is a subset of the training set and  $m\le n$.
It is easy to see that the corresponding solution is given by 
\eqal{\label{eq:nys-sol}
\tilde{f}_{m,\la}(x) &= \sum_{i=1}^m (\tilde\alpha_{m,\la})_i K(x,\tilde{x}_i), \\
\tilde{\alpha}_{m,\la} &= (K_{nm}^\top K_{nm} + \la n K_{mm})^\dag K_{nm}^\top Y,
} 
for all $x \in \X$, where $(\cdot)^\dag$ is the pseudoinverse, $\la > 0$, $K_{nm} \in \R^{n\times m}$ with $(K_{nm})_{ij} = K(x_i, \tilde{x}_j)$ and $K_{mm} \in \R^{m\times m}$ with $(K_{mm})_{i,j} = K(\tilde{x}_i, \tilde{x}_j)$. 
A more efficient formulation can also be derived. Indeed, we 
rewrite Problem~\eqref{eq:rls-problem}, restricted to $\hh_m$, as 
\eqal{
\tilde{\alpha}_{m,\la} & = \argmin_{\alpha \in \R^m} \nor{K_{nm}\alpha - Y}^2 + \la \alpha^\top K_{mm} \alpha\\
\label{eq:new-nys-problem}& = R \argmin_{\beta \in \R^k} \nor{K_{nm}R\beta - Y}^2 + \la \nor{\beta}^2,
}
where in the last step we performed the change of variable $\alpha = R\beta$ where $R \in \R^{m\times k}$ is a matrix such that $R R^\top = K_{mm}^\dag$ and $k$ is the rank of $K_{mm}$. Then, we can obtain the following closed form expression,
\eqal{\label{eq:char-nyst}
\tilde{\alpha}_{m,\la} = R(A^\top A + \la n I_n)^{-1} A^\top Y.
}
(see Proposition 2 in Section A of the appendix of \citep{camoriano2016nytro} for a complete proof).
This last formulation is convenient because  it is possible to compute $R$ by $R = S T^{-1}$ where $K_{mm} = S D$ is the economic QR decomposition of $K_{mm}$, with $S \in \R^{m \times k}$ such that $S^\top S = I_k$, $D \in \R^{k\times m}$ an upper triangular matrix and $T \in \R^{k \times k}$ an invertible triangular matrix that is the Cholesky decomposition of $S^\top K_{mm} S$. Assuming $k \approx m$, the complexity of NKRLS is then $O(nm)$ in space  and $O(nm^2 + m^3|\Lambda|)$ in time. The following known result establishes the statistical accuracy of the solution obtained by suitably choosing the points in $M$.
%
\bt[Theorem 1 of \citep{conf/colt/Bach13}]\label{thm:nys}
Let $m \leq n$ and $M = \{\tilde{x}_1,\dots, \tilde{x}_m\}$ be a subset of the training set uniformly chosen at random. Let $\tilde{f}_{m,\la}$ be as in \eqref{eq:nys-sol} and $\bar{f}_{\la}$ as in \eqref{eq:std-krls} for any $\la > 0$. Let $\delta \in (0,1)$, when
$$m \geq \left(\frac{32 \tilde{d}(\la)}{\delta} + 2\right)\log \frac{\nor{K_n} n}{\delta \la}$$
with $\tilde{d}(\la) = n \sup_{1\leq i \leq n} (K_n(K_n+\la n I_n)^{-1})_{ii}$, then the following holds
$${\mathbb E}_M {\mathbb E} R\left(\tilde{f}_{m,\la}\right) \leq (1+4\delta) {\mathbb E} R\left(\bar{f}_{\la}\right).  
$$
\et
The above result shows that the space/time complexity of   NKRLS can be  adaptive to the statistical properties of the data
while preserving the same statistical accuracy of KRLS. Indeed, using  Theorem~\ref{thm:rls}, we have that
$${\mathbb E}_M {\mathbb E} R\left(\tilde{f}_{m,\la^*}\right) \approx {\mathbb E} R\left(\bar{f}_{\la^*}\right) \leq {\mathbb E} R\left(f_\textrm{ols}\right),$$
requiring $O(n \tilde{d}(\la^*)\log\frac{n}{\la^*})$  memory and $O(n \tilde{d}(\la^*)^2(\log\frac{n}{\la^*})^2)$  time.
Hence, NKRLS is more efficient with respect to KRLS when $\tilde{d}(\la^*)$ is smaller than $\frac{n}{\log\frac{n}{\la^*}}$, that is when the problem is mildly complex.

Given the above discussion, it is natural to ask whether subsampling and early stopping ideas can be fruitfully 
combined.
Providing a positive answer to this question is the main contribution of this section, and we discuss it next. 

\subsection{Proposed Algorithm and Main Results}\label{sect:proposedAlgorithm}
We begin by describing the proposed algorithm incorporating  the \Nystrom{} approach described above
in  iterative regularization by early stopping.  The intuition is that the algorithm thus obtained could have memory and time complexity adapted to the statistical accuracy allowed by the data, while automatically computing the whole regularization path. Indeed, this intuition is then confirmed through a statistical analysis of the corresponding excess risk. Our result indicates in which regimes
 KRLS, NKRLS, Early Stopping and NYTRO are  preferable.

\subsubsection{The Algorithm}

 NYTRO is obtained considering a finite number of iterations of the  gradient descent minimization of the empirical 
 risk in Problem~\eqref{eq:ols-problem} over the space in \eqref{eq:Hm}.
 The algorithm thus obtained is given by, 
\eqal{\label{eq:nytro-sol}
 \hat{f}_{m,\gamma,t}(x) &= \sum_{i=1}^m (\hat{\alpha}_{m,\gamma,t})_i K(\tilde{x}_i, x), \\
 \hat{\beta}_{m,\gamma,t} &= \hat{\beta}_{m,\gamma,t-1} - \frac{\gamma}{n} R^\top (K_{nm}^\top(K_{nm}\hat{\beta}_{m,\gamma,t-1} - Y)), \\
 \hat{\alpha}_{m,\gamma,t} &= R\beta_{m,\gamma,t},
} 
for all $x \in \X$, where $\gamma = 1/(\sup_{1\leq i \leq n} K(x_i,x_i))$ and $\hat{\beta}_{m,0} = 0$.
Considering that the cost of computing $R$ is $O(m^3)$, the total cost for the above algorithm is $O(nm)$ in memory and $O(nmt + m^3)$ in time.

In the previous section, we have seen that NKRLS has an accuracy comparable to the one of the standard KRLS under a suitable choice of $m$. We next show that, under the same conditions, the accuracy of NYTRO is comparable with the ones of KRLS and NKRLS, for suitable choices of $t$ and $m$.

\begin{figure*}[t]
\centering
\includegraphics[width=1\linewidth]{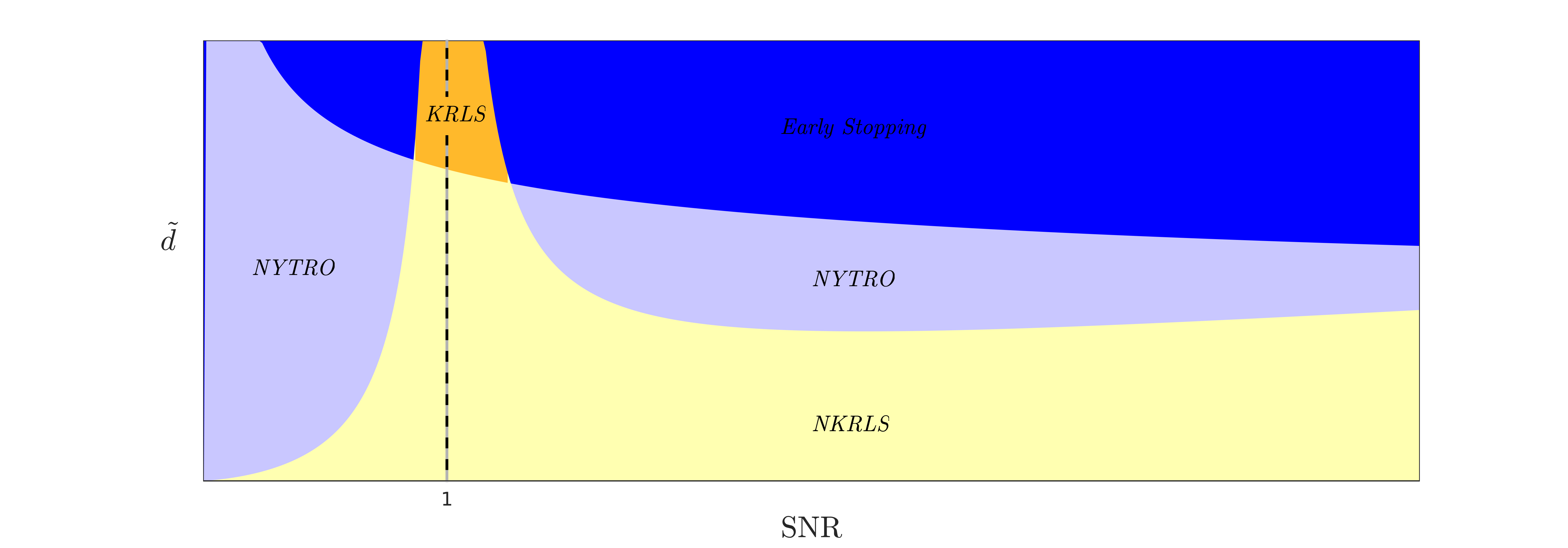}
\caption{The graph represents the family of learning problems parametrized by the dimensionality $\tilde{d}$ and the signal-to-noise ratio $\SNR$ (see \eqref{eq:tilde-d},~\eqref{eq:snr}). The four different regions represent the regimes where some algorithm is faster than the others. Purple: NYTRO is faster, Blue: Early Stopping is faster, Orange: KRLS is faster, Yellow: NKRLS is faster -- see Section~\ref{sect:discussion}.}
\label{fig:areaPlotv2_WIDE}
\end{figure*}

\subsubsection{Error Analysis}
We next establish excess risk bounds for NYTRO by providing a direct comparison  
with  NKRLS and KRLS.
\bt[NYTRO and NKRLS] \label{thm:main}
Let $m \leq n$ and $M$ be a subset of the training set. Let $\hat{f}_{m,\gamma,t}$ be the NYTRO solution as in \eqref{eq:nytro-sol}, $\tilde{f}_{m,\frac{1}{\gamma t}}$ the NKRLS solution as in \eqref{eq:nys-sol}. When $t \geq 2$ and $\gamma < \nor{K_{nm} R}^2$ (for example $\gamma = 1/\max_i K(x_i,x_i)$) the following holds
$$
{\mathbb E} R\left(\hat{f}_{m,\gamma,t}\right) \leq c_t\; {\mathbb E} R\left(\tilde{f}_{m,\frac{1}{\gamma t}}\right).  
$$
with $c_t = 4\left(1+\frac{1}{t-1}\right)^2 \leq 16$.
\et
Note that the above result 
holds for any $m \leq n$ and any selection strategy of the \Nystrom{} subset $M$. 
The proof of Theorem~\ref{thm:main} is different from the one of Theorem~\ref{thm:nys} and is based 
only on geometric properties of the estimator and tools from spectral theory and inverse problems see~\citep[][]{engl1996regularization}. 
In the next corollary we compare  NYTRO and KRLS, by combining Theorems~\ref{thm:nys} and ~\ref{thm:main}, hence considering $M$ to be chosen uniformly at random from the training set.
\bcor\label{cor:uniform}
Let $t \geq 2$, $\gamma = 1/\nor{K_n}$, $\delta \in (0,1)$ and $m$ be chosen as 
$$m \geq \left(32 \frac{\tilde{d}(1/(\gamma t))}{\delta} + 2\right)\log \frac{n\nor{K_n} \gamma t}{\delta}.$$
Let $\bar{f}_{\frac{1}{\gamma t}}$ be the KRLS solution as in \eqref{eq:std-krls} and $\hat{f}_{m,\gamma,t}$ be the NYTRO solution. When the subset $M$ is chosen uniformly at random from the training set, the following holds
$$ {\mathbb E}_M {\mathbb E} R\left(\hat{f}_{m,\gamma,t}\right) \leq c_{t,\delta}\; {\mathbb E} R\left(\bar{f}_{\frac{1}{\gamma t}}\right)
$$
where $c_{t,\delta} = 4\left(1+\frac{1}{t-1}\right)^2(1 + 4 \delta) \leq 80$.
\ecor
The above result shows that NYTRO can achieve essentially the same results as KRLS. 
In the following we compare NYTRO to the other regularization algorithms introduced so far, by discussing how their computational complexity adapts to the statical accuracy in the data. In particular, by parametrizing the learning problems with respect to their dimension and their signal-to-noise ratio, we characterize the regions of the problem space where one algorithm is more efficient than the others.

\subsubsection{Discussion}\label{sect:discussion}
In Subsection~\ref{sect:setting} we have compared the expected excess risk of different regularization algorithms. More precisely, we have seen that there exists a suitable choice of $\la$ that is $\la^* = \frac{1}{\SNR}$, where $\SNR$ is the signal-to-noise ratio associated to the learning problem,  such that the expected risk of KRLS is smaller than the one of KOLS, and indeed potentially much smaller.
For this reason, in the other result, statistical accuracy of the other methods was directly compared to that of 
 KRLS with $\la=\la^*$.

We exploit these results to analyze the complexity of the algorithms with respect to the statistical accuracy allowed by the data. If we choose $m \approx \tilde{d}(\la^*)\log (n/\la^*)$ and $t = \frac{1}{\gamma \la^*}$, then combining Theorem~\ref{thm:rls}   with Corollary~\ref{cor:uniform} and with Theorem~\ref{thm:nys}, respectively, we see that 
the expected excess risk of both NYTRO and NKRLS is in the same order of the one of KRLS. Both algorithms have a memory requirement of $O(nm)$ (compared to $O(n^2)$ for KRLS), but they differ in their time requirement. 
For NYTRO we have $O(n\frac{\tilde{d}(\la^*)}{\la^*}\log\frac{n}{\la^*})$, while for NKRLS it is $O(n\tilde{d}(\la^*)^2(\log \frac{n}{\la^*})^2)$. Now note that $\tilde{d}(\la^*)$ by definition is bounded by
$$ d_\textrm{eff}(\la) \leq \tilde{d}(\la) \leq \frac{1}{\la},\quad \forall \la > 0,$$
thus, by comparing the two computational times, two regimes can be identified,
\eqals{
\begin{cases}
d_\textrm{eff}(\la^*) \leq\; \tilde{d}(\la^*)\; \leq \frac{1}{\la^* \log \frac{n}{\la^*}}  & \implies \textrm{NKRLS faster}\\
\frac{1}{\la^* \log\frac{n}{\la^*}} \leq \tilde{d}(\la^*) \leq \frac{1}{\la^*}  & \implies \textrm{NYTRO faster}
\end{cases}
} 
To illustrate  the regimes in which different algorithms can be preferable from a computational point of view while achieving the same error as KRLS with $\la^*$ (see Figure~\ref{fig:areaPlotv2_WIDE}), it is useful to 
parametrize the family of learning problems with respect to the signal-to-noise ratio defined in \eqref{eq:snr} and to the dimensionality of the problem $\tilde{d} := \tilde{d}(\la^*)$ defined in \eqref{eq:tilde-d}. We choose $\tilde{d}$ as a measure of dimensionality with respect to $d_\textrm{eff}$, because $\tilde{d}$ directly affects the computational properties of the analyzed algorithms.
In Figure~\ref{fig:areaPlotv2_WIDE},
the parameter space describing the learning problems is 
 partitioned in regions given by the curve 
$$c_1(\SNR) = \frac{n}{|\log(n \SNR)|},$$
that separates the subsampling methods from the standard methods and
$$c_2(\SNR) = \frac{\SNR}{|\log(\SNR)|},$$ that separates the iterative from Tikhonov methods.

\begin{table}
\caption{Specifications of the datasets used in time-accuracy comparison experiments. $\sigma$ is the bandwidth of the Gaussian kernel.}
\begin{center}
\resizebox{0.6\textwidth}{!}{%
\begin{tabular}{ccccc}
\toprule 
{\em Dataset} & $n$ &  $n_{test}$ & $d$ & $\sigma$\\
\midrule
{\em InsuranceCompany} & 5822 & 4000 & 85 &      3\\ 
{\em Adult} & 32562 & 16282 & 123 &    6.6\\ 
{\em Ijcnn} & 49990 & 91701 & 22 &      1\\ 
{\em YearPrediction} & 463715 & 51630 & 90 &      1\\ 
{\em CovertypeBinary} & 522910 & 58102 & 54 &      1\\ 
\bottomrule
\hline 
\end{tabular} 
}

%

\end{center}
\label{tab:dataSpecNytro}
\end{table}

As illustrated in Figure~\ref{fig:areaPlotv2_WIDE}, NYTRO is preferable when $\SNR \leq 1$, that is when the problem is quite noisy. 
When  $\SNR > 1$, then NYTRO is faster when the dimension of the problem is sufficiently large.
Note that,  in particular,  the area of the NYTRO region on $\SNR > 1$ increases with $n$, and  the  curve $c_1$ is quite flat when $n$ is very large. 
On the opposite extremes we have early stopping and NKRLS. Indeed, one is effective when the dimensionality is very large, while the second when it is very small. There is  a peak around $\SNR \approx 1$ for which it seems that the only useful algorithm is NKRLS when the dimensionality is sufficiently large.  The only region where KRLS is more effective is when $\SNR \approx 1$ and the dimensionality is close to $n$. 

In the next subsection, the theoretical results are validated by an experimental analysis on benchmark datasets.
We add one remark first.

\begin{rem}[Empirical  parameter choices and regularization path]
Note that an important aspect that is not covered by Figure~\ref{fig:areaPlotv2_WIDE} is that iterative algorithms have the further desirable  property of computing  the regularization path. In fact, for  KRLS and NKRLS computations are  slowed by a factor of $|\Lambda|$, where $\Lambda$ is the discrete set of cross-validated $\lambda$s. This last aspect is very relevant in practice, because the optimal regularization parameter values are not known and need to be found via model selection/aggregation. 
\end{rem}

\begin{table*}
\caption{Time-accuracy comparison on benchmark datasets.}
\begin{center}
{\tiny\resizebox{\textwidth}{!}{%
\begin{tabular}{ccccccc}
\toprule 
{\em Dataset} & & \specialcell{{\em KOLS}} & \specialcell{{\em KRLS}} & \specialcell{{\em Early}\\ {\em Stopping}} & \specialcell{{\em NKRLS}} & \specialcell{{\em NYTRO}} \\
\midrule
\multirow{3}{*}{\specialcell{{\em InsuranceCompany}\\ $ n = 5822 $ \\ $m = 2000$  }}
 & $Time \, (s)$ & \textbf{1.04} & 97.48 $\pm$ 0.77 & 2.92 $\pm$ 0.04 & 20.32 $\pm$ 0.50 &   5.49 $\pm$ 0.12 \\ 
 & $RMSE$ & 5.179 & \textbf{0.4651 $\pm$ 0.0001} & \textbf{0.4650 $\pm$ 0.0002} &  {\bf 0.4651 $\pm$ 0.0003} &   {\bf 0.4651 $\pm$ 0.0003} \\ 
 & $Par.$ & NA & 3.27e-04 & 494 $\pm$ 1.7 & 5.14e-04 $\pm$ 1.42e-04 & 491 $\pm$ 3 \\ 
 \hline 
\multirow{3}{*}{\specialcell{{\em Adult}\\ $ n = 32562 $ \\ $m = 1000$  }}
 & $Time \, (s)$ & 112 & 4360 $\pm$ 9.29 & 5.52 $\pm$ 0.23 & 5.95 $\pm$ 0.31 &   {\bf 0.85 $\pm$ 0.05} \\ 
 & $RMSE$ & 1765 & \textbf{0.645 $\pm$ 0.001}  & 0.685 $\pm$ 0.002 & 0.6462 $\pm$ 0.003 &  0.6873 $\pm$ 0.003 \\ 
 & $Par.$ & NA & 4.04e-05 $\pm$ 1.04e-05 & 39.2 $\pm$ 1.1 & 4.04e-05 $\pm$ 1.83e-05 & 44.9 $\pm$ 0.3 \\ 
 \hline 
\multirow{3}{*}{\specialcell{{\em Ijcnn}\\ $ n = 49990 $ \\ $m = 5000$  }}
 & $Time \, (s)$ & 271 	& 825.01 $\pm$ 6.81 	& 154.82 $\pm$ 1.24 & 160.28 $\pm$ 1.54 &  {\bf 80.9 $\pm$ 0.4}\\ 
 & $RMSE$ & 730.62 	& 0.615 $\pm$ 0.002 & {\bf 0.457 $\pm$ 0.001}& 0.469 $\pm$ 0.003 &  {\bf 0.457 $\pm$ 0.001}\\ 
 & $Par.$ & NA 	& 1.07e-08 $\pm$ 1.47e-08 & 489 $\pm$ 7.2 & 1.07e-07 $\pm$ 1.15e-07 & 328.7 $\pm$ 2.6\\ 
 \hline 
\multirow{3}{*}{\specialcell{{\em YearPrediction}\\ $ n = 463715 $ \\ $m = 10000$  }}
 & $Time \, (s)$ &  &  & & 1188.47 $\pm$ 36.7 & {\bf 887 $\pm$ 6}\\ 
 & $RMSE$ & NA & NA & NA & {\bf 0.1015 $\pm$ 0.0002} & 0.1149 $\pm$ 0.0002\\ 
 & $Par.$ & &  &  & 3.05e-07 $\pm$ 1.05e-07 & 481 $\pm$ 6.1\\ 
 \hline 
\multirow{3}{*}{\specialcell{{\em CovertypeBinary}\\ $ n = 522910 $ \\ $m = 10000$  }}
 & $Time \, (s)$   &  &  & & 1235.21 $\pm$ 42.1 &  {\bf 92.69 $\pm$ 2.35}\\ 
 & $RMSE$ & NA & NA & NA & 1.204 $\pm$ 0.008 &  {\bf 0.918 $\pm$ 0.006}\\ 
 & $Par.$ & & & & 9.33e-09 $\pm$ 1.12e-09 & 39.2 $\pm$ 2.3\\ 
\bottomrule
\hline 
\end{tabular} 
}

}
\label{tab:testSetComparisonNytro}
\end{center}
\end{table*}

\subsection{Experiments}\label{sect:experiments}
In this subsection we present an empirical evaluation of the NYTRO algorithm, showing regimes in which it provides a significant model selection speedup with respect to NKRLS and the other exact kernelized learning algorithms mentioned above (KOLS, KRLS and Early Stopping). We  consider the Gaussian kernel and  the subsampling of the training set points for kernel matrix approximation is performed uniformly at random. All experiments have been carried out on a server with 12 $\times$ 2.10GHz Intel$^\circledR$ Xeon$^\circledR$ E5-2620 v2 CPUs and 132 GB of RAM.

We compare the algorithms on the benchmark datasets reported in Table \ref{tab:dataSpecNytro}\footnote{All the datasets are available at \texttt{http://archive.ics.uci.edu/ml} or \texttt{https://www.csie.ntu.edu.tw/$\sim$cjlin/libsvmtools/datasets/}}.
In the table we also report the bandwidth parameter $\sigma$ adopted for the Gaussian kernel computation. Following \citep[]{conf/icml/SiHD14}, we measure performance by the root mean squared error (RMSE).\\
For the \texttt{YearPredictionMSD} dataset, outputs are scaled in $\left[ 0,1 \right] $. 

For all the  algorithms, model selection is performed via hold-out cross validation, where the validation set is composed of 20\% of the training points chosen uniformly at random at each trial.
We select the regularization parameter $\lambda$ for NKRLS between $100$ guesses logarithmically spaced in $ \left[  10^{-15} ,  1 \right] $, by computing the validation error for each model and choosing the $\lambda^*$ associated with the lowest error. 
$m$ is fixed for each learning task.
NYTRO's regularization parameter is the number of iterations $t$. We select the optimal $t^*$ by considering the evolution of the validation error.
As an early stopping rule, we choose an iteration such that the validation error ceases to be decreasing up to a 
given threshold chosen to be the  $5\%$ of the relative RMSE.
After model selection, we evaluate the performance on the test set. We report the results in Table \ref{tab:testSetComparisonNytro} and discuss them further below.

\begin{figure}
\centering
\includegraphics[width = 0.6\textwidth]{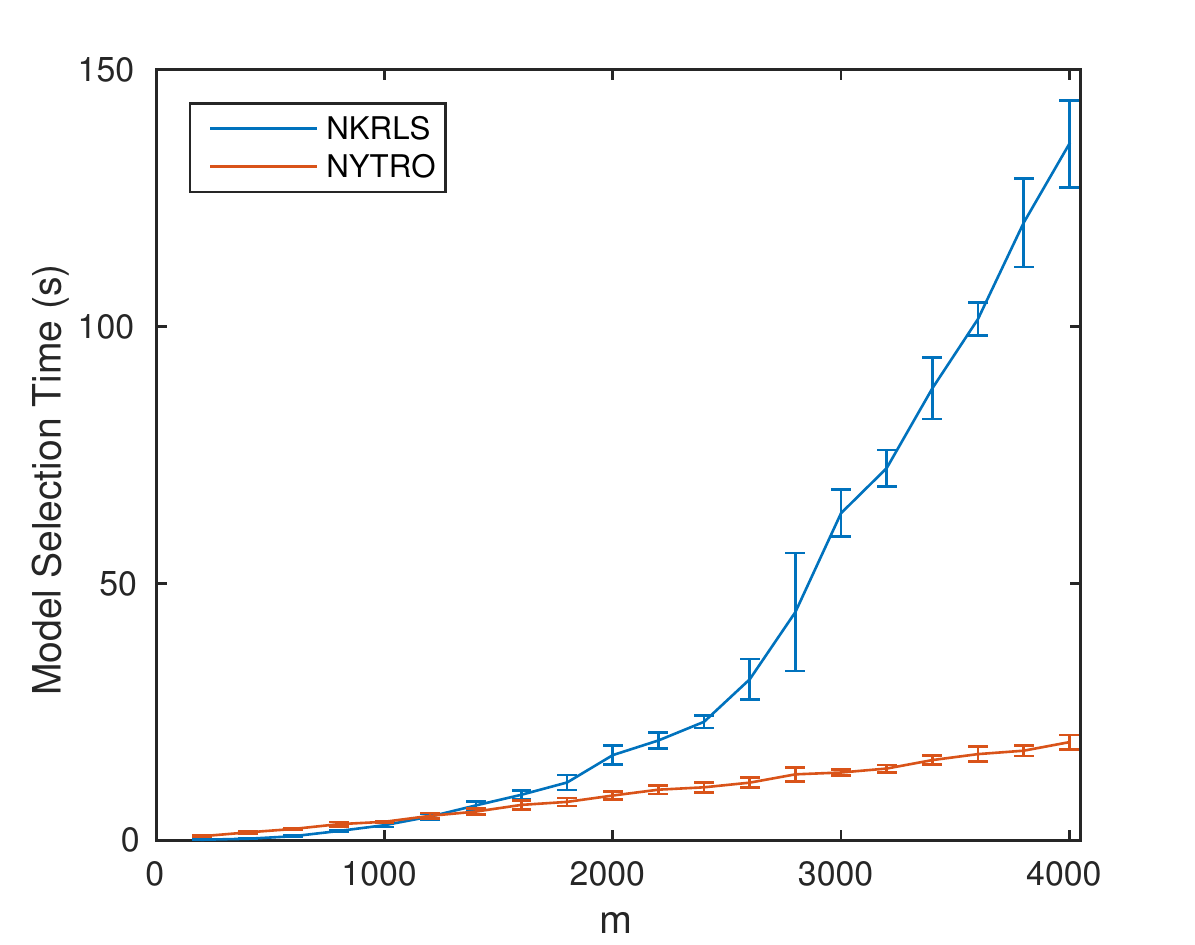}
\caption{Training time of NKRLS and NYTRO on the \texttt{cpuSmall} dataset as the subsampling level $m$ varies linearly between 100 and 4000. Experiment with 5 repetitions. mean and standard deviation reported.}
\label{fig:timeComparison}
\end{figure}

\paragraph{Time Complexity Comparison}
We start by showing how the time complexity changes with the subsampling level $m$, making NYTRO more convenient than NKRLS if $m$ is large enough. For example,  consider Figure \ref{fig:timeComparison}. We performed training on the \texttt{cpuSmall}\footnote{\texttt{http://www.cs.toronto.edu/$\sim$delve/data/datasets.html}} dataset ($n=6554,\, d=12$), with $m$ spanning between $100$ and $4000$ at $100$-points linear intervals. The experiment is repeated $5$ times, and we report the mean and standard deviation of the NYTRO and NKRLS model selection times. We consider $100$ guesses for $\lambda$, while the NYTRO iterations are fixed to a maximum of $500$. As revealed by the plot, the time complexity grows linearly with $m$ for NYTRO and quadratically for NKRLS. This is consistent with the time complexities outlined in Sections \ref{sect:setting} and \ref{sect:proposedAlgorithm} ($O(nm^2 + m^3)$ for NKRLS and $O(nmt + m^3)$ for NYTRO).

\paragraph{Time-accuracy Benchmarking}

We also compared the training time and accuracy performances for KRLS, KOLS, Early Stopping (ES), NKRLS and NYTRO, reporting the selected hyperparameter ($\lambda^*$ for KRLS and NKRLS, $t^*$ for ES and NYTRO), the model selection time and the test error in Table \ref{tab:testSetComparisonNytro}. 
All the experiments are repeated 5 times. 
The standard deviation of the results is negligible. 
Notably, NYTRO achieves comparable or superior predictive performances with respect to its counterparts in a fraction of the model selection time.
In particular, the absolute time gains are most evident on large-scale datasets such as \texttt{Covertype} and \texttt{YearPredictionMSD}, for which a reduction of an order of magnitude in cross-validation time corresponds to saving tens of minutes.
Note that exact methods such as KOLS, KRLS and ES cannot be applied to such large scale datasets due to their prohibitive memory requirements. 
Remarkably, NYTRO's predictive performance is not significantly penalized in these regimes and can even be improved with respect to other methods, as in the \texttt{Covertype} case, where it requires 90\% less time for model selection. 


\chapter{Speeding up by Data-independent Subsampling}
\label{chap:randfeats}

\section{Setting}
As we have seen in Chapter \ref{Chap:SLT}, a basic problem in machine learning is estimating a function from random noisy data \citep{vapnik1998statistical,cucker02onthe}.
The function to be learned is typically fixed, but unknown, and flexible nonlinear/nonparametric models
are needed for good results. A general class of models  is based on  functions of the form,
\eqal{\label{eq:non-par-model}
\fn(x) \; = \; \sum_{i=1}^m \, \alpha_i \, q(x, \omega_i),
}
where $q$ is a nonlinear function, $\omega_1, \dots, \omega_m \in \R^d$ are the \textit{centers},   $\alpha_1, \dots, \alpha_m \in \R$ are the coefficients, and  
$m = m_{n}$ could/should {\em grow} with the number of data points $n$. In this context, the problem of learning reduces to the problem of  computing from  data the parameters $\omega_1, \dots, \omega_m$, $\alpha_1, \dots, \alpha_m$ and $m$. Among others,   one-hidden layer networks \citep{bishop2006}, or RBF networks \citep{Poggio/Girosi/90},  are  examples of classical approaches considering these models. Here, parameters are computed by considering a non-convex optimization problem, typically hard to solve and study \citep{pinkus1999approximation}.
Kernel methods, introduced in Chapter \ref{Chap:kernelMethods}, are another notable example \citep{schlkopf2002learning}.
In this case, 
$q$ is assumed  to be a positive definite function \citep{aronszajn1950theory} (the kernel function) and it is shown that choosing the centers to be the input points, hence $m=n$, suffices for optimal statistical results \citep{kimeldorf1970correspondence,scholkopf2001generalized,caponnetto2007optimal}. In kernel methods, computations reduce to finding the coefficients $\alpha_i$, which can be  typically done by convex optimization. While theoretically sound and remarkably effective in small and medium size problems,  memory requirements make kernel methods rapidly become unfeasible as datasets grow large.  A current 
challenge is then   to design scalable non-parametric procedures,  while not giving up the nice  theoretical properties of kernel methods.\\ 
A simple, yet effective, idea is that  of sampling the centers at random, either in a data-dependent or in a data-independent way.  Notable examples of this idea include \Nystrom{} \citep{conf/icml/SmolaS00,conf/nips/WilliamsS00} and random features approaches \citep{conf/nips/RahimiR07}.
Given random centers, computations still reduce to convex optimization with potential big memory gains, if centers are fewer than data-points. 
In practice, the  choice of the number of centers  is   based on heuristics or memory  constraints and the question arises of characterizing theoretically which choices  provide optimal learning bounds. 
For data-dependent subsampling, a.k.a. \Nystrom{} methods, some results in the fixed design setting were given in \citep{journals/jmlr/CortesMT10,conf/colt/Bach13,alaoui2014fast}, while 
in the context of statistical learning a fairly exhaustive analysis was recently provided in \citep{rudi2015less}, as we will see in Chapter \ref{chap:lessismore}. These latter results show, in the statistical learning theory framework, 
that a number of centers smaller than the number of data points suffices for optimal generalization properties. \\
The results on data-independent subsampling, a.k.a. random features, are fewer and require a number of centers in the order of $n$ to attain basic generalization bounds \citep{rahimi2009weighted,journals/jmlr/CortesMT10,bach2015}. 
The study presented in this Chapter, based on \citep{rudi2016genrf}, improves on these results by
\begin{enumerate}
\item Deriving optimal learning bounds for regularized learning with random features.
\item Proving that the optimal bounds are achievable with a number of random features substantially smaller than the number of examples.
\end{enumerate} 
%
%
%
Following \citep{rudi2015less}, we further show that the number of random features/centers can  be seen as a form of 	``computational''  regularization, controlling at the same time statistical and computational aspects.  Theoretical findings are complemented by numerical experiments, validating the bounds and showing the regularization properties of random features.  
The rest of the paper is organized as follows. In Section \ref{sec:ker-LS}, we review relevant results on learning with kernels and least squares. In Section \ref{sect:random-features}, we introduce regularized learning with random features. In Section \ref{sect:main-results}, we present and discuss our main results, while proofs are deferred to the appendix.  Finally, numerical experiments are presented in Section \ref{sec:exp}. 

{\bf Notation.}~~For the sake of readability, we denote by $f(n) \lesssim g(n)$ the condition $f(n) \leq c g(n) (\log n)^s$ for any $n \in \N$ and $c > 0$, $s  \geq 0$, for two functions $f, g: \N \to \R$. We denote by $f(n) \approx g(n)$ the condition $f(n) \lesssim g(n)$ and $g(n) \lesssim f(n)$. The symbol $\gtrsim$ is defined accordingly.
 \section{Background: Generalization Properties of KRLS}\label{sec:ker-LS}
A main motivation of this chapter is showing how random features improve the computations of kernel methods, while retaining their good statistical properties.
Recalling these latter results provide the background of the chapter.
The computational aspects of Kernel Regularized Least Squares (KRLS) were already introduced in Section \ref{sec:krr}.
Thus, to derive learning bounds we only need to first recall KRLS's generalization properties in this Section.
Let $\X$ be a probability space and $\rho$ a probability distribution on   $\X\times \R$. 
For all $x\in \X$, denote by $\rho(y|x)$ the conditional probability of $\rho$, given $x$. 
We make basic assumptions on  the probability distribution and  the kernel \citep{de2005model,caponnetto2007optimal}.
\ba\label{ass:noise}
There exists $\frho(x) = \int y d\rho(y|x)$ such that $\int |\frho(x)|^2 d\rho<\infty$. For $M, \sigma > 0$,
\eqal{\label{eq:noise-on-y}
\int |y - \frho(x)|^p d\rho(y|x) \leq \frac{1}{2}p! M^{p-2} \sigma^2, \quad \forall p \geq 2,
}
holds almost surely. The data $(x_1, y_1), \dots (x_n, y_n)$ are independently and identically distributed according to $\rho$. 
\ea
Note that \eqref{eq:noise-on-y} is satisfied when the random variable $|y - \frho(x)|$,  with $y$ distributed according to $\rho(y|x)$, is uniformly bounded, subgaussian or subexponential for any $x \in \X$.
\ba\label{ass:kernel-bounded} $K$ is measurable, bounded by $\kappa^2$, $\kappa \geq 1$ and the associated RKHS $\hh$ is separable. 
\ea
We  need  one further assumption to control the approximation (bias) and the estimation (variance) properties of KRLS. 
Let $L^2(\X, \rho)=\{ f :\X\to \R~:~\nor{f}^2_\rho=\int |f(x)|^2 d\rho<\infty\}$ be the space of square integrable functions on $\X$, and define the integral operator $L:L^2(\X, \rho)\to L^2(\X, \rho)$, 
$$
 Lf(x)=\int K(x,x')f(x')d\rho, \quad \forall f\in L^2(\X, \rho), 
$$
almost everywhere.
The operator $L$ is known to be symmetric, positive definite and trace class under Assumption~\ref{ass:kernel-bounded} \citep{devito2005learning}. We make the following common assumption~\citep{caponnetto2007optimal,SteinwartHS09}.
\ba[Source \& Capacity Conditions]\label{ass:source-condition}
There exist $r \geq 1/2$ and $R \in [1,\infty)$ such that
\be\label{eq:source}
\nor{L^{-r} \frho}_\rho \leq R.
\ee
Moreover, for $\la>0$ let ${\cal N}(\la)=\text{Tr}( (L+\la I)^{-1}L)$, then
there exist  $\gamma \in (0,1]$ and  $Q > 0$ such that
\be\label{eq:capa} {\cal N}(\la) \leq Q^2\la^{-\gamma}.\ee
\ea
We add some comments. The {\em Mercer} source condition~\eqref{eq:source}  is better illustrated recalling that, according to  Mercer theorem \citep{cucker02onthe},
functions in $\hh$ are  functions $f\in L^2(\X, \rho)$ such that $f =  L^{1/2} g$ for some $g\in L^2(\X, \rho)$.
Note that,  under Assumption~\ref{ass:noise}, we only have $f_\rho\in L^2(X, \rho)$. Then, by \ref{eq:source}, 
for $r=1/2$ we are assuming $f_\rho$ to belong to $\hh$, while for larger $r$ we are  strengthening this condition assuming   $f_\rho$ to belong to smaller subspaces of $\hh$. This condition is known to control the bias of KRLS \citep{caponnetto2007optimal,SteinwartHS09}. \\
The assumption in \eqref{eq:capa} is a natural capacity condition on the space $\hh$. If $\hh$ is finite dimensional,  then ${\cal N}(\la)\leq \text{dim}(\hh)$, for all $\la\ge 0$. More  generally, it can be seen as an assumption on the  ``effective'' dimension at scale $\la$ \citep{caponnetto2007optimal}.  It is always true in the limit case $Q = \kappa$, $\gamma = 1$, sometimes referred to as the capacity independent setting \citep{cucker2007learning}. However, when $\gamma < 1$ faster learning bounds can be achieved. Assumption~\ref{ass:source-condition} corresponds to a ``polynomial'' rates regime that can be specialized common  smoothness classes in  non-parametric statistics, such as Sobolev spaces \citep{steinwart2008support}. Other regimes, e.g. exponential decays or finite dimensional cases \citep{conf/colt/ZhangDW13}, can also be derived with minor modifications. This discussion is omitted. 

Before giving the generalization bounds for KRLS, we introduce the {\em generalization error} of an estimator $f \in \Ltwo$, that will be useful in the rest of the paper and is measured by the excess risk \citep{cucker02onthe,steinwart2008support}
\eqal{\label{eq:excess-risk}
\EE(f) - \EE(f_\rho), \quad \textrm{with} \quad \EE(f)=\int (y-f(x))^2 d\rho(x,y).
}
The following theorem is the generalization bound for KRLS and is taken from \citep{caponnetto2007optimal}, see also \citep{smale2007learning,SteinwartHS09}.
\bt\label{thm:kkrupbound} Let $\tau \geq 0$.
Under Assumptions~\ref{ass:noise},~\ref{ass:kernel-bounded},~\ref{ass:source-condition}, if 
$
\la_n \approx n^{-\frac{1}{2r+\gamma}}
$
and $\widehat{f}_n=\widehat{f}_{\la_n}$ of \eqref{eq:krr-clform},
then the following holds with probability at least $1-e^{-\tau}$
$$
\EE(\widehat f _n)-\EE(f_\rho) \lesssim \tau^2 n^{- \frac{2v}{2v+\gamma}}, \quad\quad v=\min\{r, 1\}.
$$
\et
The above result can be shown to be optimal, i.e. matching a corresponding lower bound \citep{caponnetto2007optimal,SteinwartHS09}.
\bt\label{thm:lower-bound}
For any measurable estimator $\widehat{f}_n$ depending on $n$ training examples, there exists a probability measure $\rho$, satisfying  Assumptions~\ref{ass:noise},~\ref{ass:kernel-bounded},~\ref{ass:source-condition}, such that
$$
\mathbb{E} ~\EE(\widehat{f}_n) ~-~ \EE(f_\rho) ~~\gtrsim ~ ~n^{-\frac{2r}{2r+\gamma}}.
$$
\et

{\bf Beyond KRLS.}~~Summarizing, the generalization properties of KRLS are optimal, since the upper bound of Thm.~\ref{thm:kkrupbound} matches the lower bound in Thm.~\ref{thm:lower-bound}. However, for KRLS, optimal statistical properties come at the cost of high computational complexity. As seen from~\eqref{eq:krr-clform}, KRLS requires $O(n^2)$ in space, to store the matrix $\K_n$, and roughly $O(n^3)$ in time, to solve a corresponding linear system. 
While the time complexity can be reduced without hindering generalization, by considering  {\em iterative \& online techniques} \citep{engl1996regularization,bauer2007regularization,CapYao06,blanchard2010optimal}, in general lowering the space requirements is a challenge.
The question is, then, if it is possible to design a form of computational regularization controlling at once generalization,  time {\em and space} requirements. Indeed, this can be done considering random features, as we show next.
\section{Regularized Learning with Random Features}\label{sect:random-features}
Following the discussion in the introduction,  the idea is to consider a general non-linear function in \eqref{eq:non-par-model}, but now taking the centers at random according to a known probability distribution \citep{conf/nips/RahimiR07}. More precisely,  let   $(\Omega, \theta)$ be a known probability space and $\psi: \Omega \times \X \to \CC$. 
Given $\omega_1, \dots, \omega_m$  independently and identically distributed according to $\theta$, the idea is to consider functions in the linear span of the set of {\em random features} 
$$
\Big\{ \frac{1}{\sqrt{m}}\psi(\omega_1, \cdot), \dots, \frac{1}{\sqrt{m}} \psi(w_m, \cdot) \Big\} .
$$
In particular, given a dataset,  we  consider the estimator defined by, 
\be\label{eq:RF-KRLS} 
\tilde{f}_{m,\la}(x) = \sum_{j=1}^m \frac{1}{\sqrt{m}} \psi(\omega_j, x) \tilde \alpha_j, 
 \quad \tilde\alpha_{m,\la} = \Big(\tSn^\top \tSn + \la I\Big)^{-1}\tSn^\top\yn, \quad \forall x \in \X,
\ee
where $(\tilde \alpha_1, \dots, \tilde \alpha_m)=\tilde\alpha_{m,\la}$, $\yn = \bar{y}/\sqrt{n}$, and $\tSn \in \R^{n\times m}$ with entries  $(\tSn)_{i,j}= \psi(\omega_j,x_i)/\sqrt{nm}$.
Some observations are in order. First, note that the above estimator is exactly of the form discussed in the introduction (see \eqref{eq:non-par-model}), where the non-linear function is now denoted by $\psi$ and the centers can belong to an abstract probability space (see e.g. Example~\ref{example:rf} or Appendix F of \citep{rudi2016genrf}).  Second,  it is easy to see that the above estimator follows from  the minimization problem, 
\be
\label{eq:prob-base}
\min_{\alpha \in \CC^{m}} \frac{1}{n}\sum_{i=1}^n \biggl(
 \sum_{j=1}^m \frac{1}{\sqrt{m}} \psi(\omega_j, x_i) \alpha_j -y_i\biggr)^2 + \la  \nor{\alpha}_m^2,
\ee
where $\nor{\cdot}_{m}$ is the Euclidean norm in $\R^m$ (so we will refer to it as the RF-KRLS estimator). 
Finally, by comparing \eqref{eq:krr-clform} and \eqref{eq:RF-KRLS}, it is clear that if $m$ is smaller than $n$, then 
the time/space requirements of the above approach can be much smaller than those of KRLS.  Indeed, computing~\eqref{eq:RF-KRLS} requires $O(mn)$ in space and roughly $O(nm^2+m^3)$ in time.
The question is then if the potential computational gain comes at the expense of  generalization properties. While the above approach is general, to answer this latter question random features  will be  related to kernels.

{\bf Assumptions on random features.}~~ 
We first need a basic assumption on the random features.
\ba\label{ass:psi-continuous-bounded}
$\psi$ is measurable and there exists $\tkappa \geq 1$ such that $ |\psi(\omega, x)| \leq \tkappa$ almost surely.
\ea
Under Assumption~\ref{ass:psi-continuous-bounded} the integral below exists and defines the kernel induced by $\psi$,
\be\label{eq:kerrf}
\K(x,x') = \int_\Omega \psi(\omega,x){\psi(\omega,x')} d\theta(\omega), \quad \forall x, x' \in \X.
\ee
Indeed it is easy to see that $\K$ is a positive definite function, measurable and bounded almost surely. We still need a minor assumption. 
\ba\label{ass:kerrf-sep}
The RKHS  $\hh$ associated to the kernel in \eqref{eq:kerrf} is separable.
\ea
Note that, Assumptions~\ref{ass:psi-continuous-bounded},~\ref{ass:kerrf-sep} are  mild and satisfied by essentially all examples of random features (see Appendix F of \citep{rudi2016genrf}). 
\bex[Gaussian random features \citep{conf/nips/RahimiR07}]\label{example:rf}
When $\X=\R^d$, $d \in \N$, by setting $\Omega=\R^d\times \R$ and $\psi(\omega, x)= \cos(\beta^\top x+b)$, with $\omega=(\beta, b) \in \Omega$, then there exists a probability distribution on $\Omega$,  such that \eqref{eq:kerrf} holds for the Gaussian kernel $K(x,x')=e^{-\gamma\nor{x-x'}^2}$, $x,x'\in \R^d$. Then, random features are continuous, hence measurable, and  bounded. Moreover, the RKHS associated to the Gaussian kernel is separable \citep{steinwart2008support}.
\eex 
In general, the space $\Omega$ and its distribution $\theta$ need not to be related to the data space and distribution. 
However, we will see in the following that better results can be obtained if they are {\em compatible} in a sense made precise by the next assumption \citep{bach2015}. 
\ba[Random Features Compatibility] \label{ass:omegas}
For $\la>0$, let  
\be\label{eq:F-infty}
{\cal F}_\infty(\la) = \operatorname*{ess\,sup}_{\omega \in \Omega} ~\nor{(L + \la I)^{-1/2} \psi(\omega,\cdot)}_\rho^2,
\ee 
then there exists $\alpha \in (0,1]$ and $F \geq \tkappa$ such that,
 \be\label{eq:F-inftydecay} {\cal F}_\infty(\la) \leq F^2 \la^{-\alpha}, \qquad \forall \la > 0.\ee
\ea
Note that the condition above is always satisfied when $\alpha = 1$ and $F = \tkappa$. However, when $\alpha < 1$, the assumption leads to better results in terms of the number of random features required for optimal generalization (see Theorem~\ref{thm:optimal-rates-RF-RKLS}). Moreover, Assumption~\ref{ass:omegas} allows to extend our analysis to more refined ways of selecting the random features, as discussed in the next remark.
\br[Beyond uniform sampling for RF]\label{rem:imp-sampling} A natural question is if sampling the random features non-uniformely, allows to use less features.  
This approach can be cast in our setting, considering $\theta' = \theta q$ and $\psi' = \psi/\sqrt{q}$, where $q$ is the sampling distribution. For example, when considering the data-dependent distribution proposed in \citep{bach2015}, Assumption~\ref{ass:omegas} is satisfied with $\alpha = \gamma$.
\er
%
%
\subsection{Main Results}\label{sect:main-results}
We next state results proving the optimal generalization properties of learning with random features.
Our main results in Theorems~\ref{thm:opt-RF-worst-case},~\ref{thm:opt-RF-adapt-noalpha},~\ref{thm:optimal-rates-RF-RKLS} characterize the number of random features required by the RF-KRLS estimator in \eqref{eq:RF-KRLS}, to achieve the optimal learning bounds under different regularity conditions. In the theorem below, we analyze the generalization properties in a basic scenario, where the only requirement is that $\frho$ is in $\hh$.
\bt\label{thm:opt-RF-worst-case} Let $\tau > 0$. Under Assumptions~\ref{ass:noise},~\ref{ass:psi-continuous-bounded},~\ref{ass:kerrf-sep}, and that $f_\rho \in \hh$, there exists $n_0 \in \N$ (see proof, for the explicit constant) such that, when $n \geq n_0$,
\eqals{
\la_n \approx n^{-\frac{1}{2}},\quad m_n \gtrsim \la_n^{-1}
}
and $\tilde{f}_n = \tilde{f}_{m_n,\la_n}$ as in \eqref{eq:RF-KRLS}, then the following holds with probability at least $1 - e^{-\tau}$
\eqal{\label{eq:rf-opt-bound-worst-case}
{\cal E}(\tilde{f}_n) - {\cal E}(\frho) \lesssim \tau^2 n^{-\frac{1}{2}}.
}
\et
We add some comments. First, the RF-KRLS estimator in \eqref{eq:RF-KRLS} achieves the optimal learning bound in this scenario. Indeed the condition $f_\rho \in \hh$ is equivalent to Assumption~\ref{ass:source-condition} with $r=1/2, \gamma=1$ and so, the bound of Theorem~\ref{thm:opt-RF-worst-case} matches the ones in  Theorem~\ref{thm:kkrupbound},~\ref{thm:lower-bound}. 
Moreover, RF-KRLS attains the optimal bound with a number of random features that is roughly
$$m_n \approx \sqrt{n}.$$
Thus, RF-KRLS generalizes optimally by requiring $O(n^{1.5})$ in space and  $O(n^2)$ in time, compared to $O(n^2)$ in space and $O(n^3)$ in time for KRLS. 

\begin{figure}[t]
\begin{center}
\includegraphics[width=0.75\linewidth]{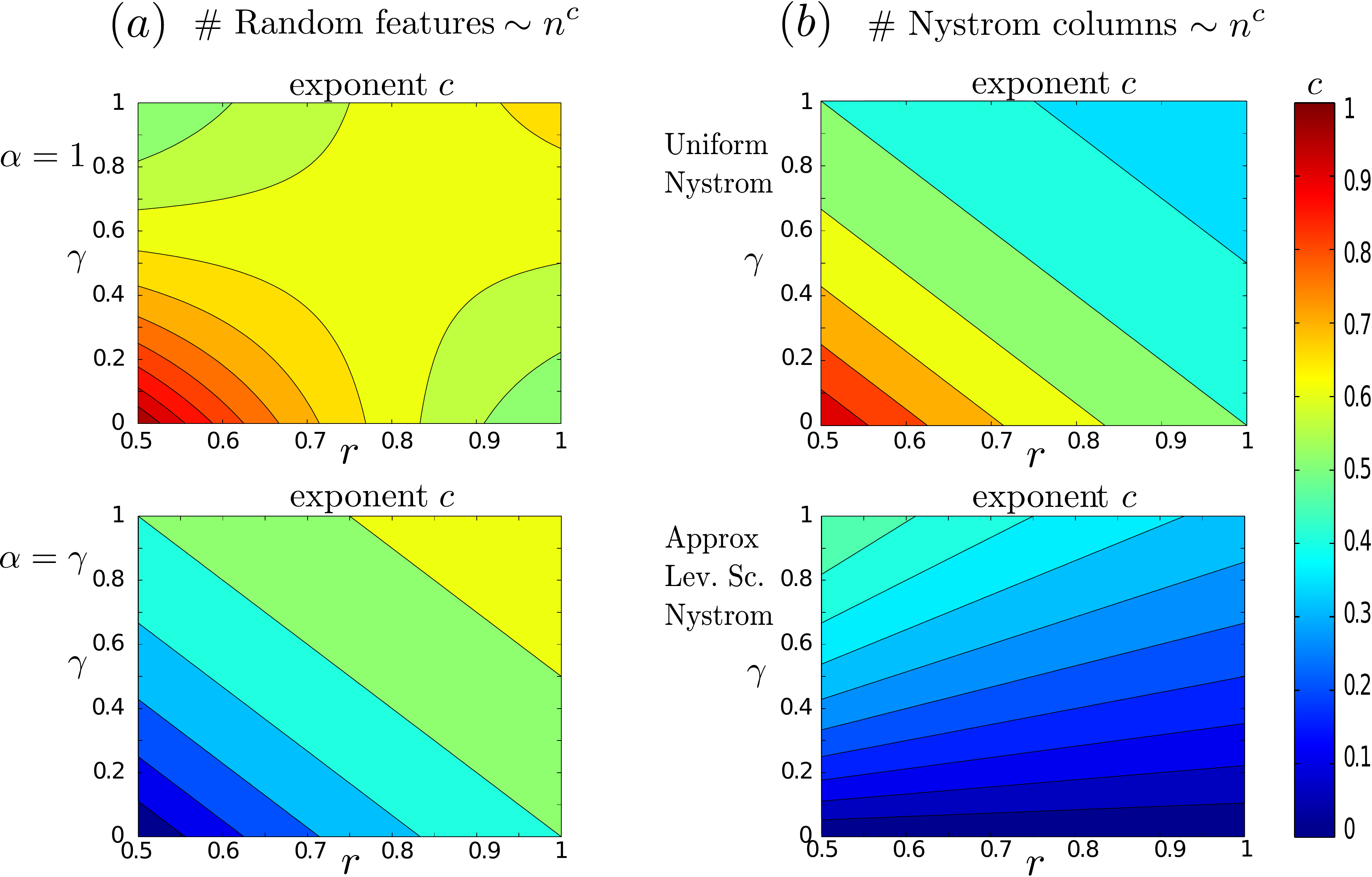}
\end{center}
\caption{Predicted behaviour of the number of features/centers required to achieve optimal learning rates for (a) random features (see Theorem~\ref{thm:optimal-rates-RF-RKLS}) and (b) \Nystrom{} (see \citep{rudi2015less}) as a function of $r, \gamma, \alpha$.\label{fig:ny-vs-randf}}
\vspace{-0.2cm}
\end{figure}

We next show that when the learning task satisfies additional regularity assumptions, it is possible to achieve faster learning rates than $n^{-1/2}$ (compare with Theorem~\ref{thm:kkrupbound}). In particular, the next theorem shows that under Assumption~\ref{ass:source-condition}, the RF-KRLS estimator achieves fast learning rates, again for a number or random features smaller than $n$.  
\bt\label{thm:opt-RF-adapt-noalpha}
Let $\tau > 0$ and $v = \min(r,1)$. Under Assumptions~\ref{ass:noise},~\ref{ass:source-condition},~\ref{ass:psi-continuous-bounded},~\ref{ass:kerrf-sep}, there exists $n_0 \in \N$ (see proof in the Appendix E of \citep{rudi2016genrf}, for the explicit constant) such that when $n \geq n_0$,
\eqals{
\la_n ~\approx~ n^{-\frac{1}{2v + \gamma}},\quad m_n ~\gtrsim ~\la_n^{-c}, \quad\textrm{with}\quad c = 1 +\gamma(2v - 1),
}
and $\tilde{f}_n = \tilde{f}_{m_n,\la_n}$ as in \eqref{eq:RF-KRLS}, then the following holds with probability at least $1 - e^{-\tau}$,
\eqals{
{\cal E}(\tilde{f}_n) - {\cal E}(\frho) \lesssim \tau^2 n^{-\frac{2v}{2v + \gamma}}.
}
\et
Comparing Theorem~\ref{thm:opt-RF-adapt-noalpha} with Theorem~\ref{thm:kkrupbound}, we see that the RF-KRLS algorithm achieves optimal learning bounds that are possibly faster than $n^{-1/2}$. In particular, when $\frho$ is regular (that is $r$ is close to $1$) and the effective dimension is small (that is $\gamma$ close to $0$), RF-KRLS achieves a fast learning bound of $n^{-1}$, with a number of random features that is only $m_n \approx n^{1/2}$. However, with other combinations of $r,\gamma$, RF-KRLS may require more random features to attain the faster rates. The relation between the number of features for optimal rates and the parameters $r, \gamma$ is illustrated in Figure~\ref{fig:ny-vs-randf} (top-left). 

Note that in Theorem~\ref{thm:opt-RF-adapt-noalpha}, the possible interaction between the feature map $\psi$, the feature distribution $\theta$ and the data distribution $\rho$ is not taken into account. In the next theorem we consider this interaction, by means of the {\em compatibility condition} in \eqref{eq:F-inftydecay}. In particular we prove that the number or random features needed to achieve the optimal bounds may be even smaller than $n^{1/2}$ when the compatibility index $\alpha$ is smaller than $1$.
\bt\label{thm:optimal-rates-RF-RKLS}
Let $\tau > 0$ and $v = \min(r,1)$. Under Assumptions~\ref{ass:noise},~\ref{ass:source-condition},~\ref{ass:psi-continuous-bounded},~\ref{ass:kerrf-sep},~\ref{ass:omegas}, there exists $n_0 \in \N$ (see proof, for the explicit constant) such that when $n \geq n_0$,
\eqals{
\la_n ~\approx~ n^{-\frac{1}{2v + \gamma}},\quad m_n ~\gtrsim ~\la_n^{-c}, \quad\textrm{with}\quad c = \alpha +(1+\gamma-\alpha)(2v - 1),
}
and $\tilde{f}_n = \tilde{f}_{m_n,\la_n}$ as in \eqref{eq:RF-KRLS}, then the following holds with probability at least $1 - e^{-\tau}$
\eqals{
{\cal E}(\tilde{f}_n) - {\cal E}(\frho) \lesssim \tau^2 n^{-\frac{2v}{2v + \gamma}}.
}
\et
Again, the RF-KRLS estimator achieves the optimal learning bounds, depending on the regularity conditions of the learning task. Note that the assumption in \eqref{eq:F-inftydecay} affects only the number of random features needed to achieve the optimal bound. In particular, in the less favorable case, that is, when $\alpha = 1$, we recover exactly the results of Theorem~\ref{thm:opt-RF-adapt-noalpha}, while in the favorable case when $\alpha$ is close to $\gamma$, a number of random features that is typically smaller than $n^{1/2}$ is required to achieve the optimal bounds.
For example, even with $r = 1/2$, it is possible to achieve the fast bound of $n^{-1/(1+\gamma)}$, with $m_n \approx n^{-\gamma/(1+\gamma)}$. This means, a fast learning rate of $n^{-3/4}$ with $m_n \approx n^{1/4}$, when $\gamma = 1/3$, and, more surprisingly, a fast learning rate of $n^{-1}$ with $m_n \approx 1$ when the effective dimension of the learning task is very small, that is $\gamma$ close to $0$. 
Figure~\ref{fig:ny-vs-randf} (bottom-left) is an illustration of the number of random features needed for optimal learning rates when $\alpha = \gamma$. Note that, in general this latter condition depends on the unknown data-distribution. In other words, recalling Remark~\ref{rem:imp-sampling}, we see that while non-uniform sampling can 
in principle have a dramatic effect on the number of features required for optimal rates, 
how to design practical sampling schemes is an open question (see also discussion in \citep{bach2015}). Before giving a sketch of the proof, we first comment on the regularization role of $m_n$.

{\bf The number of random features as a regularizer.}~ While in Theorems\ref{thm:opt-RF-worst-case},~\ref{thm:opt-RF-adapt-noalpha},~\ref{thm:optimal-rates-RF-RKLS}, the number of features $m_n$ is typically viewed as the computational budget and generalization properties are governed by $\la_n$, following \citep{rudi2015less}, an interesting observation can be derived exchanging  the 
roles of $m$ and $\la$. For example, in the worst-case scenario we can set 
 $$m_n \approx  \sqrt{n}  \quad \text{and}  \quad \la_n \approx m_n^{-1}.$$
Clearly the obtained estimator achieves the optimal error bound in \eqref{eq:rf-opt-bound-worst-case}, but the number of features can now be seen a regularization parameter  controlling  at once generalization properties and also  time/space requirements.
As noted in \citep{rudi2015less}, an advantage of this parameterization is that the solutions corresponding  to different numbers of features can be efficiently computed with an incremental strategy (see Appendix~\ref{sect:incr-algo}). Note that, in practice, parameters are tuned by cross validation and an incremental approach allows for a fast computation of the whole regularization path, resulting in efficient model selection\footnote{In this view it is also useful to consider that, using results in \citep{CapYao06}, it is possible to  prove  
that optimal rates can be achieved adaptively if hold-out is used for parameter tuning.}.

{\bf Sketch of the proof.} ~ Theorem~\ref{thm:optimal-rates-RF-RKLS} is the main technical contribution of the paper from which Theorems~\ref{thm:opt-RF-worst-case},~\ref{thm:opt-RF-adapt-noalpha} are derived. 
The extended version of Theorem~\ref{thm:optimal-rates-RF-RKLS} with explicit constants is given in Theorem 8 in the appendix of \citep{rudi2016genrf}.
Its proof relies on a novel error decomposition separates functional analytic results 
from  probabilistic estimates
.  
For the sake of conciseness, we omit to report the proofs, which can be found in the appendix of \citep{rudi2016genrf}. 
Here we illustrate some basic ideas. 
From Theorem 6 of \citep{rudi2016genrf} we have
\eqal{\label{eq:sketch}
\EE(\tilde{f}_{m,\la}) - \EE(\frho)  \;\;\;\lesssim\;\;\;(\;{\cal S}(\la, n)\;\; + \;\;{\cal C}(\la,m)\;\; + \;\;R\la^{v}\;)^2.
}
The estimation error ${\mathcal S}(\la, n)$ is bounded in Lemma 7 of \citep{rudi2016genrf} and depends on the noise, the regularization level and an empirical version of the effective dimension ${\cal N}(\la)$ which takes into account the random sampling of the data and of the random features. 
The empirical version of the effective dimension ${\cal N}(\la)$ is decomposed analytically in Proposition 8 of \citep{rudi2016genrf} and bounded in terms of its continuous version in Lemma 3 of \citep{rudi2016genrf}.
The approximation error $R \la^{v}$ is independent of the data or the random features. It depends only on the regularization level and is controlled by the Mercer source condition~\eqref{eq:source}. 
Finally, the term ${\mathcal C}(\la, m)$ is obtained by means of an interpolation inequality for linear operators in Lemma 2 of \citep{rudi2016genrf}, depending on the source condition~\eqref{eq:source}. 
It is bounded in Lemma 6 of \citep{rudi2016genrf} in terms of the effective dimension ${\cal N}(\la)$, but also the random features compatibility term ${\cal F}_\infty(\la)$. 
Learning rates (Theorems~\ref{thm:opt-RF-worst-case}-\ref{thm:optimal-rates-RF-RKLS}, and Theorem 8 of \citep{rudi2016genrf}) are obtained by optimizing the bound \eqref{eq:sketch}.
%
\begin{figure}[t]
\includegraphics[width=0.32\textwidth,height=0.23\textwidth]{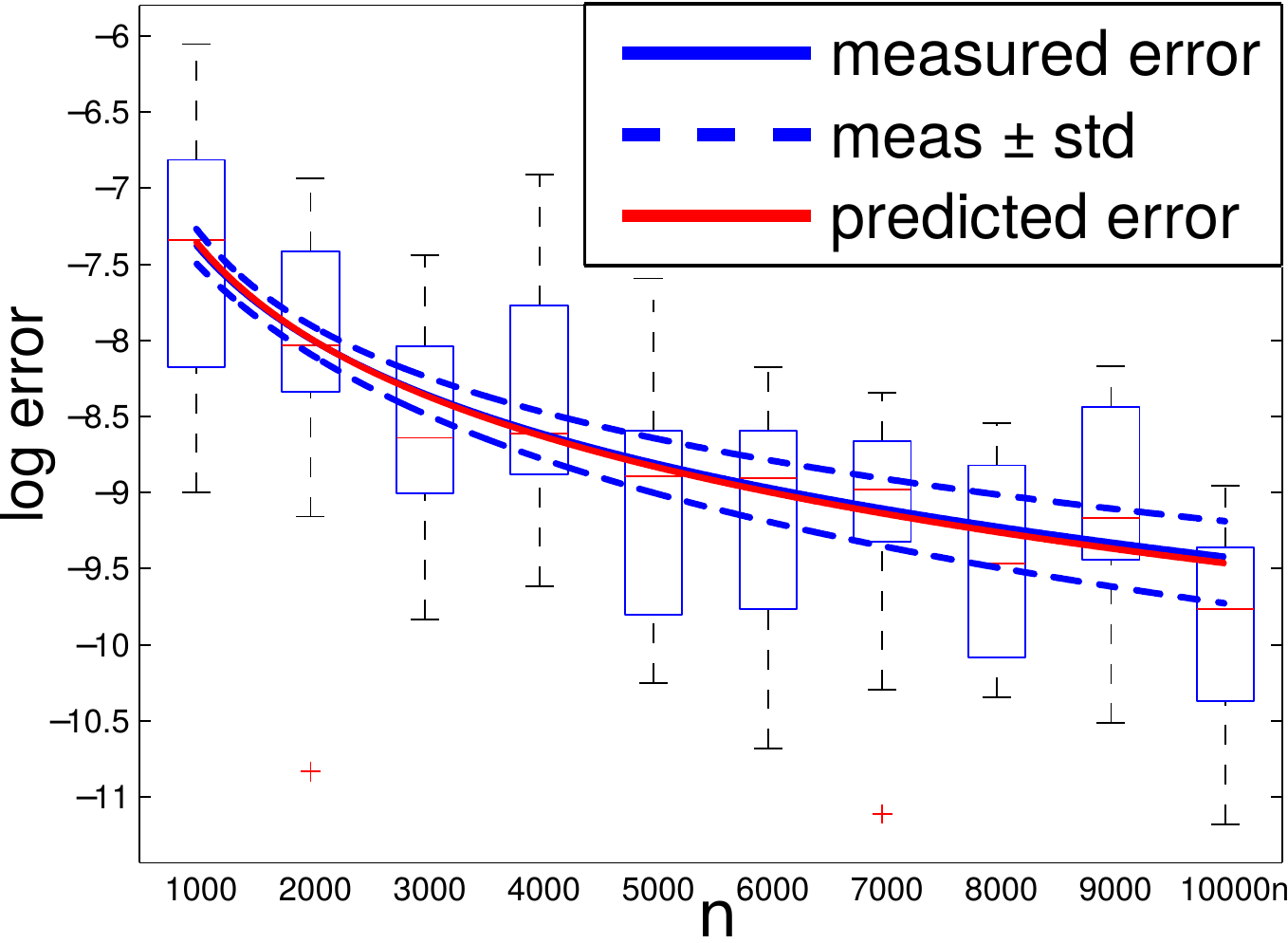}
\hfill
\includegraphics[width=0.32\textwidth,height=0.23\textwidth]{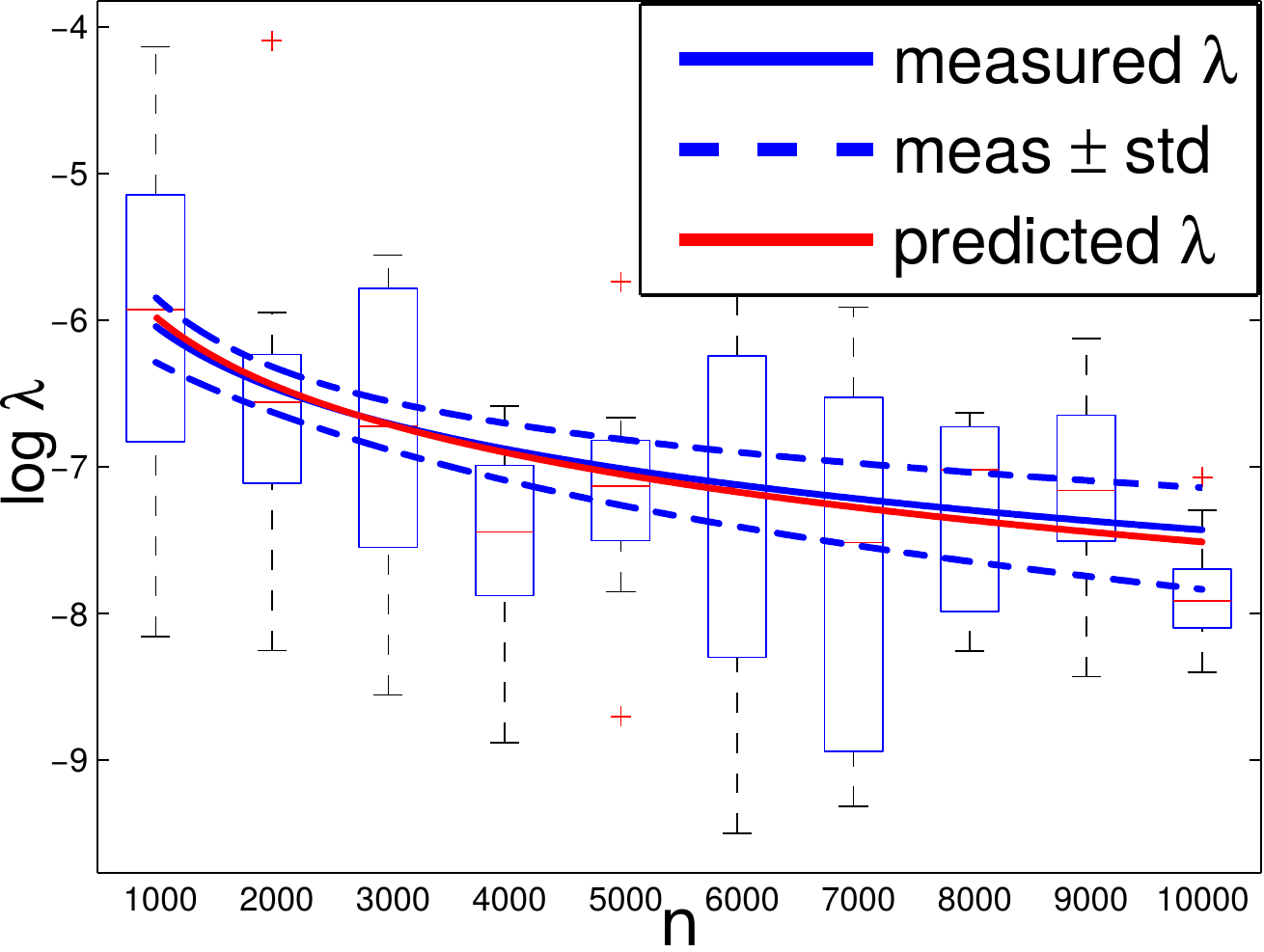}
\hfill
\includegraphics[width=0.32\textwidth,height=0.23\textwidth]{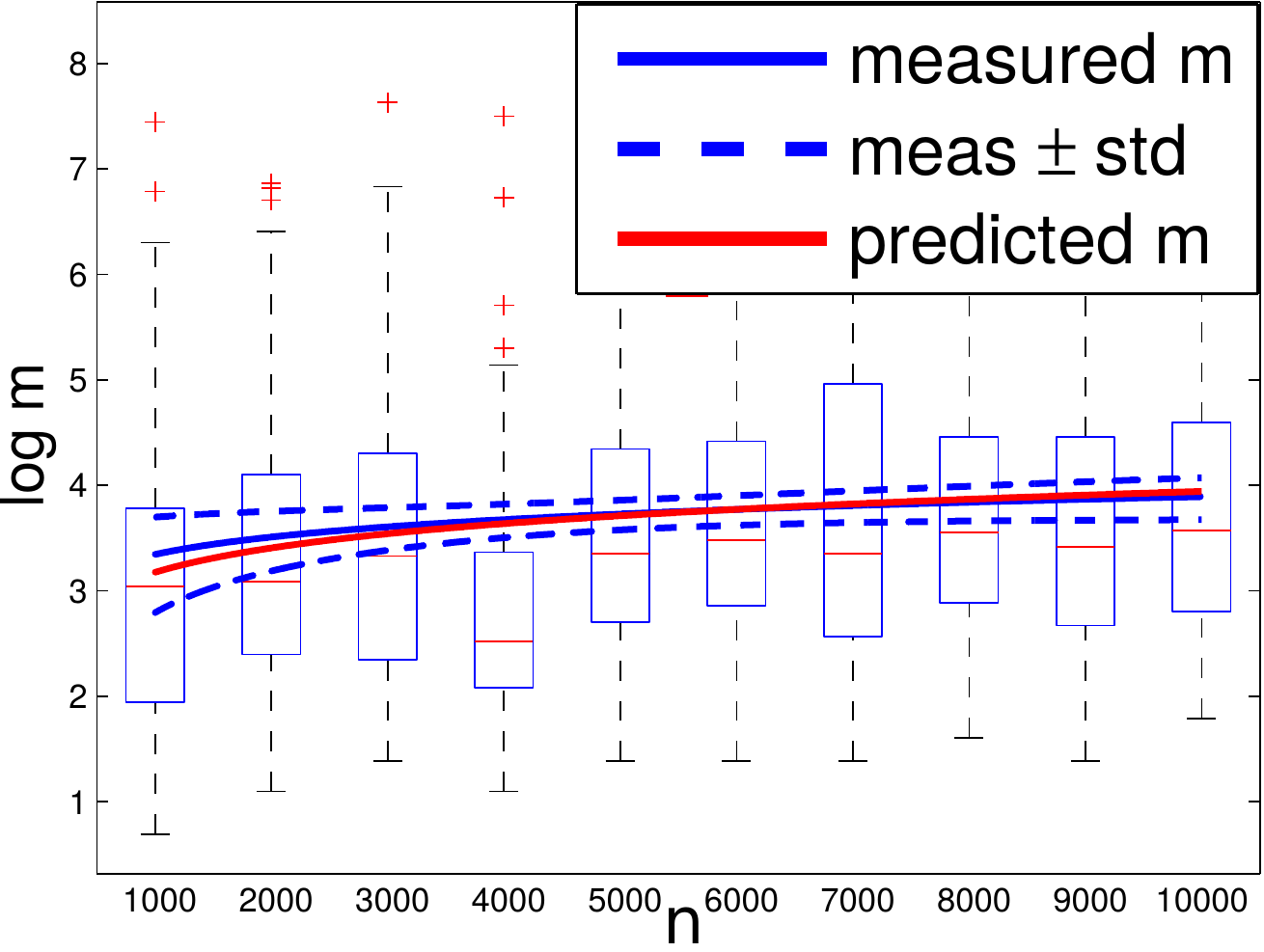}\\
\includegraphics[width=0.32\textwidth,height=0.23\textwidth]{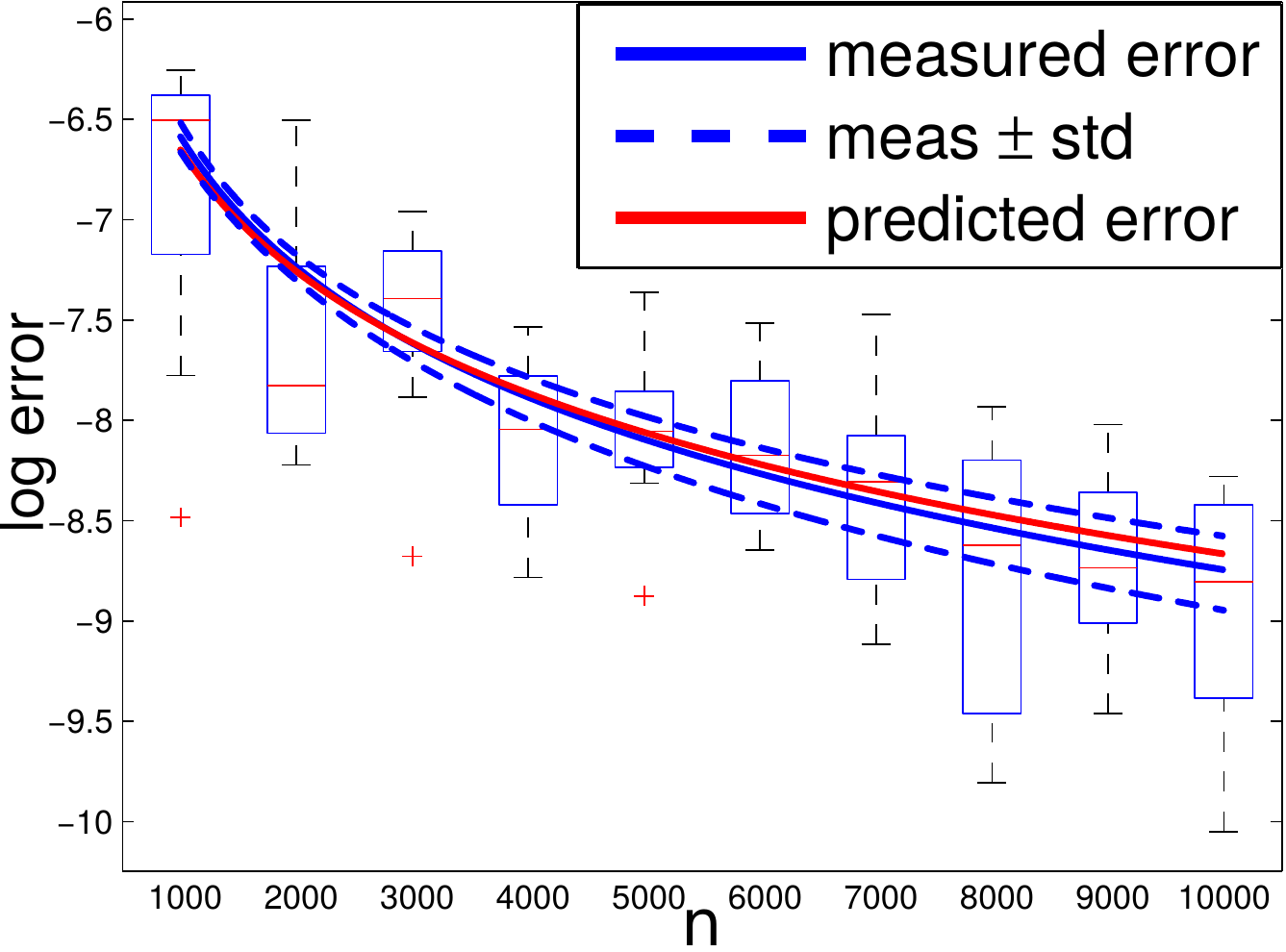}
\hfill
\includegraphics[width=0.32\textwidth,height=0.23\textwidth]{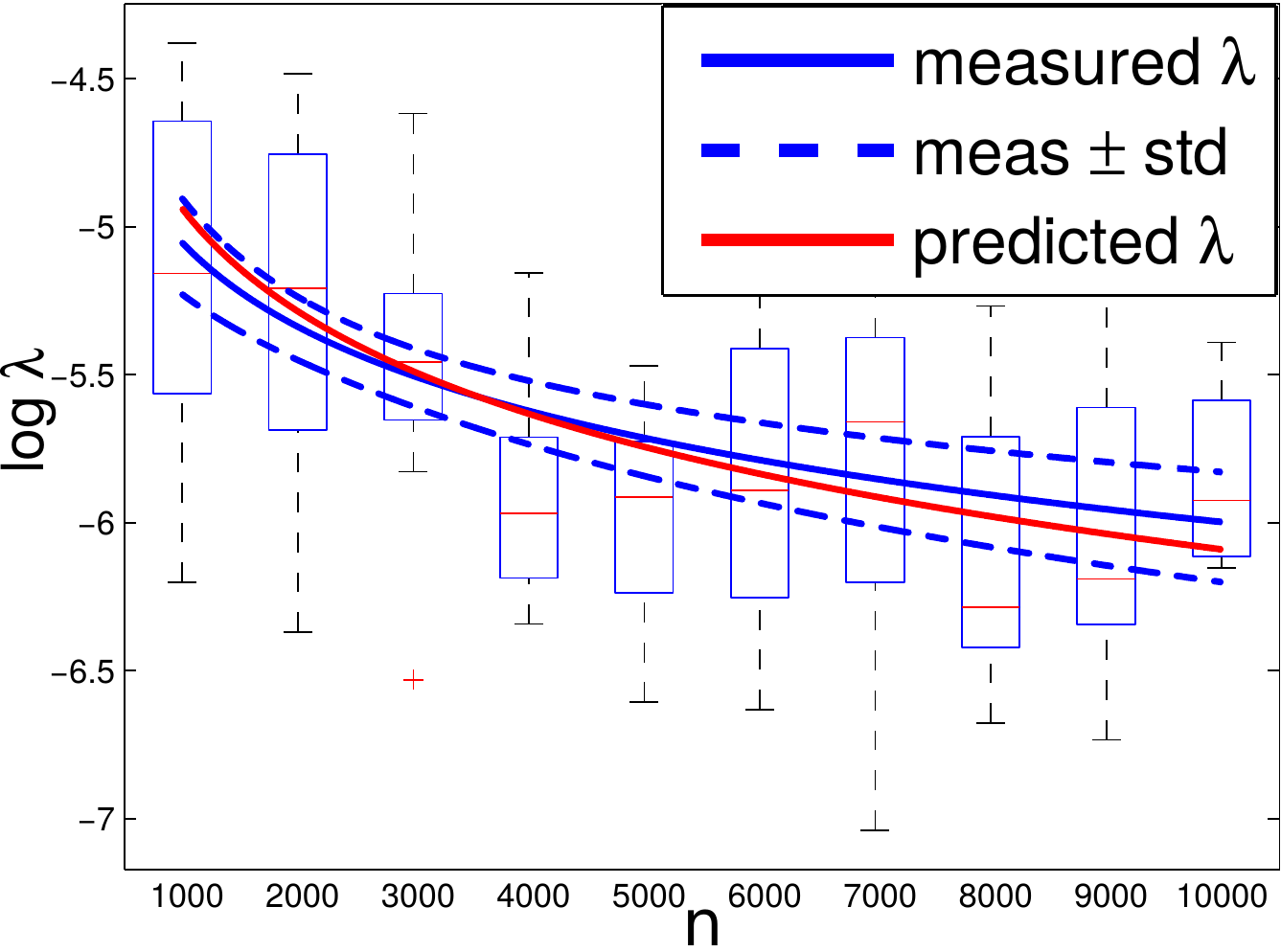}
\hfill
\includegraphics[width=0.32\textwidth,height=0.23\textwidth]{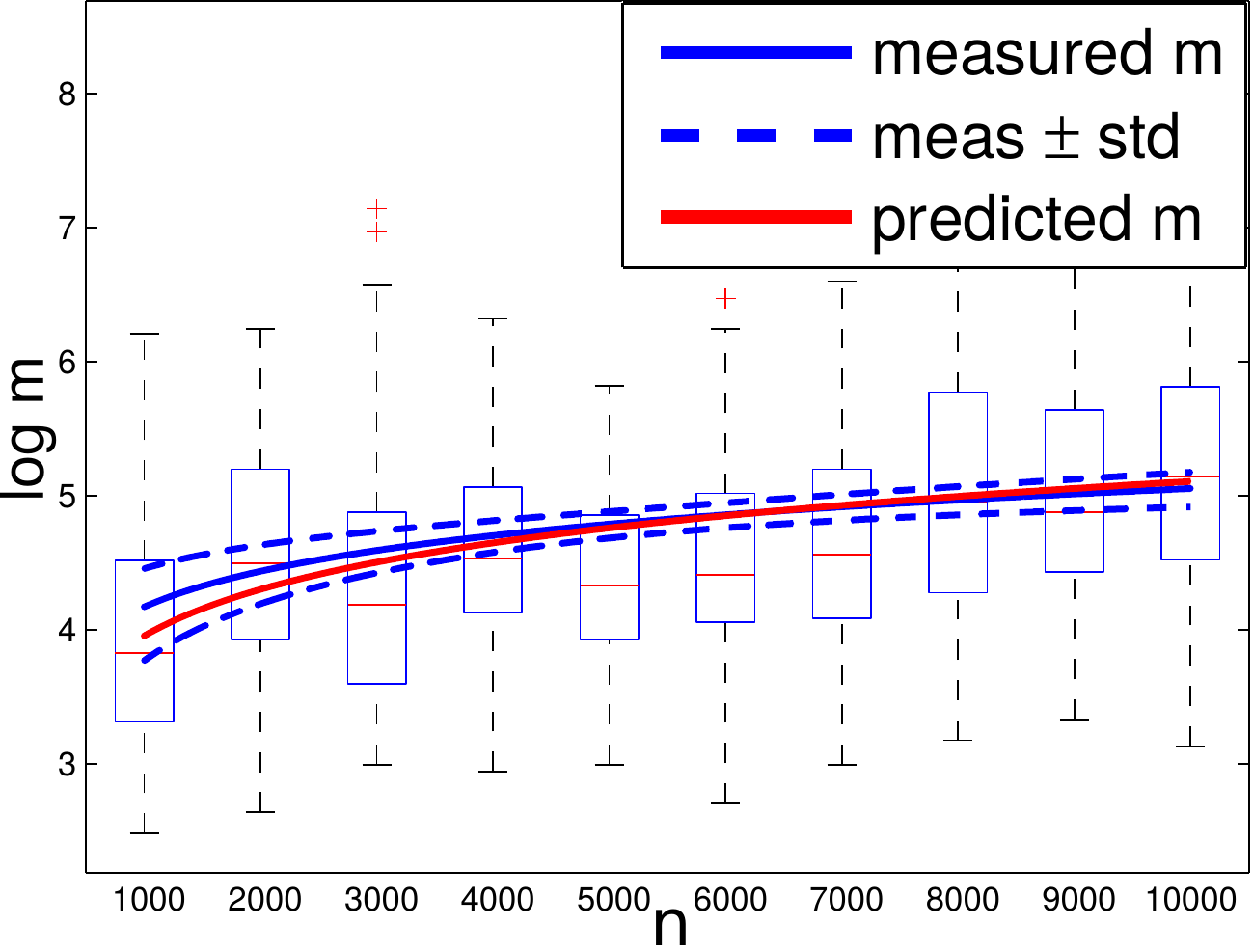}
\caption{Comparison of theoretical and simulated rates for: excess risk ${\cal E}(\tilde{f}_{\la,m}) - {\cal E}(f_\rho)$, $\la$, $m$, with respect to $n$ (100 repetitions). Model parameters $r = 11/16, \gamma = 1/8$ (top), and $r = 7/8, \gamma = 1/4$ (bottom).\label{fig:sim-1}}
\vspace{-0.3cm}
\end{figure}
\subsection{Comparison with Previous Works}\label{sect:previous-results}
Most theoretical works on random features have focused on the approximation of the kernel function 
$K$  in~\eqref{eq:kerrf} by $\tilde K(x,x')=\frac 1 m \sum_{j=1}^m \psi(\omega_j, x)\psi(\omega_j, x')$, $x,x'\in \X$, see for example 
 \citep{conf/nips/RahimiR07,raginsky2009locality,rahimi2009weighted}, and \citep{sriperumbudur2015optimal} for more refined results.
 While the question of kernel approximation is interesting in its own right, 
 it does not directly yield information on generalization properties of learning with random features.
%
%

Much fewer works have analyzed this latter  question \citep{journals/jmlr/CortesMT10,rahimi2009weighted,bach2015}.
In particular, \citep{rahimi2009weighted} is the first work to study the generalization properties of random features, considering the empirical risk minimization on the space
\eqals{ 
\hh_{R,\infty} = \Big\{f(\cdot) = \int\beta(\omega)\psi(\omega,\cdot)d\theta(\omega) ~\Big|~ \nor{\beta}_\infty \leq R, \beta: \Omega \to \R\Big\},
}
for a fixed $R > 0$. The work in \citep{rahimi2009weighted} proves the following bound 
\begin{equation}\label{benbo}
{\cal E}(\tilde{f}_n) - \inf_{f \in \hh_{R,\infty}} {\cal E}(f)\lesssim R n^{-1/2}
\end{equation}
for a number of random features $m_n \approx n$. A corresponding bound on  the excess risk can be derived considering
$$
{\cal E}(\tilde{f}_n)
- {\cal E}(f_\rho)= 
({\cal E}(\tilde{f}_n) - \inf_{f \in \hh_{R,\infty}} {\cal E}(f))+(\inf_{f \in \hh_{R,\infty}}{\cal E}(f) - {\cal E}(f_\rho))
$$ 
however in this case the approximation error needs be taken into account and  the bound in~\eqref{benbo}  leads to slow, possibly suboptimal rates. Most importantly it does not allow to derive results for a number of feature smaller than the number of points. Analogous results are given in \citep{bach2015} but   replacing  $\hh_{R, \infty}$ with a larger space corresponding to a ball in the RKHS induced by a random features kernel.  This latter paper further explores the potential benefit of non uniform sampling. 
Theorem~\ref{thm:optimal-rates-RF-RKLS} sharpen these latter results providing faster rates.

In this sense, our results are close to those for \Nystrom{} regularization given in
\citep{rudi2015less}. A graphical comparison of these results is given in Figure~\ref{fig:ny-vs-randf} and suggests that while in the worst-case  \Nystrom{} and random features behave similarly, in general \Nystrom{} methods could naturally adapt to the data distribution. However a definitive comparison would ultimately rely on deriving computational lower bounds giving the minimal possible number of features to achieve optimal rates.

\section{Numerical results}\label{sec:exp}
In this section we provide several numerical experiments complementing the theoretical results. First of all we perform some numerical simulations to validate the rates derived theoretically and verify the number of features needed for optimal rates since theoretical lower bounds are missing. 
We consider a model where the excess risk can be computed analytically. In particular, we select $\X = \Omega = [0,1]$, with both $\theta$ and $\rhox$ uniform densities, with $\rhox$ the marginal of $\rho$ on $\X$. 
Given  $\gamma \in (0,1/2), r \in [1/2,1]$ and denoting with $K_q$ the spline kernel of order $q > 1$ (\citep{Wahba/90}, Equation~2.1.7),  we define $\psi(\omega,x) = K_{1/(2\gamma)}(\omega,x)$,
$f_\rho(x) = K_{r/\gamma + 1/2 +\epsilon}(x,x_0)$, for $\epsilon > 0, x_0 \in \X$ and $\rho(y|x)$ a Gaussian density centered in $f_\rho(x)$. This setting can be shown to satisfy Assumptions~\ref{ass:noise}-\ref{ass:kerrf-sep} and \ref{ass:omegas}, with $\alpha = \gamma$. We first compute the KRLS estimator with $n \in \{10^3,\dots,10^4\}$ and select the $\la$ minimizing the excess risk (computed analytically). 
Thus, we compute the RF-KRLS estimator and select the number of features $m$ needed to obtain an excess risk within $5\%$ of the one by KRLS.
In Figure~\ref{fig:sim-1},  the theoretical and estimated behavior of
the excess risk, $\la$ and $m$ with respect to $n$ are reported together with their standard deviation over 100 repetitions. 
The experiment shows that the predictions by Theorem~\ref{thm:optimal-rates-RF-RKLS} are quite accurate, since the probabilistic estimations are within one standard deviation from the values measured in the simulation.

\begin{figure}[t]
\centering
\includegraphics[width=0.45\linewidth]{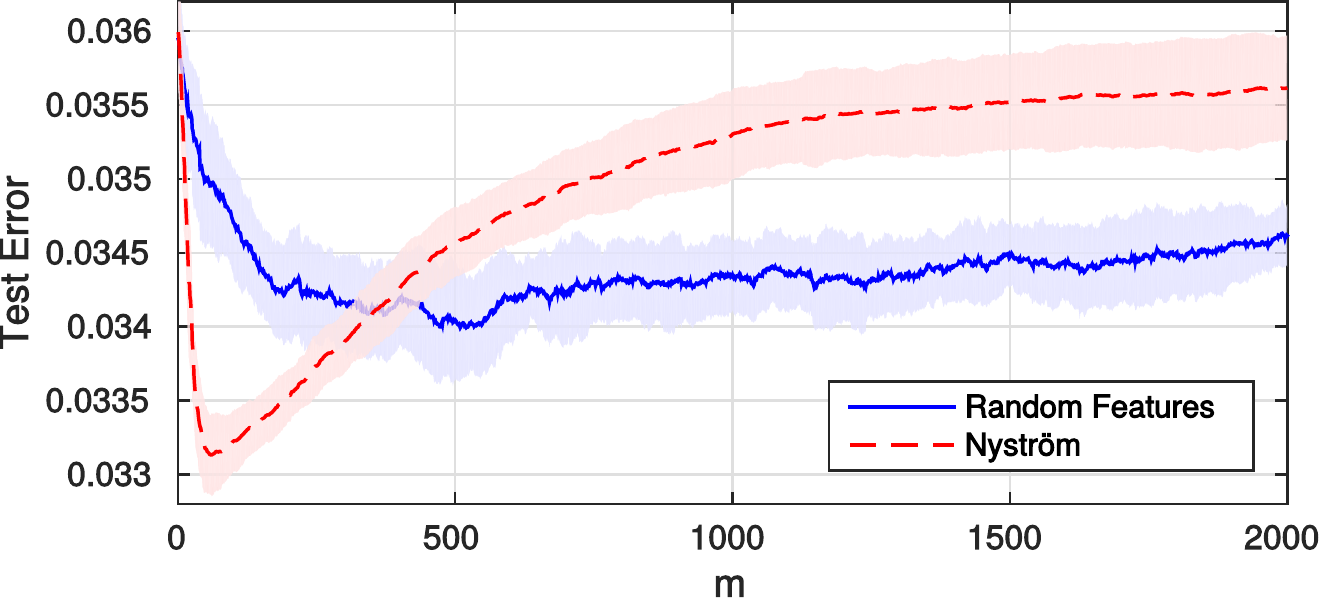}
\includegraphics[width=0.45\linewidth]{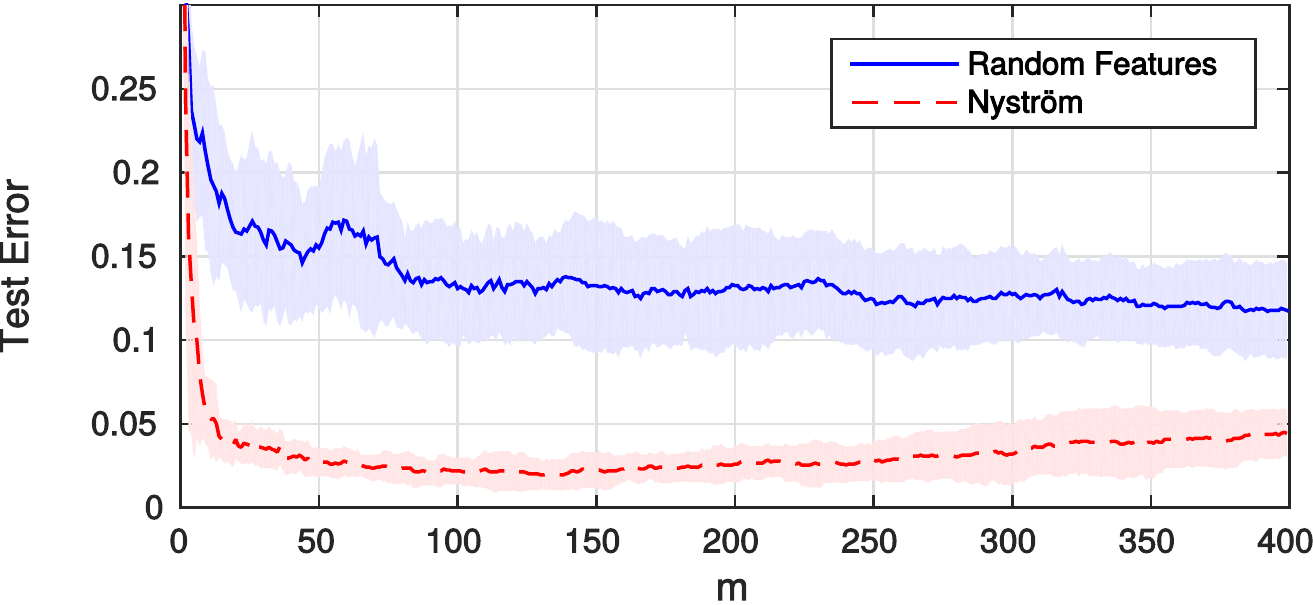}
\caption{Test error with respect to $m$ on {\em pumadyn32nh} with $\lambda = 10^{-9}$ and {\em breastCancer} with $\lambda = 10^{-2}$.\label{fig:pumadyn32nhTest}}
\vspace{-0.1cm}
\end{figure}
\begin{table*}[t]
\begin{center}
\resizebox{0.9\textwidth}{!}{%
\begin{tabular}{cccc||c||cccccc}
\toprule 
{\em Dataset} & $n_{tr}$ & $d$ & $\sigma$ & {\em \textbf{Incremental}} & {\em Incremental} &{\em KRLS}   & {\em Batch } & {\em RF} & {\em Fastfood} & {\em Fastfood}  \\ 
&  & & & {\em \textbf{RF RBF}} & {\em \Nystrom{} RBF} & {\em RBF}  & {\em \Nystrom{} RBF} & {\em RBF} & {\em RBF} & {\em FFT} \\ 
\midrule
{\em Insurance} & 5822 & 85 & 6 & $0.232$ &  $0.232$ & \textbf{0.231} & 0.232 & 0.266 & 0.264 & 0.266\\ 
{\em CPU} & 6554 & 21 & 0.904 & 8.593 & \textbf{$\mathbf{2.847}$} & 7.271 & 6.758 & 7.103 & 7.366 & 4.544 \\  
{\em CT slices} & 42800 & 384 & 5.261 & 22.627& $\mathbf{7.111}$ & NA & 60.683 & 49.491 & 43.858 & 58.425 \\ 
{\em YearPredMSD} & 463715 & 90 & 1.889 & $0.132$ & $\mathbf{0.105}$ & NA & 0.113 & 0.123 & 0.115 & 0.106 \\ 
{\em Forest} & 522910 & 54 & 1.122 &  $0.977$ & $0.964$ & NA & \textbf{0.837} & 0.840 & 0.840 & 0.838 \\
\bottomrule
\hline 
\end{tabular} 
}
\caption{Accuracy of RF-KRLS when cross-validating $m$ vs. state of the art. $d$: Input dimensionality, $\sigma$: Kernel bandwidth. \label{tab:RFaccuracy}}%
\end{center}
\vspace{-0.2cm}
\end{table*}
Secondly, we test the regularization behavior of the number random features.
In Figure~\ref{fig:pumadyn32nhTest} we plot the test error with respect to the number of random features $m$ on two datasets, by using the random features in \citep{conf/nips/RahimiR07} approximating the Gaussian kernel. The experiment shows, for $m$, the typical curve of the bias-variance trade-off associated to a regularization parameter.
The plots in Figure~\ref{fig:pumadyn32nhTest} also contrast random features and \Nystrom{} method, and seem to suggest that \Nystrom{} needs considerably fewer centers to achieve the same error. 

In Table~\ref{tab:RFaccuracy} we test, quantitatively, if considering $m$ a regularization parameter and then cross-validating on it gives a better accuracy. We compared RF-KRLS cross-validated on $\la$ and $m$ with some state of the art methods \citep{conf/nips/WilliamsS00,conf/nips/RahimiR07,conf/icml/LeSS13,rudi2015less} on several small to medium size benchmark datasets, with the Gaussian kernel. The table shows that cross-validating on the number of random features gives a gain in the accuracy of the algorithm.

\part{Incremental Approaches and Lifelong Learning}
\label{part:lifelong}

	\chapter[Generalization Properties of SGM]{Generalization Properties of Stochastic Gradient Methods}
\label{chap:sgd}
		%

\section{Setting}

The stochastic gradient method (SGM), often called stochastic gradient descent,
has become an algorithm of choice in machine learning, because of  its simplicity and small computational cost
especially when dealing with big data sets \citep{bousquet2008tradeoffs}. \\
Despite   its widespread use, the  generalization properties of the variants of SGM used in practice
are relatively little understood.  Most previous works consider generalization properties of SGM with only one pass over the data, see e.g.  \citep{nemirovski2009robust} or \citep{orab14} and references therein, while in practice  multiple passes are usually considered. The effect of multiple passes has  been studied extensively for the optimization  of an empirical objective \citep{boyd2007stochastic}, but the role for generalization is less clear.
In practice, early-stopping of the number of  iterations, for example monitoring a hold-out set error,  is a
strategy often used to regularize. 
Moreover, the  step-size is  typically  tuned to obtain the best results. 
The study in this chapter is a step towards grounding theoretically these commonly used  heuristics. \\
Our starting point are a few recent works considering the generalization properties of different variants of SGM.
One first series of results  focus on least squares, either with one \citep{ying2008online,tarres2014online,dieuleveut2014non}, or  multiple (deterministic) passes over the data \citep{rosasco2015learning}.
In the former case, it is shown that,  in general
 if  only one pass over the data is considered, then
the step-size needs to be tuned to ensure optimal results. In \citep{rosasco2015learning} it is shown that a universal step-size choice can be taken, if multiple passes are considered. In this case, it is the stopping time that needs to be tuned. \\
In our work, we are interested in general, possibly non smooth,  convex loss functions.
The analysis for least squares heavily exploits properties of the loss and does not generalize to this
more general setting. Here, our starting point are the results in \citep{lin2015iterative,hardt2015, orab14}  considering convex loss functions.
In \citep{lin2015iterative},  early stopping of a (kernelized) batch subgradient method is analyzed,  whereas in \citep{hardt2015}
the stability properties of SGM for smooth loss functions are considered in a general stochastic optimization setting
and certain convergence results derived. In \citep{orab14},  a more complex variant of SGM is analyzed and shown to achieve optimal rates.

Since we are interested in analyzing regularization and generalization properties of SGM, we consider a general non-parametric setting.
In this latter setting, the effects of regularization are typically more evident since
it can  directly  affect  the convergence rates.
In this context, the difficulty of a problem is characterized by an assumption on the approximation error. Under this condition, the need for regularization becomes clear.
Indeed, in the absence of other constraints,
%
the good performance of the algorithm relies on a bias-variance trade-off that can be controlled
by suitably choosing the  step-size and/or the number of epochs. These latter parameters can be seen to act as  regularization parameters.
Here, we refer to the regularization as ``implicit'', in the sense that it is achieved neither by
penalization nor by adding explicit constraints.
The two main variants of the algorithm  are the same as in least squares: One pass over the data with tuned step-size, or,
fixed step-size choice and number of passes appropriately tuned. While in principle  optimal parameter tuning requires
explicitly solving a bias-variance trade-off, in practice adaptive choices can be implemented by cross validation.
In this case, both algorithm variants achieve optimal results, but different computations are entailed.
In the first case, multiple single pass SGM need to be considered with different step-sizes, whereas  in the second case, early stopping is used.
Experimental results, complementing the theoretical analysis are given and provide further
insights on the properties of the algorithms.

The rest of the chapter is organized as follows. In Section \ref{sec:setting}, we describe the supervised learning setting and the algorithm, and in Section \ref{sec:theory} we state and discuss our main results.
The detailed proofs can be found in the supplementary material of \citep{lin2016}.
In Section \ref{sec:simulations} we present some numerical simulations on real datasets.

{\bf Notation}. For notational simplicity, $[n]$ denotes $\{1,2,\cdots,n\}$ for any $n\in \mN$.
The notation $a_k\lesssim b_k$ means that there exists a universal constant $C>0$ such that $a_k \leq Cb_k$ for all $k\in \mN.$ We denote by  $\lceil a \rceil$ the smallest integer greater than $a$ for any given $a \in \mR.$

\section{Learning with SGM}\label{sec:setting}
In this section, we introduce the SGM algorithm in the learning setting of interest.
\paragraph{Learning Setting.}
In the following, we use the definitions of data space, probabilistic data model and loss function introduced in Chapter \ref{Chap:SLT}.
The goal is to find a function minimizing the expected risk
given  a sample (training set) $S_n = S =\{z_i=(x_i, y_i)\}_{i=1}^n$ of size $n\in\mN$ independently drawn according to $\rho$.
It is important to note that in this Chapter we deal with spaces of functions which are linearly parameterized.
Consider a possibly non-linear feature map $\Phi: \X\to \mcF$, mapping the data space in $\mR^p$, $p\le \infty$, or more generally in a (real separable) Hilbert space  space with inner product $\scal{\cdot}{\cdot}$ and norm $\|\cdot\|$.
Then, for $w\in \mcF$ we consider functions  of the form
\be\label{eq:linfun}
f_w(x)=\scal{w}{\Phi(x)}, \quad \forall x\in \X.
\ee
Examples of the above setting include the case of infinite dictionaries, $\phi_j: \X \to \mR$, $j=1, \dots$,  so that $\Phi(x)=(\phi_j(x))_{j = 1}^\infty$, for all $x \in \X,$ $\mcF=\ell_2$
and \eqref{eq:linfun} corresponds  to $f_w=\sum_{j=1}^pw^j \phi_j$. Also, this setting includes, and indeed is equivalent to considering, functions defined by a  positive definite kernel $K: \X \times \X \to \mR$, in which case $\Phi(x)=K(x, \cdot) $, for all $x\in \X$, $\mcF=\mcH$ and \eqref{eq:linfun} corresponds to the reproducing property
\be\label{reproducingProperty}
f_w(x) = \langle w, K(x,\cdot) \rangle, \quad \forall x \in \X.
\ee
In the following, we assume the feature map to be measurable  and define expected and empirical risks over functions of the form \eqref{eq:linfun}. For notational simplicity,
we write $\mcE(f_w)$ as $\mcE(w)$, and $\En(f_w)$  as $\En(w)$.

\paragraph{Stochastic Gradient Method.}
For any fixed $y\in \Y$, assume the univariate loss function $\ell(y, \cdot)$ on $\mR$ to be  convex, hence its left-hand derivative $\ell'_- (y, a)$ exists at every $a\in \mR$ and is non-decreasing.

\begin{alg}\label{alg:SIGD}
Given a sample $S$,
 the stochastic gradient  method (SGM) is defined by  $w_1= 0$ and
 \be\label{SIGD}
w_{t+1}=w_t - \eta_t  \ell'_- (y_{j_t}, \langle w_t, \Phi(x_{j_t}) \rangle ) \Phi(x_{j_t}),\, t=1, \ldots, \bar t,
\ee
 for a non-increasing sequence of step-sizes $\{\eta_t >0 \}_{t \in \mN}$ and a stopping rule $\bar t \in \mN$.
 Here, $j_1,j_2,\cdots,j_{\bar t}$ are independent and identically distributed (i.i.d.) random variables from the uniform distribution on $[n]$.
The (weighted) averaged iterates are defined by
\begin{equation}\label{aveite}
\overline{w}_t = { \sum_{k=1}^{t} \eta_k w_k /a_t}, \quad a_t = { \sum_{k=1}^t \eta_k}, \quad t=1, \dots, \bar t.
\end{equation}
\end{alg}
Note that $\bar t$ may be greater than $n$, indicating that we can use the sample more than once.
We shall write $J(t)$ to mean $\{j_1,j_2,\cdots,j_t\}$, which will be also abbreviated as $J$ when there is no confusion.

The main purpose of the chapter is to  estimate the expected excess risk of the last iterate
 $$\mE_{S,J}[{\mathcal E}(w_{\bar t}) -
 \inf_{w\in \mcF}{\mathcal E}(w)],$$
 or similarly the   expected excess risk of the averaged iterate
 $\overline{w}_{\bar t}$, and study how different parameter settings in Algorithm~\ref{alg:SIGD} affect the estimates.

\section{Implicit Regularization for SGM}\label{sec:theory}
In this section, we present and discuss our main results.
We begin in Subsection \ref{subsec:convergence} by giving a universal convergence  result
and then  provide  finite sample bounds for smooth loss functions in Subsection \ref{subsec:smoothBound}, and for non-smooth functions in Subsection \ref{subsec:nonsmooth}.
As  corollaries of these results we derive different implicit regularization strategies for SGM.


\subsection{Convergence}\label{subsec:convergence}
We begin presenting a convergence result, involving conditions on both  the step-sizes and the number of  iterations.  We need some basic  assumptions.

\begin{as}\label{as:Boundness}
There holds
\be\label{boundedkernel}
\kappa=\sup_{x\in X}\sqrt{\langle \Phi(x),\Phi(x) \rangle }<\infty.
\ee Furthermore, the loss function
 is convex with respect to its second entry, and
$|\ell|_0 :=\sup_{y\in \Y} \ell(y, 0) < \infty$.
Moreover, its left-hand derivative $\ell'_- (y, \cdot)$ is bounded:
\be\label{boundedDeriviative} \left|\ell'_- (y, a)\right| \leq a_0, \qquad \forall a\in \mR, y\in  \Y.\ee
\end{as}
The above   conditions on $K$ and $\ell$ are common in statistical learning theory \citep{steinwart2008support,cucker2007learning}. They are satisfied for example by the Gaussian kernel and the hinge loss  $\ell(y,a)= |1-ya|_+=\max\{0, 1-ya\}$
  or the logistic loss  $\ell(y,a) = \log (1 + \mathrm{e}^{-ya})$ for all $a\in \mR$, with a bounded domain $\Y.$

The bounded derivative condition \eref{boundedDeriviative} is implied by the requirement on the loss function to be  Lipschitz in its second entry, when $\Y$ is a bounded domain. Given these  assumptions, the following result holds.

\begin{thm}
  \label{thm:convergence} If Assumption \ref{as:Boundness} holds,
   then
  \be
  \lim_{n \to \infty} \mE[\mcE(\overline{w}_{t^*(n)})] - \inf_{w\in\mcF} \mcE(w) = 0,
  \ee
  provided the sequence $\{\eta_k\}_{k}$ and the stopping rule $t^*(\cdot): \mN \to \mN$ satisfy \\
(A)  $\lim_{n\to \infty} {\sum_{k=1}^{t^*(n)} \eta_k \over n} = 0,$ \\
(B)  and $\lim_{n\to \infty} {1 + \sum_{k=1}^{t^*(n)} \eta_k^2 \over \sum_{k=1}^{t^*(n)}\eta_k} = 0.$
\end{thm}

Conditions (A) and (B) arise from the analysis of suitable sample,  computational, and approximation errors (see the appendix of \citep{lin2016}).
Condition (B) is similar to the one required by stochastic gradient methods \citep{bertsekas1999nonlinear,boyd2003subgradient,boyd2007stochastic}. The difference is that here
the limit is taken with respect to the number of points, but the number of passes on the data can be larger than one.

Theorem \ref{thm:convergence} shows that in order to achieve consistency, the step-sizes and the running iterations need to be appropriately chosen. For instance, given $n$ sample points for SGM with one pass\footnote{We slightly abuse the term ``one pass'' here. In practice, at each pass/epoch, that is   $n$ iterations, a sample point is selected randomly from the whole dataset, either with or without repetition.},
 i.e., $t^*(n) = n$, possible choices for the step-sizes are $\{\eta_k= n^{-\alpha}: k\in [n]\}$ and $\{\eta_k= k^{-\alpha}: k\in [n]\}$ for some $\alpha \in (0,1).$ One can also fix the step-sizes a priori, and then run the algorithm with a suitable stopping rule $t^*(n)$.

These different parameters choices  lead to different implicit  regularization strategies, as we  discuss next.

\subsection{Finite Sample Bounds for Smooth Loss Functions}\label{subsec:smoothBound}
In this subsection, we  give explicit finite sample bounds for smooth loss functions, considering  a  suitable assumption on the approximation error.
 \begin{as}\label{as:approximationerror}
 The approximation error associated to the triplet $(\rho, \ell, \Phi)$ is defined by
\begin{equation}\label{approxerror}
\mathcal{D}(\lambda) = \inf_{w\in \mcF} \left\{ \mathcal{E}(w)  + {\lambda \over 2} \|w\|^2\right\} - \inf_{w\in\mcF}  \mcE(w), \quad \forall \lambda \geq 0.
\end{equation}
We assume that for some $\beta \in (0,1]$ and $c_{\beta}>0$, the approximation error satisfies
\be
\mathcal{D}(\lambda) \leq c_{\beta}\lambda^{\beta}, \qquad \forall \ \lambda> 0.
\label{decayapprox}
\ee
\end{as}

Intuitively,  condition \eref{decayapprox} quantifies how hard it is to achieve the infimum of the expected risk. In particular, it is satisfied with $\beta=1$ when $\exists w^* \in \mcF$ such that $\inf_{w \in \mcF} \mcE(w) = \mcE(w^*)$.
More formally, the condition  is related to classical quantities in approximation theory, such as K-functionals and interpolation spaces \citep{steinwart2008support,cucker2007learning}.
The following remark is important for later discussions.
\begin{rem}[SGM and Implicit Regularization]
Assumption \ref{as:approximationerror}
 is standard in statistical learning theory when analyzing  Tikhonov regularization \citep{cucker2007learning,steinwart2008support}. In this view, our results show
that SGM can implicitly implement a form of Tikhonov regularization  by controlling
the step-size and the number of epochs.
\end{rem}

%
%
%
%

A further  assumption relates to the smoothness of the loss, and is satisfied for example by the  logistic loss.
\begin{as}\label{as:smooth}
 For all $y\in \Y$,  $\ell(y,\cdot)$ is differentiable and $\ell'(y,\cdot)$ is Lipschitz continuous with a constant $L>0$, i.e.
$$ |\ell'(y,b) - \ell'(y,a)| \leq L|b-a|, \quad \forall a,b\in \mR. $$
\end{as}
The following result characterizes the excess risk of both the last and the average  iterate
for any fixed step-size and stopping time.
%

\begin{thm}\label{thm:errorSmooth}
  If Assumptions \ref{as:Boundness}, \ref{as:approximationerror} and \ref{as:smooth} hold and  $\eta_t \leq 2/(\kappa^2 L)$ for all $t\in \mN$, then for all $t \in \mN,$
  \be\begin{split}
  &\mE[\mcE(\overline{w}_t) - \inf_{w\in\mcF}  \mcE(w)] \\
   \lesssim &  {\sum_{k=1}^t \eta_k \over n} + {\sum_{k=1}^t \eta_k^2 \over \sum_{k=1}^t \eta_k} +  \left( {1 \over \sum_{k=1}^t \eta_k} \right)^{\beta}\end{split}
  \ee
and
   \be\begin{split}
  \mE[\mcE(w_{t}) - \inf_{w\in\mcF}  \mcE(w)] \lesssim {\sum_{k=1}^t \eta_k \over n} \sum_{k=1}^{t-1} \frac{\eta_k}{\eta_t(t-k)}  \\
   + \left(\sum_{k=1}^{t-1} \frac{\eta_k^2}{\eta_t(t-k)}+ \eta_t\right) + {\left(\sum_{k=1}^t \eta_k\right)^{1-\beta} \over \eta_t t}.
  \end{split}\ee
\end{thm}
The proof of the above result follows  more or less directly combining ideas and results in \citep{lin2015iterative,hardt2015} (as reported in the appendix of \citep{lin2016}).
The constants in the bounds are omitted, but given explicitly in the proof.
While the error bound for the weighted average looks more concise than the one  for the last iterate,  interestingly, both  error bounds  lead to similar generalization properties.

The error bounds are composed of three terms related to sample error, computational error, and approximation error.  Balancing these three error terms to achieve the minimum total error bound leads to optimal choices for the step-sizes $\{\eta_k\}$ and total number of iterations $t^*.$
In other words, both the step-sizes $\{\eta_k\}$ and the number of iterations $t^*$ can play the role of a  regularization parameter.
Using the above theorem, general results for step-size $\eta_k = \eta t^{-\theta}$ with some $\theta \in [0,1),\eta= \eta(n)>0$ can be found (see Proposition 3 in the appendix of \citep{lin2016}).
Here, as corollaries we provide four different parameter choices
to obtain the best bounds, corresponding to four different regularization strategies.

The first two corollaries correspond to fixing the step-sizes a priori and using the number of iterations as a regularization parameter.
In the first result, the step-size is constant and depends on the number of sample points.

\begin{corollary}\label{cor:smoothExplicitC}
If  Assumptions \ref{as:Boundness}, \ref{as:approximationerror} and \ref{as:smooth} hold and
  $\eta_t = \eta_1/\sqrt{n}$  for all $t\in \mN$
 for some  positive constant $\eta_1 \leq 2/(\kappa^2L)$,  then for all $t \in \mN,$ and $g_t = \overline{w}_t$ (or $w_t$),
\be\label{smoothExplicitC}
  \mE [\mcE(g_t) - \inf_{w\in\mcF}  \mcE(w)]
  \lesssim { t \log t  \over \sqrt{n^3} } +  { \log t \over \sqrt{n}} + \left({ \sqrt{n} \over t}\right)^{\beta} .
  \ee
In particular, if we choose $t^* = \lceil n^{\beta+3 \over 2(\beta+1)} \rceil,$
\be\label{optimalboundsSmooth}
  \mE[\mcE(g_{t^*}) - \inf_{w\in\mcF}  \mcE(w)] \lesssim   n^{-{\beta \over \beta+1}}\log n .
  \ee
\end{corollary}
In the second result the step-sizes decay with the iterations.

\begin{corollary}\label{cor:smoothExplicit}
  If Assumptions \ref{as:Boundness}, \ref{as:approximationerror} and \ref{as:smooth} hold and
  $\eta_t = \eta_1/\sqrt{t}$  for all $t\in \mN$
  with some  positive constant $\eta_1 \leq 2/(\kappa^2L)$, then for all $t \in \mN,$ and $g_t = \overline{w}_t$ (or $w_t$),
\be\label{smoothExplicit}
  \mE [\mcE(g_t) - \inf_{w\in\mcF}  \mcE(w)]
  \lesssim { \sqrt{t} \log t  \over n} +  { \log t\over \sqrt{t}} + { 1 \over t^{\beta / 2}} .
  \ee
Particularly, when $t^* = \lceil n^{2 \over \beta+1} \rceil,$ we have \eref{optimalboundsSmooth}.
\end{corollary}
	

In both the above  corollaries  the step-sizes are fixed a priori, and  the  number of iterations becomes the  regularization parameter controlling the total error. Ignoring the logarithmic factor, the dominating terms in the bounds  \eref{smoothExplicitC}, \eref{smoothExplicit} are the sample  and approximation errors, corresponding to the first and third terms of the right hand side. Stopping too late may lead to a  large sample error, while stopping too early may lead to a large approximation error. The ideal stopping time arises from a form of bias variance trade-off and requires in general more than one pass over the data. Indeed, if we reformulate the results in terms of number of epochs, we have that   $\lceil n^{1 - \beta \over 2(1 + \beta)} \rceil$ epochs are needed for the constant step-size $\{\eta_t =\eta_1 / \sqrt{n} \}_t$, while
 $\lceil n^{1 - \beta \over 1 + \beta} \rceil$ epochs are needed for the  decaying step-size  $\{\eta_t =\eta_1 / \sqrt{t} \}_t$. These observations suggest in particular that  while both step-size choices achieve the same bounds, the constant step-size can have a computational advantage since it requires less iterations.

 Note that one pass over the data suffices only  in the limit case when $\beta=1$, while in general it will be suboptimal, at least if the step-size is fixed. 
In fact,
Theorem \ref{thm:errorSmooth} suggests that  optimal results could be recovered if the step-size is suitably tuned. The next corollaries show that this is indeed the case.
%
The first result corresponds to a suitably tuned constant step-size.
\begin{corollary}\label{cor:smoothExplicitCSingle}
  If Assumptions \ref{as:Boundness}, \ref{as:approximationerror} and \ref{as:smooth} hold and  $\eta_t = \eta_1 n^{-{\beta \over \beta+1}}$  for all $t\in \mN$  for some  positive constant $\eta_1 \leq 2/(\kappa^2L)$, then for all $t \in \mN,$ and $g_t = \overline{w}_t$ (or $w_t$),
\be\label{smoothExplicitCSingle}
\begin{split}
  &\mE [\mcE(g_t) - \inf_{w\in\mcF}  \mcE(w)] \\
  \lesssim & \, { n^{-{\beta +2 \over \beta+1}} t \log t   } +  { n^{-{\beta \over \beta+1}} \log t } + n^{{\beta^2 \over \beta+1}} t^{-\beta} .
  \end{split} \ee
In particular, we have \eref{optimalboundsSmooth} for $t^*=n.$
\end{corollary}

	The second result corresponds to tuning the decay rate for a decaying step-size.
	
\begin{corollary}\label{cor:smoothExplicitDSingle}
If Assumptions \ref{as:Boundness}, \ref{as:approximationerror} and \ref{as:smooth}
hold and  $\eta_t = \eta_1 t^{-{\beta \over \beta+1}}$  for all $t\in \mN$  for some  positive constant $\eta_1 \leq 2/(\kappa^2L)$,  then for all $t \in \mN,$ and $g_t = \overline{w}_t$ (or $w_t$),
\be\label{smoothExplicitDSingle}
\begin{split}
  &\mE [\mcE(g_t) - \inf_{w\in\mcF}  \mcE(w)] \\
  \lesssim& \,  n^{-1} { t^{{1 \over \beta+1}} \log t }  +  { t^{-{\beta \over \beta+1}} \log t } + { t^{-{\beta \over \beta+1}}} .
  \end{split}\ee
  In particular, we have \eref{optimalboundsSmooth} for $t^* = n.$
\end{corollary}

The above two results confirm that good performances can be attained with only one pass over the data, provided the step-sizes are suitably chosen, that is using  the step-size
as a regularization parameter.


Finally,  the following remark relates the above results to data-driven parameter tuning used in practice.
\begin{rem}[Bias-Variance and Cross Validation]
The above results show how the number of iterations/passes controls a bias-variance trade-off, and in this sense act as a regularization parameter.
In practice, the approximation properties of the algorithm are unknown and the question arises of how the parameter can be chosen.
As it turns out, cross validation can be used to achieve adaptively the best rates, in the sense that the rate in \eref{optimalboundsSmooth} is achieved by cross validation or more precisely by hold-out cross validation.
These results follow by an argument similar to that in Chapter 6 from \citep{steinwart2008support} and are omitted.
\end{rem}

\subsection{Finite Sample Bounds for Non-smooth Loss Functions}\label{subsec:nonsmooth}
Theorem \ref{thm:errorSmooth} holds for smooth loss functions, and it is natural to ask if
a similar result holds for non-smooth losses such as the hinge loss.
%
Indeed, analogous results hold albeit current bounds are not as sharp.
\begin{thm}\label{thm:errorGeneral}
  Instate Assumptions \ref{as:Boundness} and \ref{as:approximationerror}. Then for all $t \in \mN,$
  \be
  \begin{split}
  &\mE[\mcE(\overline{w}_t) - \inf_{w\in\mcF}  \mcE(w)] \\
  \lesssim&   \sqrt{\sum_{k=1}^t \eta_k \over n} + {\sum_{k=1}^t \eta_k^2 \over \sum_{k=1}^t \eta_k} +  \left( {1 \over \sum_{k=1}^t \eta_k} \right)^{\beta} ,
  \end{split}
  \ee
  and
  \be\begin{split}
  &\mE[\mcE(w_{t}) - \inf_{w\in\mcF}  \mcE(w)]  \lesssim\sqrt{\sum_{k=1}^t \eta_k \over n}  \sum_{k=1}^{t-1} \frac{\eta_k}{\eta_t(t-k)} \\
&  + \sum_{k=1}^{t-1} \frac{\eta_k^2}{\eta_t(t-k)} + \eta_t + {\left(\sum_{k=1}^t \eta_k\right)^{1-\beta} \over \eta_t t} .
  \end{split}\ee
\end{thm}

The proof of the above theorem is based on ideas from \citep{lin2015iterative}, where tools from Rademacher complexity \citep{bartlett2003rademacher,meir2003generalization} are employed
(the proof is reported in the appendix of \citep{lin2016}).

Using the above result with concrete step-sizes as those for smooth loss functions, we have the following explicit error bounds and corresponding stopping rules.
\begin{corollary}\label{cor:genExplicitC}
   Instate Assumptions \ref{as:Boundness} and \ref{as:approximationerror}. 
   Let $\eta_t = 1/\sqrt{n}$  for all $t\in \mN$. Then for all $t \in \mN,$ and $g_t = \overline{w}_t$ (or $w_t$),
\be
\begin{split}
  \mE [\mcE(g_t) - \inf_{w\in\mcF}  \mcE(w)] \lesssim { \sqrt{t} \log t  \over n^{3/4} } +  { \log t \over \sqrt{n}} + \left({ \sqrt{n} \over t}\right)^{\beta} .
  \end{split}
\ee
In particular, if we choose $t^* = \lceil n^{2\beta+3 \over 4\beta+2} \rceil,$
\be\label{optimalboundsNonSmooth}
  \mE[\mcE(g_{t^*}) - \inf_{w\in\mcF}  \mcE(w)] \lesssim   n^{-{\beta \over 2\beta+1}} \log n .
  \ee
\end{corollary}

\begin{corollary}\label{cor:genExplicit}
   Instate Assumptions \ref{as:Boundness}  and \ref{as:approximationerror}. Let $\eta_t = 1/\sqrt{t}$  for all $t\in \mN$. Then for all $t \in \mN,$ and $g_t = \overline{w}_t$ (or $w_t$),
\be
  \mE [\mcE(g_t) - \inf_{w\in\mcF}  \mcE(w)] \lesssim { t^{1/4}\log t \over \sqrt{n}}+ { \log t\over \sqrt{t}} + { 1 \over t^{\beta / 2}} .
\ee
In particular, if we choose $t^* = \lceil n^{2 \over 2\beta+1} \rceil,$ there holds \eref{optimalboundsNonSmooth}.
\end{corollary}
From the above two corollaries, we  see that the algorithm with constant step-size $1/\sqrt{n}$ can stop earlier
than the one with decaying step-size $1/\sqrt{t}$ when $\beta\leq 1/2,$ while they have the same convergence rate,
since $n^{2\beta+3 \over 4\beta+2} / n^{2 \over 2\beta+1} = n^{2\beta-1 \over 4\beta+1}.$
Note that the bound in \eref{optimalboundsNonSmooth} is slightly worse than that in \eref{optimalboundsSmooth}, see discussion in Subsection \ref{sec:discussion}.

Similar to the smooth case, we also have the following results for SGM with one pass where regularization is realized by step-size.

\begin{corollary}\label{cor:genExplicitCSingle}
   Instate Assumptions \ref{as:Boundness} and \ref{as:approximationerror}. Let $\eta_t = n^{-{2\beta \over 2\beta+1}}$  for all $t\in \mN$. Then for all $t \in \mN,$ and $g_t = \overline{w}_t$ (or $w_t$),
\be\begin{split}
  &\mE [\mcE(g_t) - \inf_{w\in\mcF}  \mcE(w)] \\
  \lesssim& \, { n^{-{4\beta+1 \over 4\beta+2}}\sqrt{t} \log t  } +  n^{-{2\beta \over 2\beta+1}} \log t  + n^{{2\beta^2 \over 2\beta+1}} t^{-\beta}.
  \end{split}
\ee
In particular, \eref{optimalboundsNonSmooth} holds for $t^* = n.$
\end{corollary}

\begin{corollary}\label{cor:genExplicitSingle}
   Instate Assumptions \ref{as:Boundness}  and \ref{as:approximationerror}. Let $\eta_t = t^{-{2\beta \over 2\beta+1}}$  for all $t\in \mN$. Then for all $t \in \mN,$ and $g_t = \overline{w}_t$ (or $w_t$),
\be
\begin{split}
  &\mE [\mcE(g_t) - \inf_{w\in\mcF}  \mcE(w)] \\
  \lesssim& \, n^{- {1 \over 2}}t^{1 \over 4\beta+2}\log t   + { t^{-{\min(2\beta,1) \over 2\beta+1}} \log t} + t^{-{\beta \over 2\beta+1}}  .
\end{split}
\ee
In particular, \eref{optimalboundsNonSmooth} holds for $t^* = n.$
\end{corollary}

\subsection{Discussion and Proof Sketch}
\label{sec:discussion}

As mentioned in the introduction,  the literature on theoretical  properties of the iteration in Algorithm~\ref{alg:SIGD}
is vast, both in learning theory and in optimization. A first line of works focuses on a single pass and convergence
of the expected risk. Approaches in this sense include classical results in optimization (see \citep{nemirovski2009robust} and references therein), but also approaches based on so called ``online to batch" conversion (see \citep{orab14} and references therein). 
The latter are based on analyzing a  sequential prediction setting  and then on considering the averaged iterate to turn regret bounds in expected risk bounds. A second line of works focuses on multiple passes, but measures the quality of the corresponding iteration in terms of the minimization of the empirical risk.
In this view, the iteration in Algorithm~\ref{alg:SIGD} is seen as an instance of incremental methods for the minimization of objective functions that are sums of a finite, but possibly large, number of terms \citep{bertsekas2011incremental}.  These latter works, while interesting in their own right, do not yield any direct information on the generalization properties of considering multiple passes.

Here,  we follow the approach in \citep{bousquet2008tradeoffs} advocating the combination of statistical and computational errors. The general proof strategy is to consider several intermediate steps to relate the expected risk of the empirical iteration to the minimal expected risk.
The argument we sketch below is a simplified and less sharp version with respect to the one used in the actual proof,
but it is easier to illustrate and still carry some important aspects useful to compare with related results.

Consider an intermediate element $\tilde w\in \mcF$ and decompose the excess risk as
\begin{eqnarray*}
&& {\mathbb E}
\mcE(w_t)-\inf_{w\in \mcF}\mcE =\\
&& {\mathbb E} (\mcE(w_t)- \En(w_t))
+{\mathbb E}(\En(w_t)-\En(\tilde w))\\
&&+{\mathbb E}\En(\tilde w)-\inf_{w\in \mcF} \mcE.
\end{eqnarray*}
The first term in the right hand side is the generalization error of the iterate. The second term
can be seen as a computational error. To discuss the last term, it is useful to consider a few different choices for  $\tilde w$.
Assuming the empirical and expected risks to have  minimizers $\widehat w^*$ and $w^*$,  a possibility is to set $\tilde w=\widehat w^*$, this can be seen to be  the choice made in \citep{hardt2015}. In this case, it is immediate to see that the last term is negligible since,
$$
{\mathbb E}\En(\tilde w)={\mathbb E} \min_{w\in \mcF}\En( w) \le
\min_{w\in \mcF} {\mathbb E}\En( w)
\le \min_{w\in \mcF} {\mathbb E}\mcE( w),
$$
and hence,
$$
{\mathbb E}\En(\tilde w)  -\min_{w\in \mcF} \mcE\le 0.
$$
On the other hand,  in this case the computational error depends on the norm $\|\widehat w ^*\|$ which is in general hard  to estimate.   A more convenient choice  is to set $\tilde w=w^*$. A reasoning similar to the one above shows that the last term is still  negligible and the computational error  can still be controlled depending on $\|w^*\|$. In a non-parametric setting, the existence of a minimizer is not ensured and corresponds to a limit case where there is small approximation error.
Our approach  is then to consider an {\em almost} minimizer of the expected risk with a prescribed accuracy.
Following \citep{lin2015iterative}, we do this introducing Assumption~\ref{as:approximationerror} and
choosing $\tilde w=w_\lambda$ the unique minimizer of $\mcE+\lambda\|\cdot\|^2$, $\lambda>0$.
Then the last term in the error decomposition corresponds to the approximation error.\\
For the generalization error,  the stability results from \citep{hardt2015} provide sharp estimates
for  smooth loss functions and in  the ``capacity independent'' limit, that is under no assumptions on the covering numbers of the considered function space. For this setting, the obtained bound is optimal in the sense that it  matches the best available bound for Tikhonov regularization \citep{steinwart2008support,cucker2007learning}.   For the non-smooth case a standard Rademacher complexity argument can be used and easily extended to be capacity dependent. However, the corresponding bound is not sharp and improvements are likely to hinge on
deriving  better norm estimates  for  the iterates. The question does not seem to be  straightforward and is deferred to a future work.\\
The computational error for the averaged iterates can be controlled using classic arguments \citep{boyd2007stochastic}, whereas for the last iterate the arguments in \citep{lin2015iterative,shamir2013stochastic} are needed. Finally, Theorems~\ref{thm:errorSmooth}, ~\ref{thm:errorGeneral}  result from estimating and balancing the various error terms with respect to the choice of the step-size and number of passes.

We conclude this section with some perspective on the results in the chapter.
We note that  since  the primary goal of this study was to  analyze the implicit  regularization effect of step-size and number of passes, we have considered a very simple iteration.  However, it would be very interesting to consider  more sophisticated, ``accelerated'' iterations \citep{schmidt2013minimizing}, and assess the potential advantages in terms of computational and generalization aspects.
Similarly, we chose to keep our analysis relatively simple, but several improvements can be considered
for example deriving high probability bounds and sharper errors under further assumptions. Some of these improvements are relatively straightforward,
see e.g.  \citep{lin2015iterative}, but others will require non-trivial extensions of results developed for Tikhonov regularization in the last few years. Finally, here we only referred to  a simple cross validation approach to parameter tuning, but
it would clearly be very interesting to find ways to tune parameters  online. A remarkable result in this direction
is derived in  \citep{orab14},  where it is shown that, in the capacity independent setting, adaptive online parameter tuning is indeed possible.



\section{Numerical Simulations} \label{sec:simulations}
\begin{figure}[t!]
    \centering
    \subfigure[]{ 	\label{fig:Adult1000_SIGD1}\includegraphics[width=0.6\textwidth]{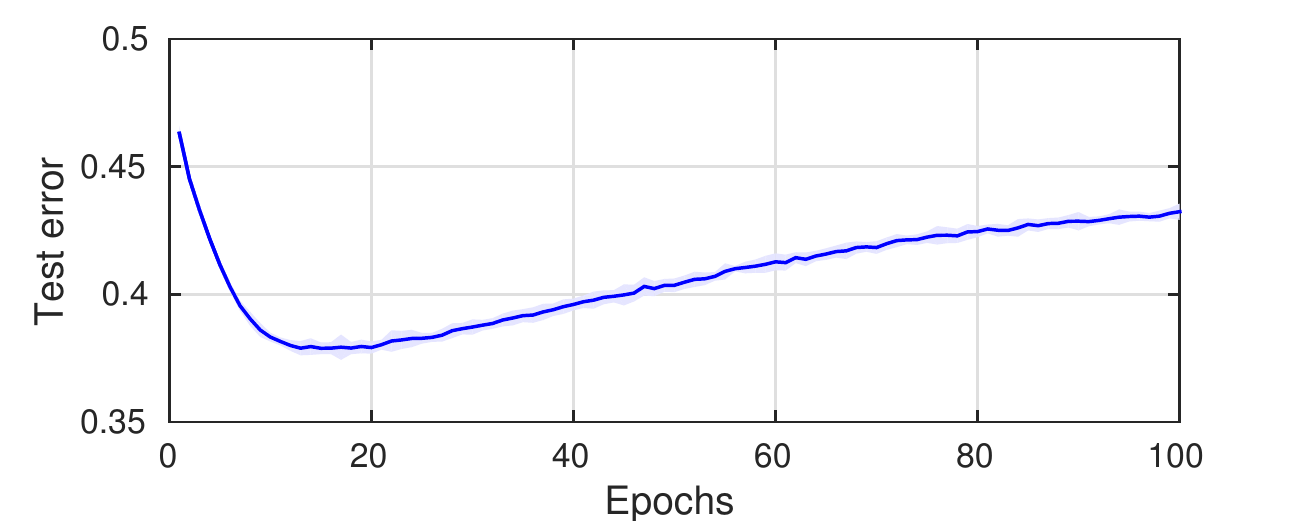}
    }\\
    ~ 
    \subfigure[]{	\label{fig:Adult1000_SIGD2}\includegraphics[width=0.6\textwidth]{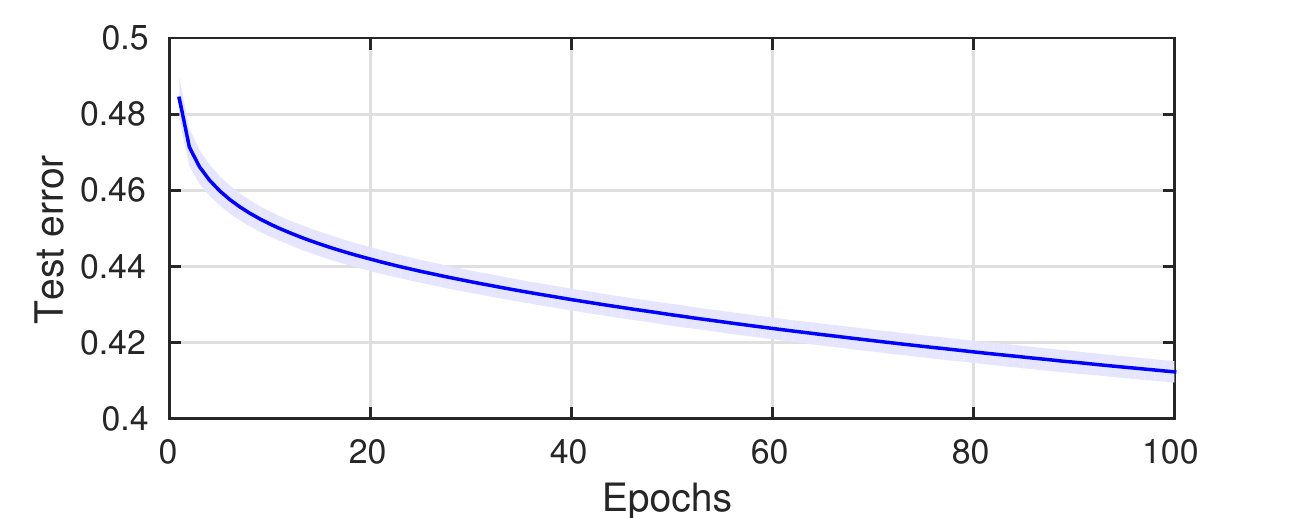}
    }
    \caption{Test error for SIGM with fixed (a) and decaying (b) step-size with respect to the number of epochs on {\em adult} ({\em n = 1000}).}
    \label{fig:Adult1000_SIGD}
\end{figure}

We carry out some numerical simulations to illustrate our results.
The experiments are executed 10 times each, on the benchmark datasets\footnote{The datasets can be downloaded from \texttt{archive.ics.uci.edu/ml} and \texttt{www.csie.ntu.edu.tw/\texttildelow cjlin/libsvmtools/} \texttt{datasets/}} reported in Table \ref{tab:dataSpecSGM}, in which the Gaussian kernel bandwidth $\sigma$ used by SGM and SIGM\footnote{In what follows, we name one pass SGM and multiple passes SGM as SGM and SIGM (Stochastic Incremental Gradient Method), respectively.}
 for each learning problem is also shown. Here, the chosen loss function is the hinge loss.
The experimental platform is a server with 12 $\times$ Intel$^\circledR$ Xeon$^\circledR$ E5-2620 v2 (2.10GHz) CPUs and 132 GB of RAM.
Some of the experimental results, as specified in the following, have been obtained by running the experiments on subsets of the data samples chosen uniformly at random.
In order to apply hold-out cross validation, the training set is split in two parts: One for empirical risk minimization and the other for validation error evaluation (80\% - 20\%, respectively).
All the data points are randomly shuffled at each repetition.

\begin{table}
\caption{Benchmark datasets and Gaussian kernel width $\sigma$ used in our experiments.}
\begin{center}

\resizebox{0.5\textwidth}{!}{%
\begin{tabular}{ccccc}
\toprule 
{\em Dataset} & $n$ &  $n_{test}$ & $d$ & $\sigma$ \\
\midrule
{\em breastCancer} & 400 & 169 & 30 & 0.4 \\
{\em adult} & 32562 & 16282 & 123 &    4 \\
{\em ijcnn1} & 49990 & 91701 & 22 &      0.6 \\
\bottomrule
\hline 
\end{tabular} 
}


\end{center}
\label{tab:dataSpecSGM}
\end{table}

\subsection{Regularization in SGM and SIGM}
\label{sec:simulation_regularization}
\begin{figure}[t!]
    \centering
    \subfigure[]{	\label{fig:Adult1000_SGD1}\includegraphics[width=0.6\textwidth]{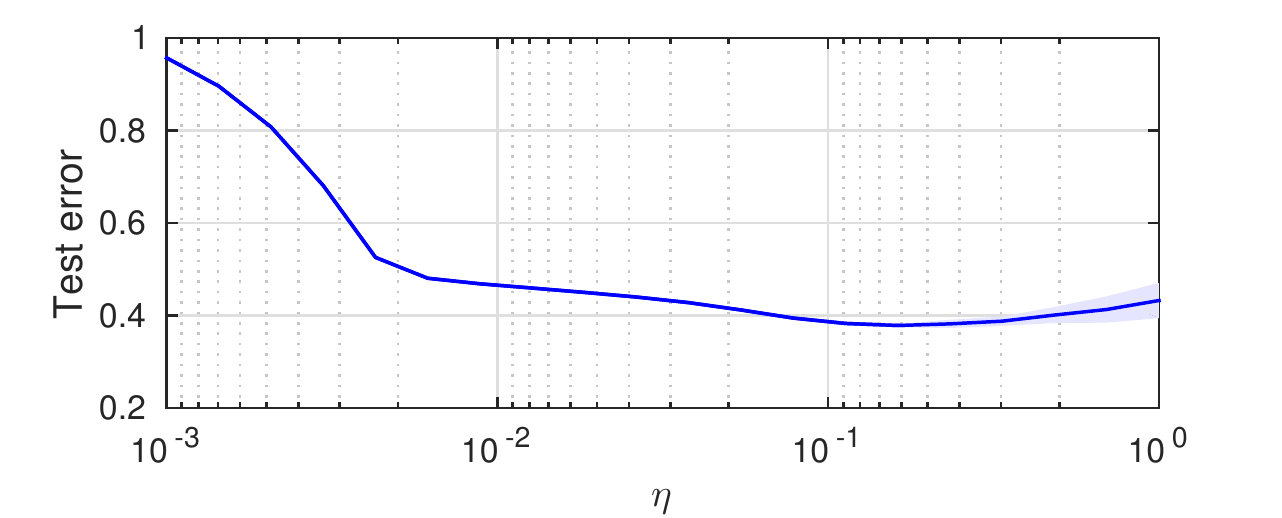}
    }\\
    ~ 
    \subfigure[]{	\includegraphics[width=0.6\textwidth]{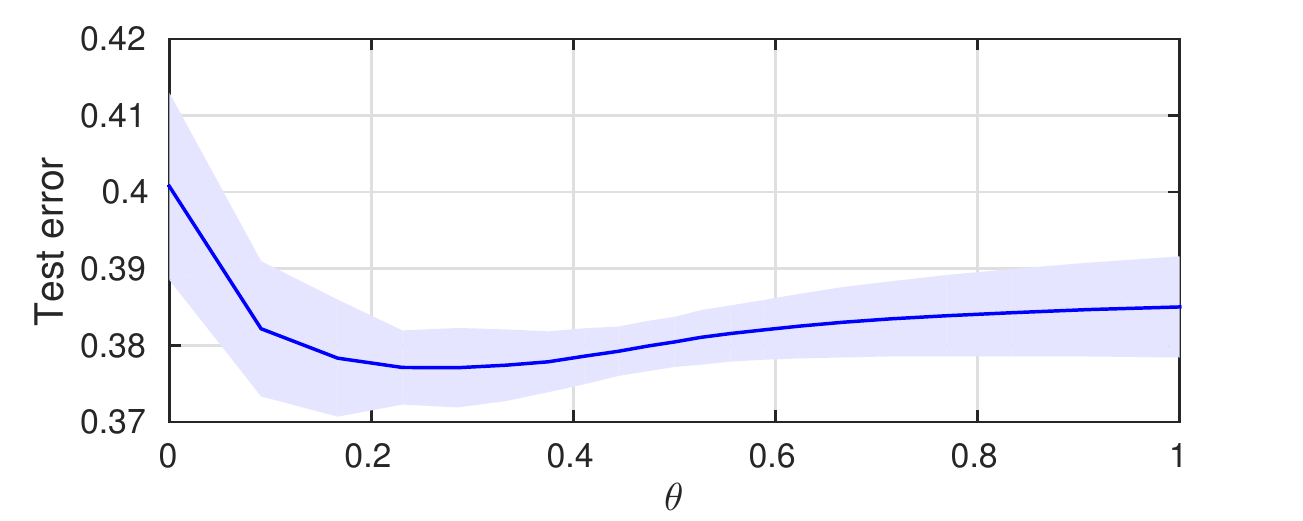}\label{fig:Adult1000_SGD2}
    }
    \caption{Test error for SGM with fixed (a) and decaying (b) step-size cross validation on {\em adult} ({\em n = 1000}).}
	\label{fig:Adult1000_SGD}
\end{figure}

In this subsection, we illustrate four concrete examples showing different regularization effects of the step-size in SGM and the number of passes in SIGM. In all these four examples, we consider the {\em adult} dataset with sample size $n = 1000$. \\
In the first experiment, the SIGM step-size is fixed as $\eta = 1/\sqrt{n}$.
 The test error computed with respect to the hinge loss at each epoch is reported in Figure \ref{fig:Adult1000_SIGD1}.
Note that the minimum test error is reached for a number of epochs smaller than 20, after which it significantly increases, a so-called overfitting regime.
This result clearly illustrates the regularization effect of the number of epochs.
In the second experiment, we
consider SIGM with decaying step-size ($\eta = 1/4$ and $\theta = 1/2$).
As shown in Figure \ref{fig:Adult1000_SIGD2}, overfitting is not observed in the first 100 epochs. In this case, the convergence to the optimal solution appears slower than the one in the fixed step-size case.
In the last two experiments, we consider SGM and show that the step-size plays the role of a regularization parameter.
 For the  fixed step-size case, i.e., $\theta = 0$, we perform SGM with different $ \eta \in ( 0, 1 ]$ (logarithmically scaled).
We plot the errors in Figure \ref{fig:Adult1000_SGD1}, showing that a large step-size ($\eta = 1$) leads to overfitting, while a smaller one (e. g., $\eta = 10^{-3}$) is associated to oversmoothing.
 For the decaying step-size case, we fix $\eta_1 = 1/4$, and run SGM with different $\theta \in [0,1]$. The errors are plotted in Figure \ref{fig:Adult1000_SGD2}, from which we see that the exponent $\theta$ has a regularization effect. In fact, a more ``aggressive'' choice (e. g., $\theta = 0$, corresponding to a fixed step-size) leads to overfitting, while for a larger $\theta$ (e. g., $\theta = 1$) we observe oversmoothing.

\subsection{Accuracy and Computational Time Comparison}
\label{sec:experiments_accuracy}
\begin{table}
\caption{Comparison of SGM and SIGM with cross-validation with decaying (D) and constant (C) step-sizes, in terms of computational time and accuracy. SGM performs cross validation on 30 step-sizes, while SIGM achieves implicit regularization via early stopping.}
\begin{center}

\resizebox{0.9\textwidth}{!}{%
\begin{tabular}{cccccc}
\toprule 
{\em Dataset} & {\em Algorithm} & {\em Step Size} & \specialcell{{\em Test Error}\\ {\em (hinge loss)}} &\specialcell{{\em Test Error}\\ {\em (class. error)}} & \specialcell{{\em Training}\\ {\em Time (s)}}  \\
\midrule
\multirow{2}{*}{\specialcell{ \\ {\em breastCancer}\\ $ n = 400 $ }}
 & $SGM$ & C &  $0.127 \pm 0.022$ & $3.1 \pm 1.1\%$  & 1.7 $\pm$ 0.2    \\
 & $SGM$ & D &  $0.135 \pm 0.024$ & $3.0 \pm 1.1\%$  & 1.4 $\pm$ 0.3    \\
 & $SIGM$ & C & 0.131 $\pm$ 0.023 & $3.2 \pm 1.1\%$ & $1.4 \pm 0.8$  \\
 & $SIGM$ & D & 0.204 $\pm$ 0.017 & $3.9 \pm 1.0\%$ & $1.8 \pm 0.5$  \\
 & $LIBSVM$ & & & $2.8 \pm 1.3\%$ & $0.2 \pm 0.0$  \\
 \hline
\multirow{2}{*}{\specialcell{ \\ {\em adult}\\ $ n = 1000 $ }}
 & $SGM$ & C& 0.380 $\pm$ 0.003 & $16.6 \pm 0.3\%$ & 5.7 $\pm$ 0.6  \\
 & $SGM$ & D&  0.378 $\pm$ 0.002 & $16.2 \pm 0.2\%$ & 5.4$\pm$ 0.3  \\
 & $SIGM$ & C & $0.383 \pm 0.002$ & $16.1 \pm 0.0\%$& $3.2 \pm 0.4$ \\
 & $SIGM$ & D & $0.450 \pm 0.002$ & $23.6 \pm 0.0\%$& $1.6 \pm 0.2$ \\
 & $LIBSVM$ & & & $18.7 \pm 0.0\%$ & $5.8\pm 0.5$  \\
 \hline
\multirow{2}{*}{\specialcell{ \\ {\em adult}\\ $ n = 32562 $ }}
 & $SGM$ & C & 0.342 $\pm$ 0.001 & $15.2 \pm 0.8 \%$ & 320.0 $\pm$ 3.3    \\
 & $SGM$ & D & 0.340 $\pm$ 0.001 & $15.1 \pm 0.7 \%$ & 332.1 $\pm$ 3.3    \\
 & $SIGM$ & C & $0.343 \pm 0.001$ & $15.7 \pm 0.9 \%$& $366.2 \pm 3.9$  \\
 & $SIGM$ & D& $0.364 \pm 0.001$ & $17.1 \pm 0.8 \%$& $442.4 \pm 4.2$     \\
 & $LIBSVM$ & &  & $ 15.3 \pm 0.7\%$ & $ 6938.7\pm 171.7$  \\
 \hline
\multirow{2}{*}{\specialcell{ \\ {\em ijcnn1}\\ $ n = 1000 $ }}
 & $SGM$ & C & 0.199 $\pm$ 0.016 & $8.4 \pm 0.8 \%$ & 3.9 $\pm$ 0.3   \\
 & $SGM$ & D & 0.199 $\pm$ 0.009 & $9.1 \pm 0.1 \%$ & 3.8$\pm$ 0.3   \\
 & $SIGM$ & C & $0.205 \pm 0.010$ & $9.3 \pm 0.5 \%$ & $1.7 \pm 0.4$   \\
 & $SIGM$ & D & $0.267 \pm 0.006$ & $9.4\pm 0.6 \%$ & $2.2 \pm 0.4$   \\
 & $LIBSVM$ & &  & $7.1 \pm 0.7\%$ & $0.6 \pm 0.1$  \\
 \hline
\multirow{2}{*}{\specialcell{ \\ {\em ijcnn1}\\ $ n = 49990 $ }}
 & $SGM$ & C & $0.041 \pm 0.002$ & $ 1.5 \pm 0.0 \%$ & $564.9 \pm 6.3$ \\
 & $SGM$ & D & $0.059 \pm 0.000$ & $ 1.7 \pm 0.0 \%$ & $578.9 \pm 1.8$\\
 & $SIGM$ & C & 0.098 $\pm$ 0.001 & $ 4.7 \pm 0.1\%$ & 522.2 $\pm$ 20.7 \\
 & $SIGM$ & D & 0.183 $\pm$ 0.000 & $ 9.5 \pm 0.0\%$ & 519.3 $\pm$ 25.8 \\
 & $LIBSVM$ & &  & $ 0.9 \pm 0.0\%$ & $ 770.4 \pm 38.5$ \\
\bottomrule
\hline 
\end{tabular} 
}

\label{tab:testSetComparison}
\end{center}
\end{table}
In this subsection, we compare SGM with cross validation and SIGM with benchmark algorithm LIBSVM~\citep{CC01a}, both in terms of accuracy and computational time.
For SGM, with 30 parameter guesses, we use cross-validation to tune the step-size (either setting $\theta=0$ while tuning $\eta$, or setting $\eta = 1/4$ while tuning $\theta$).
For SIGM, we use two kinds of step-size suggested by Section \ref{sec:theory}: $\eta= 1/\sqrt{n}$ and $\theta=0$, or $\eta=1/4$ and $\theta=1/2,$ using early stopping via cross-validation.
  The testing errors with respect to the hinge loss, the testing relative misclassification errors and the computational times are collected in Table \ref{tab:testSetComparison}.

We first start comparing accuracies. The results in Table \ref{tab:testSetComparison} indicate that SGM with constant and decaying step-sizes and SIGM with fixed step-size reach comparable test errors, which are in line with the LIBSVM baseline.
 Observe that SIGM with decaying step-size attains consistently higher test errors, a phenomenon already illustrated in Subsection \ref{sec:simulation_regularization} in theory.

We now compare the computational times for cross-validation.
We see from Table \ref{tab:testSetComparison} that the training times of SIGM and SGM, either with constant or decaying step-sizes, are roughly the same.
We also observe that SGM and SIGM are significantly faster than LIBSVM  on large datasets ({\em adult} with $n = 32562$, and {\em ijcnn1} with $n = 49990$). Moreover, for small datasets ({\em breastCancer} with $n = 400$, {\em adult} with $n = 1000$, and {\em ijcnn1} with $n = 1000$), SGM and SIGM are comparable with or slightly slower than LIBSVM.
		
	\chapter[Incremental Classification]{Incremental Classification: Adding New Classes in Constant Time}
	\label{chap:incclass}
		\section{Setting}

In order for autonomous robots to operate in unstructured environments, several perceptual capabilities are required. Most of these skills cannot be hard-coded in the system beforehand, but need to be developed and learned over time as the agent explores and acquires novel experience.
As a prototypical example of this setting, in this chapter we consider the task of visual object recognition in robotics: Images depicting different objects are received one frame at a time and the system needs to incrementally update the internal model of known objects as new examples are gathered.

In the last few years, machine learning has achieved remarkable results in a variety of applications for robotics and computer vision \citep{Krizhevsky2012,simonyan2014,schwarz2015,eitel2015multimodal}. 
However, most of these methods have been developed for off-line (or ``batch'') settings, where the entire training set is available beforehand. 
The problem of updating a learned model on-line has been addressed in the literature \citep{french1999,Sayed:2008:AF:1370975,duchi2011,goodfellow2013}, but most algorithms proposed in this context do not take into account challenges that are characteristic of realistic lifelong learning applications. 
Specifically, in on-line classification settings, a major challenge is to cope with the situation in which a novel class is added to the model. 
Indeed:
\begin{enumerate}
\item Most learning algorithms require the number of classes to be known beforehand and not grow indefinitely.
\item The imbalance between the few examples (potentially even a single one) of the new class  and the many examples of previously learned classes can lead to unexpected and undesired behaviors \citep{elkan2001}.
\end{enumerate}
More precisely, in this chapter we observe both theoretically and empirically that the new and under-represented class is likely to be ignored by the learning model in favor of classes for which more training examples have already been observed, until a sufficient number of examples are provided also for such class.

Several methods have been proposed in the literature to deal with class imbalance in the batch setting by ``rebalancing'' the misclassification errors accordingly \citep{elkan2001,steinwart2008support,he2009}. 
However, as we will point out in the following, exact rebalancing cannot be applied to the on-line setting without re-training the entire model from scratch every time a novel example is acquired. 
This would incur in computational learning times that increase at least linearly in the number of examples, which is clearly not feasible in scenarios where training data grow indefinitely. 
We propose a novel method that learns incrementally with respect to both the number of examples and classes, and accounts for potential class unbalance. 
Our algorithm builds on a recursive version of RLS for classification (RLSC) \citep{rifkin2002everything,rifkin2003regularized} to achieve constant incremental learning times both when adding new examples to the model and when dealing with imbalance between classes. 
We evaluate our approach on a standard machine learning benchmark for classification and two challenging visual object recognition datasets for robotics. 
Our results highlight the clear advantages of our approach when classes are learned incrementally.

The chapter is organized as follows: Section~\ref{sec:related_work} reviews related work. Section~\ref{sec:classprob_unbal} introduces the learning setting, discusses the impact of class imbalance and considers two approaches to deal with this problem. Section~\ref{sec:backgroundIncClass} reviews the incremental RLS algorithm, from which the approach proposed in this work is derived in Section~\ref{sec:proposed_algorithm}. Section~\ref{sec:incClassExperiments} reports the empirical evaluation of our method.

\section{Related Work}\label{sec:related_work}

{\bf Incremental Learning.} The problem of learning from a continuous stream of data has been addressed in the literature from multiple perspectives. 
The simplest strategy is to re-train the system on the updated training set, whenever a new example is received~\citep{xiao2014,jain2014}. 
The model from the previous iteration can be used as an initialization to learn the new predictor, reducing training times. These approaches require to store all the training data, and to retrain over all the points at each iteration. 
Their computational complexity increases at least linearly with the number of examples.

Incremental approaches that do not require to keep previous data in memory can be divided in {\it stochastic} and {\it recursive} methods. 
Stochastic techniques assume training data to be randomly sampled from an unknown distribution and offer asymptotic convergence guarantees to the ideal predictor~\citep{duchi2011}. 
However, it has been empirically observed that these methods do not perform well when seeing each training point only once, hence requiring to perform ``multiple passes'' \citep{hardt2015,lin2016} over the data, as discussed in Chapter \ref{chap:sgd} for SGD.
This effect has been referred to as the ``catastrophic effect of forgetting'' \citep{french1999} and has recently attracted the attention of the Neural Networks literature \citep{srivastava2013,goodfellow2013}.  

Recursive techniques are based, as the name suggests, on a recursive formulation of batch learning algorithms.
This formulation typically allows to compute the current model in closed form (or with few operations independent from the number of examples) as a combination of the previous model and the new observed example~\citep{Sayed:2008:AF:1370975,laskov2006}. 
As we discuss in more detail in Subsection \ref{sec:rls}, the algorithm proposed in this work is based on a recursive method.

{\bf Learning with an Increasing Number of Classes.} Most classification algorithms have been developed for the batch settings and therefore require the number of classes to be known a priori. 
This assumption is often broken in incremental settings, since new examples could belong to previously unknown classes. 
The problem of dealing with an increasing number of classes has been addressed in the contexts of transfer learning or {\it learning to learn}~\citep{thrun1996}. 
These settings consider a scenario where $T$ linear predictors have been learned to model $T$ classes. Then, when a new class is observed, the associated predictor is learned with the requirement of being ``close'' to a linear combination of the previous ones~\citep{tommasi2010,kuzborskij2013}. Other approaches have been recently proposed where a class hierarchy is built incrementally as new classes are observed, allowing to create a taxonomy (useful for instance for visual tasks) and exploit possible similarities among different classes \citep{xiao2014,sunneol2016}. 
However, all these methods are not incremental in the number of examples and require to retrain the system every time a new point is received.

{\bf Class Imbalance.} The problems related to class imbalance were previously studied in the literature~\citep{elkan2001,he2009,steinwart2008support} and are addressed in Section~\ref{sec:classprob_unbal}. Methods to tackle this issue have been proposed, typically re-weighting the misclassification loss~\citep{tommasi2010} to account for class imbalance. 
However, as we discuss in Subsection~\ref{sec:incremental_recoding} for the case of the square loss, these methods cannot be implemented incrementally. 
This is problematic, since imbalance among multiple classes often arises in on-line settings, even if temporarily, for instance when examples of a new class are observed for the first time.

\section[Classification Setting and Class Imbalance]{The Classification Setting and the Effect of Class Imbalance}
\label{sec:classprob_unbal}

We introduce the learning framework adopted in this work and then proceed to describe the disrupting effects of imbalance among class labels. We refer the reader to Chapter \ref{Chap:SLT} and \citep{steinwart2008support} for more details about Statistical Learning Theory for classification problems. For the sake of simplicity, in the following we consider a binary classification setting, postponing the extension to multiclass classification to the end of this section.

\subsection{Optimal Bayes Classifier and its Least Squares Surrogate}

Let us consider a binary classification problem where input-output examples are sampled randomly according to a distribution $\rho$ over $\mathcal{X}\times\{-1,1\}$. The goal is to learn a function $b_*:\mathcal{X}\to\{-1,1\}$ minimizing the overall {\it expected} classification error
\begin{equation}\label{eq:expected_classification_error}
b_* = \argmin_{b:\mathcal{X}\to\{-1,1\}} \ \ \ \int_{\mathcal{X}\times\{-1,1\}} \mathbf{1}(b(x) - y) ~ d\rho(x,y),
\end{equation}
given a finite set of observations $\{x_i,y_i\}_{i=1}^n$, $x_i\in\mathcal{X}$, $y_i\in\{-1,1\}$ randomly sampled from $\rho$. Here $\mathbf{1}(x)$ denotes the binary function taking value $0$ if $x=0$ and $1$ otherwise. The solution to \eqref{eq:expected_classification_error} is called the {\it optimal Bayes classifier} and it can be shown to satisfy the equation
\begin{equation}\label{eq:optimal_bayes}
b_*(x) = \left\{\begin{array}{cc} 1 & \mbox{if} \ \rho(1|x) > \rho(-1|x) \\ 
-1 & \mbox{otherwise}
\end{array}\right.,
\end{equation}
for all $x\in\mathcal{X}$. Here we have denoted with $\rho(y|x)$ the {\it conditional} distribution of $y$ given $x$. In this work we will denote $\rho(x)$ the {\it marginal} distribution of $x$, such that by Bayes' rule $\rho(x,y) = \rho(y|x)\rho(x)$.

Computing good estimates of $\rho(y|x)$ typically requires large training datasets and is often unfeasible in practice. Therefore, a so-called {\it surrogate} problem (see \citep{steinwart2008support,bartlett2006}) is usually adopted to simplify the optimization problem at \eqref{eq:expected_classification_error} and asymptotically recover the optimal Bayes classifier. In this sense, one well-known surrogate approach is to consider the least squares expected risk minimization  
\begin{equation}\label{eq:expected_risk_minimization}
	f_* = \argmin_{f:\mathcal{X}\to\mathbb{R}} \ \ \ \int_{\mathcal{X}\times\{-1,1\}} (y  - f(x))^2 ~ d\rho(x,y).
\end{equation}
The solution to \eqref{eq:expected_risk_minimization} allows to recover the optimal Bayes classifier. Indeed, for any $f:\mathcal{X}\to\mathbb{R}$ we have
\begin{align*}
	\int (y - f(x))^2d \rho(x,y) = \int \int (y - f(x))^2 d\rho(y|x) d\rho(x)\\
    = \int (1-f(x))^2\rho(1|x) - (f(x)+1)^2\rho(-1|x) ~ d\rho(x),
\end{align*}
which implies that the minimizer of \eqref{eq:expected_risk_minimization} satisfies
\begin{equation}\label{eq:least-squares_solution}
    f_*(x) = 2\rho(1|x) - 1 = \rho(1|x) - \rho(-1|x)
\end{equation}
for all $x\in\mathcal{X}$. The optimal Bayes classifier can be recovered from $f^*$ by taking its sign: $b_*(x) = sign(f_*(x))$. Indeed, $f_*(x)>0$ if and only if $\rho(1|x)>\rho(-1|x)$.\\

\noindent{\bf Empirical Setting}. When solving the problem in practice, we are provided with a finite set $\{x_i,y_i\}_{i=1}^n$ of training examples. In these settings the typical approach is to find an estimator $\hat{f}$ of $f^*$ by minimizing the {\it regularized empirical risk}
\begin{equation}\label{eq:erm}
	\hat{f} = \argmin_{f:\mathcal{X}\to\mathbb{R}} \ \ \ \frac{1}{n} \sum_{i=1}^n (y_i - f(x_i))^2 + R(f),
\end{equation}
where $R$ is a so-called {\it regularizer} preventing the solution $\hat{f}$ to overfit. Indeed, it can be shown~\citep{steinwart2008support,shawe-taylor2004} that, under mild assumptions on the distribution $\rho$, it is possible for $\hat{f}$ to converge in probability to the ideal $f_*$ as the number of training points grows indefinitely. In Section~\ref{sec:backgroundIncClass} we review a method to compute $\hat{f}$ in practice, both in the batch and in the on-line settings.

\subsection{The Effect of Unbalanced Data}

The classification rule at \eqref{eq:optimal_bayes} associates every $x\in\mathcal{X}$ to the class $y$ with highest likelihood $\rho(y|x)$. However, in settings where the two classes are not balanced this approach could lead to unexpected and undesired behaviors. To see this, let us denote $\gamma = \rho(y=1) = \int_\mathcal{X} d\rho(y=1,x)$ and notice that, by \eqref{eq:optimal_bayes} and the Bayes' rule, an example $x$ is labeled $y=1$ whenever
\begin{equation}\label{eq:unbalanced_rule}
	\rho(x|1)>\rho(x|-1)\frac{(1-\gamma)}{\gamma}.
\end{equation}
Hence, when $\gamma$ is close to one of its extremal values $0$ or $1$ (i.e. $\rho(y=1)\gg\rho(y=-1)$ or vice-versa), one class becomes clearly preferred with respect to the other and is almost always selected.

In Figure~\ref{fig:example_rebalance} we report an example of the effect of unbalanced data by showing how the decision boundary (white dashed curve) of the optimal Bayes classifier from \eqref{eq:optimal_bayes} varies as $\gamma$ takes values from $0.5$ (balanced case) to $0.9$ (very unbalanced case).
As it can be noticed, while the classes maintain the same shape, the decision boundary is remarkably affected by the value of $\gamma$.

Clearly, in an on-line robotics setting this effect could be critically suboptimal for two reasons: $1$) We would like the robot to recognize with high accuracy even objects that are less common to be seen. $2$) In incremental settings, whenever a novel object is observed for the first time, only few training examples are available (in the extreme case even just one) and we need a loss weighting fairly also underrepresented classes.

\subsection{Rebalancing the Loss}

We consider a general approach to ``rebalancing'' the classification loss of the standard learning problem of \eqref{eq:expected_classification_error}, similar to the ones in~\citep{elkan2001,steinwart2008support}. 
We begin by noticing that in the balanced setting, namely for $\gamma = 0.5$, the classification rule at \eqref{eq:unbalanced_rule} is equivalent to assigning class $1$ whenever $\rho(x|1) > \rho(x|-1)$ and vice-versa. 
Here we want to slightly modify the misclassification loss in \eqref{eq:expected_classification_error} to recover this same rule also in unbalanced settings. 
To do so, we propose to apply a weight $w(y)\in\mathbb{R}$ to the loss $\mathbf{1}(b(x) - y)$, obtaining the problem
\begin{equation*}\label{eq:weighted_expected_classification_error}
b^w_* = \argmin_{b:\mathcal{X}\to\{-1,1\}} \ \ \ \int_{\mathcal{X}\times\{-1,1\}} w(y)\mathbf{1}(b(x) -  y) ~ d\rho(x,y) .
\end{equation*}
Analogously to the non-weighted case, the solution to this problem is 
\begin{equation}\label{eq:optimal_bayes_weighted_pre}
b_*^w(x) = \left\{\begin{array}{cc} 1 & \mbox{if} \ \rho(1|x)w(1) > \rho(-1|x)w(-1) \\ 
-1 & \mbox{otherwise}
\end{array}\right . .
\end{equation}
In this work we take the weights $w$ to be $w(1) = 1/\gamma$ and $w(-1) = \gamma-w(1)$. Indeed, from the fact that $\rho(y|x) = \rho(x|y)(\rho(y)/\rho(x))$ we have that the rule at \eqref{eq:optimal_bayes_weighted_pre} is equivalent to
\begin{equation}\label{eq:optimal_bayes_weighted}
b_*^w(x) = \left\{\begin{array}{cc} 1 & \mbox{if} \ \rho(x|1) > \rho(x|-1) \\ 
-1 & \mbox{otherwise}
\end{array}\right . , 
\end{equation}
which corresponds to the (unbalanced) optimal Bayes classifier in the case $\gamma = 0.5$, as desired.

\begin{figure}[t]
\vspace{2mm}
\centering
\includegraphics[width=0.4\linewidth]{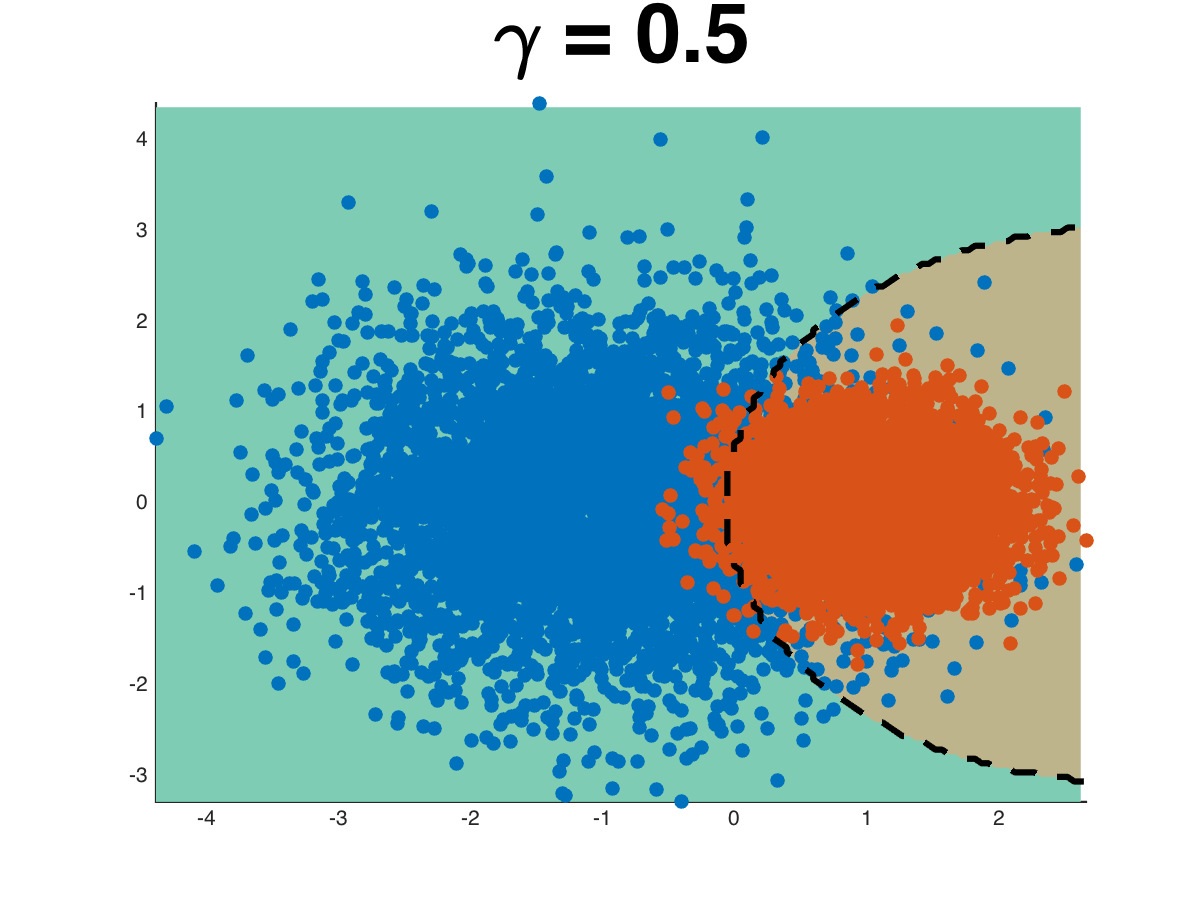}
\includegraphics[width=0.4\linewidth]{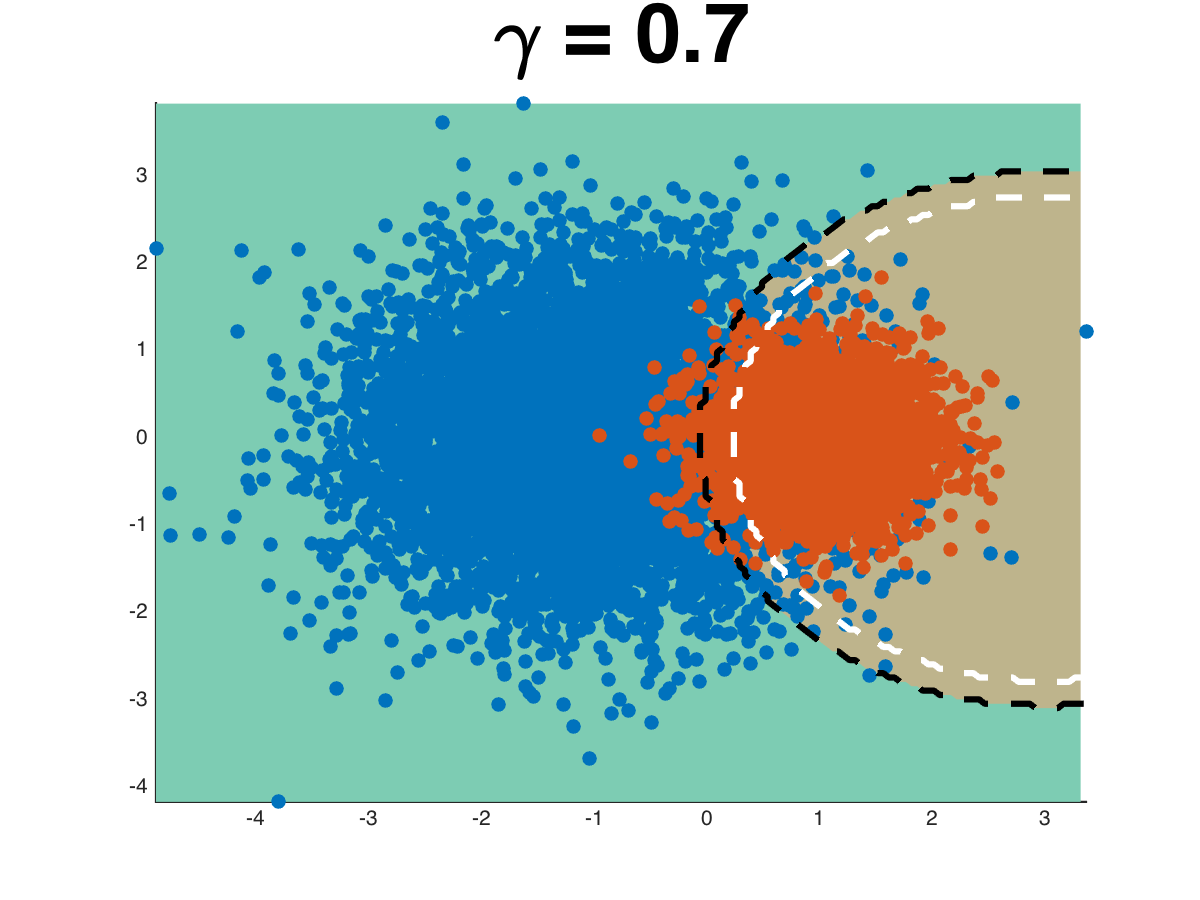}
\includegraphics[width=0.4\linewidth]{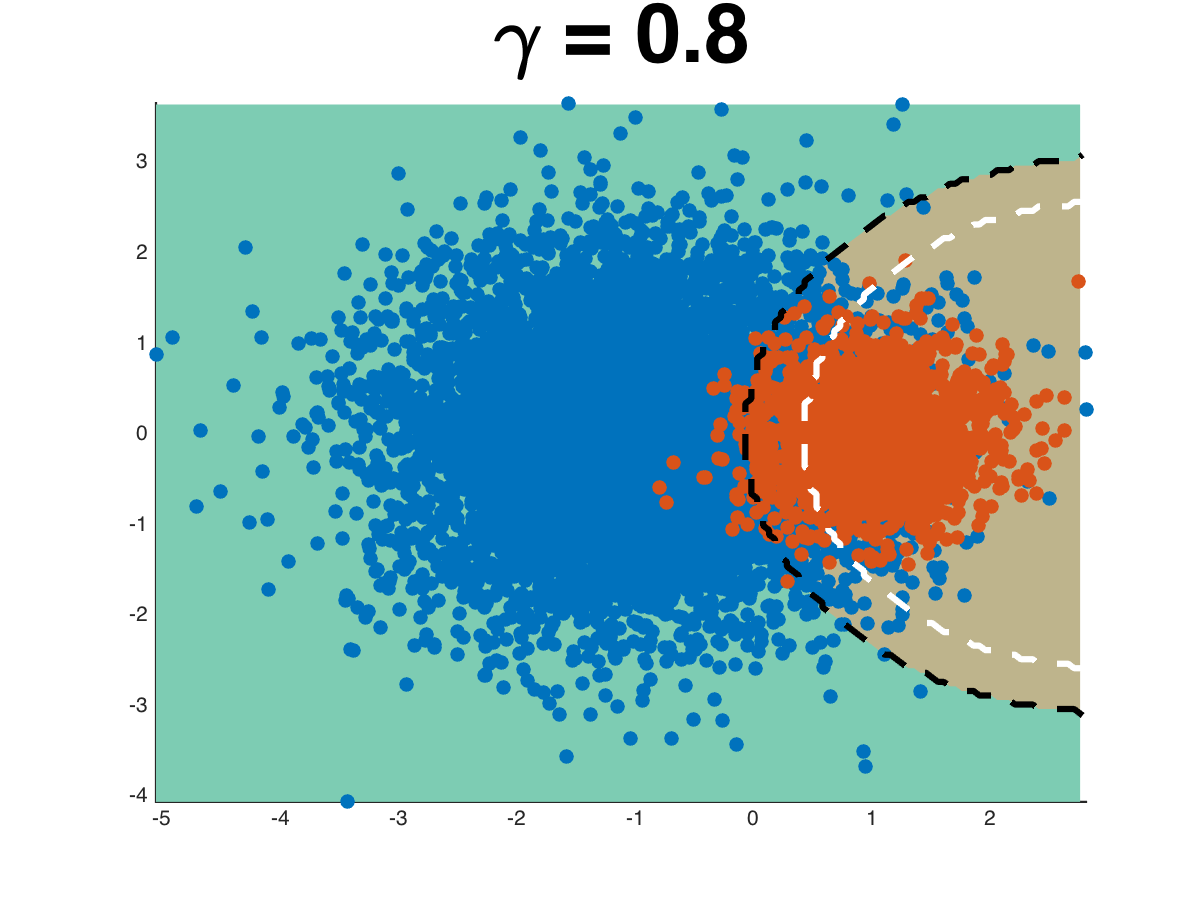}
\includegraphics[width=0.4\linewidth]{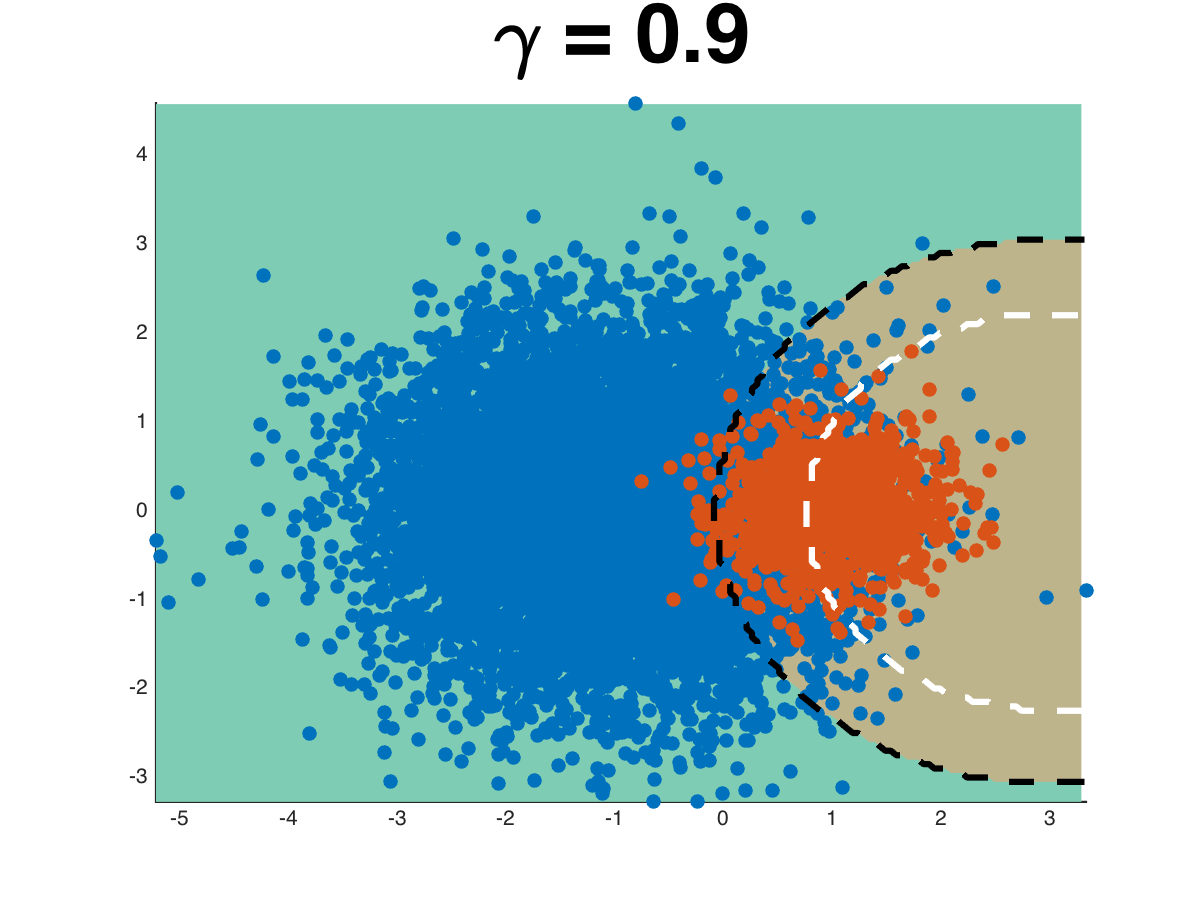}
\caption{Bayes decision boundaries for standard (dashed white line) and rebalanced (dashed black line) binary classification loss for multiple values of $\gamma=\rho(y=1)$ from $0.5$ to $0.9$. Data are sampled according to a Gaussian $\rho(x|y)\sim\mathcal{N}(\mu_y,\sigma_y)$ with $\mu_{1} = (-1,0)^\top$, $\mu_{-1} = (1,0)^\top$, $\sigma_{1}=1$ and $\sigma_{-1}=0.3$. The boundaries coincide when $\gamma=0.5$ (balanced data), while they separate as $\gamma$ increases.}
\label{fig:example_rebalance}
\end{figure}

Figure~\ref{fig:example_rebalance} compares the unbalanced and rebalanced optimal Bayes classifiers for different values of $\gamma$. Notice that rebalancing leads to solutions that are invariant to the value of $\gamma$ (compare the black decision boundary with the white one).

\subsection{Rebalancing and Recoding the Square Loss}\label{sec:rebalancing_recoding_least_squares}

Interestingly, the strategy of changing the weight of the classification error loss can be naturally extended to the least squares surrogate. 
Indeed, if we consider the weighted least squares problem, 
\begin{equation}\label{eq:weigthed_least_squares_expectation}
	f^w_* = \argmin_{f:\mathcal{X}\to\mathbb{R}} \ \ \ \int_{\mathcal{X}\times\{-1,1\}} w(y)(y  - f(x))^2 d\rho(x,y),
\end{equation}
we can again recover the (weighted) rule $b^w_*(x) = sign(f^w_*(x))$ as in the non-weighted setting. 
Indeed, by direct calculation it follows that \eqref{eq:weigthed_least_squares_expectation} has solution 
\begin{equation}\label{eq:sol_weighted_rls}
f^w_*(x) = \frac{\rho(1|x)w(1) - \rho(-1|x)w(-1)}{\rho(1|x)w(1) + \rho(-1|x)w(-1)}.
\end{equation}
If we assume $w(1)>0$ and $w(-1)>0$ (as in this work), the denominator of \eqref{eq:sol_weighted_rls} is always positive and therefore $sign(f^w_*(x))>0$ if and only if $\rho(1|x)w(1)>\rho(-1|x)w(-1)$, as desired.\\

\noindent{\bf Coding}. An alternative approach to recover the rebalanced optimal Bayes classifier via least squares surrogate is to apply a suitable coding function to the class labels $y= \{-1,1\}$, namely 
\begin{equation}\label{eq:coded_least_squares_expectation}
	f^c_* = \argmin_{f:\mathcal{X}\to\R} \ \ \ \int_{\mathcal{X}\times\{-1,1\}} (c(y)  - f(x))^2 d\rho(x,y),
\end{equation}
where $c:\{-1,1\}\to\mathbb{R}$ maps the labels $y$ into scalar codes $c(y)\in\mathbb{R}$. Analogously to the unbalanced (and uncoded) case, the solution to \eqref{eq:coded_least_squares_expectation} is
\begin{equation}\label{eq:coding_solution}
f^c_*(x) = c(1)\rho(1|x) - c(-1)\rho(-1|x) ,
\end{equation}
which, for $c(y) = w(y)$, corresponds to the numerator of \eqref{eq:sol_weighted_rls}. Therefore, the optimal (rebalanced) Bayes classifier is recovered again by $b^w_*(x) = sign(f^c_*(x))$.\\


\subsection{Multiclass Rebalancing and Recoding}\label{sec:multiclass}

In the multiclass setting, the optimal Bayes decision rule corresponds to the function $b_*:\mathcal{X}\to\{1,\dots,T\}$, assigning a label $t \in \{1,\dots,T\}$ to $x\in\mathcal{X}$ when $\rho(t|x)>\rho(s|x)$ $\forall s \neq t$, with $t,s \in \{1,\dots,T\}$. Consequently, the rebalanced decision rule would assign class $t$, whenever $\rho(y=t|x)w(t)>\rho(y=s|x)w(s)$ $\forall s \neq t$, where the function $w:\{1,\dots,T\}\to\mathbb{R}$ assigns a weight to each class. Generalizing the binary case, in this work we set $w(t) = 1/\rho(y=t)$, where we denote $\rho(y=t) = \int_\mathcal{X} d\rho(t,x)$, for each $t \in \{1,\dots,T\}$.

In multiclass settings, the surrogate least squares classification approach is recovered by adopting a {\it 1-vs-all} strategy, formulated as the vector-valued problem
\begin{equation}\label{eq:coded_multi_least_squares_expectation}
	f_* = \argmin_{f:\mathcal{X}\to\mathbb{R}^T} \ \ \ \int_{\mathcal{X}\times\{-1,1\}} \|e_y  - f(x)\|^2 d\rho(x,y) ,
\end{equation}
where $e_t\in\mathbb{R}^T$ is a vector of the canonical basis $\{e_1,\dots,e_T\}$ of $\mathbb{R}^T$ (with the $t$-th coordinate equal to $1$ and the remaining $0$). Analogously to the derivation of \eqref{eq:least-squares_solution}, it can be shown that the solution to this problem corresponds to $f_*(x) = (\rho(1|x),\dots,\rho(T|x))^\top\in\mathbb{R}^T$ for all $x\in\mathcal{X}$. Consequently, we recover the optimal Bayes classifier by
\begin{equation}\label{eq:multiclass_bayes_recover}
	b_*(x) = \argmax_{t=1,\dots,T} f(x)_t.
\end{equation}
where $f(x)_t$ denotes the $t$-th entry of the vector $f(x)\in\mathbb{R}^T$.

The extensions of recoding and rebalancing approaches to this setting follow analogously to the binary setting discussed in Subsection~\ref{sec:rebalancing_recoding_least_squares}.
In particular, the coding function $c:\{e_1,\dots,e_T\}\to\mathbb{R}$ consists in mapping a vector of the basis $e_t$ to $c(e_t) = e_t/\rho(y=t)$.\\

\noindent{\bf Note}. In the previous sections we presented the analysis on the binary case by considering a $\{-1,1\}$ coding for the class labels. This was done to offer a clear introduction to the classification problem, since we need to solve a single least squares problem to recover the optimal Bayes classifier. Alternatively, we could have followed the approach introduced in this section where classes have labels $y=1,2$ and adopt surrogate labels $e_1 = [1,0]^\top$ and $e_2 = [0,1]^\top$. This would have led to train two distinct classifiers and choose the predicted class as the $argmax$ of their scores, according to \eqref{eq:multiclass_bayes_recover}. The two approaches are clearly equivalent since the Bayes classifier corresponds respectively to the inequalities $\rho(1|x)>\rho(-1|x)$ or $\rho(1|x)>\rho(2|x)$.

\section{Incremental Multiclass Classification}
\label{sec:backgroundIncClass}

In this section we review the standard algorithm for Regularized Least Squares Classification (RLSC) and its recursive formulation used for incremental updates.

\subsection{Regularized Least Squares for Classification}
\label{sec:rls}

Here we address the problem of solving the empirical risk minimization introduced at \eqref{eq:erm} in the multiclass setting. Let $\{x_i,y_i\}_{i=1}^n$ be a finite training set, with inputs $x_i\in\mathcal{X}=\mathbb{R}^d$ and labels $y_i\in\{1,\dots,T\}$. In this work, we will assume a linear model for the classifier $\hat{f}$, namely $f(x) = W^\top x$, with $W$ a matrix in $\mathbb{R}^{d \times T}$. We can rewrite \eqref{eq:erm} in matrix notation as 
\begin{equation}\label{eq:matrix-ls}
	\widehat{W} = \argmin_{W\in\R^{d \times T}} \|Y - XW\|_F^2 + \lambda \|W\|_F^2
\end{equation}
with $\lambda>0$ the {\it regularization parameter} and $X\in\mathbb{R}^{n \times d}$ and $Y\in\mathbb{R}^{n \times T}$ the matrices whose $i$-th row correspond respectively to $x_i\in\mathbb{R}^d$ and $e_{y_i}\in\mathbb{R}^T$. We denote by $\|\cdot\|_F^2$ the squared Frobenius norm of a matrix (i.e. the sum of its squared entries).

The solution to \eqref{eq:matrix-ls} is
\begin{equation}\label{eq:W_solution}
\widehat{W} = (X^\top X + \lambda I_{d})^{-1}X^\top Y \in \mathbb{R}^{d \times T} ,
\end{equation}
where $I_{d}$ is the $d \times d$ identity matrix (see for instance \citep{boyd2004}).\\

\noindent{\bf Prediction}. According to the rule introduced in Subsection~\ref{sec:multiclass}, a given $x\in\mathbb{R}^d$ is classified according to 
\begin{equation}
\hat{b}(x) = \argmax_{i=1,\dots,T} \ \ \hat{f}(x)_i = (\widehat{W}^{(i)})^\top x
\end{equation}
with $\widehat{W}^{(i)}\in\mathbb{R}^d$ denoting the $i$-th column of $\widehat{W}$.

\subsection{Recursive Regularized Least Squares}
\label{sec:incClassRecupdate}

The closed form for the solution at \eqref{eq:W_solution} allows to derive a recursive formulation to incrementally update $\widehat{W}$ in constant time as new training examples are observed~\citep{Sayed:2008:AF:1370975}. Consider a learning process where training data are provided one at a time to the system. At iteration $k$ we need to compute $W_{k} = (X_{k}^\top X_{k} + \lambda I_d)^{-1} X_{k}^\top Y_{k}$, where $X_{k} \in \mathbb{R}^{k \times d}$ and $Y_{k} \in \mathbb{R}^{k \times T}$ are the matrices whose rows correspond to the first $k$ training examples. The computational cost for evaluating $W_{k+1}$ according to \eqref{eq:least-squares_solution} is $O(kd^2)$ (for the matrix products) and $O(d^3)$ (for the inversion). This is undesirable in the on-line setting, since $k$ can grow indefinitely. Here we show that we can compute $W_k$ incrementally from $W_{k-1}$ in $O(Td^2)$.

To see this, first notice that by construction
\[
X_{k} = [ X_{k-1}^\top , x_{k} ]^\top \qquad Y_{k} = [ Y_{k-1}^\top , e_{y_{k}} ]^\top ,
\]
and therefore, if we denote $A_k = X_k^\top X_k + \lambda I_d$ and  $b = X_k^\top Y_k$, we obtain the recursive formulations
\begin{align}
A_{k} &= X_{k}^\top X_{k} + \lambda I_{d} \nonumber \\
&=X_{k-1}^\top X_{k-1} + x_{k}^\top x_{k} + \lambda I_{d} \nonumber \\
&=A_{k-1} + x_{k}^\top x_{k} + \lambda I_{d}
\end{align}
and
\begin{equation}
b_{k} = X_{k}^\top Y_{k} = X_{k-1}^\top Y_{k-1} + x_{k}e_{y_k}^\top = b_{k-1} + x_k e_{y_k}^\top.
\end{equation}

Now, computing $b_k$ from $b_{k-1}$ requires $O(d)$ operations (since $e_{y_k}$ has all zero entries but one), computing $A_k$ from $A_{k-1}$ requires $O(d^2)$, while the inversion $A_k^{-1}$ requires $O(d^3)$. To reduce the cost of the (incremental) inversion, we recall that for a positive definite matrix $A_k$ for which its Cholesky decomposition $A_k = R_k^\top R_k$ is known (with $R^k\in\mathbb{R}^{d \times d}$ upper triangular), the inversion $A_k^{-1}$ can be computed in $O(d^2)$ \citep{Golub1996}. In principle, computing the Cholesky decomposition of $A_k$ still requires $O(d^3)$, but we can apply a rank-one update to the Cholesky decomposition at the previous step, namely $A_k = R_k^\top R_k = R_{k-1}^\top R_{k-1} + x_k x_k^\top = A_{k-1} + x_k x_k^\top$, which is known to require $O(d^2)$~\citep{bjoerck_least_squares96}. Several implementations are available for the Cholesky rank-one updates; in our experiments we used the {\sc MATLAB} routine {\sc cholupdate}.

Therefore, the update $W_k$ from $W_{k-1}$ can be computed in $O(Td^2)$, since the most expensive operation is the multiplication $A_k^{-1}b_k$. In particular, this computation is independent of the current number $k$ of training examples seen so far, making this algorithm suited for the on-line setting.


\section[Incremental Class Extension and Recoding]{Incremental Multiclass Classification with Class Extension and Recoding}
\label{sec:proposed_algorithm}
In this section, we present our approach to incremental multiclass classification where we account for the possibility to extend the number of classes incrementally and apply the recoding approach introduced in Section~\ref{sec:classprob_unbal}. The algorithm is reported in Algorithm~\ref{alg:incremental_learning}.

\subsection{Class Extension}
We introduce a modification of recursive RLSC, allowing to extend the number of classes in constant time with respect to the number of examples seen so far. Let $T_{k}$ denote the number of classes seen up to iteration $k$.  We have two possibilities:
\begin{enumerate}
\item The new sample $(x_k, y_k)$ belongs to a known class, this meaning that $e_{y_k} \in \mathbb{R}^{T_{k-1}}$ and $T_k = T_{k-1}$.
\item $(x_k, y_k)$ belongs to a previously unknown class, implying that $y_k = T_k = T_{k-1} + 1$.
\end{enumerate}
In the first case, the update rules for $A_k$, $b_k$ and $W_k$ explained in Subsection \ref{sec:incClassRecupdate} can be directly applied. In the second case, the update rule for $A_k$ remains unchanged, while the update of $b_k$ needs to account for the increase in size (since $b_k\in\mathbb{R}^{k \times (T_{k-1} + 1)}$). However, we can modify the update rule for $b_k$ without increasing its computational cost by first adding a new column of zeros $\mathbf{0}\in\mathbb{R}^d$ to $b_{k-1}$, namely
\begin{equation}
b_k = [b_{k-1}, \mathbf{0}] + x_k^\top e_{y_k} ,
\end{equation}
which requires $O(d)$ operations. Therefore, with the strategy described above it is indeed possible to extend the classification capabilities of the incremental learner during on-line operation, without re-training it from scratch. In the following, we address the problem of dealing with class imbalance during incremental updates by performing incremental recoding.

\begin{algorithm}[t]
\caption{Incremental RLSC with Class Recoding}
  \begin{algorithmic}
   \Statex \textbf{Input:} Hyperparameters $\lambda > 0$, $\alpha \in [0,1]$
    \Statex \textbf{Output:} Learned weights $W_k$ at each iteration
    \Statex \textbf{Initialize:} $R_0 \gets \sqrt{\lambda}I_d, ~ b_0 \gets \emptyset, ~ \gamma_0 \gets \emptyset, ~ T \gets 0$ \\
    
    \algblock[Increment]{Increment}{EndIncrement}
    
    \algnewcommand\algorithmicincrement{\textbf{Increment: }}
    \algrenewtext{Increment}{\algorithmicincrement}
    \algnewcommand\algorithmicendincrement{\textbf{end\ increment}}
    \algrenewtext{EndIncrement}{}

    \Increment Observed input $x_k\in\mathbb{R}^d$ and output label $y_k$:
        \If{  ($y_k  =  T + 1$) }
            \State $T \gets T + 1$
            \State $ \gamma_{k-1} \gets [ \gamma_{k-1}^\top, 0]^\top$
            \State $b_{k-1} \gets \left[b_{k-1}, \mathbf{0}\right]$, with $\mathbf{0}\in\mathbb{R}^d$
        \EndIf  
        \State $\gamma_k \gets \gamma_{k-1} + e_{y_k}$
        \State $\Gamma_k \gets k \cdot \text{diag}(\gamma_k)^{-1}$
        \State $b_k \gets b_{k-1} + x_k^\top e_{y_k}$
        \State $R_k \gets$ \textsc{choleskyUpdate}$(R_{k-1}, x_k)$
        \State $W_k \gets R_k^{-1} (R_k^\top)^{-1} b_k \Gamma_k^\alpha$
    \EndIncrement \Return $W_k$
  \end{algorithmic}
  \label{alg:incremental_learning}
\end{algorithm}

\subsection{Incremental Recoding}
\label{sec:incremental_recoding}

The main algorithmic difference between standard RLSC and the variant with recoding is in the matrix $Y$ containing output training examples. Indeed, according to the recoding strategy, the vector $e_{y_k}$ associated to an output label $y_k$ is coded into $c(e_{y_k}) = e_{y_k}/\rho(y=y_k)$. In the batch setting, this can be formulated in matrix notation as 
$$
W = (X^\top X + \lambda I_d)^{-1} X^\top Y \Gamma
$$
where the original output matrix is substituted by its encoded version $Y^{(c)}= Y\Gamma\in \R^{n \times T}$, with $\Gamma$ the $T \times T$ diagonal matrix whose $t$-th diagonal element is $\Gamma_{tt} = 1/\rho(y=t)$. Clearly, in practice the $\rho(y=t)$ are estimated empirically (e.g. by $\hat{\rho}(y=t) = n_t/n$, the ratio between the number $n_t$ of training examples belonging to class $t$ and the total number $n$ of examples).

The above formulation is favorable for the on-line setting. Indeed, we have 
\begin{equation}\label{eq:incremental_coding}
X_k^\top Y_k \Gamma_k = b_k \Gamma_k = (b_{k-1} + x_k^\top y_k) \Gamma_k,
\end{equation}
where $\Gamma_k$ is the diagonal matrix of the (inverse) class distribution estimators $\hat\rho$ up to iteration $k$. $\Gamma_k$ can be computed incrementally in $O(T)$ by keeping track of the number $k_t$ of examples belonging to $t$ and then computing $\hat\rho_k(y=t) = k_t/k$ (see Algorithm~\ref{alg:incremental_learning} for how this update was implemented in our experiments). 

Note that the above step requires $O(dT)$, since updating the (uncoded) $b_k$ from $b_{k-1}$ requires $O(d)$ and multiplying $b_k$ by a diagonal matrix requires $O(dT)$. All the above computations are dominated by the product $A_k^{-1}b_k$, which requires $O(Td^2)$. Therefore, our algorithm is computationally equivalent to the standard incremental RLSC approach.
\\

\noindent{\bf Coding as a Regularization Parameter.} Depending on the amount of training examples seen so far, the estimator $k_{t}/k$ could happen to not approximate $\rho(y=t)$ well. In order to mitigate this issue, we propose to add a parameter $\alpha\in[0,1]$ on $\Gamma_k^\alpha$, regularizing the effect of coding. Clearly, for $\alpha = 0$ we recover the (uncoded) standard RLSC, since $\Gamma_k^0 = I_T$, while $\alpha=1$ applies full recoding. In Subsection~\ref{sec:model_selection} we discuss an efficient heuristic to find $\alpha$ in practice.
\\

\noindent{\bf Incremental Rebalancing.} Note that the incremental loss-rebalancing algorithm (Subsection~\ref{sec:rebalancing_recoding_least_squares}) cannot be implemented incrementally. Indeed, the solution of the rebalanced empirical RLS is
\begin{equation}\label{sec:incr_rebalancing}
W_k = (X_k^\top \Sigma_k X_k + \lambda I_d)^{-1} X_k^\top \Sigma_K Y_k
\end{equation}
with $\Sigma_k$ a diagonal matrix whose $i$-th entry is equal to $(\Sigma_k)_{ii} = 1/\hat{\rho}(y = t_i)$ and $t_i$ is the class to which the $i$-th training example $y_i$ belongs. Since $\Sigma_k$ changes at every iteration, it is not possible to derive a rank-one update rule for $(X_k^\top \Sigma_k X_k + \lambda I_d)^{-1}$ as for standard RLSC.

\section{Experiments}
\label{sec:incClassExperiments}

We empirically assessed the performance of Algorithm\ref{alg:incremental_learning} on a standard benchmark for machine learning and on two visual recognition tasks in robotics. We compared the classification accuracy of the proposed method with standard RLSC, to evaluate the improvement provided by incremental recoding, and with the rebalanced approach~\eqref{sec:incr_rebalancing}, which cannot be implemented incrementally, but is a competitor in terms of accuracy when classes are imbalanced.\\

\subsection{Experimental Protocol}\label{sec:protocol}

We adopted the following experimental protocol:

\begin{enumerate}
\item Given a dataset with $T$ classes, we simulated a scenario where a new class is observed by selected $T-1$ of them to be ``balanced'' and the remaining one to be under-represented. 
\item We trained a classifier for the balanced classes, using a randomly sampled dataset containing $n_{bal}$ examples per class. We sampled also a validation set with $n_{bal}/5$ examples per class. $n_{bal}$ is specified below for each dataset.
\item We incrementally trained the classifier from the previous step by sampling on-line $n_{imb}$ examples for the $T$-th class. Model selection was performed {\it using exclusively the validation set of the balanced classes}, following the strategy described in Subsection~\ref{sec:model_selection}.
\item To measure performance, we sampled a separate test set containing $n_{test}$ examples per class (both balanced and under-represented).
\end{enumerate}

For each dataset, we performed $10$ independent trials to account for statistical variability. We measured the performance of the algorithms on the test set while they were trained incrementally. In particular, in Table~\ref{tab:incClassAccuracy} we report the average test accuracy on the imbalanced class and on the entire test set to asses both the relative and overall performance.

\subsection{Datasets}\label{sec:datasets}

{\bf MNIST} \citep{lecun1998gradient} is a benchmark dataset composed of $60000$ $28 \times 28$ centered greyscale pictures of digits from $0$ to $9$. In our experiment, we considered a much smaller subset of MNIST, with $n_{bal}=1000$ per balanced class. The test set was obtained by sampling $n_{test}=200$ examples per class for all digits. The imbalanced class was chosen randomly at each trial. We used the raw pixels of the images as inputs for the linear classifiers (i.e. $x\in\mathbb{R}^{784}$).

{\bf iCubWorld28} \citep{pasquale2015teaching} is a dataset for visual object recognition in robotics. It was collected during a series of sessions where a human teacher showed different objects to the iCub humanoid \citep{Metta2010}. The task is to discriminate between $28$ objects {\it instances} and contains $\sim 1800$ images per class. We used $n_{bal}=700$ and $n_{test} = 700$ and chose the $28$-th class to be under-represented. We performed feature extraction on {\it iCubWorld28} images by taking the output of the {\it fc$7$} activation layer of a {\sc CaffeNet}~\citep{Jia2014} Convolutional Neural Network (CNN), pre-trained on the ImageNet dataset~\citep{ILSVRC15} (see \citep{pasquale2015teaching} for feature extraction details)

{\bf RGB-D} \citep{lai2011} is a dataset for visual object {\it categorization} in robotics settings. The dataset has been collected while $300$ objects from $51$ categories were observed on a table from multiple points of view and scales. An average of $\sim 900$ images are available for each object category. While also depth information is available, in this work we focused only on $RGB$ data. Feature extraction was performed analogously to {\it iCubWorld28}. We used $n_{bal}=500$ and $n_{test}=400$ and chose the {\it tomato} category to be under-represented.

\subsection{Model Selection}
\label{sec:model_selection}

In traditional batch learning settings for RLSC, model selection for the hyperparameter $\lambda$ is typically performed via hold-out, k-fold or similar cross-validation techniques. In the incremental setting these strategies cannot be directly applied since examples are observed on-line, but a simple approach to create a validation set is to hold out every $i$-th example without using it for training (in our experiments we set $i=6$). At each iteration, multiple candidate models are trained incrementally each for a different value of $\lambda$, and the one with highest validation accuracy is selected for prediction.

However, following the same argument of Section~\ref{sec:classprob_unbal}, in presence of class imbalance this strategy would often select classifiers that ignore the under-represented class. Rebalancing the validation loss (see Section~\ref{sec:classprob_unbal}) does not necessarily solve the issue, but could rather lead to overfit the under-represented class, degrading the accuracy on other classes since errors count less on them. Motivated by empirical evidence discussed below, in this work we have adopted a model selections heuristic for $\lambda$ and $\alpha$ in Algorithm~\ref{alg:incremental_learning}, which guarantees to not degrade accuracy on well-represented classes while at the same time achieving higher or equal accuracy on the under-represented one.

\begin{figure}[t]
\vspace*{2mm}
\centering
\includegraphics[width = 0.8\textwidth]{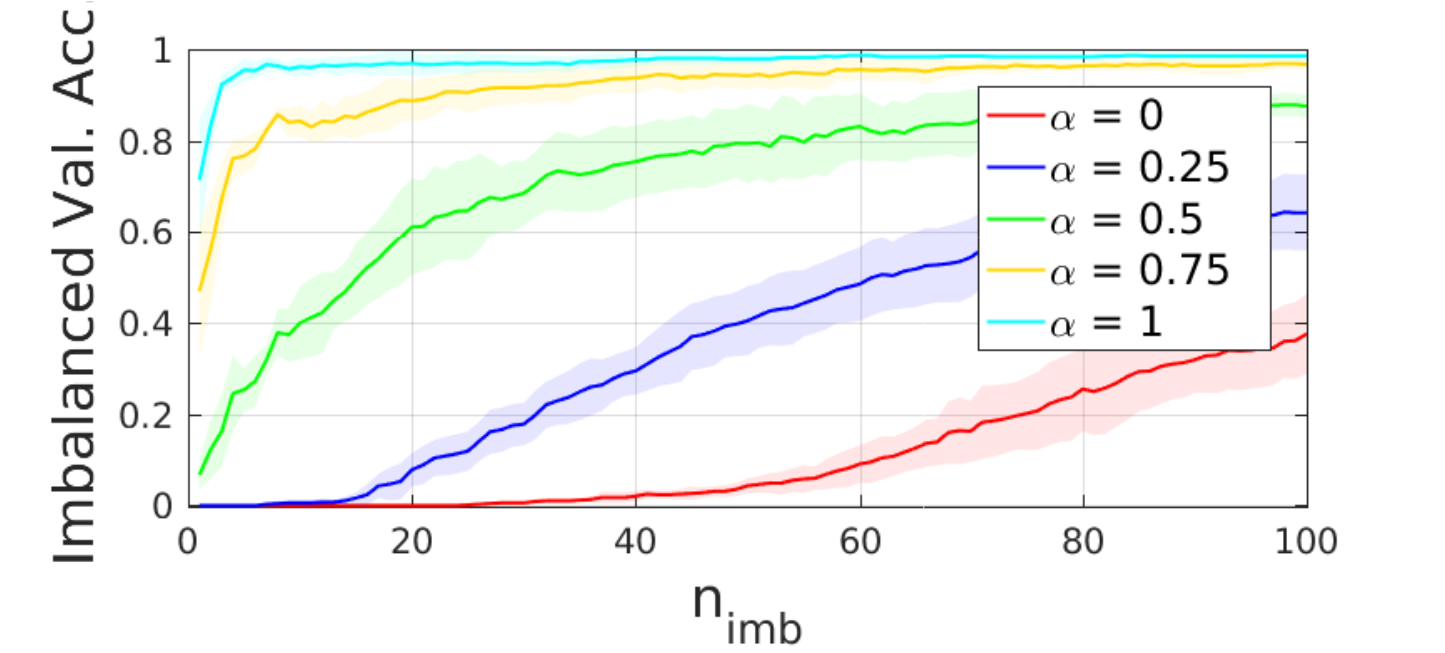}\\
\includegraphics[width = 0.8\textwidth]{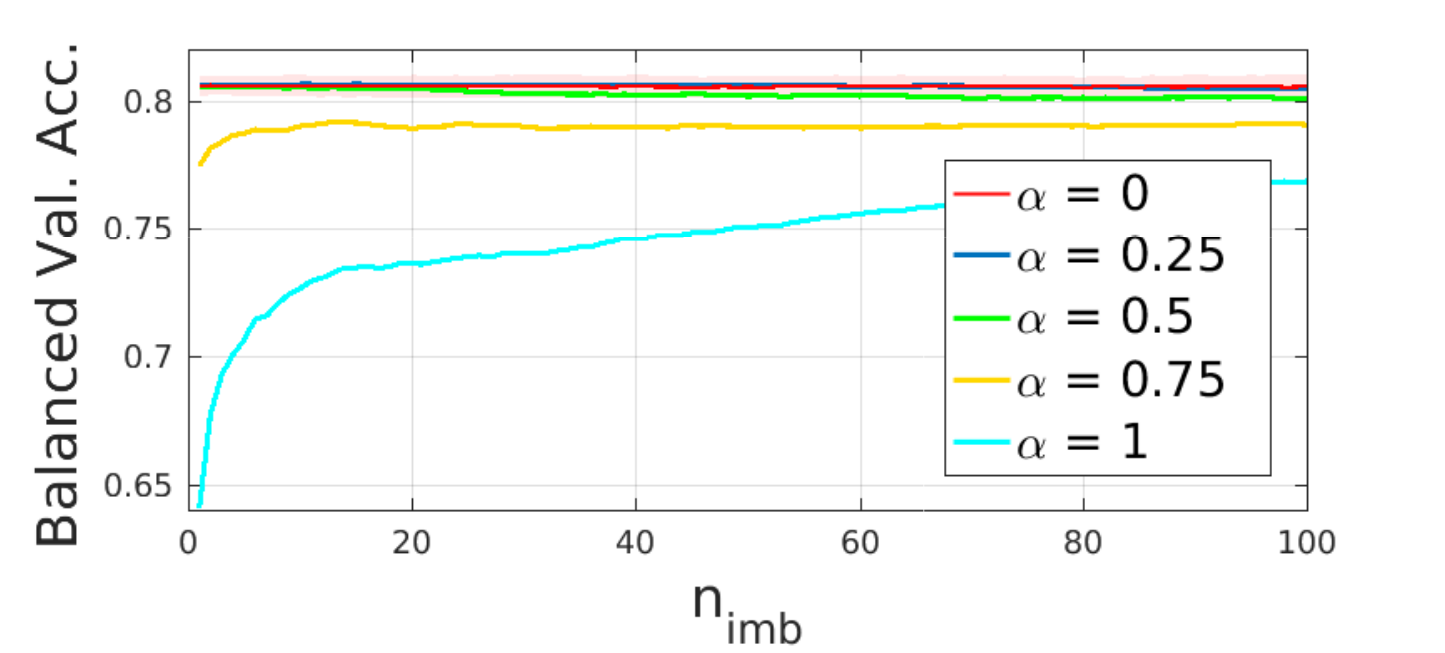}
\caption{Classification accuracy on \textit{iCubWorld28} imbalanced (top) and balanced (bottom) test classes for models trained according to Algorithm~\ref{alg:incremental_learning} with varying $\alpha$ and best $\lambda$ within a pre-defined range (chosen at each iteration and for each $\alpha$). Growing $\alpha$ from $0$ to $1$ allows to find a model that maintains same performance on known classes while improving on the under-represented one.}
\label{fig:acc_modsel}
\end{figure}

Our strategy evaluates the accuracy of the candidate models on the incremental validation set, but {\it only for classes that have a sufficient number of examples} (e.g. classes with less examples than a pre-defined threshold are not used for validation). 
Then, we choose the model with largest $\alpha \in [0,1]$ for which such accuracy is higher or equal to the one measured for $\alpha = 0$, namely without coding. 
Indeed, as can be seen in Figure~\ref{fig:acc_modsel} for validation experiments on {\it iCubWorld28}, as $\alpha$ grows from $0$ to $1$, the classification accuracy on the under-represented class increases, Figure~\ref{fig:acc_modsel} (Top), while it decreases on the remaining ones, Figure~\ref{fig:acc_modsel} (Bottom).
Our heuristic chooses the best trade-off for $\alpha$ so that performance does not degrade on well-known classes, but at the same time it will often improve on the under-represented one.



\begin{table}                                     
\vspace*{2mm}
\centering        
\caption{Incremental classification accuracy on for the na\"ive (N) RLSC, Rebalanced (RB) and recoding Algorithm~\ref{alg:incremental_learning} (RC).}                                
\begin{adjustbox}{width=\linewidth,center}
\begin{tabular}{c c c c >{\columncolor{Gray}}c c c >{\columncolor{Gray}}c }                              
\toprule
\multirow{2}{*}{\textbf{Dataset}} & \multirow{2}{*}{$n_{imb}$} & \multicolumn{3}{c}{\textbf{Total Acc. (\%)}}  & \multicolumn{3}{c}{\textbf{Imbalanced Acc. (\%)}} \\ \cline{3-8}  
 \\ [-0.9em]
 & &  \textbf{N}  & \textbf{RB} & \textbf{RC}&  \textbf{N}  & \textbf{RB} & \textbf{RC} \\ 
\hline
\multirow{6}{*}{\shortstack{\textit{\textbf{MNIST}} \\ \\$n_{bal}$ \\ $=$ \\ $1000$ }}%
  & 1 & 79.2 $\pm$ 0.3 & \textbf{79.7 $\pm$ 0.4} & \textbf{79.7 $\pm$ 0.6} &     0.0 $\pm$     0.0 &  7.4 $\pm$ 7.7 & \textbf{9.5 $\pm$ 4.9} \\ 
  & 5 & 79.1 $\pm$ 0.3 & \textbf{82.5 $\pm$ 0.7} & 80.3 $\pm$ 0.6 &     0.0 $\pm$     0.0 & \textbf{39.6 $\pm$ 6.2 }& 17.5 $\pm$ 6.6 \\ 
  & 10 & 79.2 $\pm$ 0.3 & \textbf{83.6 $\pm$ 0.7} & 81.0 $\pm$ 0.6 &     0.0 $\pm$     0.0 & \textbf{49.5 $\pm$ 5.7} & 25.1 $\pm$ 5.3 \\ 
  & 50 & 79.2 $\pm$ 0.3 &  \textbf{85.5 $\pm$ 0.3} & 83.9 $\pm$ 0.5 &     0.0 $\pm$     0.0 & \textbf{73.5 $\pm$ 3.3} & 49.1 $\pm$ 3.5 \\ 
 & 100 & 79.2 $\pm$ 0.4 & \textbf{85.9 $\pm$ 0.4} & 85.1 $\pm$ 0.5 & 2.0 $\pm$ 0.9 & \textbf{75.5 $\pm$ 2.7} & 62.7 $\pm$ 2.9 \\ 
 & 500 & 85.5 $\pm$ 0.3 &\textbf{ 86.2 $\pm$ 0.3 }& \textbf{86.1 $\pm$ 0.3} & 66.9 $\pm$ 1.1 & \textbf{78.5 $\pm$ 0.9} & \textbf{77.8 $\pm$ 1.1} \\ 
\hline
\multirow{6}{*}{\shortstack{\textit{\textbf{iCub}} \\ \\$n_{bal}$ \\$=$ \\ $700$}}%
  &  1  & 77.6 $\pm$ 0.3 & 76.8 $\pm$ 0.1 & \textbf{77.7 $\pm$ 0.3} &     0.0 $\pm$     0.0 & 0.4 $\pm$ 0.6 &  \textbf{8.0 $\pm$ 11.4} \\ 
  & 5 & 77.6 $\pm$ 0.3 & 77.9 $\pm$ 0.1 & \textbf{78.6 $\pm$ 0.3} &     0.0 $\pm$     0.0 & 8.1 $\pm$ 3.9  & \textbf{38.5 $\pm$ 9.7} \\ 
  & 10 & 77.6 $\pm$ 0.3 & 78.3 $\pm$ 0.4 & \textbf{78.9 $\pm$ 0.2} &     0.0 $\pm$     0.0 & 23.7 $\pm$ 10.8 & \textbf{49.6 $\pm$ 5.6} \\ 
  & 50 & 77.7 $\pm$ 0.2 &   \textbf{80.0 $\pm$ 0.2} & \textbf{80.0 $\pm$ 0.1} & 5.4 $\pm$ 4.1 & 73.9 $\pm$ 7.3 &  \textbf{75.0 $\pm$ 5.5} \\ 
 & 100 & 78.6 $\pm$ 0.1 & \textbf{80.2 $\pm$ 0.1} & 80.1 $\pm$ 0.2 & 39.1 $\pm$ 3.6 & 85.9 $\pm$  4.0 & \textbf{86.5 $\pm$  3.0} \\ 
 & 500 & 80.2 $\pm$ 0.2 & \textbf{80.1 $\pm$ 0.1} & \textbf{80.1 $\pm$ 0.2} & 89.3 $\pm$ 2.5 & 93.8 $\pm$  2.0 & \textbf{94.8 $\pm$ 1.9} \\ 
\hline
\multirow{6}{*}{\shortstack{\textit{\textbf{RGB-D}} \\ \\$n_{bal} $\\ $=$ \\ $500$}} %
  & 1 & 80.4 $\pm$ 2.2 & 78.6 $\pm$ 3.2 & \textbf{83.3 $\pm$ 3.2} &     0.0 $\pm$     0.0 &  62.0 $\pm$ 42.1 & \textbf{72.2 $\pm$ 26.3} \\ 
  & 5 & 80.4 $\pm$ 2.2 &  83.0 $\pm$ 2.1 & \textbf{83.9 $\pm$ 2.6} &     0.0 $\pm$     0.0 & 91.7 $\pm$ 12.8 & \textbf{99.9 $\pm$ 0.3} \\ 
  & 10 & 80.4 $\pm$ 2.2 & \textbf{83.8 $\pm$ 1.8} & \textbf{83.6 $\pm$ 2.6} & 2.8 $\pm$ 2.4 & 94.7 $\pm$ 8.4 &     \textbf{100.0 $\pm$     0.0} \\ 
  & 50 & 82.3 $\pm$ 2.2 & \textbf{84.3 $\pm$ 1.9} & 83.5 $\pm$ 2.9 & 96.6 $\pm$ 3.7  &     \textbf{100.0$\pm$     0.0}  &     \textbf{100.0 $\pm$     0.0} \\ 
 & 100 & 82.4 $\pm$ 2.1 & \textbf{84.4 $\pm$ 2.0} & 83.5 $\pm$ 2.8 &     100.0 $\pm$     0.0 &    \textbf{ 100.0 $\pm$     0.0 } &     \textbf{100.0 $\pm$     0.0} \\ 
 & 500 & 82.3 $\pm$ 2.1 & \textbf{84.1 $\pm$ 2.0} & \textbf{84.1 $\pm$ 2.8}  &     100.0 $\pm$     0.0 &     \textbf{100.0 $\pm$     0.0}  &     \textbf{100.0 $\pm$     0.0} \\ 
\bottomrule
\end{tabular}                                     
\end{adjustbox}
\label{tab:incClassAccuracy}                        
\end{table}

\subsection{Results}

\begin{figure}[t]
\centering
\textit{MNIST} \qquad\qquad\quad \textit{iCubWorld28} \qquad\qquad\quad \textit{RGB-D}\\
\includegraphics[width = 0.3\textwidth]{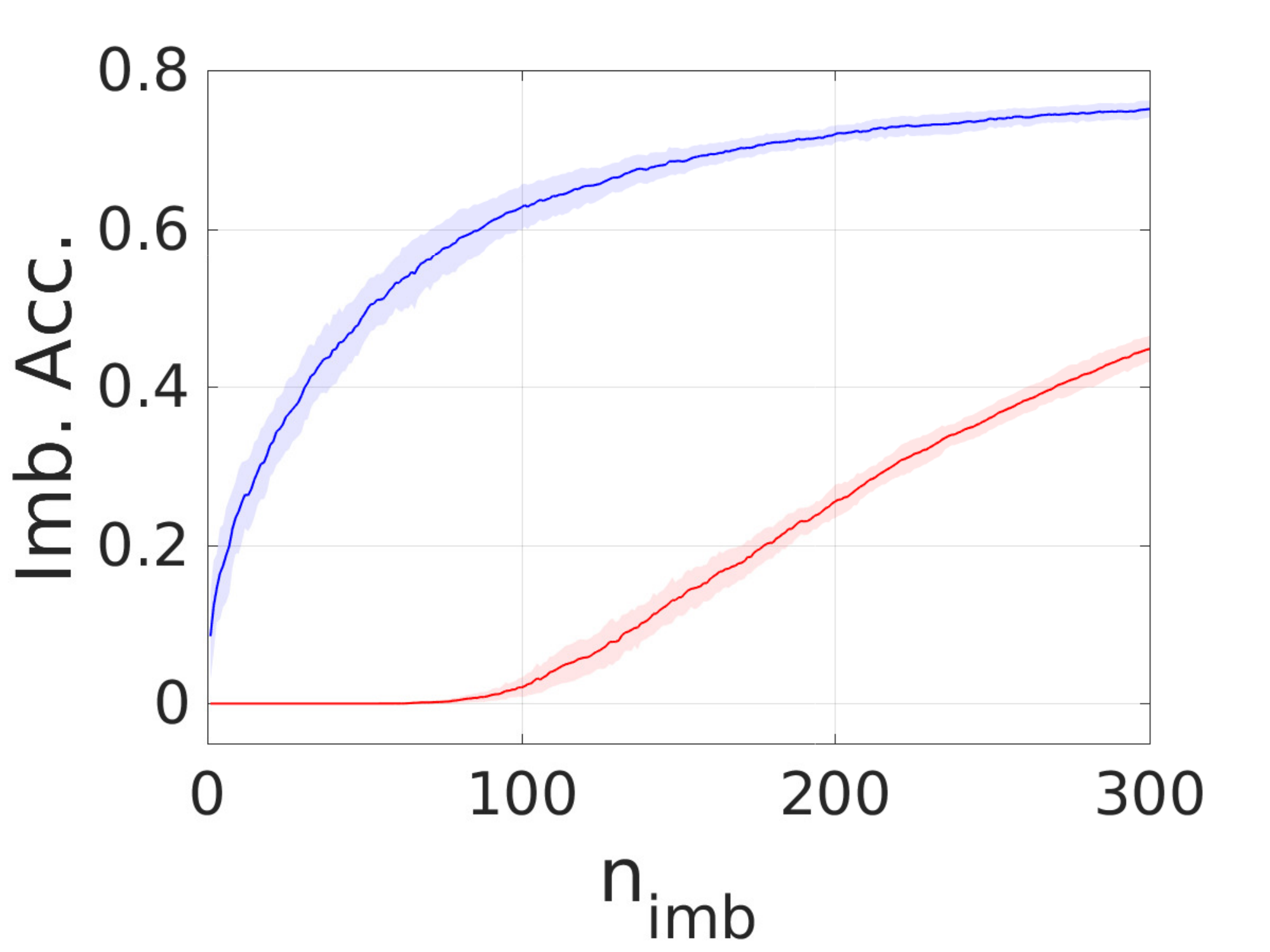}%
\includegraphics[width = 0.3\textwidth]{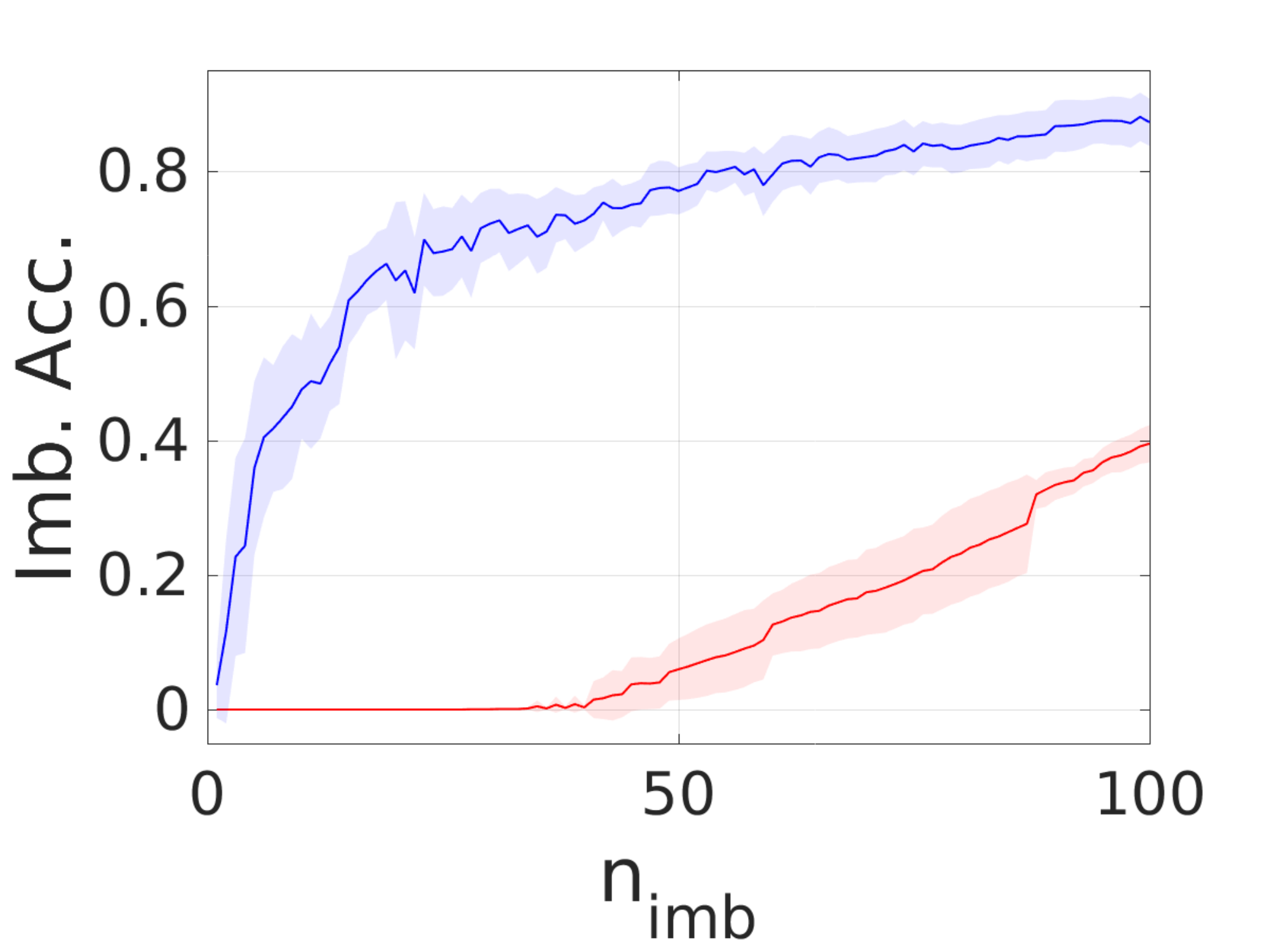}%
\includegraphics[width = 0.3\textwidth]{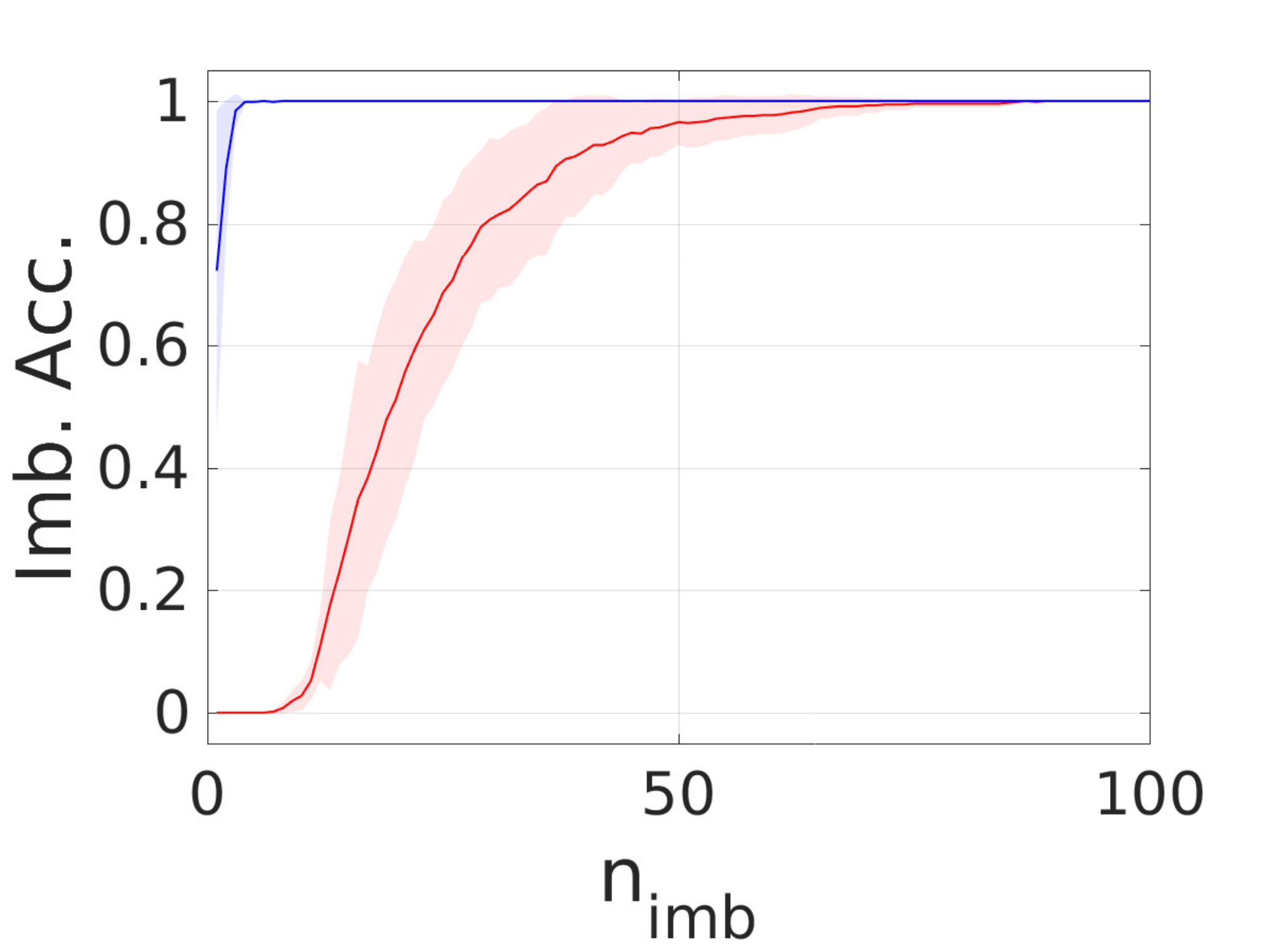}\\
\includegraphics[width = 0.3\textwidth]{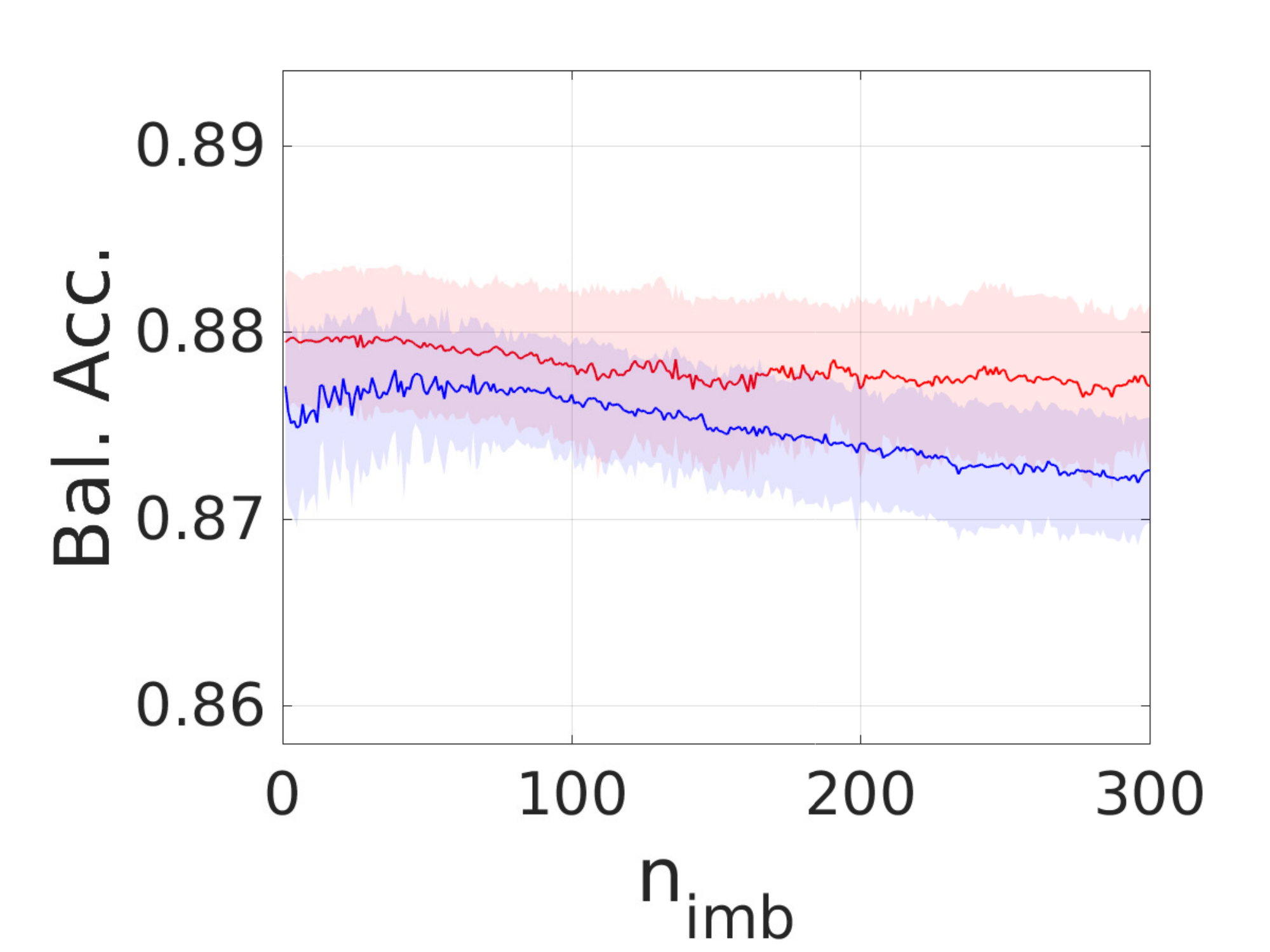}%
\includegraphics[width = 0.3\textwidth]{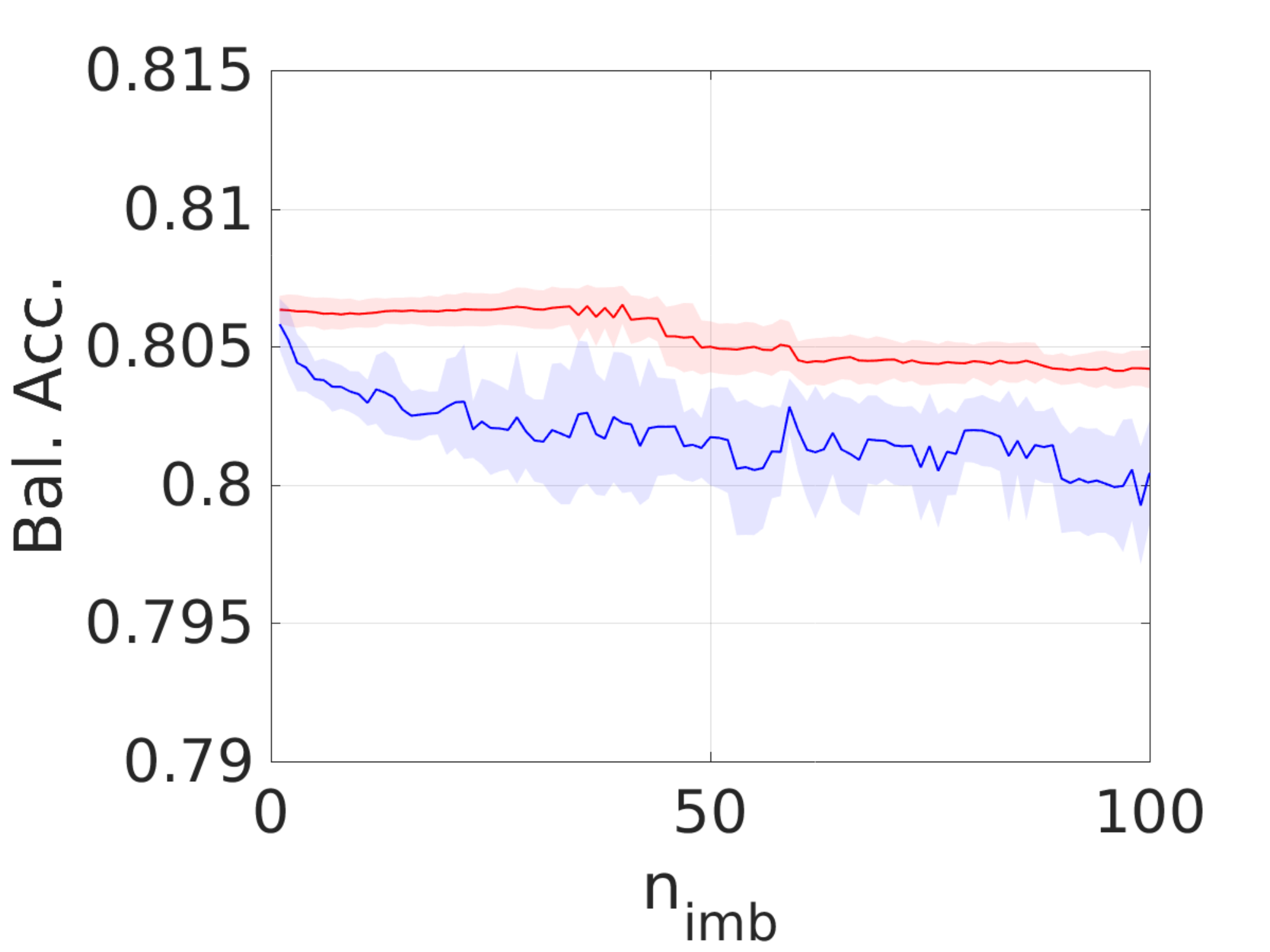}%
\includegraphics[width = 0.3\textwidth]{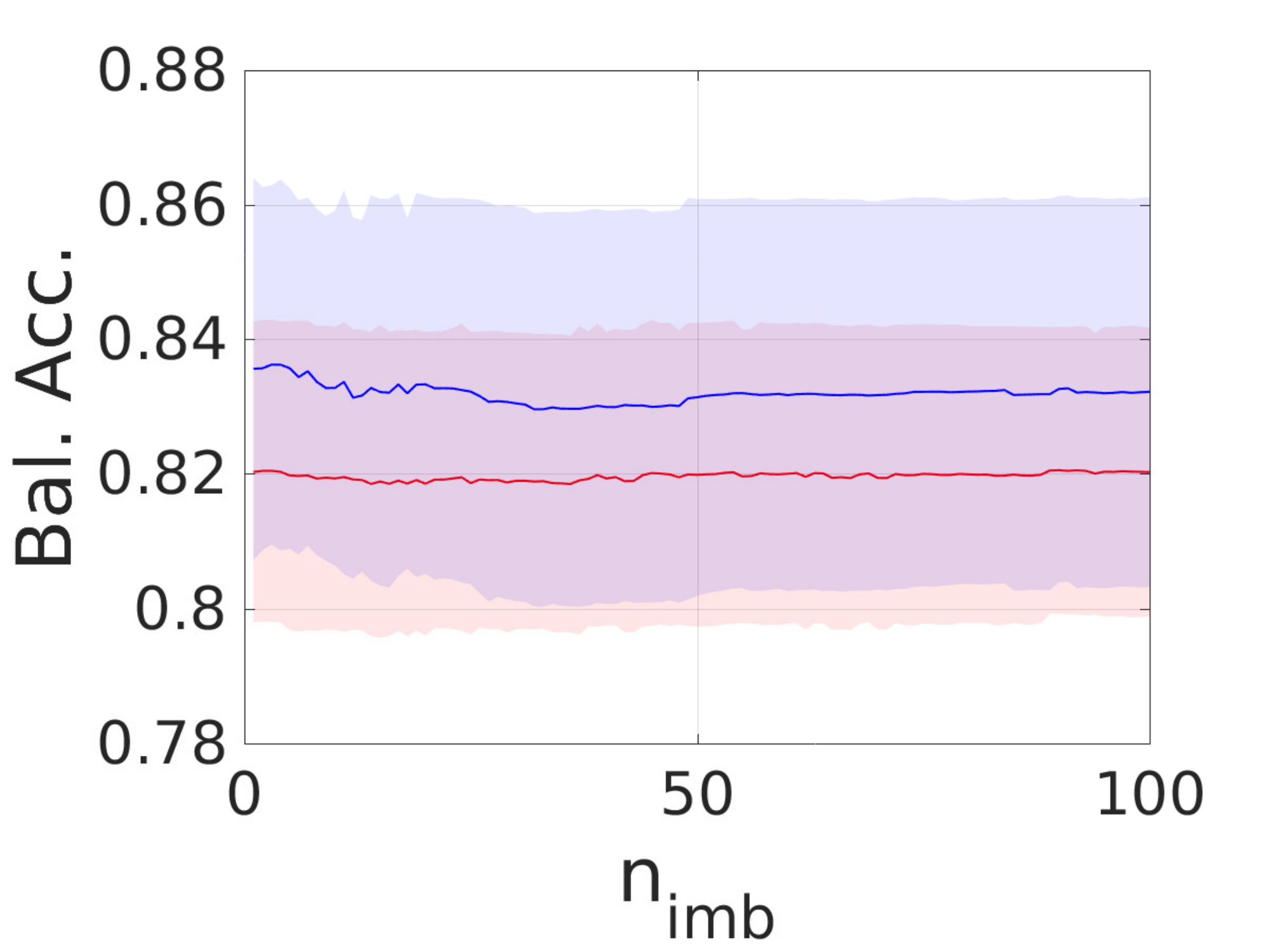}%
\caption{Average test classification accuracy of the standard incremental RLSC (red) and the variant proposed in this work (blue) over the imbalanced (top) and balanced (bottom) classes. The models are incrementally trained as $n_{imb}$ grows, as described in Subsection~\ref{sec:protocol}.}
\label{fig:acc}
\end{figure}

In Table~\ref{tab:incClassAccuracy} we report the classification accuracy of the three models evaluated in our experiments over {\it MNIST}, {\it iCubWorld28} and {\it RGB-D}. 
The RLSC baseline and Algorithm~\ref{alg:incremental_learning} were trained incrementally as new examples of the under-represented class were fed to the system, while the balanced approach was trained from scrach at each iteration. As can be noticed, Algorithm~\ref{alg:incremental_learning} consistently outperforms the RLSC baseline, which does not account for class imbalance and learns models which ignore the under-represented class. 
To offer a clear intuition of the improvement provided by our approach, in Figure~\ref{fig:acc} we report the test accuracy of both RLSC and Algorithm~\ref{alg:incremental_learning} on the balanced and under-represented classes as they are trained on new examples. Interestingly, on the two robotics tasks, our method significantly outperforms also the rebalanced approach. This is favorable, since the rebalanced approach cannot be implemented incrementally (Subsection~\ref{sec:incremental_recoding}). 

We care to point out that for the \textit{RGB-D} dataset, the overall classification accuracy of Algorithm~\ref{alg:incremental_learning} when trained with all available examples for all classes (last row in Table~\ref{tab:incClassAccuracy}) is approximately $84\%$, which is comparable with the state of the art on this dataset~\citep{schwarz2015}. Moreover, notice that the under-represented category ({\it tomato}) is in general easy to be distinguished, since all classifiers end up achieving $100\%$ accuracy on this class when sufficient examples are observed. However, RLSC completely ignores this class when only few examples are available, showing that even for ``easy'' classes the imbalance can heavily affect accuracy and that our approach successfully addresses this issue.\\

{\bf On On-line Backpropagation and Fine-tuning.} In principle, for the experiments performed on {\it iCubWorld28} and {\it RGB-D}, we could have directly trained the CaffeNet convolutional neural network on-line by stochastic gradient descent (backpropagation) rather than using it just as a feature extractor. Indeed, it has been empirically observed that in batch settings, {\it fine-tuning} a network previously trained on a large-scale dataset (such as ImageNet~\citep{ILSVRC15} for the case of CaffeNet) typically leads to remarkable performance improvements~\citep{chatfield2014}. However, our preliminary results in this direction showed that performing fine-tuning on-line without using training data from previous classes led to overfit the new class and overall extremely poor results, which we did not include in this work. This is in line with the previous literature studying the ``catastrophic effect of forgetting'' \citep{french1999,srivastava2013,goodfellow2013}, which occurs when training a stochastic model only on new examples while ignoring previous ones.



	\chapter{Incremental Inverse Dynamics Learning}
	\label{chap:invdyn}

\section{Setting}

In order to control a robot a model describing the relation between the actuator inputs, 
the interactions with the world and bodies accelerations is required. This model is called the \emph{dynamics} model of the robot. 
A dynamics model can be obtained from first principles in mechanics, using the techniques of Rigid Body Dynamics (RBD) \citep{reference/robo/FeatherstoneO08}, 
resulting in a \emph{parametric model} in which the values of physically meaningful parameters 
must be provided to complete the fixed structure of the model. 
Alternatively, the dynamical model can be obtained from experimental data using black-box machine learning techniques, resulting in a \emph{nonparametric model}.

Traditional dynamics parametric methods are based on several assumptions, such as rigidity of links or that friction has
a simple analytical form, which may not be accurate in real systems.  
On the other hand, nonparametric methods based on algorithms such as  Kernel Ridge Regression (KRR) \citep{RidgeRegression,SaundersGV98,cristianini2000introduction}, Kernel Regularized Least Squares (KRLS --- see Section \ref{sec:krr}) \citep{rifkin2003regularized} or Gaussian Processes \citep{rasmussen2006}
can model dynamics by extrapolating the input-output relationship directly from the available data\footnote{Note that KRR and KRLS have a very similar formulation, and that these are also equivalent to the techniques derived from Gaussian Processes, as explained for instance in Chapter 6 of  \citep{cristianini2000introduction}.}. 
If a suitable kernel function is chosen, then the nonparametric model is a universal approximator 
which can account for the dynamics effects which are not considered by the parametric model. 
Still, nonparametric models have no prior knowledge about the target function to be approximated. Therefore, they need a sufficient amount of training examples in order to produce accurate predictions on the entire input space.
If the learning phase has been performed offline, both approaches are susceptible to the variation of the mechanical properties over long time spans, which are mainly caused by temperature shifts and wear. Even the inertial parameters can change over time. For example if the robot grasps a heavy object, the resulting change in dynamics can be described by a change of the inertial parameters of the hand.  
A solution to this problem is to address the variations of the identified system properties by learning \emph{incrementally}, continuously
updating the model as long as new data becomes available. 
In this Chapter, we propose a novel technique that joins parametric and nonparametric model learning in an incremental fashion.

\begin{table}[t] 
\caption{Schematic comparison of this work with related ones on semiparametric or incremental robot dynamics learning.}
\begin{center}
\resizebox{0.6\textwidth}{!}{%
\begin{tabular}{| c | c c |} 
\hline
         \rowcolor[gray]{.9}                                                                                                        
          \textbf{Author, Year}                                   & \textbf{Parametric}     & \textbf{Nonparametric}  \\ 
          [0.5ex] \hline
         \citep{Nguyen-TuongP10}      & Batch  &  Batch   \\
         \rowcolor[gray]{.9}                                                                                                        
         \citep{GijsbertsM11}            &  -      &  Incremental   \\
         \citep{conf/iros/wu12}         &  Batch &  Batch   \\
         \rowcolor[gray]{.9}                                                                                                        
         \citep{conf/ifac/delacruz12}   &  CAD$^*$  &  Incremental  \\
         \citep{um2014independent}   &  CAD  &  Batch  \\
         \rowcolor[gray]{.9}                                                                                                        
         \citep{camoriano2016incremental}&  Incremental &  Incremental \\
         \hline
\end{tabular} 
}
\end{center}

\label{table:soa} 
$^*$ In \citep{conf/ifac/delacruz12} the parametric part is used only for initializing the nonparametric model. 
\end{table}


Classical methods for physics-based dynamics modeling can be found in \citep{reference/robo/FeatherstoneO08}. 
These methods require to identify the mechanical parameters of the rigid 
bodies composing the robot \citep{conf/humanoids/Yamane11,conf/humanoids/TraversaroPMNN13,conf/humanoids/OgawaVO14,hollerbach2008model}, 
which can then be employed in model-based control and state estimation schemes.

 In \citep{Nguyen-TuongP10}, the authors present a learning  technique which 
 combines prior knowledge about the physical structure of the mechanical
 system and learning from available data with Gaussian Process Regression (GPR) \citep{rasmussen2006}. 
 Similar approaches are presented in \citep{conf/iros/wu12,um2014independent}. Both techniques 
  require an offline training phase and are not incremental, limiting them to scenarios in which the properties of the system do not change significantly over time. 
 
In \citep{conf/ifac/delacruz12}, an incremental semiparametric robot dynamics
learning scheme based on Locally Weighted Projection Regression (LWPR) \citep{conf/icml/VijayakumarS00} is presented, that is initialized using a linearized parametric model. However, this approach uses a fixed parametric model, 
that is not updated as new data becomes available. Moreover, LWPR has been shown to underperform with respect to other methods (e.g. \citep{GijsbertsM11}).


In \citep{GijsbertsM11}, a fully nonparametric incremental approach for inverse dynamics learning with constant
 update complexity is presented, based on kernel methods \citep{schlkopf2002learning} (in particular KRR) and random features \citep{conf/nips/RahimiR07} (see Chapter \ref{chap:randfeats}). The incremental nature of this 
 approach allows for adaptation to changing conditions in time. The authors also show that the proposed algorithm outperforms other methods such as LWPR, GPR and Local Gaussian Processes (LGP) \citep{lgpr}, both in terms of accuracy and prediction time. Nevertheless, the fully nonparametric nature of this approach undermines the interpretability of the inverse dynamics model.




In this chapter we propose a method that is incremental with fixed update complexity (as \citep{GijsbertsM11}) and 
semiparametric (as  \citep{Nguyen-TuongP10} and \citep{conf/iros/wu12}). The fixed update complexity and prediction time are key properties of our method, enabling real-time performances. Both the parametric and nonparametric parts can be updated, 
as opposed to \citep{conf/ifac/delacruz12} in which only the nonparametric part is. A comparison between the existing literature and our incremental
method is reported in Table \ref{table:soa}.
We validate the proposed method with experiments performed on an arm of the iCub humanoid robot \citep{Metta2010}.
\begin{figure}[ht!]
\vspace{3mm}
\centering
\includegraphics[width=0.95\linewidth]{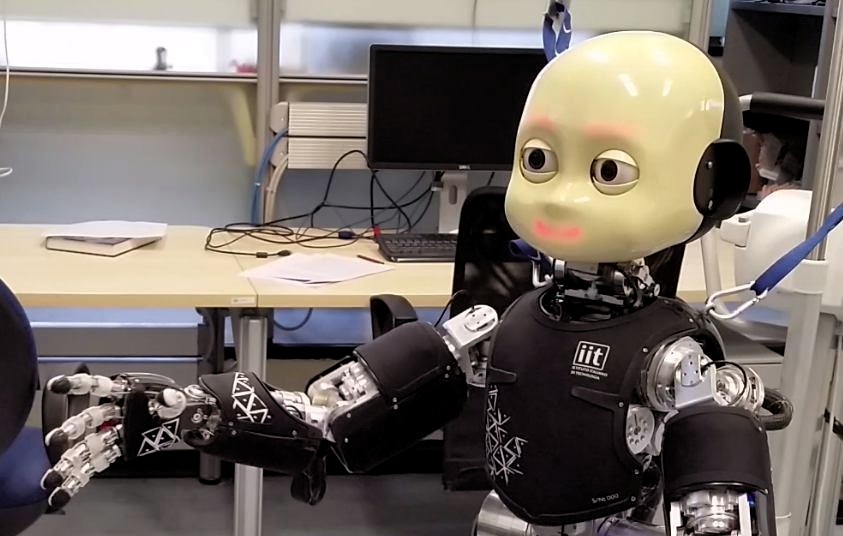}
\caption{iCub learning its right arm's dynamics.\label{overflow}}
\end{figure}

The chapter is organized as follows. 
Section \ref{sec:background} introduces the existing techniques for parametric and nonparametric robot dynamics learning.
In Section \ref{sec:semiparametric}, a complete description of the proposed semiparametric incremental learning technique is introduced. 
Section \ref{sec:experiments} presents the validation of our approach on the iCub humanoid robotic platform.

\section{Background}
\label{sec:background}


\subsection{Parametric Models of Robot Dynamics}
\label{sec:parmod}
Robot dynamics parametric models are used to represent the relation
connecting the geometric and inertial parameters with some dynamic quantities 
that depend uniquely on the robot model.
A typical example is obtained by writing the robot's inverse dynamics equation in linear form with respect to the 
robot inertial parameters $\boldsymbol\pi$:
\begin{equation}
\boldsymbol\tau = M(\mathbf{q})\mathbf{\ddot{q}} + C(\mathbf{q},\mathbf{\dot{q}})\mathbf{\dot{q}} + g(\mathbf{q}) = \regressor \boldsymbol\pi,
\end{equation}
where:
$\mathbf{q} \in \mathbb{R}^{n_{dof}}$ is the vector of joint positions,
$\boldsymbol\tau \in \mathbb{R}^{n_{dof}}$ is the vector of joint torques,
$\boldsymbol\pi  \in \mathbb{R}^{n_p}$ is the vector of the identifiable (base) inertial parameters \citep{reference/robo/FeatherstoneO08},
$\regressor \in \mathbb{R}^{n_{dof} \times n_p}$ is the ``regressor'', i.e. a matrix that depends only 
on the robot kinematic parameters.
In the rest of the chapter we will indicate with $x \in \mathbb{R}^{3n_{dof}}$ the input triple given by $(\mathbf{q},\mathbf{\dot{q}},\mathbf{\ddot{q}})$.
Other parametric models write different measurable quantities as a product 
of a regressor and a vector of parameters, for example  
the total energy of the robot \citep{gautier1988identification}, the instantaneous power provided to the robot \citep{gautier1997dynamic}, the sum of all external forces acting on the robot \citep{journal/ayusawa2014} or 
the center of pressure of the ground reaction forces \citep{conf/baelemans2013}. Regardless of the choice of the  measured variable $\outputQuant \in \mathbb{R}^{T}$ (row vector), 
the structure of the regressor is similar:
\begin{equation}
\label{eq:parametricModel}
\outputQuant^\top = \Phi(\mathbf{q},\mathbf{\dot{q}},\mathbf{\ddot{q}}) \boldsymbol\pi = \regressor \boldsymbol\pi.
\end{equation}
The $\inertialParameters$ vector is composed of certain linear combinations of the inertial parameters of the links, the \emph{base inertial parameters} \citep{khalil2004modeling}. 
In particular, the inertial parameters of a single body are the mass $m$, 
the first moment of mass $m\mathbf{c} \in \mathbb{R}^3$ expressed in a body fixed frame and the inertia matrix $I\in \mathbb{R}^{3 \times 3}$ expressed 
in the orientation of the body fixed frame and with respect to its origin.  
In \emph{parametric modeling} of robot dynamics, the regressor structure depends on the kinematic parameters 
of the robot, that are obtained from CAD models of the robot through kinematic calibration techniques. 
Similarly, the inertial parameters $\boldsymbol\pi$ can also be obtained from CAD models of the robot, 
however these models may be unavailable, for example because the manufacturer of the robot does not provide them. 
In this case the usual approach is to estimate $\boldsymbol\pi$ from experimental data \citep{hollerbach2008model}.
To do that, given $\nrOfSamples$ measures of the measured quantity $\outputQuant_i$ (with $i = 1 \dots \nrOfSamples$), stacking \eqref{eq:parametricModel} for the
$\nrOfSamples$ samples it is possible to write:

\begin{equation}
 \label{eq:LSParametric}
 \begin{bmatrix}
 \outputQuant_1^\top \\
 \outputQuant_2^\top \\
 \vdots       \\
 \outputQuant_{\nrOfSamples}^\top 
 \end{bmatrix}
 = 
 \begin{bmatrix}
 \regressorNum{1} \\
 \regressorNum{2} \\
 \vdots           \\
 \regressorNum{\nrOfSamples} 
 \end{bmatrix}
 \boldsymbol\pi .
\end{equation}

This equation can then be solved in least squares (LS) sense to find an estimate $\estimatedInertialParameters$ of the base inertial 
parameters. 
Given the training trajectories, it is possible that not all parameters in $\inertialParameters$ can be estimated well as the problem in \eqref{eq:LSParametric}
can be ill-posed, hence this equation is usually solved as a Regularized Least Squares (RLS --- see Section \ref{sec:RLS}) problem. Defining 
$$
\overline{\outputSet}_{\nrOfSamples} = \begin{bmatrix}
 \outputQuant_1^\top \\
 \outputQuant_2^\top \\
 \vdots       \\
 \outputQuant_{\nrOfSamples}^\top 
 \end{bmatrix} ,
 \hspace{1em}
 \overline{\bm{\Phi}}_{\nrOfSamples} = \begin{bmatrix}
 \regressorNum{1} \\
 \regressorNum{2} \\
 \vdots           \\
 \regressorNum{\nrOfSamples} 
 \end{bmatrix} ,$$
 the RLS problem that is solved for the parametric identification is:
\begin{equation}
 \label{eq:parametricRLS}
  \estimatedInertialParameters = \argmin\limits_{\inertialParameters \in  \mathbb{R}^{n_p}} \left( \|\overline{\bm{\Phi}}_{\nrOfSamples} \inertialParameters -\overline{\outputSet}_{\nrOfSamples}\|^2 + \lambda \| \inertialParameters \|^2 \right) , \lambda > 0.
\end{equation}


\subsection{Nonparametric Modeling with Kernel Methods}
\label{sec:np_modeling}

Kernel methods in the Statistical Learning Theory framework are the form of nonparametric modeling of interest for this chapter, and have already been discussed in Chapters \ref{Chap:SLT} and \ref{Chap:kernelMethods}.
In a nonparametric modeling setting, the goal is to find a function $f^*: \mathcal{X} \rightarrow \mathcal{Y}$ 
such that
\begin{equation}
\label{eq:rm}
f^* = \argmin\limits_{f \in \mathcal{H}} 
\underbrace{\int_{\mathcal{X} \times \mathcal{Y}} \ell(f(x),y) d\rho(x,y)}_{\mathcal{E}(f)},
\end{equation}
In the rest of this work, we will consider the squared loss $\ell(f(x),y) = \|f(x)-y\|^2$.
%
The optimization problem outlined in \eqref{eq:rm} can be approached empirically by means of many different algorithms, among which one of the most widely used is Kernel Regularized Least Squares (KRLS) \citep{SaundersGV98,rifkin2003regularized}.
In KRLS, as discussed in Section \ref{sec:krr}, a regularized solution $\widehat{f}_\la : \mathcal{X} \rightarrow \mathcal{Y}$ is found solving
\begin{equation}
\label{eq:tik}
\widehat{f}_\la = \argmin\limits_{f \in \mathcal{H}}  \sum_{i=1}^n \|f(x_i) - y_i\|^2 + \lambda \|f\|_{\mathcal{H}}^2 ,\quad \lambda > 0 ,
\end{equation}
where for simplicity we do not divide the sum of the point-wise errors by $n$, and it can be conveniently expressed as
\begin{equation}
\widehat{f}_{\la}(x) = \sum_{i = 1}^n \alpha_i K(x_i,x)
\end{equation}
with $\alpha = (
K_n+\lambda I_n)^{-1}Y \in \mathbb{R}^{n \times T}$, $\alpha_i$ $i$-th row of $\alpha$ and $Y = \left[ y^\top_1, \ldots, y^\top_n \right]^\top$.
As we know, it is necessary to invert and store the kernel matrix $K_n \in \mathbb{R}^{n \times n}$, which implies $O(n^3)$ and $O(n^2)$ time and memory complexities, respectively.
Such complexities render the above-mentioned KRLS approach prohibitive in settings where $n$ is large, including the one treated in this work. This limitation can be dealt with by resorting to approximated methods such as random features (see Chapter \ref{chap:randfeats}), which will now be recalled in this setting.

\subsubsection{Random Feature Maps for Kernel Approximation}
\label{sec:randfeats}
The empirical kernel matrix $K_n$ can become too cumbersome to invert and store as $n$ grows.
In this chapter, we introduce a random feature map $\tilde{\phi}: \mathbb{R}^d \rightarrow \mathbb{R}^D$
directly approximating the infinite-dimensional Gaussian kernel feature map $\phi$, so that
\begin{equation}
K(x,x') = 
e^{-\frac{\|x-x'\|^2}{2\sigma^2}} =
\left\langle \phi(x) , \phi(x') \right\rangle _{\mathcal{H}} \approx 
\tilde{\phi}(x)\tilde{\phi}(x')^\top.
\end{equation}
The number of random features $D$ can be chosen according to the desired approximation accuracy, as guaranteed by the convergence bounds reported in \citep{conf/nips/RahimiR07,rahimi2008uniform}.
The approximated feature map in this case is $\tilde{\phi}(x) = \left[ e^{i x {\omega}_1}, \ldots, e^{i x{\omega}_D} \right]$, where
\begin{equation}
{\omega} \sim p({\omega}) = (2 \pi)^{-\frac{D}{2}}e^{-\frac{\|{\omega}\|^2}{2\sigma^2}},
\end{equation}
with ${\omega} \in \mathbb{R}^d$ column vector. 
Therefore, it is possible to map the input data as $\tilde{x} = \tilde{\phi}(x) \in \mathbb{R}^{D}$, with $\tilde{x}$ row vector, to obtain a nonlinear and nonparametric model of the form
\begin{equation}
\tilde{f}(x) = \tilde{x} \widetilde{W} \approx \hat{f}_{\lambda}(x) = \sum_{i = 1}^n \alpha_i K(x_i,x)
\end{equation}
approximating the exact kernelized solution $\hat{f}_{\lambda}(x)$, with $\widetilde{W} \in \mathbb{R}^{D \times t}$. Note that the approximated model is nonlinear in the input space, but linear in the random features space. We can therefore introduce the regularized linear regression problem in the random features space as follows:
\begin{equation}
\label{eq:randfeatsminimizationprob}
\widetilde{W}^\lambda = \argmin\limits_{\widetilde{W} \in  \mathbb{R}^{d \times T}} \left( \| \widetilde{X} \widetilde{W} - Y \|^2 + \lambda \| \widetilde{W} \|^2 \right) , \quad \lambda > 0,
\end{equation}
where $\widetilde{X} \in \mathbb{R}^{n \times D}$ is the matrix of the training inputs where each row has been mapped by $\tilde{\phi}$.
The main advantage of performing a random feature mapping is that it allows us to obtain a nonlinear model by applying linear regression methods.
For instance, Regularized Least Squares (RLS --- see Section \ref{sec:RLS}) can compute the solution $\widetilde{W}^\lambda$ of \eqref{eq:randfeatsminimizationprob} with $O(nD^2)$ time and $O(D^2)$ memory complexities.
Once $\widetilde{W}^\lambda$ is known, the prediction $\hat{y} \in \mathbb{R}^{1 \times T}$ for a mapped sample $\tilde{x} $ can be computed as $\hat{y} = \tilde{x} \widetilde{W}^\lambda$.

\subsection{RLS for Parametric and Nonparametric Learning}
%
%
We now see in detail how RLS can be applied to solve both the parametric RBD and nonparametric random features-based dynamics learning problems.
Let $\RLSinputMat\in \mathbb{R}^{a\times b}$ and 
$\RLSoutputMat \in \mathbb{R}^{a\times c}$ be two matrices of real numbers, with $a,b,c \in \mathbb{N}^+$.
RLS computes a solution $W^\lambda \in \mathbb{R}^{b\times c}$ 
of the potentially ill-posed problem $\RLSinputMat W = \RLSoutputMat $. 
Considering the Tikhonov regularization scheme, $W^\lambda \in \mathbb{R}^{b\times c}$ is the solution to the problem
\begin{equation}
\label{eq:rlsminimizationprob}
W^\lambda = \argmin\limits_{W \in  \mathbb{R}^{b \times c}} \underbrace{ \left( \| \RLSinputMat W - \RLSoutputMat \|^2 + \lambda \| W \|^2 \right)}_{J(W,\lambda)}  , \quad \lambda>0 .
\end{equation}
As we know, the minimizing solution is
\begin{equation}
\label{eq:rlsSolution}
W^\lambda = (\RLSinputMat^\top \RLSinputMat + \lambda I_{b})^{-1}\RLSinputMat^\top \RLSoutputMat .
\end{equation}

Both the parametric identification problem \eqref{eq:parametricRLS} and the nonparametric random features problem \eqref{eq:randfeatsminimizationprob} are \textit{specific
instances} of the general problem \eqref{eq:rlsminimizationprob}.
In particular, the parametric problem \eqref{eq:parametricRLS} is equivalent to \eqref{eq:rlsminimizationprob} with
$$
W^\lambda = \hat{\boldsymbol\pi}, \hspace{1em} Z = \overline{\Phi}_n, \hspace{1em} U = \overline{\mathbf{y}}_n,
$$
while the random features learning problem \eqref{eq:randfeatsminimizationprob} is equivalent to \eqref{eq:rlsminimizationprob} with
$$
W^\lambda = \widetilde{W}^\lambda , \hspace{1em} Z = \widetilde{X} , \hspace{1em} U = Y.
$$
Hence, both problems for a given treining set can be solved applying \eqref{eq:rlsSolution}.

\subsection{Recursive RLS with Cholesky Update}
\label{sec:recupdate}

In scenarios in which supervised samples become available sequentially, a very useful extension of the RLS algorithm consists in the definition of an update rule for the model which allows it to be incrementally trained, increasing adaptivity to changes of the system properties through time.
This algorithm is called Recursive Regularized Least Squares (RRLS), already discussed in Subsection \ref{sec:recupdate} in its classification variant.
We consider RRLS with the Cholesky update rule \citep{bjoerck_least_squares96}, which is numerically more stable than others (e.g. the Sherman-Morrison-Woodbury update rule).
In adaptive filtering, this update rule is known as the \textit{QR algorithm} \citep{Sayed:2008:AF:1370975}.

Let us define $A = Z^\top Z + \lambda I_{b}$, with $\lambda > 0$, and $B = Z^\top U$. Our goal is to update the model (fully described by $A$ and $B$)
with a new supervised sample $({z}_{k+1}, {u}_{k+1})$, with ${z}_{k+1} \in \mathbb{R}^b$, ${u}_{k+1}\in \mathbb{R}^c$ row vectors.
Consider the Cholesky decomposition $A = R^\top R$. It can always be obtained, since $A$ is positive definite for $\lambda > 0$.
Thus, we can express the update problem at step $k+1$ as:
\begin{equation}
\begin{array}{r@{}l}
A_{k+1} &{}= R^\top_{k+1} R_{k+1}\\
&{}= A_{k} + \RLSinputSample_{k+1}^\top \RLSinputSample_{k+1}\\
&{}= R^\top_{k} R_{k} + \RLSinputSample_{k+i}^\top \RLSinputSample_{k+1} , 
\end{array}
\end{equation}
where $R$ is full rank and unique, and $R_0 = \sqrt{\lambda} I_b$.\\
By defining
\begin{equation}
\widetilde{R}_{k} =
\left[ \begin{array}{c}
R_{k} \\
 \RLSinputSample_{k+1} \end{array} \right] \in \mathbb{R}^{b+1 \times b},
\end{equation}
we can write $A_{k+1}=\widetilde{R}_{k}^\top \widetilde{R}_{k}$. However, in order to compute $R_{k+1}$ from the obtained $A_{k+1}$ it would
be necessary to recompute its Cholesky decomposition, requiring $O(b^3)$ computational time. There exists a procedure, based on Givens
rotations, which can be used to compute $R_{k+1}$ from $\widetilde{R}_{k}$ with $O(b^2)$ time complexity.
A recursive expression can  be obtained also for $B_{k+1}$ as follows:

\begin{equation}
\label{eq:rrlsUpdate}
\begin{array}{r@{}l}
B_{k+1} &{}= \RLSinputMat_{k+1}^\top \RLSoutputMat_{k+1}\\
&{}= \RLSinputMat_{k}^\top \RLSoutputMat_{k} + \RLSinputSample_{k+1}^\top \RLSoutputSample_{k+1}.
\end{array}
\end{equation}
Once $R_{k+1}$ and $B_{k+1}$ are known, the updated weights matrix $W_k$ can be obtained via back and forward substitution as 
\begin{equation}
W_{k+1} = R_{k+1} \setminus (R^\top_{k+1} \setminus B_{k+1}) .
\end{equation}
The time complexity for updating $W$ is $O(b^2)$.

As for RLS, the RRLS incremental solution can be applied to both the parametric \eqref{eq:parametricRLS} and nonparametric  with random features \eqref{eq:randfeatsminimizationprob} problems, assuming $\lambda > 0$. 
In particular, RRLS can be applied to the parametric case by noting that the arrival of a new sample $\left( \Phi_r , {y}_r \right)$ adds 
$T$ rows to $Z_k = \overline{\Phi}_{r-1}$ and $U_k = \overline{\mathbf{y}}_{r-1}$. Consequently, the update of $A$ must be decomposed 
in $T$ update steps using \eqref{eq:rrlsUpdate}. For each one of these $T$ steps we consider only one row of $\Phi_{r}$ and $y^\top_r$, namely: 
$$
\RLSinputSample_{k+i} = (\Phi_r)_i , \hspace{1em} \RLSoutputSample_{k+i} = (y^\top_r)_i , \hspace{1em} i = 1 \dots T 
$$
where $(V)_i$ is the $i$-th row of the matrix $V$ or the $i$-th element of vector $V$.

On the other hand, in the nonparametric random features case, RRLS can be applied considering
$$
\RLSinputSample_{k+1} = \tilde{x}_r , \hspace{1em} \RLSoutputSample_{k+1} = y_r ,
$$
where $\left(\tilde{ x}_r,  y_r \right)$ is the supervised sample at step $r$.

\section{Semiparametric Incremental Dynamics Learning}
\label{sec:semiparametric}

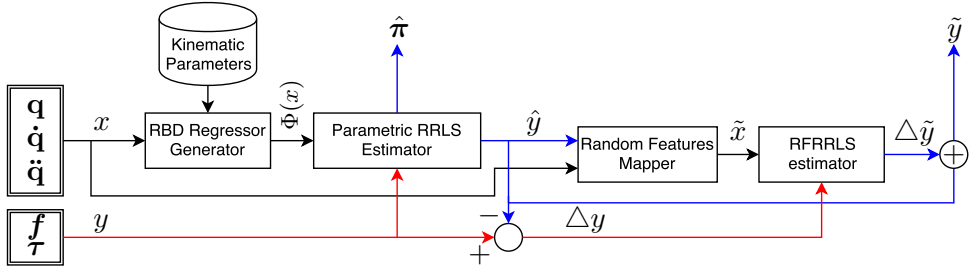
\begin{figure*}
\centering
\vspace{3mm}
\begin{overpic}[width=\textwidth]{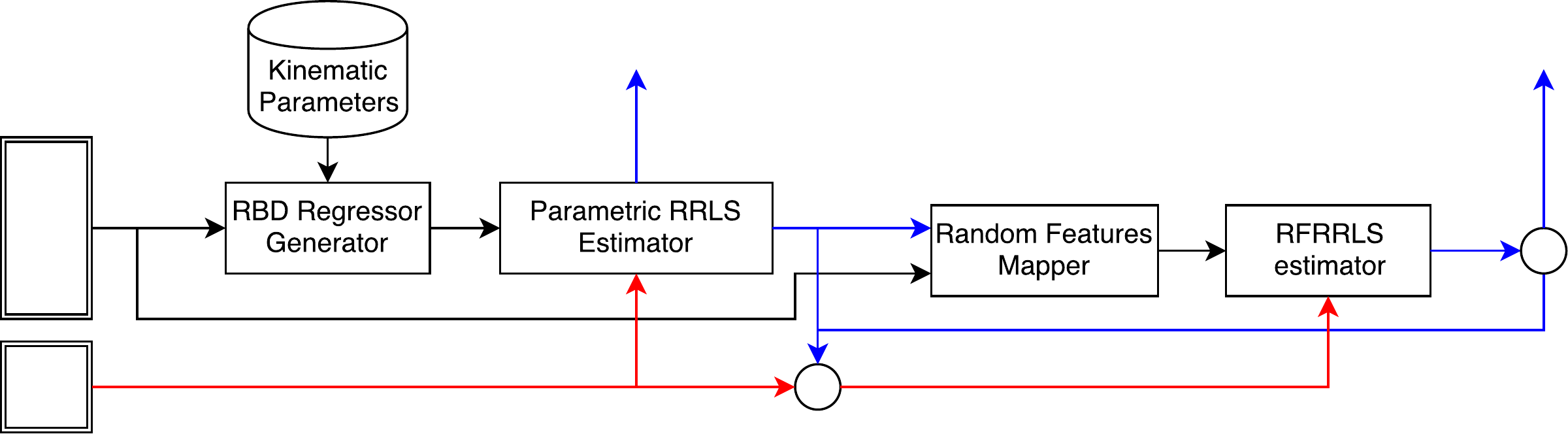}
\put (2.1,16) {$ \mathbf{q} $}
\put (2.1,12.5) {$ \mathbf{\dot{q}}  $}
\put (2.1,9) {$ \mathbf{\ddot{q}} $}
\put (2.1,3.3) {$ \bm{f} $}
\put (2.1,1.1) {$ \bm{\tau} $}
\put (9,14.2) {$ {x} $}
\put (9,4) {$ {y} $}
\put (28.4,14.2) {\rotatebox{90}{\footnotesize $ \Phi({x}) $}}
\put (39.73,24) {$ \hat{\bm{\pi}}$}
\put (54,14.2) {$ \hat{{y}} $}
\put (75.2,13.2) {$ \tilde{{x}} $}
\put (92.2,13.2) {$ \triangle \tilde{{y}} $}
\put (49,4.5) {$ - $}
\put (48,0.4) {$ + $}
\put (58,4) {$ \triangle {y} $}
\put (97.3,10.9) {$ + $}
\put (97.8,24) {$ \tilde{{y}}$}
\end{overpic}
\caption{Block diagram displaying the functioning of the proposed prioritized semiparametric inverse dynamics estimator. $ \bm{f} $ and $ \bm{\tau} $ indicate measured force and torque components, concatenated in the measured output vector ${y}$. The parametric part is composed of the RBD regressor generator and of the parametric estimator based on RRLS. Its outputs are the estimated parameters $\hat{\bm{\pi}}$ and the predicted output $\hat{{y}}$. The nonparametric part maps the input to the random features space with the Random Features Mapper block, and the RFRRLS estimator predicts the residual output $\triangle \tilde{{y}} $, which is then added to the parametric prediction $\hat{{y}}$ to obtain the semiparametric prediction $\tilde{{y}}$.}
\label{fig:blockdia}
\end{figure*}

We propose a semiparametric incremental inverse dynamics estimator, designed to have better generalization properties with respect to fully parametric and nonparametric ones, both in terms of accuracy and convergence rates.
The estimator, whose functioning is illustrated by the block diagram in Figure \ref{fig:blockdia}, is composed of two main parts.
The first one is an incremental parametric estimator taking as input the rigid body dynamics regressors $ \Phi(x) $ and computing two quantities at each step:
\begin{itemize}
\item An estimate $\hat{{y}}$ of the output quantities of interest.
\item An estimate $\hat{\bm{\pi}}$ of the base inertial parameters of the links composing the rigid body structure.
\end{itemize}
The employed learning algorithm is RRLS. Since it is supervised, during the model update step the real measured output ${y}$ is used by the learning algorithm as ground truth.
The parametric estimation is performed first, and is independent of the nonparametric part. 
This property is desirable in order to give priority to the identification of the inertial parameters $\bm{\pi}$. 
Moreover, since the estimator is incremental, the estimated inertial parameters $\hat{\bm{\pi}}$ adapt to changes in the inertial properties of the links, which can occur if the end-effector is holding a heavy object. 
Still, this adaptation cannot address changes in nonlinear effects which do not respect the rigid body assumptions.\\
The second estimator is also RRLS-based, fully nonparametric and incremental. It leverages the approximation of the kernel function via random Fourier features, as outlined in Subsection  \ref{sec:randfeats}, to obtain a nonlinear model that can be updated incrementally with constant update complexity $O(D^2)$ (see Subsection \ref{sec:recupdate}).
This estimator receives as inputs the current vectorized ${x}$ and the parametric estimation $\hat{{y}}$, normalized and mapped to the random features space approximating an infinite-dimensional feature space introduced by the Gaussian kernel.
The supervised output, used for training the nonparametric part, is the residual $ \triangle {y} = {y} - \hat{{y}}$.
The nonparametric estimator provides as output the estimate $ \triangle \tilde{{y}} $ of the residual, which is then added to $\hat{{y}}$ to obtain the final semiparametric estimate $ \tilde{{y}}$.
Similarly to the parametric part, in the nonparametric one the estimator's internal nonlinear model can be updated during operation, which constitutes an advantage in the case in which the robot has to explore a previously unseen area of the state space, or when the mechanical conditions change (e.g. due to wear, tear or temperature shifts).

\section{Experimental Results}
\label{sec:experiments}

\subsection{Software}
For implementing the proposed algorithm we used two existing 
open source libraries. 
For the RRLS learning part we used GURLS \citep{tacchetti2013gurls}, 
a regression and classification library based on the Regularized Least Squares (RLS) algorithm, available for 
Matlab and C++. 
For the computations of the regressors $\Phi(\mathbf{q},\mathbf{\dot{q}},\mathbf{\ddot{q}})$
we used iDynTree%
\footnote{\url{https://github.com/robotology/idyntree}} (see~\citep{10.3389/frobt.2015.00006}), a C++ dynamics library designed for free floating robots. Using SWIG~\citep{beazley1996swig}, iDynTree supports 
calling its algorithms in several programming languages, such as Python, Lua and Matlab. 
For producing the presented results, we used the Matlab interfaces of iDynTree and GURLS. 

\subsection{Robotic Platform}
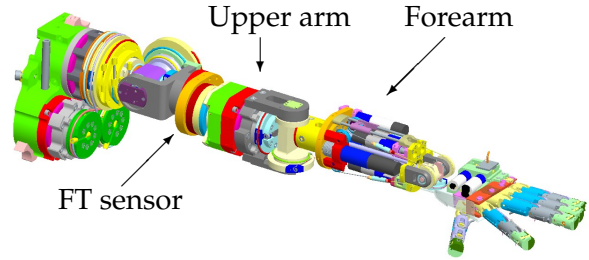
\begin{figure}[htb]
\centering
\begin{overpic}[width=0.6\textwidth]{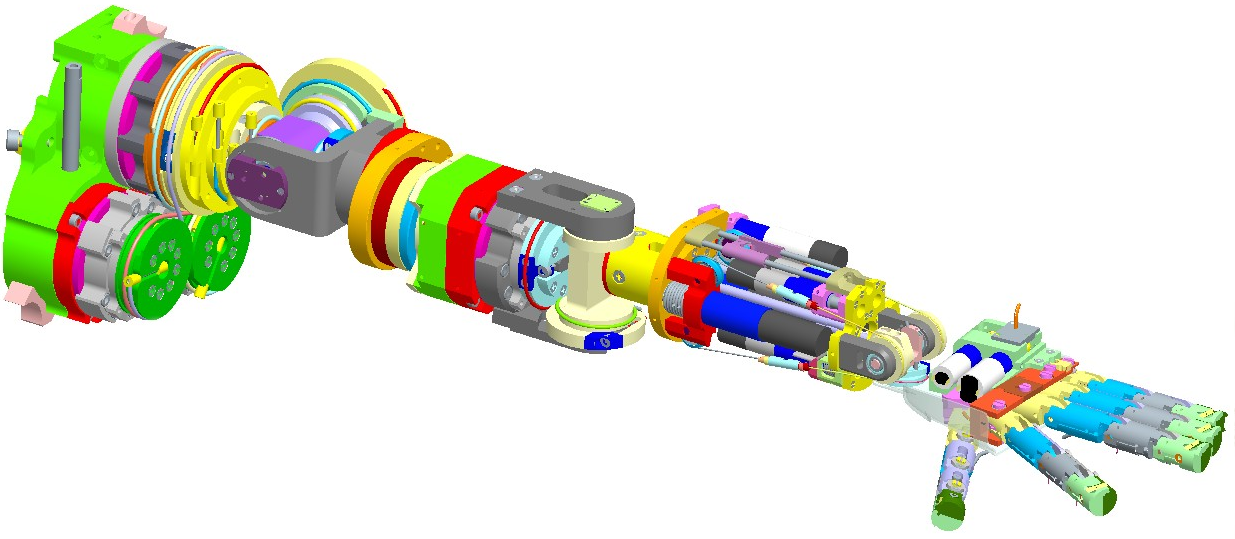}\put(8,9){FT sensor}
\put(19,13){\vector(1,1){9}}
\put(34,41){Upper arm}
\put(43,39){\vector(0,-1){6}}
\put(68,41){Forearm}
\put(75,39){\vector(-1,-1){10}}
\end{overpic}
\caption{CAD drawing of the iCub arm used in the experiments. The six-axis F/T sensor used for validation is visible in the middle of the upper arm link.}
\label{fig:cadArm}
\end{figure}

iCub is a full-body humanoid with 53 degrees of freedom \citep{Metta2010}. For validating 
the presented approach, we learned the dynamics of the right arm of the iCub as measured
from the proximal six-axis force/torque (F/T) sensor embedded in the arm (see Figure \ref{fig:cadArm}). 
The considered output ${y}$ is the reading of the F/T sensor, and the inertial
parameters $\boldsymbol\pi$ are the base parameters of the arm~\citep{traversaro2015inertial}.
As ${y}$ is not a input variable for the system, the output of the dynamic model is not directly usable for control, 
but it is still a proper benchmark for the dynamics learning problem, as also shown in ~\citep{GijsbertsM11}.
Nevertheless, the joint torques could be computed seamlessly from the F/T sensor readings  if needed for control purposes, by applying the method presented in \citep{6100813}.

\subsection{Validation}
\label{sec:validation}
\begin{figure}[th!]
\centering
\includegraphics[width=0.95\linewidth]{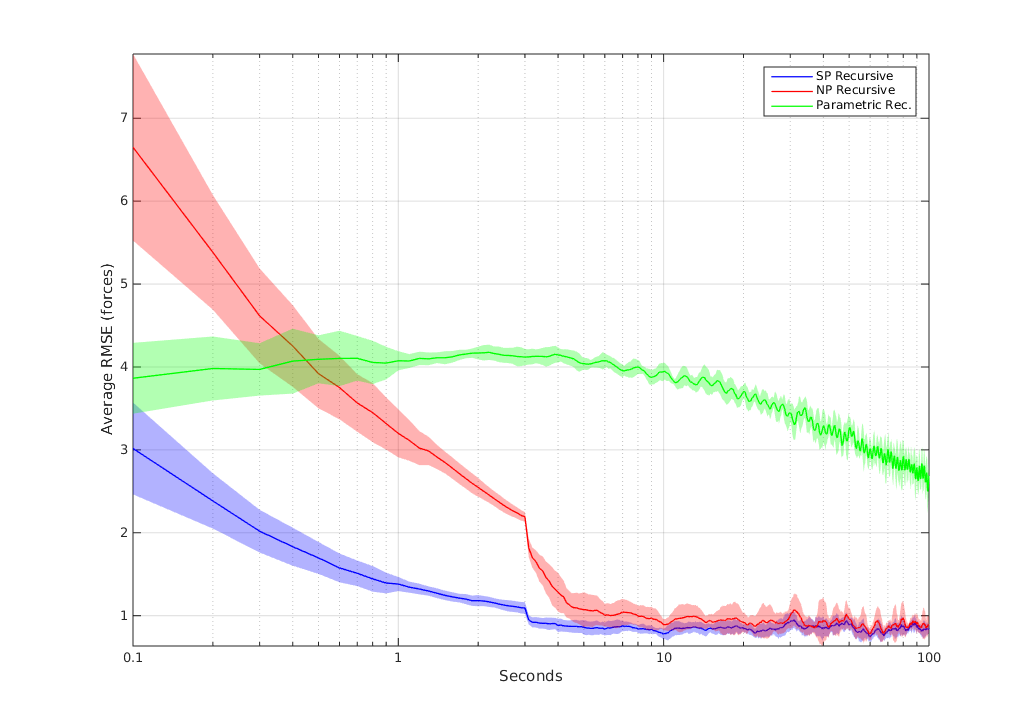}\\
\includegraphics[width=0.95\linewidth]{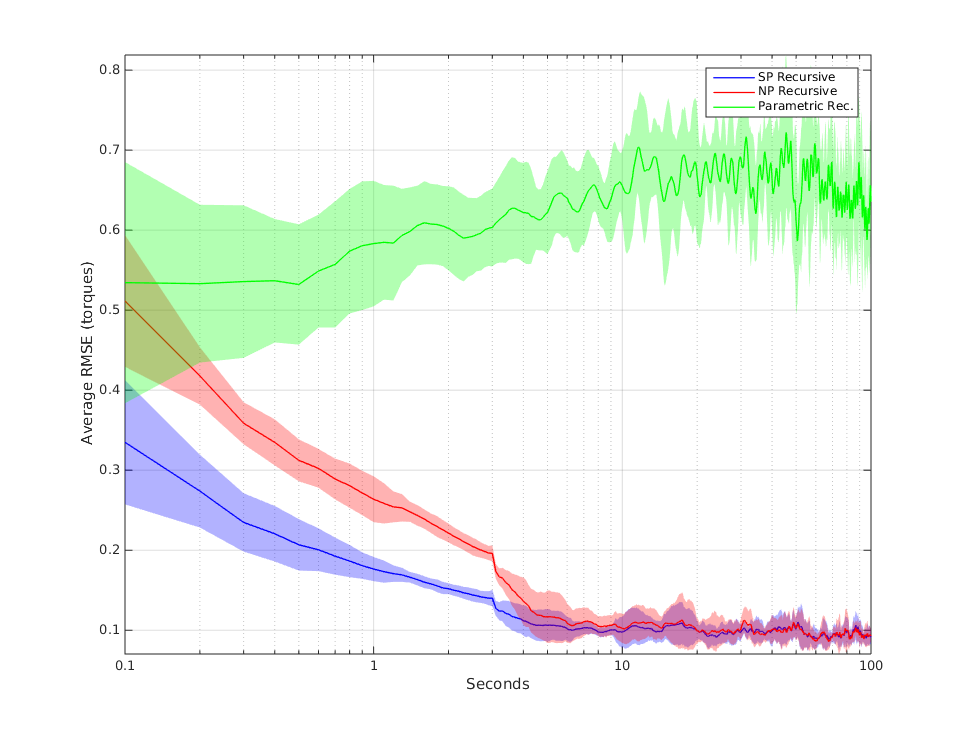}
\caption{Predicted forces (top) and torques (bottom) components average RMSE ($y$ axis) in time ($x$ axis), averaged over a 30-samples window for the recursive SP (blue), NP (red) and P (green) estimators. The transparent areas correspond to the standard deviation over 10 repetitions.}
\label{fig:exp1}
\end{figure}
We now present the results of the experimental validation of the proposed semiparametric model.
The model includes a parametric part, based on physical modeling.
This part is expected to provide acceptable prediction accuracy for the force components in the whole workspace of the robot, since it is based on prior knowledge about the structure of the robot itself, which does not abruptly change as the trajectory changes.
On the other hand, the nonparametric part can provide higher prediction accuracy in specific areas of the input space for a given trajectory, since it also models nonrigid body dynamics effects by learning directly from data.
To provide empirical foundations to the above insights, a validation experiment has been set up using the right arm of the iCub humanoid robot, considering as input the positions, velocities and accelerations of the 3 shoulder joints and of the elbow joint, and as outputs the 3 force and 3 torque components measured by the six-axis F/T sensor in-built in the upper arm.
We employ two datasets for this experiment, collected at $10 Hz$ as the end-effector tracks (using the Cartesian controller presented in \citep{5650851}) circumferences with $10cm$ radius on the transverse ($XY$) and sagittal ($XZ$) planes\footnote{For more information on the iCub reference frames, see \url{http://eris.liralab.it/wiki/ICubForwardKinematics}} at approximately $0.6 m/s$. The total number of points for each dataset is $10000$, corresponding to approximately $17$ minutes of continuous operation.
The steps of the validation experiment are the following:
\begin{enumerate}
\item Initialize the recursive parametric, nonparametric and semiparametric models to zero. The inertial parameters are also initialized to zero
\item Train the models on the whole $XY$ dataset (10000 points)
\item Split the $XZ$ dataset in 10 sequential parts of 1000 samples each. Each part corresponds to 100 seconds of continuous operation
\item Test and update the models independently on the 10 splitted datasets, one sample at a time.
\end{enumerate}
In Figure \ref{fig:exp1} we present the means and standard deviations of the average root mean squared error (RMSE) of the predicted force and torque components on the 10 different test sets for the three models, averaged over a 3-seconds sliding window. 
At steady state, the nonparametric (NP) and the semiparametric (SP) models provide more accurate predictions than the parametric (P) model with statistical significance.
In fact, their force prediction error is approximately $1N$, while the one of the P model is approximately 2 to 3 times larger. 
Similarly, the torque prediction error is around $0.1Nm$ for SP and NP, which is considerably better than the $0.5Nm$ to $0.7Nm$ average RMSE of the P model.
The SP model outperforms the P model, both in the transient and at regime, since it makes use of both the prior physical information and of the available data.
It is particularly interesting to note that the error of the SP model is considerably lower than the NP one \textit{in the initial transient}, both for forces and torques.
This indicates that the SP model takes advantage of the prior knowledge about the physical structure of the system in an area of the state space that is new for the fully data-driven NP model.\\
Given these experimental results, we can conclude that in terms of predictive accuracy the proposed incremental semiparametric method outperforms the incremental parametric one and matches the fully nonparametric one.
The SP method also shows a smaller standard deviation of the error with respect to the competing methods.
Considering the previous results and observations, the proposed method has been shown to be able to combine the main advantages of parametric modeling (i.e. interpretability) with the ones of nonparametric modeling (i.e. capacity of modeling nonrigid body dynamics phenomena).
The incremental nature of the algorithm, in both its P and NP parts, allows for adaptation to changing conditions of the robot itself and of the surrounding environment.
Note that the introduction of a forgetting factor to assign smaller weights to older or less relevant samples might be necessary to avoid saturation phenomena in long time spans and properly track the changing system dynamics (see \citep{paleologu2008robust} and references therein).
This interesting aspect will be addressed in future work.
%
%



\pagebreak
\chapter{Conclusion}
\label{chap:concl}
In this work, we address several open problems in large-scale machine learning and propose some applications of our methods to the solution of challenging lifelong  robot learning tasks.
In particular, we focus on randomized methods to scale up kernel machines to tackle learning problems with a large (and potentially growing) number of training examples.
We analyze the generalization properties of randomized learning algorithms in the statistical learning theory framework, outlining the tight relationship between statistics and optimization giving rise to novel computational regularization schemes, provably allowing for optimal generalization with reduced computational burden.\\
First, in Part \ref{part:largeScale} we analyze the generalization properties of two of the most widely used randomized subsampling schemes for kernel methods, namely Nystr\"om methods and random features.
Theoretical results are accompanied by numerical simulations and experiments on large-scale benchmark datasets.
In particular, Chapter \ref{chap:lessismore} is focused on data-dependent subsampling schemes, namely Nystr\"om methods.
In this context, in Section \ref{sec:lessismore} we prove novel optimal generalization bounds for Nystr\"om-based kernel methods with Tikhonov regularization (NKRLS), provided that the subsampling level is appropriately chosen.
Moreover, we show that the subsampling level controls regularization and computations at the same time.
In Section \ref{sec:nytro} we investigate if iterative regularization/early stopping and Nystr\"om methods can be fruitfully combined.
We answer this question by introducing the NYTRO (NYst\"om iTerative RegularizatiOn) algorithm, and by analyzing its statistical and computational properties in different regimes.
On the other hand, in Chapter \ref{chap:randfeats} we show that a number of random features smaller than the number of examples can be sufficient to achieve optimal learning rates.
We also show that random features mapping can be interpreted as a form of regularization, in which the number of random features is not limited to be a computational efficiency parameter, but also controls a bias-variance trade-off.\\
Part \ref{part:lifelong} is devoted to the investigation of the generalization performance of online/incremental learning methods, and to their application in lifelong learning scenarios of interest for robotics, such as visual object recognition and dynamics learning.
Chapter \ref{chap:sgd} is concerned with the generalization properties of the stochastic gradient method (SGM) for learning with convex loss functions and linearly parametrized functions.
We show that the stability and approximation properties of the algorithm can be controlled by the number of passes over the training data or the step-size, without any additional penalization or constraints, thus implementing a form of implicit regularization.
Interestingly, depending on the statistical properties of the data distribution, multiple passes might be required for optimal performance.
Numerical results complement our theoretical findings.\\
In Chapter \ref{chap:incclass} we address the problem of learning on-line with an increasing number of classes. 
Motivated by the visual object recognition applications in a lifelong learning robotic setting, we focus on the problems related to class imbalance, which naturally arises when a new class/object is observed for the first time. 
To address these issues, we propose a variant of the incremental Regularized Least Squares for Classification (RLSC) algorithm that incorporates new classes and dynamically applies class recoding when new examples are observed. 
Updates are performed in constant time with respect to the number of training examples seen so far. 
We evaluate the proposed algorithm on a standard machine learning benchmark and on two datasets for visual object recognition in robotics, showing that our approach is indeed favorable in on-line settings when classes are imbalanced.
In line with the literature on ``learning to learn'' and transfer learning~\citep{thrun1996}, future research will focus on strategies to exploit knowledge of previous, well-represented classes, to improve classification accuracy on novel and under-represented ones. Indeed, as empirically observed in recent work~\citep{tommasi2010,kuzborskij2013,sunneol2016}, sharing information and structures among classification tasks can dramatically improve performance.\\
Finally, in Chapter \ref{chap:invdyn} we present a novel incremental semiparametric modeling approach for inverse dynamics learning, joining together the advantages of parametric modeling derived from rigid body dynamics equations and of nonparametric learning methods.
A distinctive trait of the proposed approach lies in its incremental nature, encompassing both the parametric and nonparametric parts and allowing for the prioritized update of both the identified base inertial parameters and the nonparametric weights.
This feature is key to enabling robotic systems to adapt to mutable conditions of the environment and their own mechanical properties throughout extended time periods.
We validate our approach on the iCub humanoid robot by analyzing the performances of a semiparametric inverse dynamics model of its right arm and comparing them with the ones obtained by state of the art fully nonparametric and parametric approaches.

\pagebreak
\bibliography{thesisBib}
\addcontentsline{toc}{chapter}{Bibliography}

\begin{appendices}

\chapter[Loss Functions and Target Functions]{Different Loss Functions Yield Different Target Functions}
\label{app:lossTarget}
The choice of the loss function reflects itself on the target function resulting from the associated minimization of the expected risk.
We show this in practice by a simple binary classification example, first with the square loss and then with the logistic loss.

\begin{ex}[Square loss]
Recall that the square loss is defined as $\ell(a,b) = ( a-b )^2$, with $a,b \in \R$.
Thus, the minimization of the expected error takes the form
\begin{align}
\begin{split}
{\mathcal E}(f)&=\int \rho(x,y) \ell(y,f(x)) dxdy  \\ 
&=\int \rho(y|x)\rho_{\mathcal X}(x)dxdy (y-f(x))^2 \\
&= \int \rho_{\mathcal X}(x)dx\int \rho(y|x)dy (y-f(x))^2.
\end{split}
\end{align}
To simplify the computation, it can be shown that it is possible to disregard the first term and write the target function as
$$
f^*(x)=\arg\min_{a\in \mathbb R} \int \rho(y|x)dy (y-a)^2 \quad \forall x \in \mathcal{X},
$$
which corresponds to
$$
f^*(x) = a \quad \text{s. t.} \quad \dfrac{d}{da} \left[ \int \rho(y|x)dy (y-a)^2 \right] = 0 \quad \forall x \in \mathcal{X}.
$$
Therefore, 
\begin{gather*}
\dfrac{d}{da} \left[ \int \rho(y|x)dy (y-a)^2 \right] = 0\\
\int \rho(y|x) \dfrac{d}{da} \left[ (y-a)^2 \right] dy = 0 \\
\int \rho(y|x)  2(a-y) dy = 0 \\
2a \underbrace{\int \rho(y|x)  dy}_{1} - 2 \int \rho(y|x)  y dy  = 0\\
a = \int \rho(y|x)  y dy 
\end{gather*}
Thus, the target function is
$$f^*(x) = \int \rho(y|x)  y dy.$$
Since we are considering the binary classification setting, we can further decompose $f^*(x)$ as
$$
f^*(x)= (+1)(p) + (-1)(1-p) = 2p-1,
$$ 
with $p=p(1|x), 1-p=p(-1|x)$.
Thus, if $p(1|x) = 1$ then $f^*(x)=1$ while if $p(-1|x) = 1$ then $f^*(x)=-1$, which justifies taking the sign of the learned estimator $f$ to obtain the output class label.
\end{ex}

\begin{ex}[Logistic loss]
Consider the {\em logistic loss}
$$
\ell(a,b)\log(1+e^{-ab})
$$
with $a,b \in \R$.
The target function can be written as
\begin{align}
\begin{split}
f^*(x)&=\argmin_{a\in \mathbb R} \int \rho(y|x) \log(1+e^{-ya}) dy = \\
&= \argmin_{a\in \mathbb R} p\log(1+e^{-a})+(1-p)\log(1+e^{a}).
\end{split}
\end{align}

By performing similar calculations to the ones of the previous example, we obtain that $f^*(x) = a$ such that
\begin{gather*}
p\frac{-e^{-a}}{(1+e^{-a})}+(1-p)\frac{e^{a}}{(1+e^{a})}=0\\
-p\frac{1}{(1+e^{a})}+(1-p)\frac{e^{a}}{(1+e^{a})} = 0\\
p=\frac{e^{a}}{(1+e^{a})}\\
a=\log{\frac{p}{1-p}}.
\end{gather*}
\end{ex}

\chapter{Linear Systems}
Consider the problem 
$$
Ma=b, 
$$
where $M$ is a $d \times d$ matrix and $a,b$ vectors in $\mathbb R ^d$. 
We are interested in determing $a$ satisfying the above equation given $M,b$. 
If $M$ is invertible, the solution to the problem is 
$$
a=M^{-1}b.
$$
\begin{itemize}
\item  If $M$ is a diagonal $M=diag(\sigma_1, \dots, \sigma_d)$
where $\sigma_i\in (0, \infty)$ for all $i=1, \dots, d$, then 
$$
M^{-1}=diag(1/\sigma_1, \dots, 1/\sigma_d),\quad (M+\lambda I)^{-1}=diag(1/(\sigma_1+\lambda), \dots, 1/(\sigma_d+\lambda)
$$
\item If $M$ is symmetric and positive definite, then 
considering the eigendecomposition 
$$
M=V \Sigma V^\top ,\quad \Sigma=diag(\sigma_1, \dots, \sigma_d), ~ VV^\top =I,
$$ 
then 
$$
M^{-1}=V \Sigma^{-1}V^\top ,\quad \Sigma^{-1}=diag(1/\sigma_1, \dots, 1/\sigma_d),$$ 
and 
$$
(M+\lambda I)^{-1}=V\Sigma_\lambda V^\top , \quad  \Sigma_\lambda=diag(1/(\sigma_1+\lambda), \dots, 1/(\sigma_d+\lambda)
$$
\end{itemize}
The ratio $\sigma_d/\sigma_1$ is called the {\it condition number} of $M$. 


\chapter[Incremental Random Features]{Incremental Algorithm for Random Features Regularization}
\label{sect:incr-algo}

The algorithm below was used in Section~\ref{sec:exp} and is a variation of the incremental algorithm introduced in \citep{rudi2015less} for KRLS with \Nystrom{} approximation. 
It computes the regularized least squares solution for random features (see \eqref{eq:RF-KRLS}) incrementally in the number of random features, by means of Cholesky updates. In this way, for a given regularization parameter $\la$ and a number of random features $m_{\max}$, the algorithm computes all the possible estimators $\tilde{f}_{1,\la}, \dots \tilde{f}_{m_{\max},\la}$ at the cost of computing only $\tilde{f}_{m_{\max},\la}$. Thus it is possible to explore the whole regularization path in the number of random features with computational cost $O(n m_{\max}^2 + m_{\max}^3)$, instead of $O(n m_{\max}^3 + L m_{\max}^4)$.

\begin{center}
\centering
\begin{algorithmic}
 \State {{\bf Input:} Dataset $(x_i, y_i)_{i=1}^n$, Feature mappings $(\psi(\omega_j,\cdot))_{j=1}^m$, Regularization Parameter $\la$.}
 \State  {{\bf Output:} coefficients $\{\tilde\alpha_{1,\lambda},\dots,\tilde\alpha_{m,\lambda}\}$, see \eqref{eq:RF-KRLS}.}
 \State $A_0 \gets ()$;~~ $R_1 \gets ();$\\
 {\bf for} $t \in \{1,\dots, m\}$ {\bf do}\\
  \qquad $a_t = (\psi(\omega_t,x_1),\dots ,\psi(\omega_t,x_n))$;~~$A_t = (A_{t-1}~~a_t)$; \\
  \qquad $\gamma_t = a_t^\top a_t + \la n$;~~$c_t = A_{t-1}^\top a_t$;~~$g_t = (\sqrt{3} - 1) \gamma_t$; \\
  \qquad $u_t \gets (2c_t/g_t,\,\sqrt{ 3 \gamma_t / 4})$;~~$v_t = (-2c_t/g_t,\,\sqrt{\gamma_t / 4})$;\\
  \qquad $R_t \gets \begin{pmatrix} R_{t-1} & 0\\ 0 & 0\end{pmatrix}$;\\
  \qquad $R_t \gets {\tt choleskyupdate}(R_t, u_t,  \text{`}+\text{'})$;\\
  \qquad $R_t \gets {\tt choleskyupdate}(R_t, v_t, \text{`}+\text{'})$;\\
  \qquad $\tilde\alpha_{t,\la} \gets R_t^{-1} (R_t^{-\top} (A_t^\top y));$\\
 {\bf end}
\end{algorithmic}
\end{center}

\chapter{Incremental Algorithm for Nystr\"om Computational Regularization}
\label{sect:algNysCompReg}

Let $(x_i, y_i)_{i=1}^{n}$ be the dataset and $(\tilde{x}_i)_{i=1}^{m}$ be the selected \Nystrom{} points. 
We want to compute $\tilde \alpha$ of \eqref{eq:repny}, incrementally in $m$. Towards this goal we compute an incremental Cholesky decomposition $R_t$ for $t \in \{1,\dots,m\}$ of the matrix $G_t = K_{n t}^\top K_{n t} + \la n K_{t t}$, and the coefficients $\tilde{\alpha}_t$ by $\tilde{\alpha}_t = R_t^{-1} R_t^{-\top}K_{n t}^\top Y$. Note that, for any $1 \leq t \leq m-1$, by assuming $G_t = R_t^\top R_t$ for an upper triangular matrix $R_t$, we have
\eqals{
& G_{t+1} = \begin{pmatrix} G_t & c_{t+1}\\ c_{t+1}^\top & \gamma_{t+1}\end{pmatrix} = \begin{pmatrix} R_t & 0\\ 0 & 0\end{pmatrix}^\top \begin{pmatrix} R_t & 0\\ 0 & 0\end{pmatrix} + C_{t+1},
}
with
\eqals{
C_{t+1} = \begin{pmatrix} 0 & c_{t+1}\\ c_{t+1}^\top & \gamma_{t+1}\end{pmatrix}, 
}
and $c_{t+1}$, $\gamma_{t+1}$ as in Subsection~\ref{sect:eff-inc-updates}.
Note moreover that $G_1 = \gamma_1$. Thus if we decompose the matrix $C_{t+1}$ in the form $C_{t+1} = u_{t+1} u_{t+1}^\top - v_{t+1} v_{t+1}^\top$ we are able compute $R_{t+1}$, the Cholesky matrix of $G_{t+1}$, by updating a bordered version of $R_t$ with two rank-one Cholesky updates. This is exactly Algorithm~\ref{alg:incr-nys-krls} with $u_{t+1}$ and $v_{t+1}$ as in Subsection~\ref{sect:eff-inc-updates}.
Note that the rank-one Cholesky update requires $O(t^2)$ at each call, while the computation of $c_t$ requires $O(n t)$ and the ones of $\tilde\alpha_t$ requires to solve two triangular linear systems, that is $O(t^2 + nt)$. Therefore the total cost for computing $\tilde\alpha_2,\dots,\tilde{\alpha}_m$ is $O(nm^2 + m^3)$.

\end{appendices}

\end{document}